%% file: main.tex
\documentclass{article}

\usepackage[preprint]{neurips_2022}
\usepackage[utf8]{inputenc}

\usepackage[title]{appendix}

\usepackage{hyperref}
\usepackage{url}

\input{macros}

\title{An Image is Worth One Word: Personalizing Text-to-Image Generation using Textual Inversion}
\author{Rinon Gal$^{1,2}$\thanks{\,\,Work was done during an internship at NVIDIA} \And Yuval Alaluf$^{1}$ \And Yuval Atzmon$^{2}$ \And Or Patashnik$^{1}$ \vspace{-20pt} \AND Amit H. Bermano$^{1}$ \And Gal Chechik$^{2}$ \And Daniel Cohen-Or$^{1}$ \vspace{-15pt} \AND
$^{1}$Tel-Aviv University \And $^{2}$NVIDIA}

\date{July 2022}

\begin{document}

\maketitle

\input{abstract}
\input{intro}

\input{related}
\input{method}
\input{experiments}
\input{analysis}
\input{future}

\bibliography{main}
\bibliographystyle{iclr2022_conference}

\clearpage

\title{Supplementary Materials \\ An Image is Worth One Word: Personalizing Text-to-Image Generation using Textual Inversion}
\author{Rinon Gal$^{1,2}$ \And Yuval Alaluf$^{1}$ \And Yuval Atzmon$^{2}$ \And Or Patashnik$^{1}$ \AND \vspace{-12pt} Amit H. Bermano$^{1}$ \And Gal Chechik$^{2}$ \And Daniel Cohen-Or$^{1}$ \AND \vspace{-17pt}
$^{1}$Tel-Aviv University \And $^{2}$NVIDIA}

\settitle

\appendix
\vspace{-15pt}
\input{appendix}

\end{document}

%% file: macros.tex
\usepackage{amsmath}
\usepackage{comment}
\usepackage{multirow,bigdelim}
\usepackage{lipsum}
\usepackage{array}
\usepackage{wrapfig}

\usepackage{adjustbox}
\usepackage[percent]{overpic}
\usepackage{makecell}
\usepackage[normalem]{ulem}
\usepackage{blindtext}
\usepackage{xcolor}
\usepackage{soul}
\usepackage{cleveref}
\usepackage{amsfonts}

\makeatletter
\newcommand{\settitle}{\@maketitle}
\makeatother

\newcolumntype{C}[1]{>{\centering\let\newline\\\arraybackslash\hspace{0pt}}m{#1}}

\newif\ifdraft
\drafttrue

\ifdraft
\definecolor{darkpink}{rgb}{0.561, 0.282, 0.427}

\newcommand{\dcc}[1]{{\color{red}[\textbf{DC:} #1]}}
\newcommand{\rgc}[1]{{\color{purple}[\textbf{RG:} #1]}}
\newcommand{\opc}[1]{{\color{blue}[\textbf{OP:} #1]}}

\newcommand{\abc}[1]{{\color{green}[\textbf{AB:} #1]}}

\newcommand{\drop}[1]{}

\else
\newcommand{\dcc}[1]{}
\newcommand{\rgc}[1]{}
\newcommand{\opc}[1]{}
\newcommand{\gcc}[1]{}
\newcommand{\hmc}[1]{}
\newcommand{\abc}[1]{}

\fi

\def\naive{na\"{\i}ve\xspace}

\makeatletter
\DeclareRobustCommand\onedot{\futurelet\@let@token\@onedot}
\def\@onedot{\ifx\@let@token.\else.\null\fi\xspace}

\def\eg{\emph{e.g}\onedot}

\def\ie{\emph{i.e}\onedot}

\def\etal{\emph{et al}\onedot}
\def\pholder{$S_*$}
\def\pholdercolor{{\color{blue}$S_*$}}

\def\mathpholdercolor{{\color{blue}S_*}}

\makeatother

\usepackage[bottom]{footmisc}
\usepackage{arydshln}

\makeatletter
\def\blfootnote{\xdef\@thefnmark{}\@footnotetext}
\makeatother

\DeclareMathOperator*{\argmin}{arg\,min}

%% file: abstract.tex
\vspace{-12pt}
\begin{abstract}

Text-to-image models offer unprecedented freedom to guide creation through natural language. Yet, it is unclear how such freedom can be exercised to generate images of specific unique concepts, modify their appearance, or compose them in new roles and novel scenes.
In other words, we ask: how can we use language-guided models to turn \textit{our} cat into a painting, or imagine a new product based on \textit{our} favorite toy? 
Here we present a simple approach that allows such creative freedom. Using only $3$-$5$ images of a user-provided concept, like an object or a style, we learn to represent it through new ``words" in the embedding space of a frozen text-to-image model.
These ``words" can be composed into natural language sentences, guiding \textit{personalized} creation in an intuitive way.
Notably, we find evidence that a \textit{single} word embedding is sufficient for capturing unique and varied concepts. 
We compare our approach to a wide range of baselines, and demonstrate that it can more faithfully portray the concepts across a range of applications and tasks. 

Our code, data and new words will be available at: \url{https://textual-inversion.github.io}

\end{abstract}

%% file: intro.tex
\vspace{-5pt}
\input{resources/figures/teaser}

\section{Introduction}

In a famous scene from the motion picture ``Titanic'', Rose makes a request of Jack: ``...draw me like one of your French girls".
Albeit simple, this request contains a wealth of information. It indicates that Jack should produce a drawing; It suggests that its style and composition should match those of a subset of Jack's prior work; Finally, through a single word, ``me'', Rose indicates that this drawing should portray a specific, unique subject: Rose herself.
In making her request, Rose relies on Jack's ability to reason over these concepts --- both broad and specific --- and bring them to life in a new creation.

Recently, large-scale text-to-image models~\citep{rombach2021highresolution,ramesh2021zero,ramesh2022hierarchical,nichol2021glide,yu2022scaling,saharia2022photorealistic} have demonstrated an unprecedented capability to reason over natural language descriptions. They allow users to synthesize novel scenes with unseen compositions and produce vivid pictures in a myriad of styles. These tools have been used for artistic creation, as sources of inspiration, and even to design new, physical products~\citep{yacoubian2022avocadobag}. Their use, however, is constrained by the user's ability to describe the desired target through text. Turning back to Rose, one could then ask: How might she frame her request if she were to approach one of these models? How could we, as users, ask text-to-image models to craft a novel scene containing a cherished childhood toy? Or to pull our child's drawing from its place on the fridge, and turn it into an artistic showpiece? 

Introducing new concepts into large scale models is often difficult. Re-training a model with an expanded dataset for each new concept is prohibitively expensive, and fine-tuning on few examples typically leads to catastrophic forgetting~\citep{ding2022don,Li2022}. More measured approaches freeze the model and train transformation modules to adapt its output when faced with new concepts~\citep{zhou2021learning,gao2021clip,skantze2022collie}. However, these approaches are still prone to forgetting prior knowledge, or face difficulties in accessing it concurrently with newly learned concepts~\citep{kumar2022fine,cohen2022my}.

We propose to overcome these challenges by \textit{finding} new words in the textual embedding space of pre-trained text-to-image models. We consider the first stage of the text encoding process (\Cref{fig:embedding}). Here, an input string is first converted to a set of tokens. Each token is then replaced with its own embedding vector, and these vectors are fed through the downstream model. Our goal is to find new embedding vectors that represent new, specific concepts.

We represent a new embedding vector with a new \textit{pseudo-word}~\citep{rathvon2004early} 
which we denote by $S_*$. This pseudo-word is then treated like any other word, and can be used to compose novel textual queries for the generative models. One can therefore ask for ``a photograph of $S_*$ on the beach", ``an oil painting of a $S_*$ hanging on the wall", or even compose two concepts, such as ``a drawing of $S^1_*$ in the style of $S^2_*$". Importantly, this process leaves the generative model untouched. In doing so, we retain the rich textual understanding and generalization capabilities that are typically lost when fine-tuning vision and language models on new tasks.

To find these pseudo-words, we frame the task as one of inversion. We are given a fixed, pre-trained text-to-image model and a small (3-5) image set depicting the concept. We aim to find a single word embedding, such that sentences of the form ``A photo of $S_*$" will lead to the reconstruction of images from our small set. This embedding is found through an optimization process, which we refer to as ``Textual Inversion''. 

We further investigate a series of extensions based on tools typically used in Generative Adversarial Network (GAN) inversion. Our analysis reveals that, while some core principles remain, applying the prior art in a \naive way is either unhelpful or actively harmful.

We demonstrate the effectiveness of our approach over a wide range of concepts and prompts, showing that it can inject unique objects into new scenes, transform them across different styles, transfer poses, diminish biases, and even imagine new products.

In summary, our contributions are as follows:
\begin{itemize}
    \item We introduce the task of personalized text-to-image generation, where we synthesize novel scenes of user-provided concepts guided by natural language instruction.
    \item We present the idea of ``Textual Inversions" in the context of generative models. Here the goal is to find new pseudo-words in the embedding space of a text encoder that can capture both high-level semantics and fine visual details. 
    \item We analyze the embedding space in light of GAN-inspired inversion techniques and demonstrate that it also exhibits a tradeoff between distortion and editability. We show that our approach resides on an appealing point on the tradeoff curve.
    \item We evaluate our method against images generated using user-provided captions of the concepts and demonstrate that our embeddings provide higher visual fidelity, and also enable more robust editing.
\end{itemize}

%% file: resources/figures/teaser.tex
\begin{figure}[!hbt]
    \centering
    \setlength{\abovecaptionskip}{6.5pt}
    \setlength{\belowcaptionskip}{-3.5pt}
    \setlength{\tabcolsep}{0.55pt}
    \renewcommand{\arraystretch}{1.0}
    {
    \fontsize{8pt}{8pt}\selectfont %
    
    \begin{tabular}{c@{\hskip 5pt} c@{\hskip 5pt} c c c c}
    
        \begin{tabular}{c c}
            \includegraphics[width=0.09\linewidth,height=0.09\linewidth]{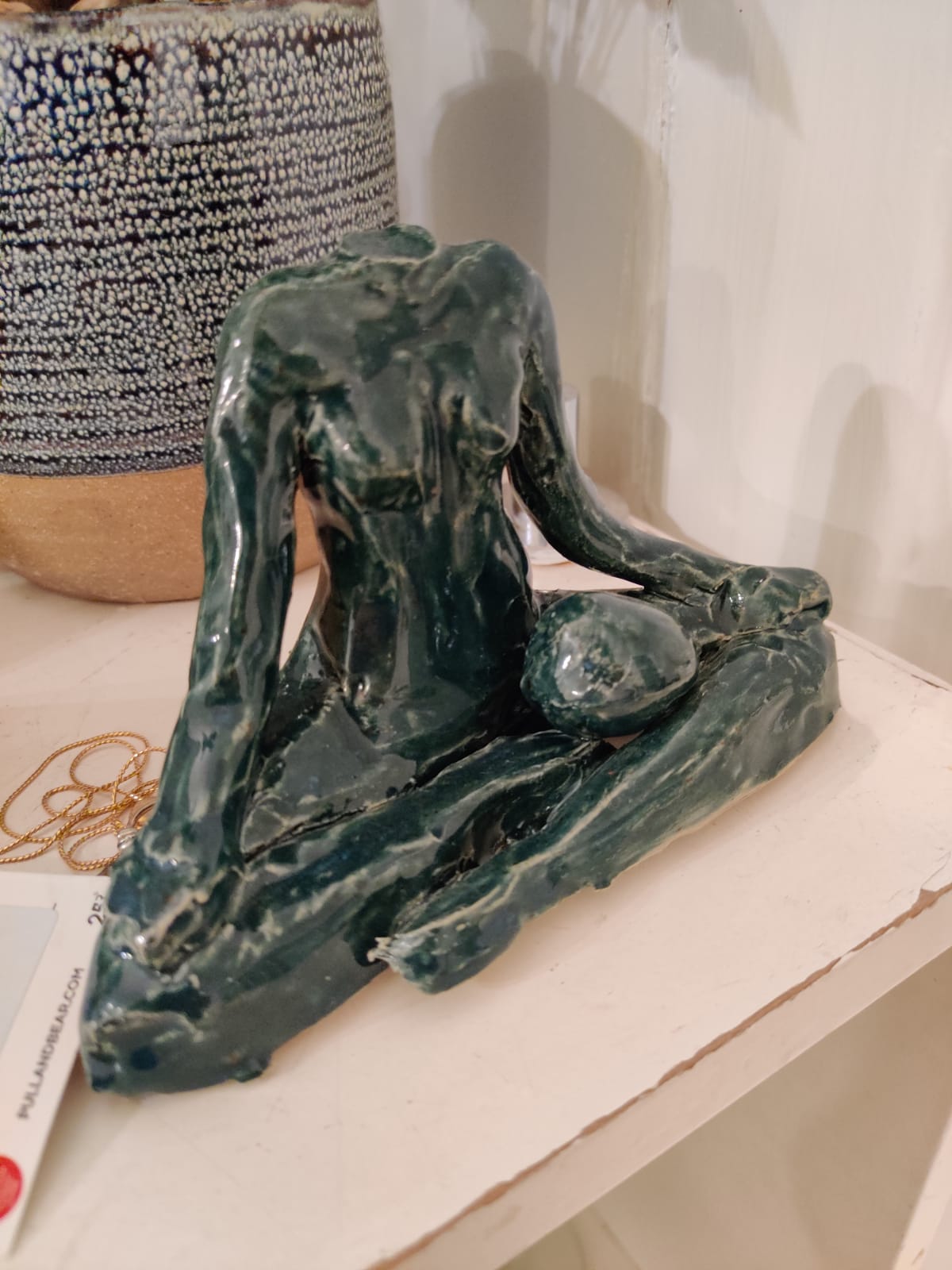} & 
            \includegraphics[width=0.09\linewidth,height=0.09\linewidth]{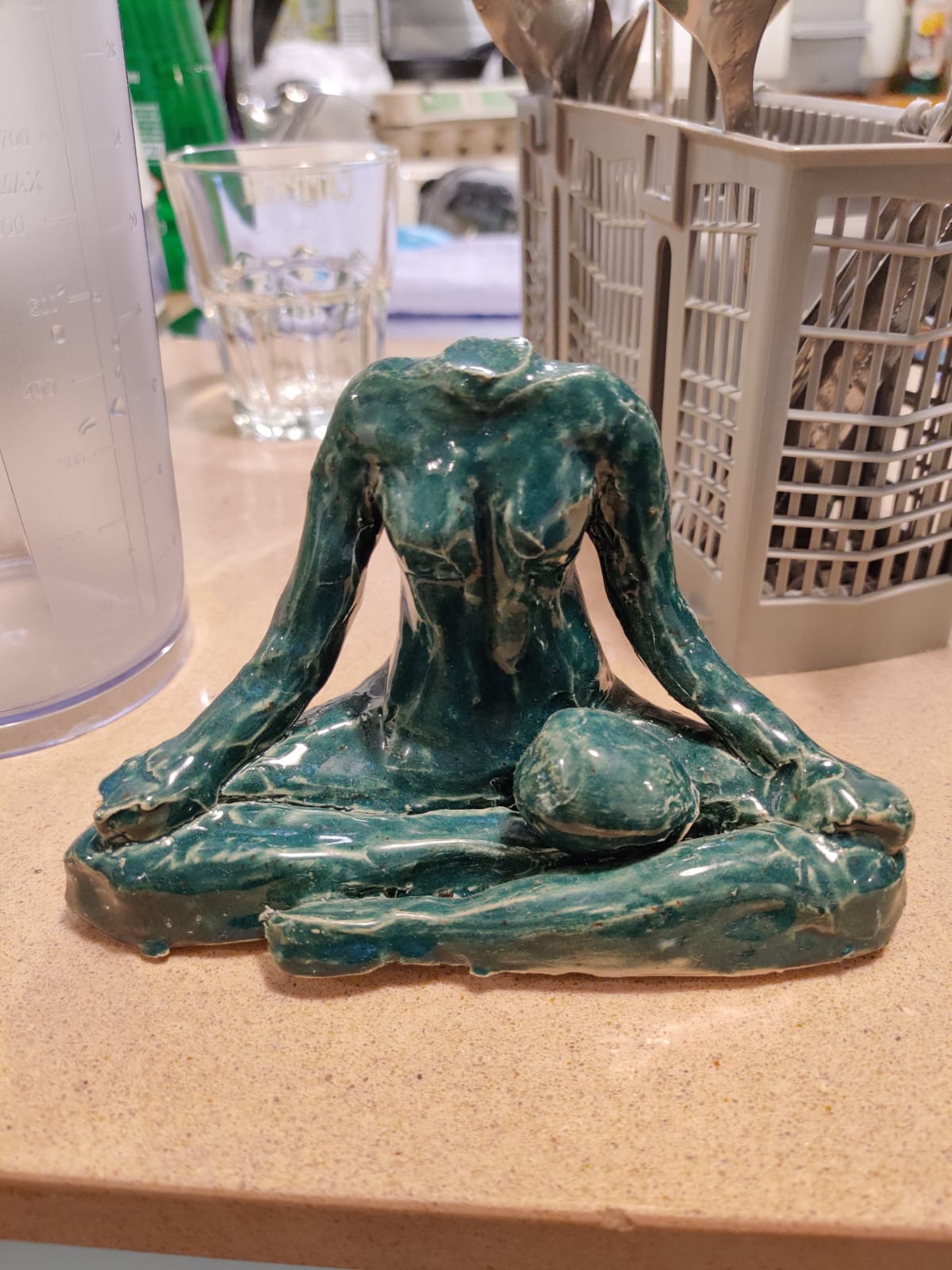} \\
            \includegraphics[width=0.09\linewidth,height=0.09\linewidth]{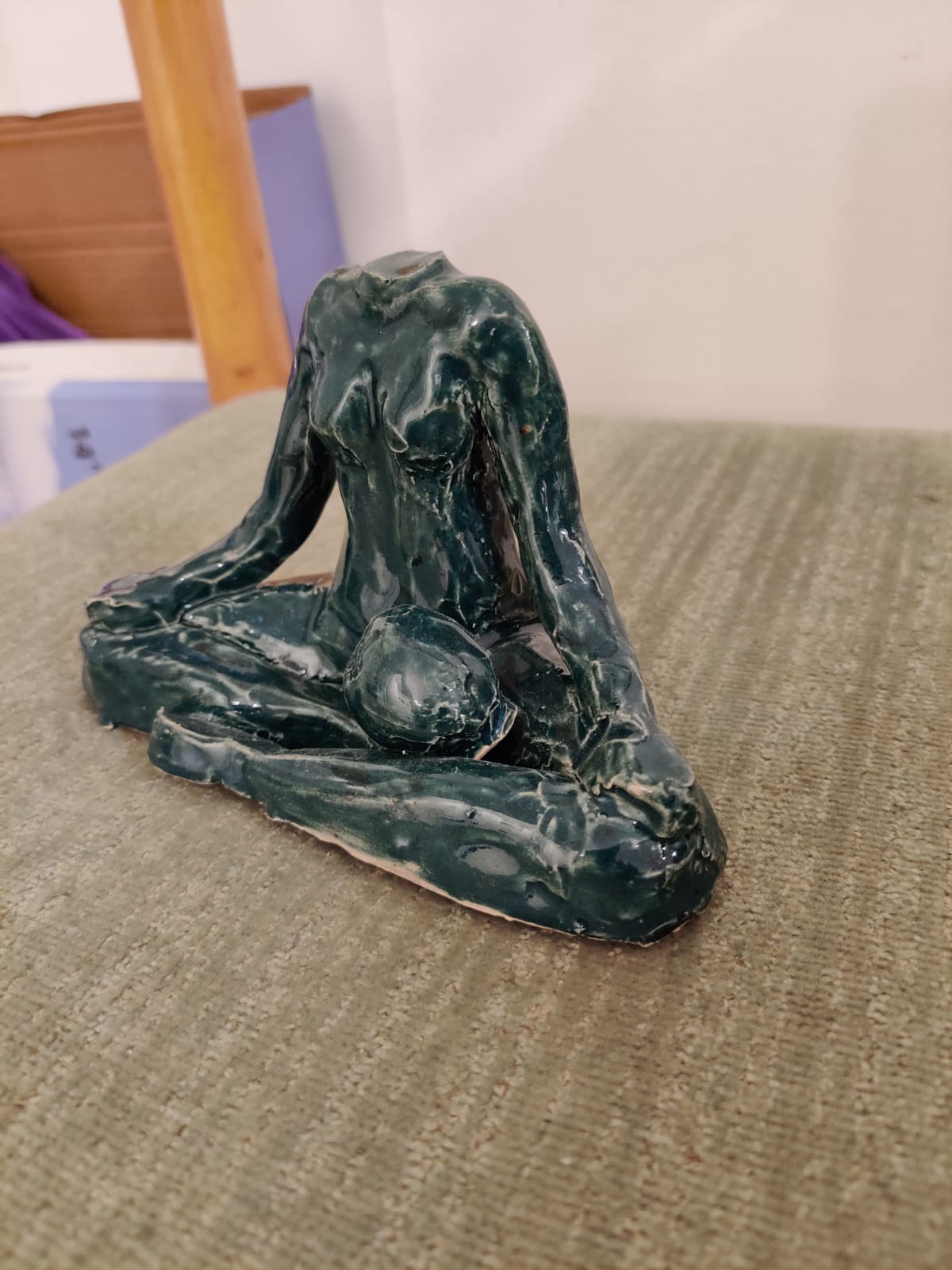} & 
            \includegraphics[width=0.09\linewidth,height=0.09\linewidth]{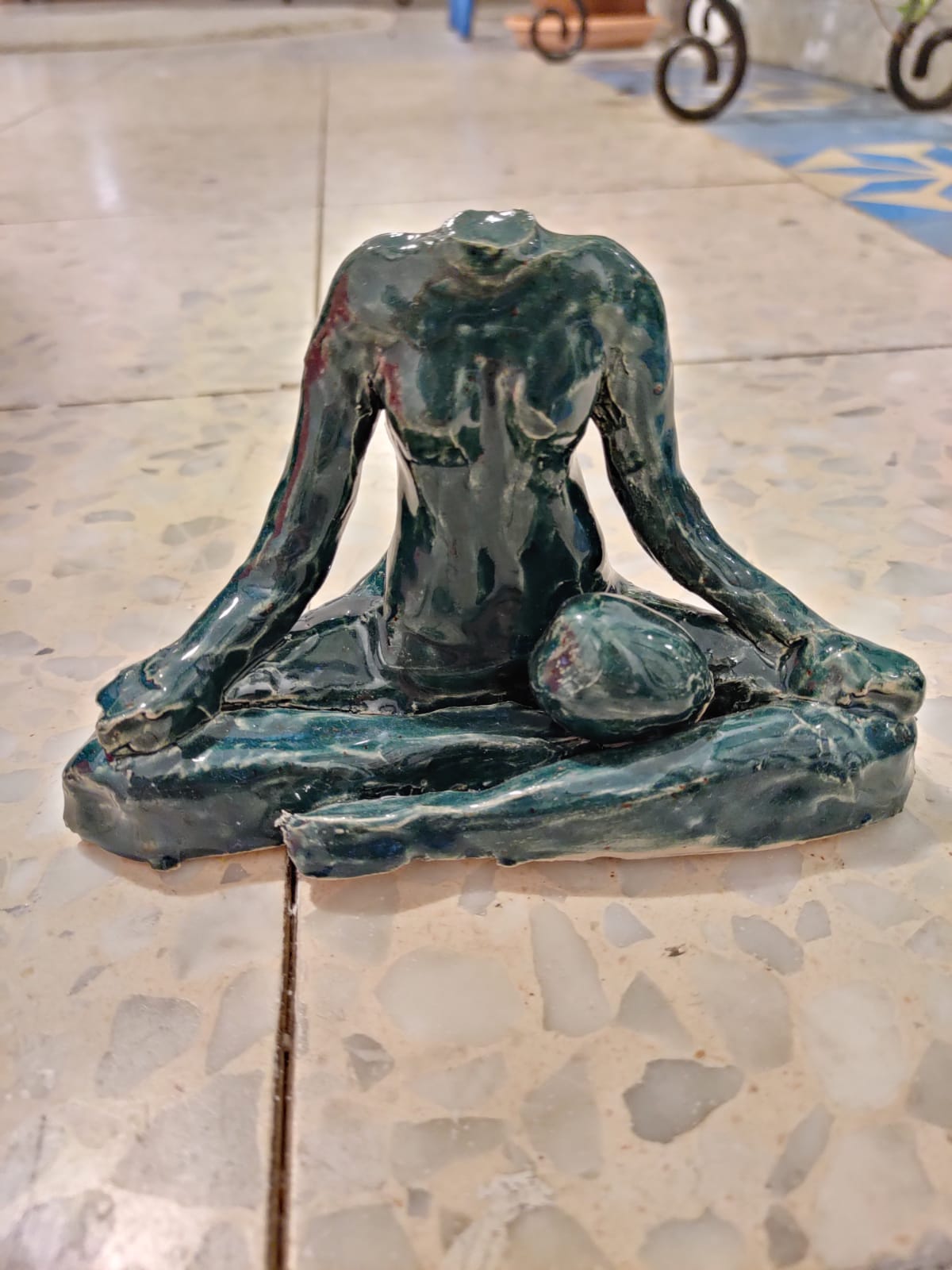}
        \end{tabular}
        
        &
        $\rightarrow$
        &
        \begin{tabular}{c}
        \includegraphics[width=0.184\linewidth]{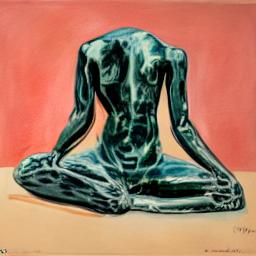}
        \end{tabular} &
        \begin{tabular}{c}
        \includegraphics[width=0.184\linewidth]{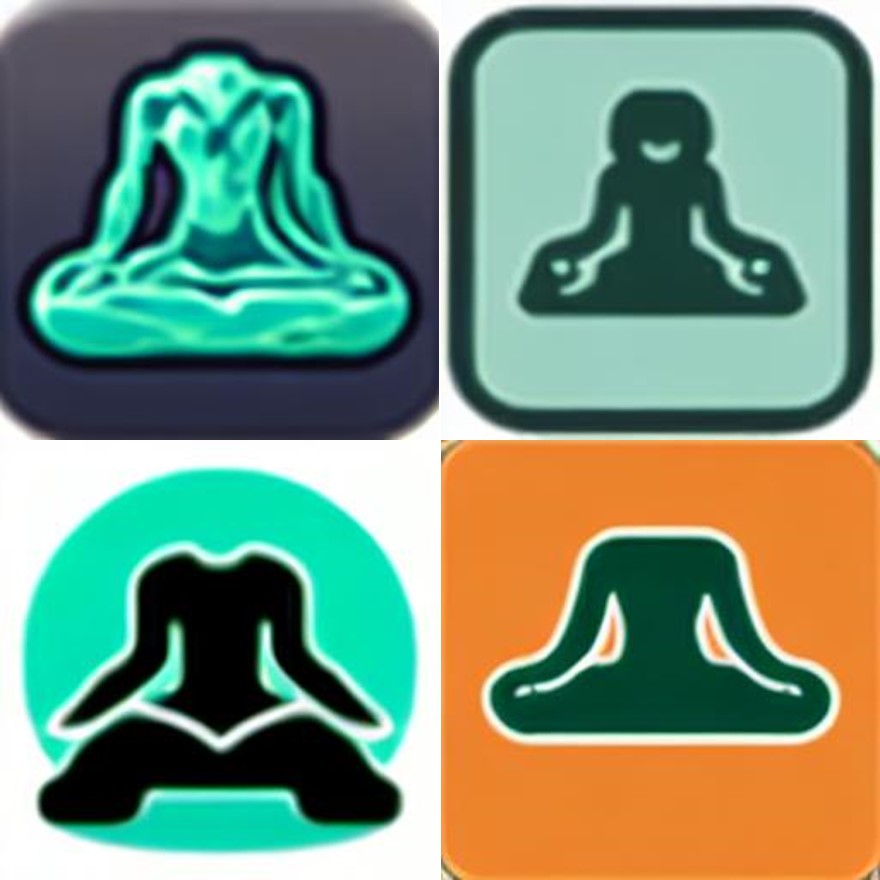} 
        \end{tabular} &
        \begin{tabular}{c}
        \includegraphics[width=0.184\linewidth]{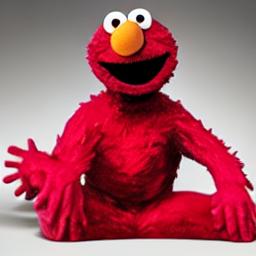} 
        \end{tabular} &
        \begin{tabular}{c}
        \includegraphics[width=0.184\linewidth]{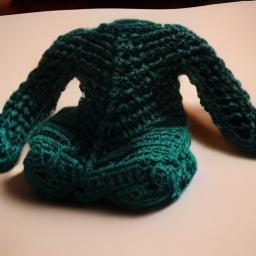}
        \end{tabular} \\
        
        { %
        \fontsize{8pt}{8pt}\selectfont %
        Input samples $\xrightarrow{invert} ``\mathpholdercolor"$} & & {\begin{tabular}{c@{}c@{}c@{}c@{}} ``An oil painting of \pholdercolor" \end{tabular}} & {\begin{tabular}{c@{}c@{}c@{}c@{}} ``App icon of \pholdercolor" \end{tabular}} & 
        {\begin{tabular}{c@{}c@{}c@{}c@{}} ``Elmo sitting in \\ the same pose as \pholdercolor" \end{tabular}} &
        {\begin{tabular}{c@{}c@{}c@{}c@{}} ``Crochet \pholdercolor" \end{tabular}} \\ \\
        
        \begin{tabular}{c c}
            \includegraphics[width=0.09\linewidth,height=0.09\linewidth]{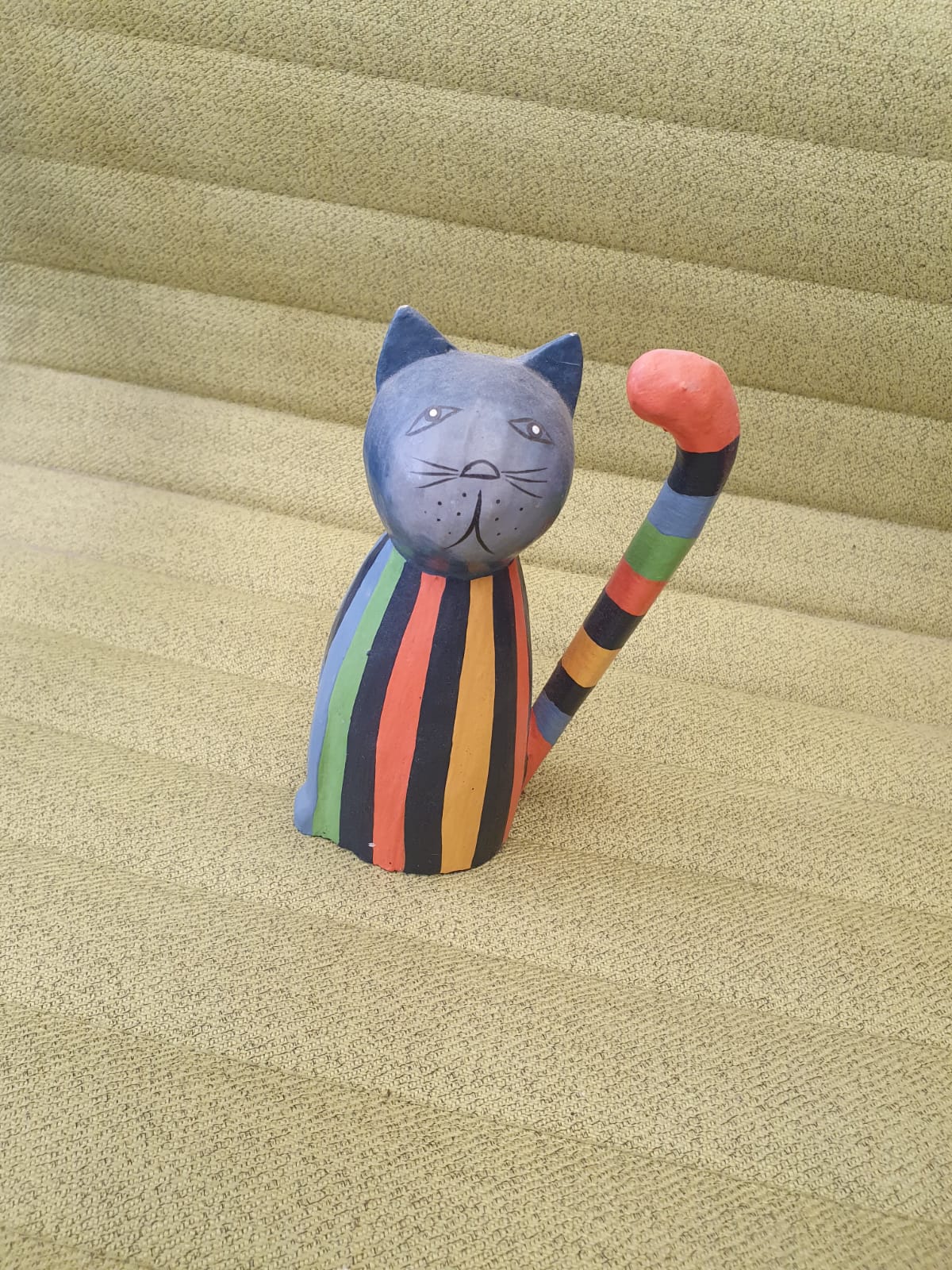} & 
            \includegraphics[width=0.09\linewidth,height=0.09\linewidth]{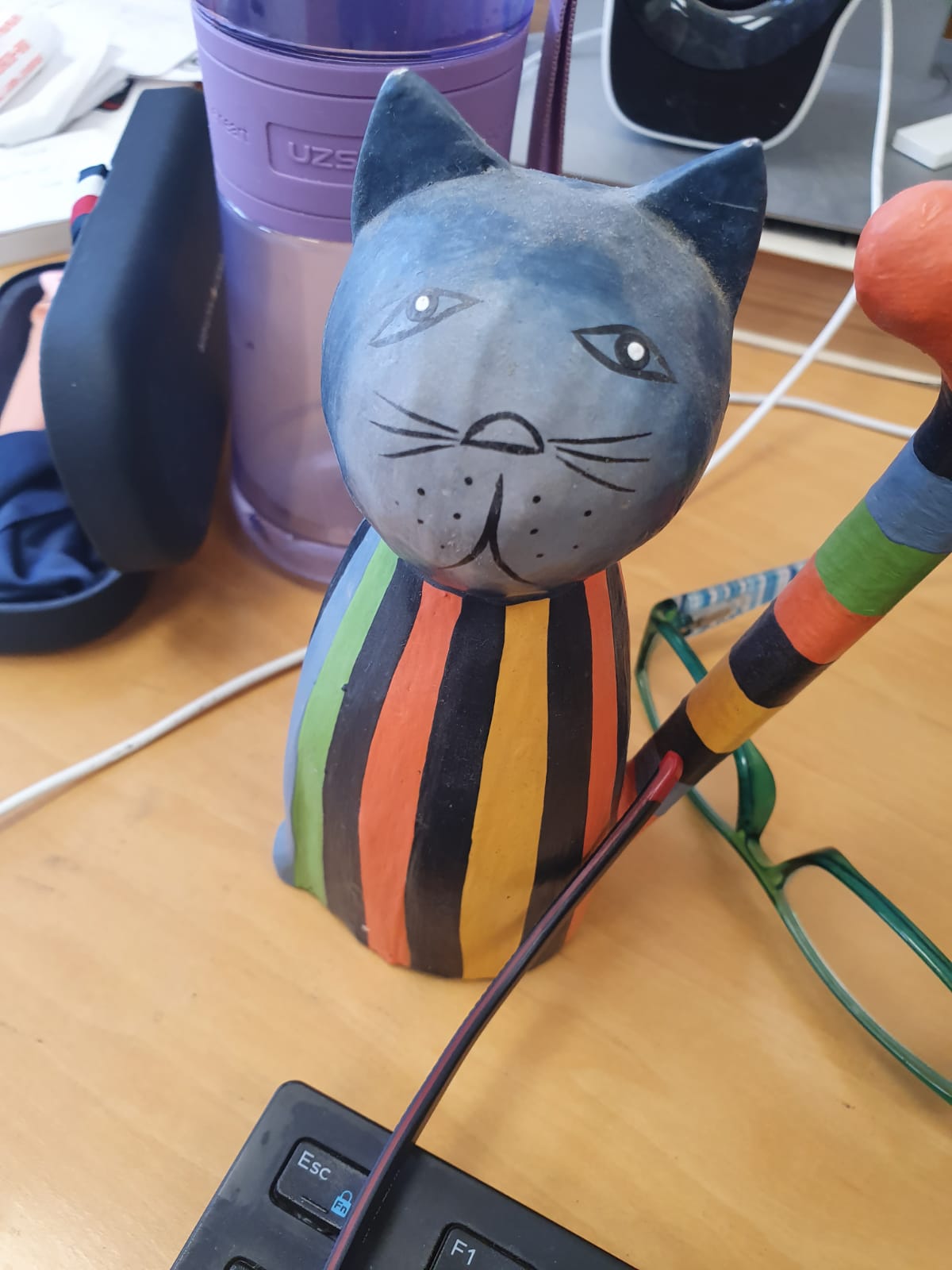} \\
            \multicolumn{2}{c}{\includegraphics[width=0.09\linewidth,height=0.09\linewidth]{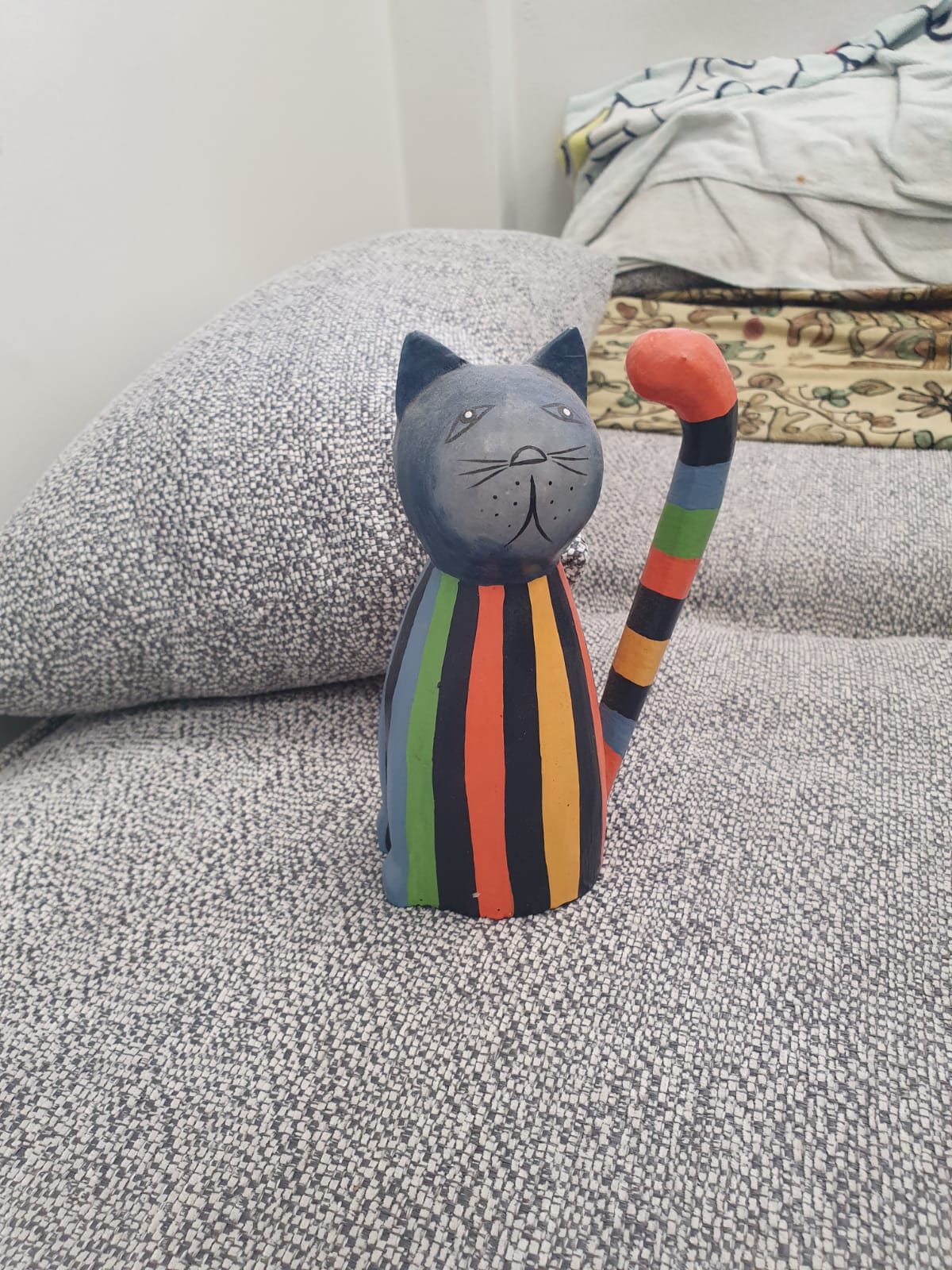}}
        \end{tabular}
        &
        $\rightarrow$
        &
        \begin{tabular}{c}
        \includegraphics[width=0.184\linewidth]{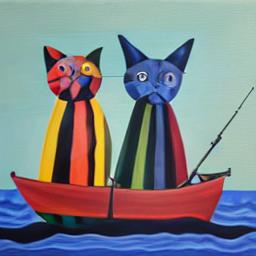} 
        \end{tabular} &
        \begin{tabular}{c}
        \includegraphics[width=0.184\linewidth]{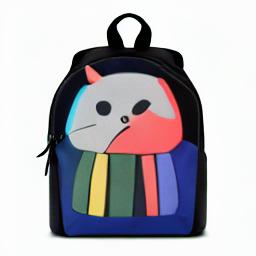} 
        \end{tabular} &
        \begin{tabular}{c}
        \includegraphics[width=0.184\linewidth]{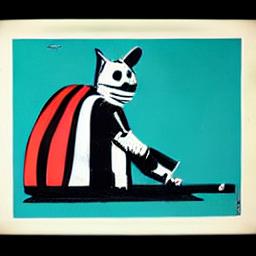} 
        \end{tabular} &
        \begin{tabular}{c}
        \includegraphics[width=0.184\linewidth]{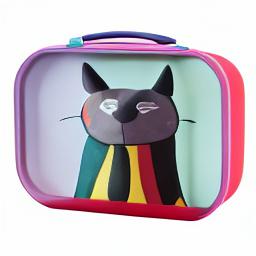} 
        \end{tabular} \\
        
        { %
        \fontsize{8pt}{8pt}\selectfont %
        Input samples $\xrightarrow{invert} ``\mathpholdercolor"$} & & {\begin{tabular}{c@{}c@{}c@{}c@{}} ``Painting of two \pholdercolor \\ fishing on a boat" \end{tabular}} & {\begin{tabular}{c@{}c@{}c@{}c@{}} ``A \pholdercolor{} backpack" \end{tabular}} & {\begin{tabular}{c@{}c@{}c@{}c@{}} ``Banksy art of \pholdercolor" \end{tabular}} & {\begin{tabular}{c@{}c@{}c@{}c@{}} ``A \pholdercolor{} themed lunchbox" \end{tabular}} \\

    \end{tabular}}
    \caption{(left) We find new pseudo-words in the embedding space of a pre-trained text-to-image model which describe specific concepts. (right) These pseudo-words can be composed into new sentences, placing our targets in new scenes, changing their style or composition, or ingraining them into new products.}
    \label{fig:teaser} 
    \vspace{-5pt}
\end{figure}

%% file: related.tex
\section{Related work}

\paragraph{Text-guided synthesis.}

Text-guided image synthesis has been widely studied in the context of GANs~\citep{goodfellow2014generative}. Typically, a conditional model is trained to reproduce samples from given paired image-caption datasets~\citep{zhu2019dm,tao2020df}, leveraging attention mechanisms~\citep{xu2018attngan} or cross-modal contrastive approaches~\citep{zhang2021cross,ye2021improving}. More recently, impressive visual results were achieved by leveraging large scale auto-regressive~\citep{ramesh2021zero,yu2022scaling} or diffusion models~\citep{ramesh2022hierarchical,saharia2022photorealistic,nichol2021glide,rombach2021highresolution}.

Rather than training conditional models, several approaches employ test-time optimization to explore the latent spaces of a pre-trained generator~\citep{crowson2022vqgan,murdock2021bigsleep,crowson2021diffusion}. These models typically guide the optimization to minimize a text-to-image similarity score derived from an auxiliary model such as CLIP~\citep{radford2021learning}.

Moving beyond pure image generation, a large body of work explores the use of text-based interfaces for image editing~\citep{patashnik2021styleclip,abdal2021clip2stylegan,avrahami2022blended}, generator domain adaptation~\citep{gal2021stylegan,kim2022diffusionclip}, video manipulation~\citep{tzaban2022stitch,bar2022text2live}, motion synthesis~\citep{tevet2022motionclip,petrovich22temos}, style transfer~\citep{kwon2021clipstyler,liu2022name} and even texture synthesis for 3D objects~\citep{text2mesh}.

Our approach builds on the open-ended, conditional synthesis models. Rather than training a new model from scratch, we show that we can expand a frozen model's vocabulary and introduce new pseudo-words that describe specific concepts.

\paragraph{GAN inversion.}
Manipulating images with generative networks often requires one to find a corresponding latent representation of the given image, a process referred to as \textit{inversion}~\citep{zhu2016generative,xia2021gan}.
In the GAN literature, this inversion is done through either an optimization-based technique~\citep{abdal2019image2stylegan,abdal2020image2stylegan++,zhu2020improved,gu2020image} or by using an encoder~\citep{richardson2020encoding,zhu2020domain,pidhorskyi2020adversarial,tov2021designing}. Optimization methods directly optimize a latent vector, such that feeding it through the GAN will re-create a target image. Encoders leverage a large image set to train a network that maps images to their latent representations. 

In our work, we follow the optimization approach, as it can better adapt to unseen concepts. Encoders face harsher generalization requirements, and would likely need to be trained on web-scale data to offer the same freedom. We further analyze our embedding space in light of the GAN-inversion literature, outlining the core principles that remain and those that do not.

\paragraph{Diffusion-based inversion.}
In the realm of diffusion models, inversion can be performed \naive{ly} by adding noise to an image and then de-noising it through the network. However, this process tends to change the image content significantly. \citet{choi2021ilvr} improve inversion by conditioning the denoising process on noised low-pass filter data from the target image.
\citep{dhariwal2021diffusion} demonstrate that the DDIM~\citep{song2020denoising} sampling process can be inverted in a closed-form manner, extracting a latent noise map that will produce a given real image. In DALL-E 2~\citep{ramesh2022hierarchical}, they build on this method and demonstrate that it can be used to induce changes in the image, such as cross-image interpolations or semantic editing. The later relies on their use of CLIP-based codes to condition the model, and may not be applicable to other methods.

Whereas the above works invert a given \textit{image} into the model's latent space, we invert a user-provided \textit{concept}. Moreover, we represent this concept as a new pseudo-word in the model's vocabulary, allowing for more general and intuitive editing.

\paragraph{Personalization.}
Adapting models to a specific individual or object is a long-standing goal in machine learning research. Personalized models are typically found in the realms of recommendation systems~\citep{benhamdi2017personalized,fernando2018artwork,martinez2009s,cho2002personalized} or in federated learning~\citep{mansour2020three,jiang2019improving,fallah2020personalized,shamsian2021personalized}.

More recently, personalization efforts can also be found in vision and graphics. There it is typical to apply a delicate tuning of a generative model to better reconstruct specific faces or scenes~\citep{semantic2019bau,roich2021pivotal,alaluf2021hyperstyle,dinh2022hyperinverter,cao2022authentic,nitzan2022mystyle}. 

Most relevant to our work is PALAVRA~\citep{cohen2022my}, which leverages a pre-trained CLIP model for retrieval and segmentation of personalized objects. PALAVRA identifies pseudo-words in the textual embedding space of CLIP that refer to a specific object. These are then used to describe images for retrieval, or in order to segment specific objects in a scene. However, their task and losses are both discriminative, aiming to separate the object from other candidates. As we later show (\Cref{fig:baseline_comp}), their approach fails to capture the details required for plausible reconstructions or synthesis in new scenes.

%% file: method.tex
\section{Method}
\label{sec:method}

\input{resources/figures/fig_embedding}

Our goal is to enable language-guided generation of new, user-specified concepts. To do so, we aim to encode these concepts into an intermediate representation of a pre-trained text-to-image model. 
Ideally, this should be done in a manner that would allow us to leverage the rich semantic and visual prior represented by such a model, and use it to guide intuitive visual transformations of the concepts.

It is natural to search for candidates for such a representation in the word-embedding stage of the text encoders typically employed by text-to-image models. There, the discrete input text is first converted into a continuous vector representation that is amenable to direct optimization.

Prior work has shown that this embedding space is expressive enough to capture basic image semantics~\citep{cohen2022my,tsimpoukelli2021multimodal}. However, these approaches leveraged contrastive or language-completion objectives, neither of which require an in-depth visual understanding of the image. As we demonstrate in \Cref{sec:applications}, those methods fail to accurately capture the appearance of the concept, and attempting to employ them for synthesis leads to considerable visual corruption. Our goal is to find pseudo-words that can guide \textit{generation}, which is a \textit{visual} task. As such, we propose to find them through a \textit{visual} reconstruction objective.

Below, we outline the core details of applying our approach to a specific class of generative models --- Latent Diffusion Models~\citep{rombach2021highresolution}. In \Cref{sec:analysis}, we then analyze a set of extensions to this approach, motivated by GAN-inversion literature. However, as we later show, these additional complexities fail to improve upon the initial representation, presented here.

\paragraph{Latent Diffusion Models.}
We implement our method over Latent Diffusion Models (LDMs)~\citep{rombach2021highresolution}, a recently introduced class of Denoising Diffusion Probabilistic Models (DDPMs)~\citep{ho2020denoising} that operate in the latent space of an autoencoder. 

LDMs consist of two core components. First, an autoencoder is pre-trained on a large collection of images. An encoder $\mathcal{E}$ learns to map images $x\in \mathcal{D}_x$ into a spatial latent code $z = \mathcal{E}(x)$, regularized through either a KL-divergence loss or through vector quantization \citep{van2017neural,agustsson2017soft}. The decoder $D$ learns to map such latents back to images, such that $D\left(\mathcal{E}(x)\right)\approx x$.

The second component, a diffusion model, is trained to produce codes within the learned latent space. This diffusion model can be conditioned on class labels, segmentation masks, or even on the output of a jointly trained text-embedding model. Let $c_\theta(y)$ be a model that maps a conditioning input $y$ into a conditioning vector. The LDM loss is then given by:
\begin{equation}
    L_{LDM} := \mathbb{E}_{z\sim\mathcal{E}(x), y, \epsilon \sim \mathcal{N}(0, 1), t }\Big[ \Vert \epsilon - \epsilon_\theta(z_{t},t, c_\theta(y)) \Vert_{2}^{2}\Big] \, ,
    \label{eq:LDM_loss}
\end{equation}
where $t$ is the time step, $z_t$ is the latent noised to time $t$, $\epsilon$ is the unscaled noise sample, and $\epsilon_\theta$ is the denoising network. 
Intuitively, the objective here is to correctly remove the noise added to a latent representation of an image. While training, $c_\theta$ and $\epsilon_\theta$ are jointly optimized to minimize the LDM loss. At inference time, a random noise tensor is sampled and iteratively denoised to produce a new image latent, $z_0$. Finally, this latent code is transformed into an image through the pre-trained decoder $x' = D(z_0)$.

We employ the publicly available 1.4 billion parameter text-to-image model of \citet{rombach2021highresolution}, which was pre-trained on the LAION-400M dataset~\citep{schuhmann2021laion}. Here, $c_\theta$ is realized through a BERT~\citep{devlin2018bert} text encoder, with $y$ being a text prompt. 

We next review the early stages of such a text encoder, and our choice of inversion space.

\paragraph{Text embeddings.}

Typical text encoder models, such as BERT, begin with a text processing step (\Cref{fig:embedding}, left). First, each word or sub-word in an input string is converted to a token, which is an index in some pre-defined dictionary. 
Each token is then linked to a unique embedding vector that can be retrieved through an index-based lookup. These embedding vectors are typically learned as part of the text encoder $c_\theta$.

In our work, we choose this embedding space as the target for inversion. Specifically, we designate a placeholder string, $S_*$, to represent the new concept we wish to learn. 
We intervene in the embedding process and replace the vector associated with the tokenized string with a new,  \textit{learned} embedding $v_*$, in essence ``injecting'' the concept into our vocabulary. In doing so, we can then compose new sentences containing the concept, just as we would with any other word. %

\paragraph{Textual inversion.}
To find these new embeddings, we use a small set of images (typically $3$-$5$), which depicts our target concept across multiple settings such as varied backgrounds or poses. We find $v_*$ through direct optimization, by minimizing the LDM loss of \Cref{eq:LDM_loss} over images sampled from the small set. To condition the generation, we randomly sample neutral context texts, derived from the CLIP ImageNet templates~\citep{radford2021learning}. These contain prompts of the form ``A photo of $S_*$", ``A rendition of $S_*$", etc. The full list of templates is provided in the supplementary materials.

Our optimization goal can then be defined as:
\begin{equation}
   v_* = \argmin_v \mathbb{E}_{z\sim\mathcal{E}(x), y, \epsilon \sim \mathcal{N}(0, 1), t }\Big[ \Vert \epsilon - \epsilon_\theta(z_{t},t, c_\theta(y)) \Vert_{2}^{2}\Big] \, ,
    \label{eq:v_opt}
\end{equation}
and is realized by re-using the same training scheme as the original LDM model, while keeping both $c_\theta$ and $\epsilon_\theta$ fixed. Notably, this is a reconstruction task. As such, we expect it to motivate the learned embedding to capture fine visual details unique to the concept.

\paragraph{Implementation details.}
Unless otherwise noted, we retain the original hyper-parameter choices of LDM~\citep{rombach2021highresolution}. Word embeddings were initialized with the embeddings of a single-word coarse descriptor of the object (\eg ``sculpture" and ``cat" for the two concepts in \Cref{fig:teaser}). Our experiments were conducted using $2\times$V100 GPUs with a batch size of 4. The base learning rate was set to $0.005$. Following LDM, we further scale the base learning rate by the number of GPUs and the batch size, for an effective rate of $0.04$. All results were produced using $5,000$ optimization steps. We find that these parameters work well for most cases. However, we note that for some concepts, better results can be achieved with fewer steps or with an increased learning rate.

%% file: resources/figures/fig_embedding.tex
\begin{figure}[t]
\setlength{\abovecaptionskip}{6.5pt}
\setlength{\belowcaptionskip}{-5pt}
\setlength{\tabcolsep}{1pt}
    \centering
    \includegraphics[width=0.95\linewidth]{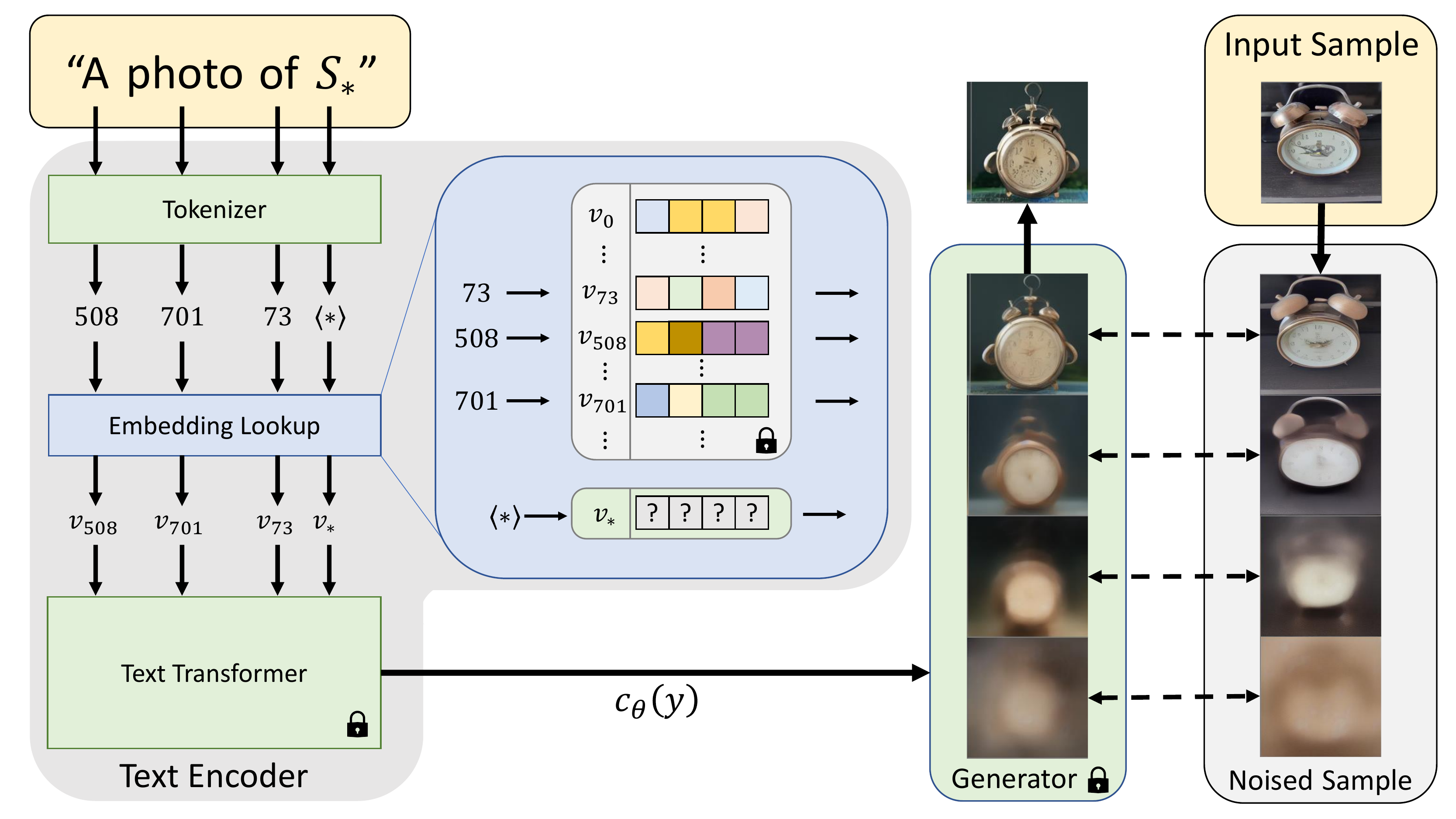}
    \caption{Outline of the text-embedding and inversion process. A string containing our placeholder word is first converted into tokens (\ie word or sub-word indices in a dictionary). These tokens are converted to continous vector representations (the ``embeddings", $v$). Finally, the embedding vectors are transformed into a single conditioning code $c_\theta(y)$ which guides the generative model. We optimize the embedding vector $v_*$ associated with our pseudo-word \pholder, using a reconstruction objective.}
    \label{fig:embedding}
    \vspace{-5pt}
\end{figure}

%% file: experiments.tex
\section{Qualitative comparisons and applications} 

In the following section, we demonstrate a range of applications enabled through Textual Inversions, and provide visual comparisons to the state-of-the-art and human-captioning baselines. 

\label{sec:applications}

\subsection{Image variations}
\label{sec:image_variations}

\input{resources/figures/user_sentences_comparison}

We begin by demonstrating our ability to capture and recreate variations of an object using a single pseudo-word. In \Cref{fig:user_sentences_comp} we compare our method to two baselines: LDM guided by a human caption and DALLE-2 guided by either a human caption or an image prompt. Captions were collected using Mechanical Turk. Annotators were provided with four images of a concept and asked to describe it in a manner that could allow an artist to recreate it. We asked for both a short ($\leq12$ words) and a long ($\leq30$ words) caption. In total, we collected $10$ captions per concept --- five short and five long. \Cref{fig:user_sentences_comp} shows multiple results generated with a randomly chosen caption for each setup. Additional large-scale galleries showing our uncurated reconstructions are provided in the supplementary.

As our results demonstrate, our method better captures the unique details of the concept. Human captioning typically captures the most prominent features of an object, but provides insufficient detail to reconstruct finer features like color patterns (\eg of the teapot). In some cases (\eg the skull mug) the object itself may be exceedingly difficult to describe through natural language. When provided with an image, DALLE-2 is able to recreate more appealing samples, particularly for well-known objects with limited detail (Aladdin's lamp). However, it still struggles with unique details of personalized objects that the image encoder (CLIP) is unlikely to have seen (mug, teapot). In contrast, our method can successfully capture these finer details, and it does so using only a single word embedding. However, note that while our creations are more similar to the source objects, they are still variations that may differ from the source.

\subsection{Text-guided synthesis}

\input{resources/figures/generation}

In \Cref{fig:conditional_gen,fig:teaser} we show our ability to compose novel scenes by incorporating the learned pseudo-words into new conditioning texts. For each concept, we show exemplars from our training set, along with an array of generated images and their conditioning texts. As our results demonstrate, the frozen text-to-image model is able to jointly reason over both the new concepts and its large body of prior knowledge, bringing them together in a new creation. Importantly, despite the fact that our training goal was generative in nature, our pseudo-words still encapsulate semantic concepts that the model can then leverage. For example, observe the bowl's ability (row four) to contain other objects like food, or the ability to preserve the Furby's bird-like head and crown while adapting his palette to better match a prompt (album cover, row three). Additional concepts and texts are provided in the supplementary materials.

\input{resources/figures/baseline_comparisons}

To better evaluate our ability to compose objects into new scenes, we compare our method to several personalization baselines (\Cref{fig:baseline_comp}). In particular, we consider the recent PALAVRA~\citep{cohen2022my}, which is most similar to our own work. PALAVRA encodes object sets into the textual embedding space of CLIP, using a mix of contrastive learning and cyclic consistency goals. We find a new pseudo-word using their approach and use it to synthesize new images by leveraging VQGAN-CLIP~\citep{crowson2022vqgan} and CLIP-Guided Diffusion~\citep{crowson2021diffusion}. As a second baseline, we apply the CLIP-guided models of Crowson~\etal while trying to jointly minimize the CLIP-based distances to both the training set images and to the target text (VQGAN-CLIP) or by initializing the optimization with an input image from our set (Guided Diffusion). For the latter, we chose image-based initializations as we observed that they outperform the use of images in the optimization loss. Similar observations were reported in Disco Diffusion~\citep{lettsDiscoDiffusion2021}.

The images produced by PALAVRA (rows $2$, $3$) typically contain elements from the target prompt (\eg a beach, a moon) but they fail to accurately capture the concept and display considerable visual corruption. This is unsurprising, as PALAVRA was trained with a discriminative goal. In their case, the model needs to only encode enough information to distinguish between two typical concepts (\eg it may be sufficient to remember the mug was black-and-white with text-like symbols). Moreover, their word-discovery process had no need to remain in regions of the embedding space that contain embedding vectors that can be mapped to outputs on the natural image manifold. In the case of the text-and-image guided synthesis methods (rows $4$, $5$), results appear more natural and closer to the source image, but they fail to generalize to new texts. Moreover, as our method builds upon pre-trained, large-scale text-to-image synthesis models, we can optimize a single pseudo-word and re-use it for a multitude of new generations. The baseline models, meanwhile, use CLIP for test-time optimization and thus require expensive optimization for every new creation.

\subsection{Style transfer}

A typical use-case for text-guided synthesis is in artistic circles, where users aim to draw upon the unique style of a specific artist and apply it to new creations. Here, we show that our model can also find pseudo-words representing a specific, unknown style. To find such pseudo-words, we simply provide the model with a small set of images with a shared style, and replace the training texts with prompts of the form: ``A painting in the style of \pholdercolor". Results are shown in \Cref{fig:styles}. They serve as further demonstration that our ability to capture concepts extends beyond simple object reconstructions and into more abstract ideas.

Note that this differs from traditional style transfer, as we do not necessarily wish to maintain the content of some input image. Instead, we offer the network the freedom to decide how to depict the subject, and merely ask for an appropriate style.

\input{resources/figures/styles}

\subsection{Concept compositions}

\input{resources/figures/compositions}
In \Cref{fig:compositions} we demonstrate compositional synthesis, where the guiding text contains multiple learned concepts. We observe that the model can concurrently reason over multiple novel pseudo-words at the same time. However, it struggles with relations between them (\eg it fails to place two concepts side-by-side). We hypothesize that this limitation arises because our training considers only single concept scenes, where the concept is at the core of the image. Training on multi-object scenes may alleviate this shortcoming. However, we leave such investigation to future work.

\subsection{Bias reduction}

A common limitation of text-to-image models is that they inherit the biases found in the internet-scale data used to train them. These biases then manifest in the generated samples. For example, the DALLE-2 system card~\citep{mishkin2022risks} reports that their baseline model tends to produce images of people that are white-passing and male-passing when provided with the prompt ``A CEO". Similarly, results for ``wedding", tend to assume Western wedding traditions, and default to heterosexual couples.

Here, we demonstrate that we can utilize a small, curated dataset in order to learn a new ``fairer" word for a biased concept, which can then be used in place of the original to drive a more inclusive generation.

Specifically, in \Cref{fig:bias} we highlight the bias encoded in the word ``Doctor'', and show that this bias can be reduced (\ie we increase perceived gender and ethnic diversity) by learning a new embedding from a small, more diverse set.

\input{resources/figures/bias}

\subsection{Downstream applications}

Finally, we demonstrate that our pseudo-words can be used in downstream models that build on the same initial LDM model. Specifically, we consider the recent Blended Latent Diffusion~\citep{avrahami2022blendedlatent} which enables localized text-based editing of images via a mask-based blending process in the latent space of an LDM. In \Cref{fig:blended} we demonstrate that this localized synthesis process can also be conditioned on our learned pseudo-words, without requiring any additional modifications of the original model. 

\input{resources/figures/belnded_diffusion}

\subsection{Image curation}

Unless otherwise noted, results in this section are partially curated. For each prompt, we generated $16$ candidates (or six for DALLE-2) and manually selected the best result. We note that similar curation processes with larger batches are typically employed in text-conditioned generation works~\citep{avrahami2022blended,ramesh2021zero,yu2022scaling}, and that one can automate this selection process by using CLIP to rank images. In the supplementary materials, we provide large-scale, uncurated galleries of generated results, including failure cases.

%% file: resources/figures/user_sentences_comparison.tex
\begin{figure}[!hbt]
    \centering
    \setlength{\abovecaptionskip}{6.5pt}
    \setlength{\belowcaptionskip}{-3.5pt}
    \setlength{\tabcolsep}{0.55pt}
    \renewcommand{\arraystretch}{1.0}
    {
    
    \begin{tabular}{c}
    
    \begin{tabular}{c@{\hskip 5pt} c c c @{\hskip 10pt} c c c@{\hskip 10pt} c c c}
    & \multicolumn{3}{c}{Skull Mug} & \multicolumn{3}{c}{Teapot} & \multicolumn{3}{c}{Aladdin Lamp} \\

    \raisebox{0.045\linewidth}{\footnotesize\begin{tabular}{c@{}c@{}} Input \\ samples \end{tabular}} & 
    \includegraphics[width=0.09\linewidth,height=0.09\linewidth]{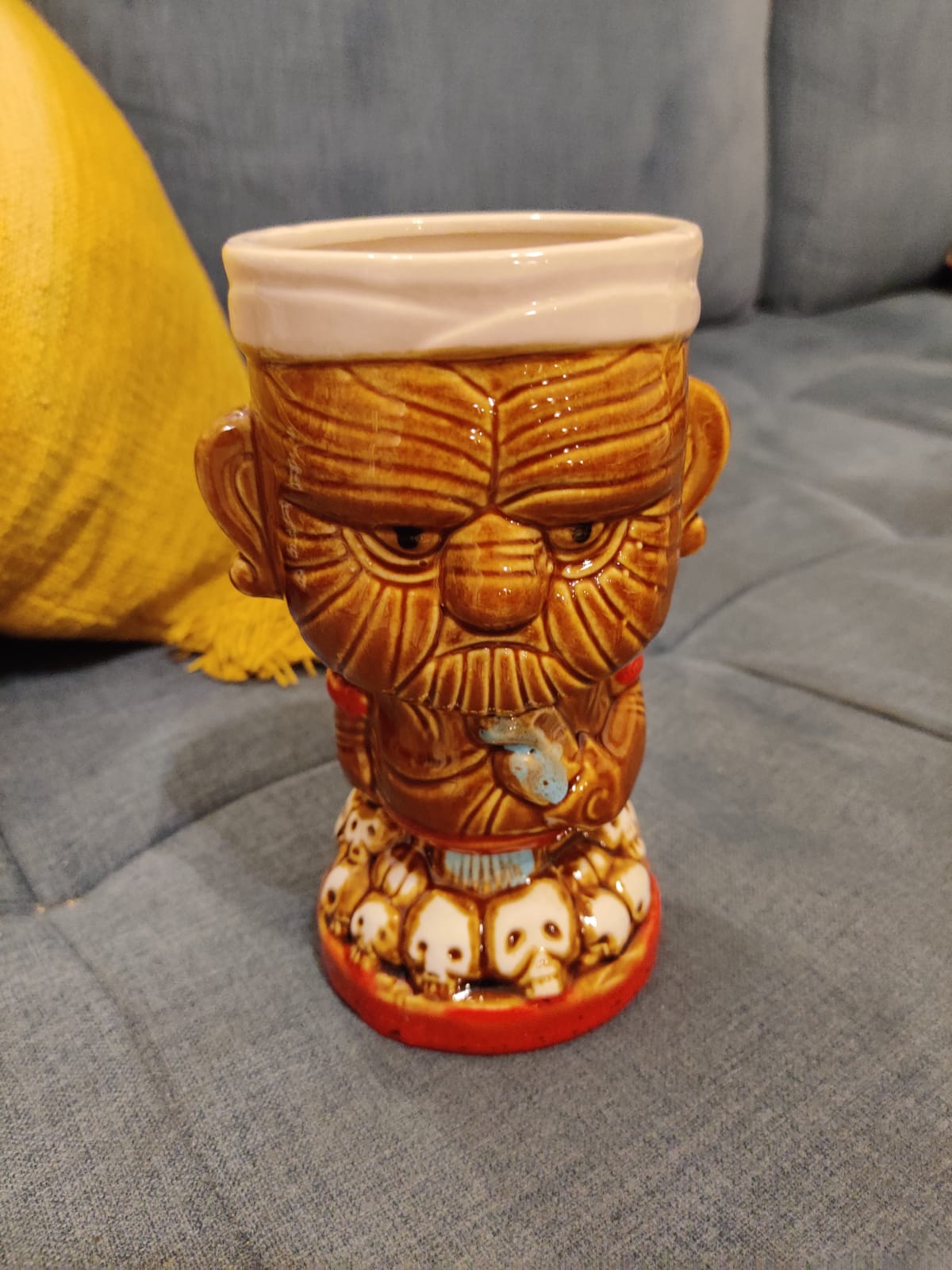} & 
    \includegraphics[width=0.09\linewidth,height=0.09\linewidth]{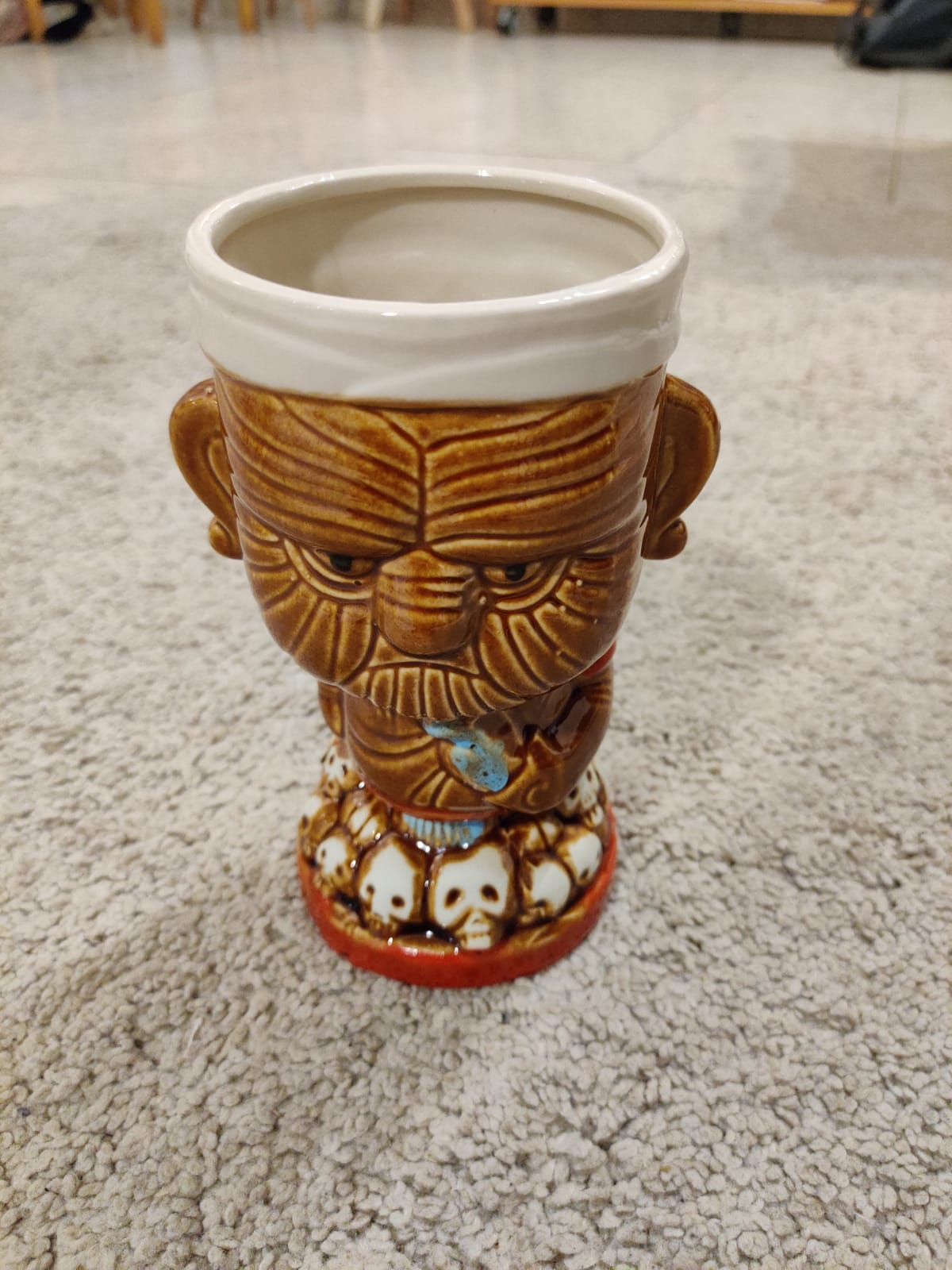} & 
    \includegraphics[width=0.09\linewidth,height=0.09\linewidth]{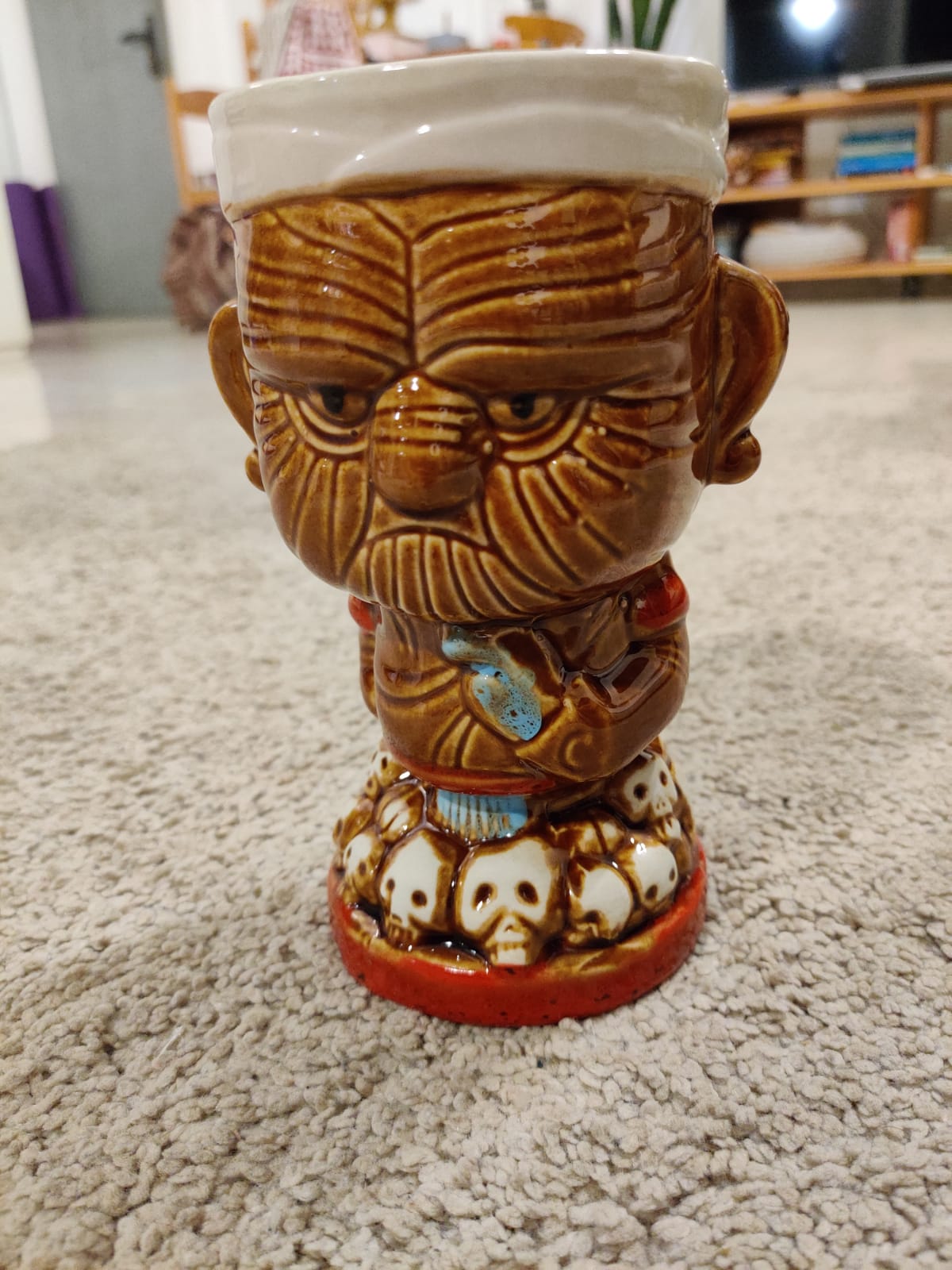} & 
    \includegraphics[width=0.09\linewidth,height=0.09\linewidth]{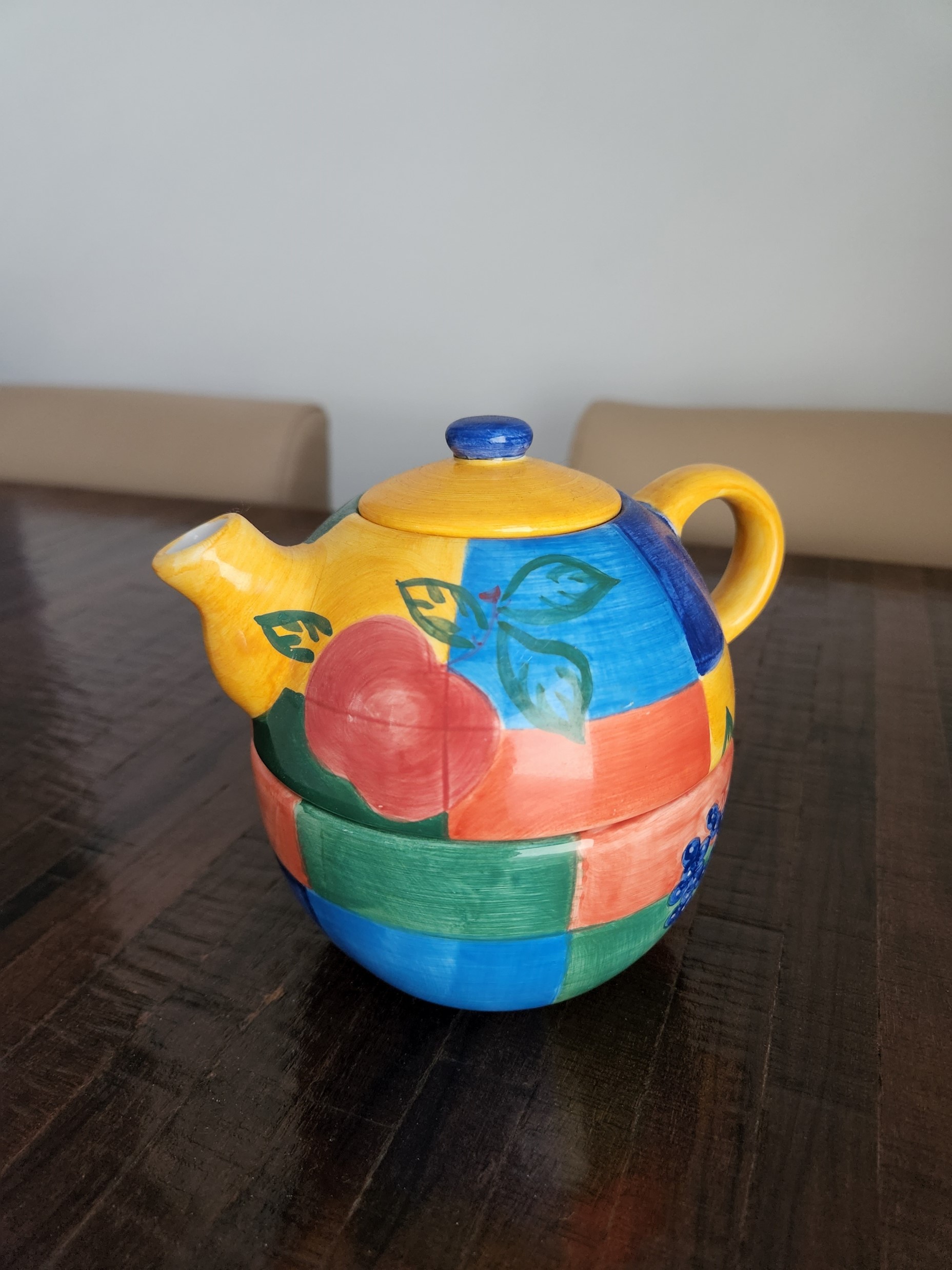} & 
    \includegraphics[width=0.09\linewidth,height=0.09\linewidth]{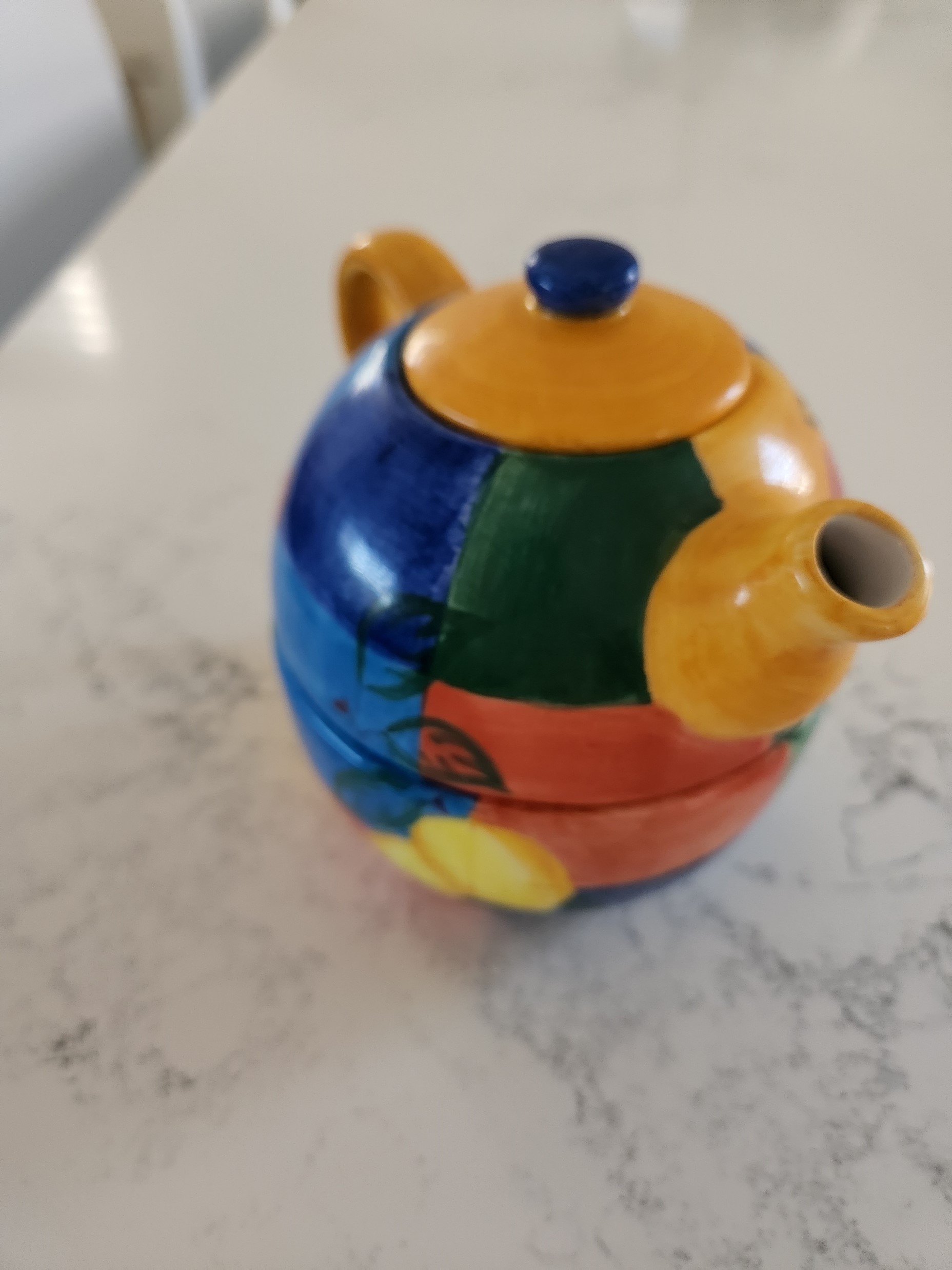} &
    \includegraphics[width=0.09\linewidth,height=0.09\linewidth]{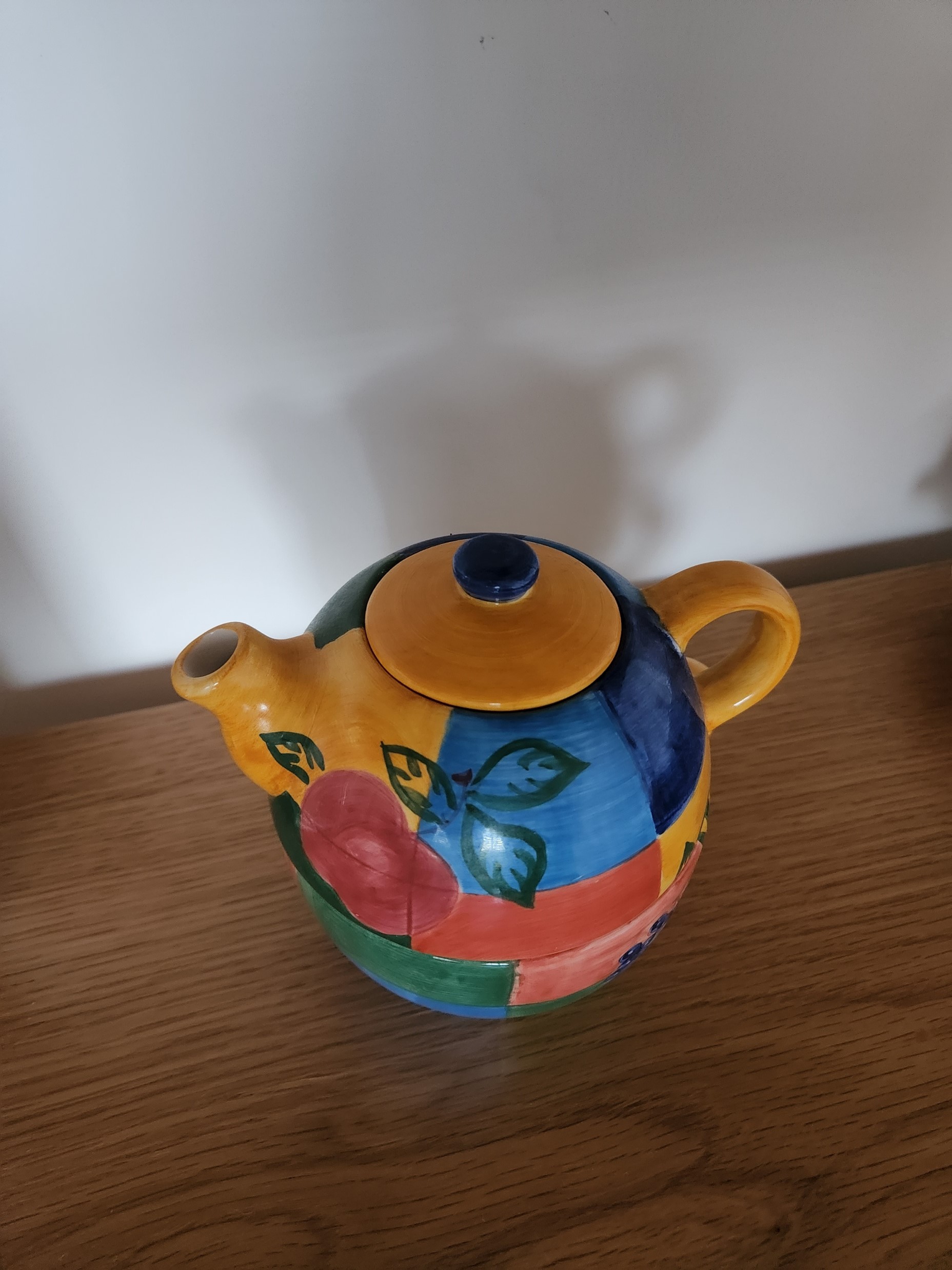} &
    \includegraphics[width=0.09\linewidth,height=0.09\linewidth]{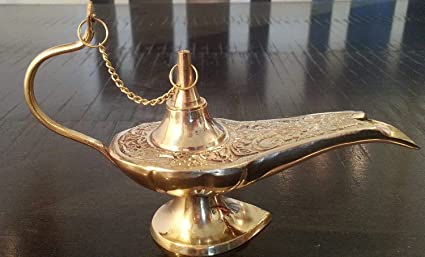} & 
    \includegraphics[width=0.09\linewidth,height=0.09\linewidth]{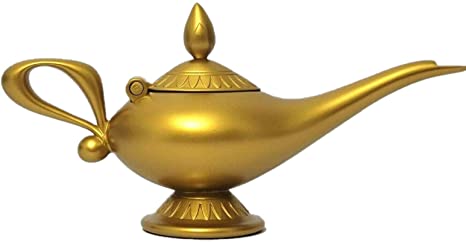} & 
    \includegraphics[width=0.09\linewidth,height=0.09\linewidth]{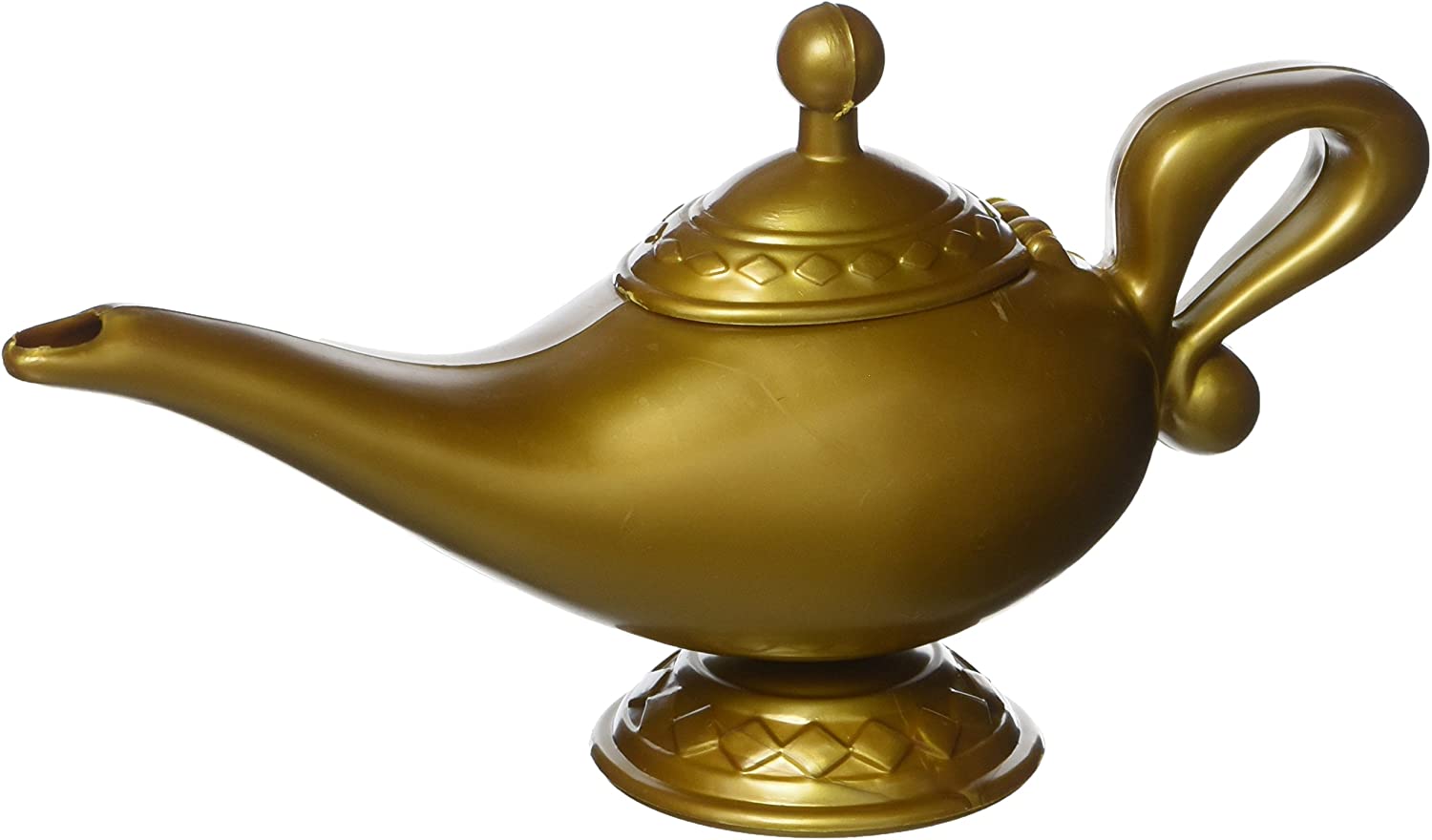} \\[2pt]
    
    \raisebox{0.045\linewidth}{\footnotesize\begin{tabular}{c@{}c@{}c@{}c@{}} Ours \end{tabular}}  &
    \includegraphics[width=0.09\linewidth,height=0.09\linewidth]{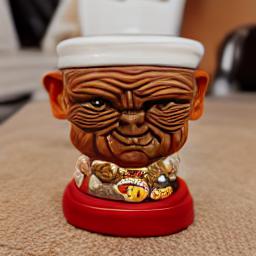} & 
    \includegraphics[width=0.09\linewidth,height=0.09\linewidth]{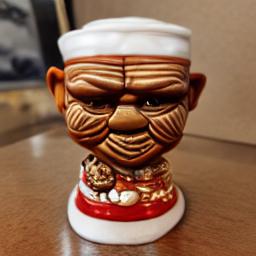} & 
    \includegraphics[width=0.09\linewidth,height=0.09\linewidth]{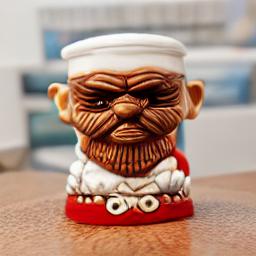} & 
    \includegraphics[width=0.09\linewidth,height=0.09\linewidth]{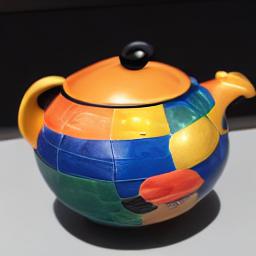} & 
    \includegraphics[width=0.09\linewidth,height=0.09\linewidth]{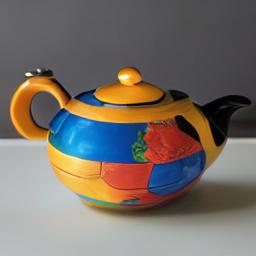} &
    \includegraphics[width=0.09\linewidth,height=0.09\linewidth]{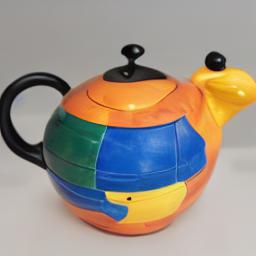} &
    \includegraphics[width=0.09\linewidth,height=0.09\linewidth]{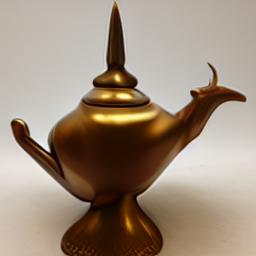} & 
    \includegraphics[width=0.09\linewidth,height=0.09\linewidth]{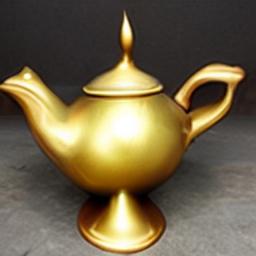} & 
    \includegraphics[width=0.09\linewidth,height=0.09\linewidth]{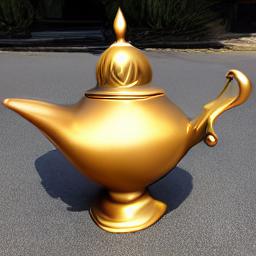} \\[-4pt]
        &
    \multicolumn{3}{c}{\tiny\begin{tabular}{c@{}c@{}c@{}} ``A photo of \pholdercolor" \end{tabular}} &
    \multicolumn{3}{c}{\tiny\begin{tabular}{c@{}c@{}c@{}} ``A photo of \pholdercolor" \end{tabular}} & 
    \multicolumn{3}{c}{\tiny\begin{tabular}{c@{}c@{}c@{}} ``A photo of \pholdercolor" \end{tabular}} \\[3pt]
    
    \raisebox{0.045\linewidth}{\footnotesize\begin{tabular}{c@{}c@{}c@{}c@{}} DALLE-2 \\ (Image \\ Inputs) \end{tabular}}  &
    \includegraphics[width=0.09\linewidth,height=0.09\linewidth]{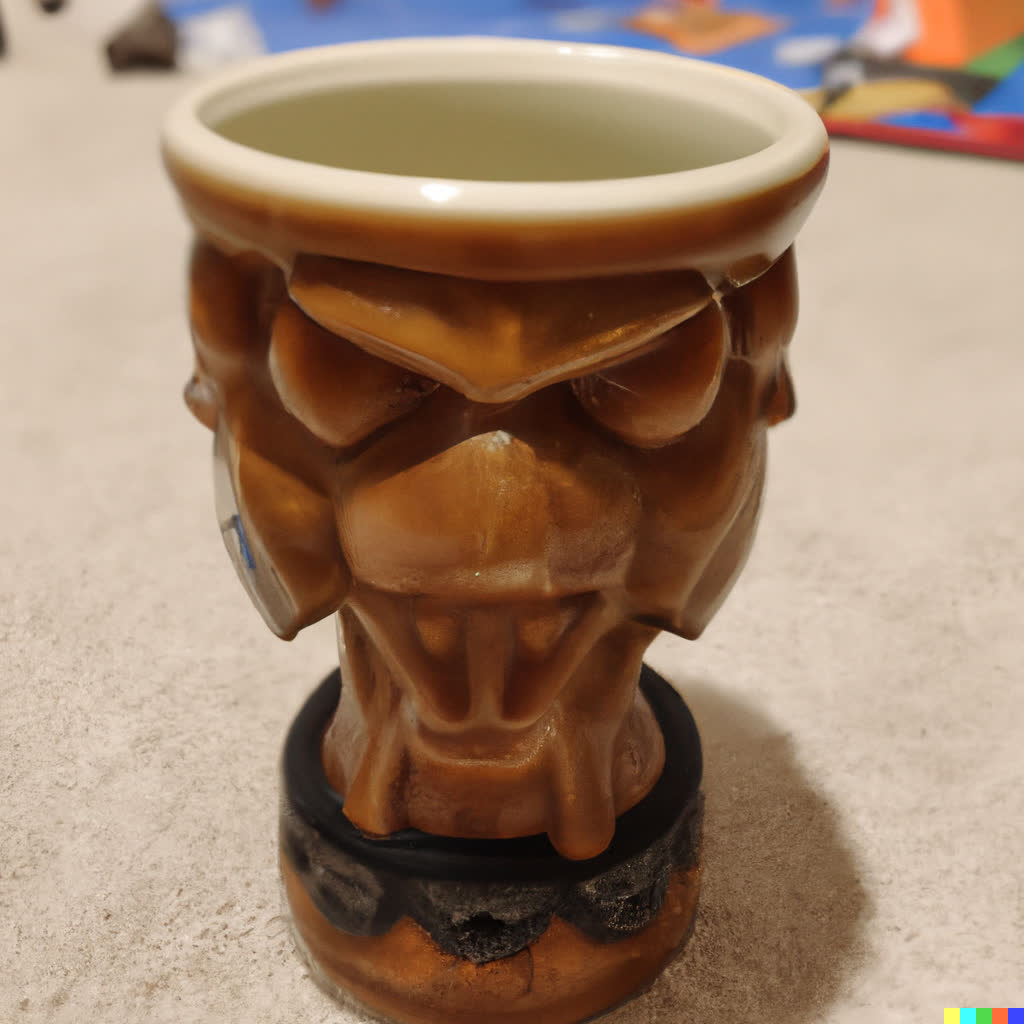} & 
    \includegraphics[width=0.09\linewidth,height=0.09\linewidth]{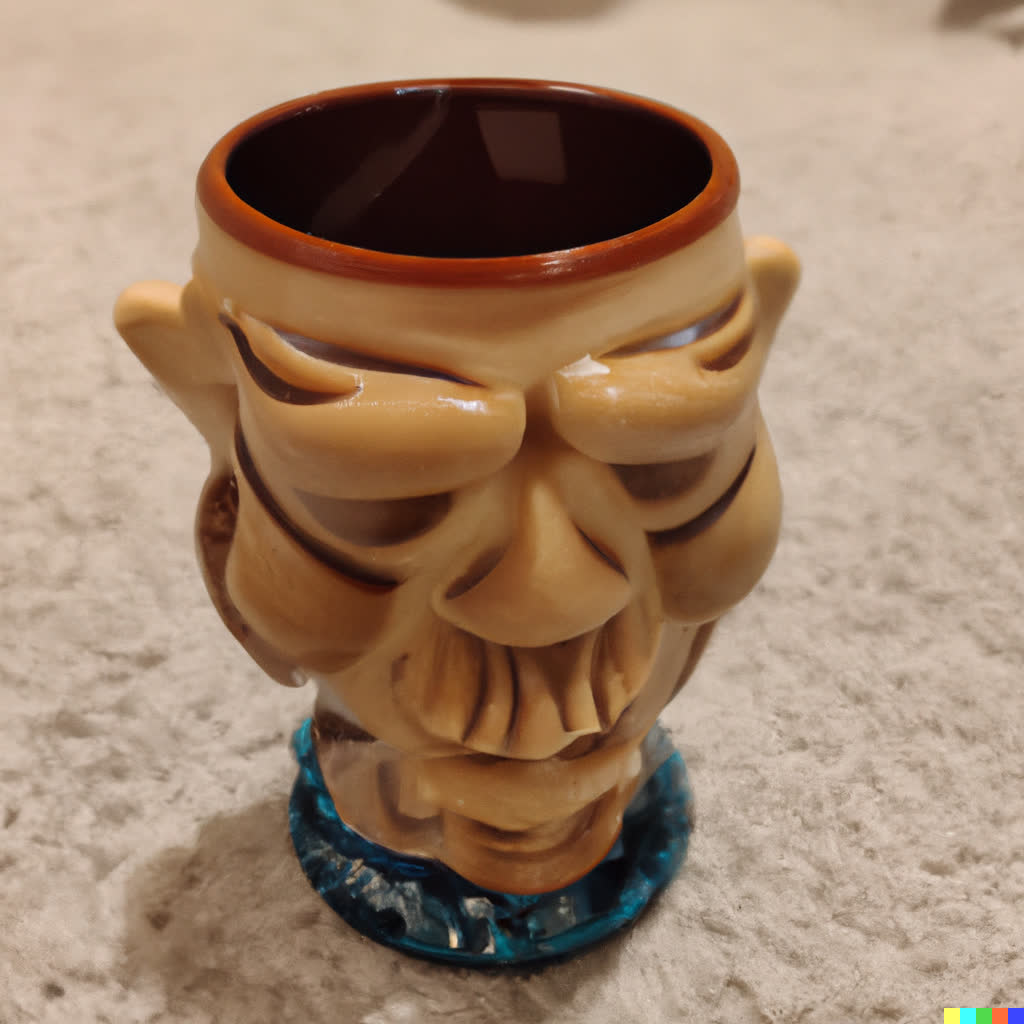} & 
    \includegraphics[width=0.09\linewidth,height=0.09\linewidth]{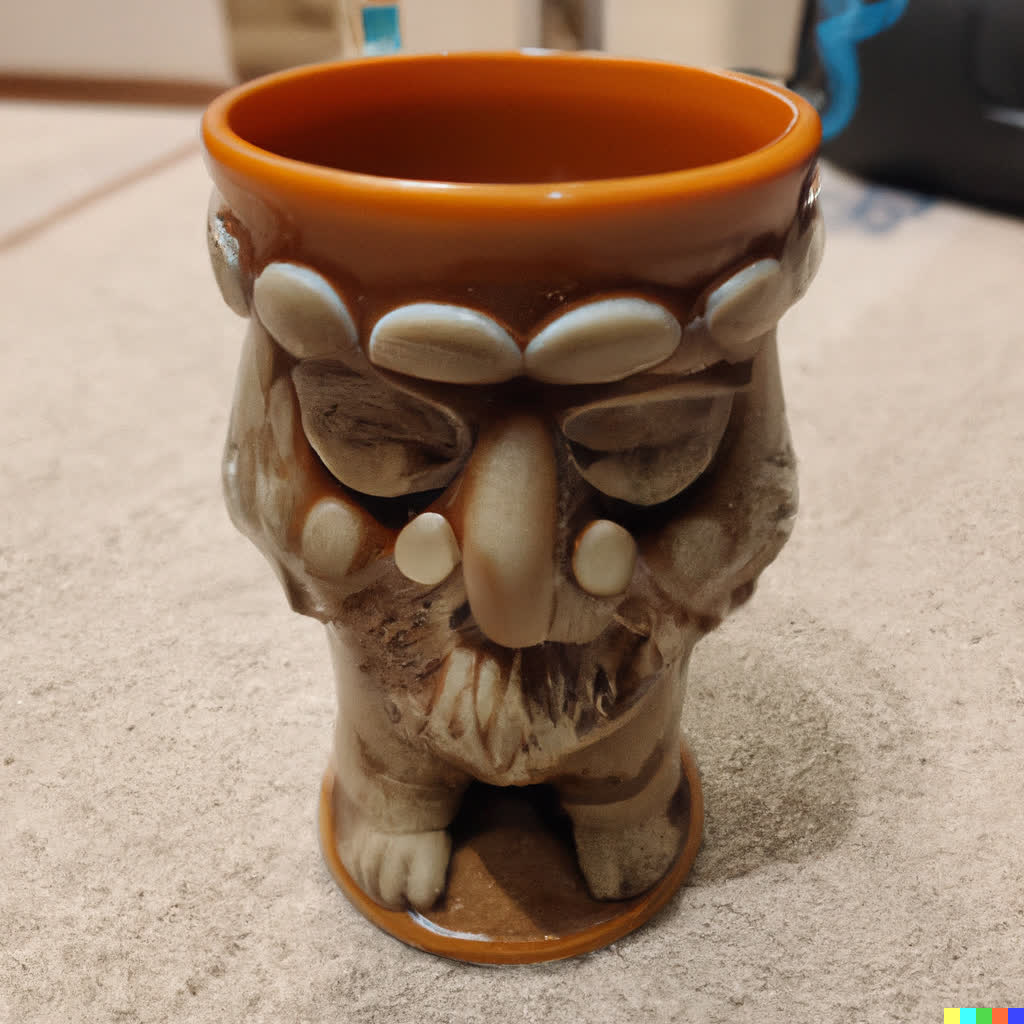} & 
    \includegraphics[width=0.09\linewidth,height=0.09\linewidth]{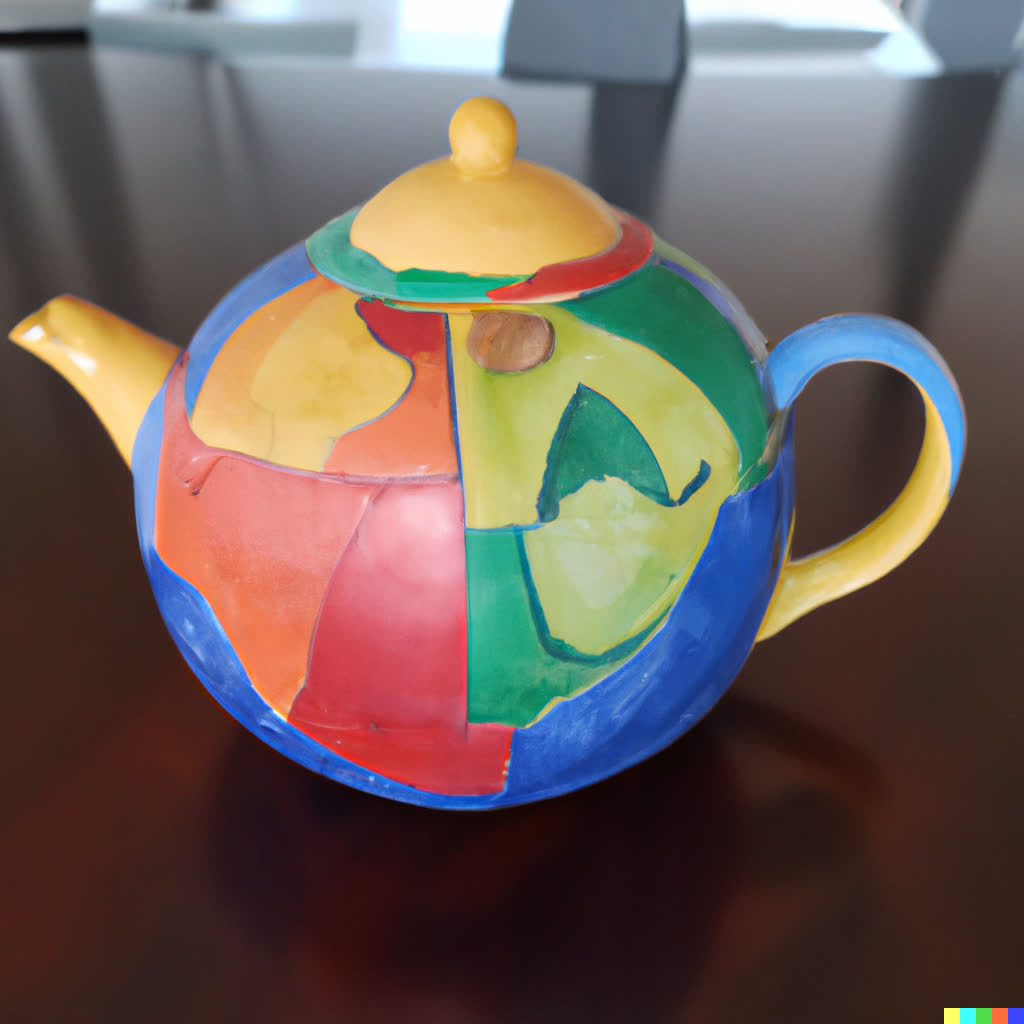} & 
    \includegraphics[width=0.09\linewidth,height=0.09\linewidth]{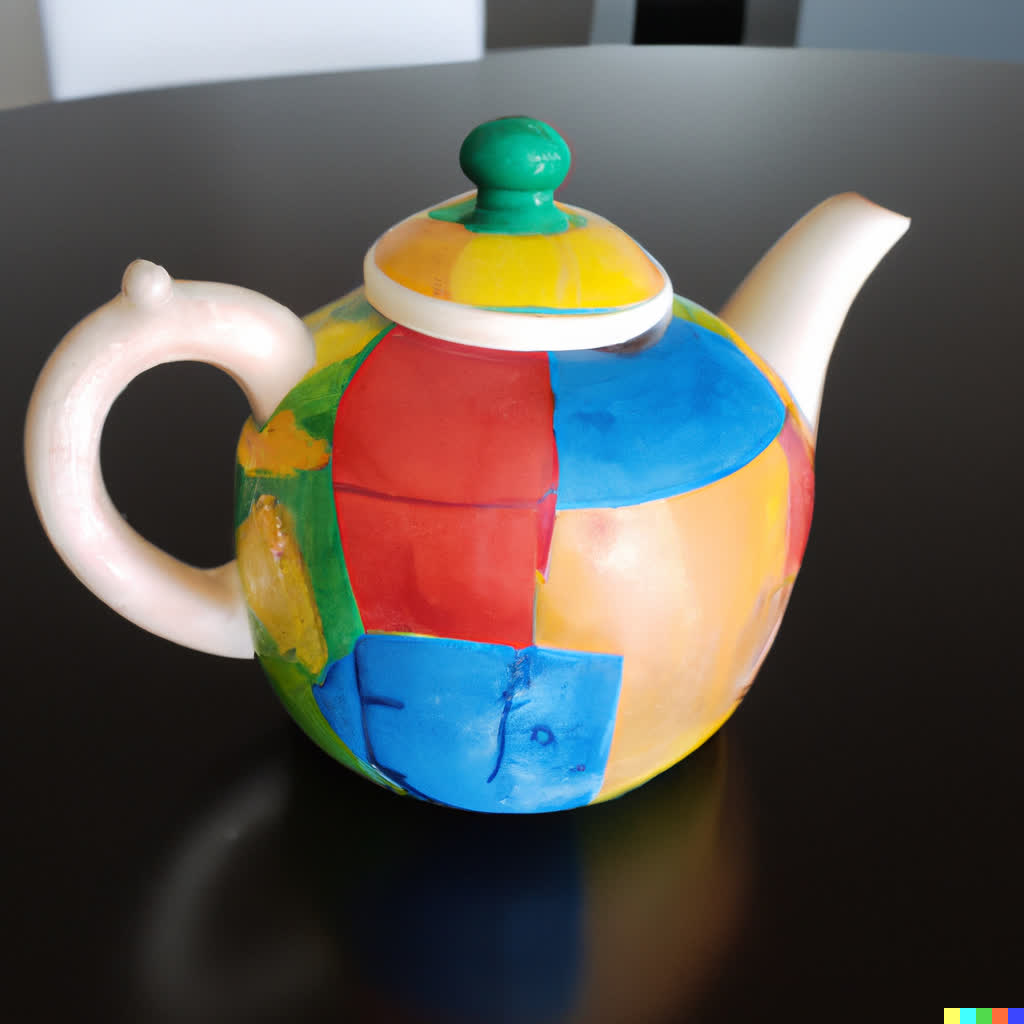} &
    \includegraphics[width=0.09\linewidth,height=0.09\linewidth]{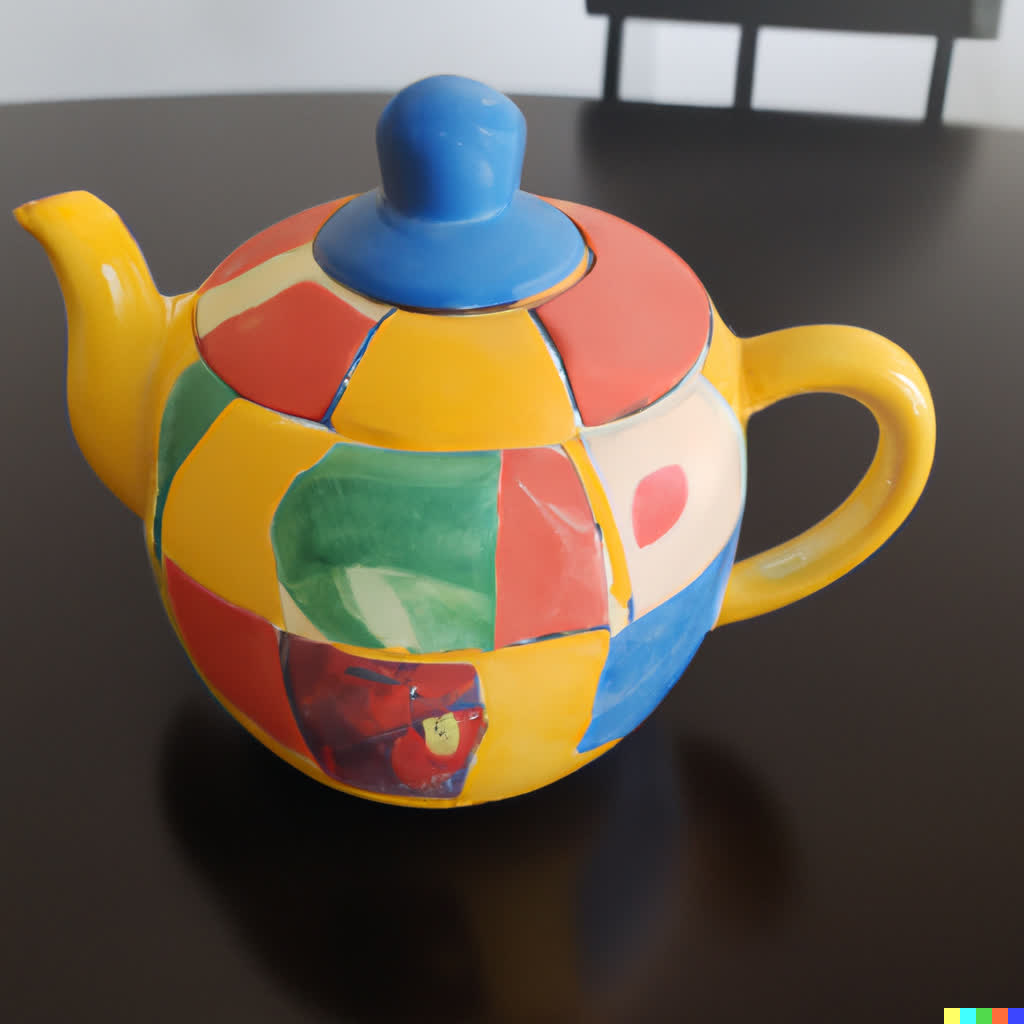} & 
    \includegraphics[width=0.09\linewidth,height=0.09\linewidth]{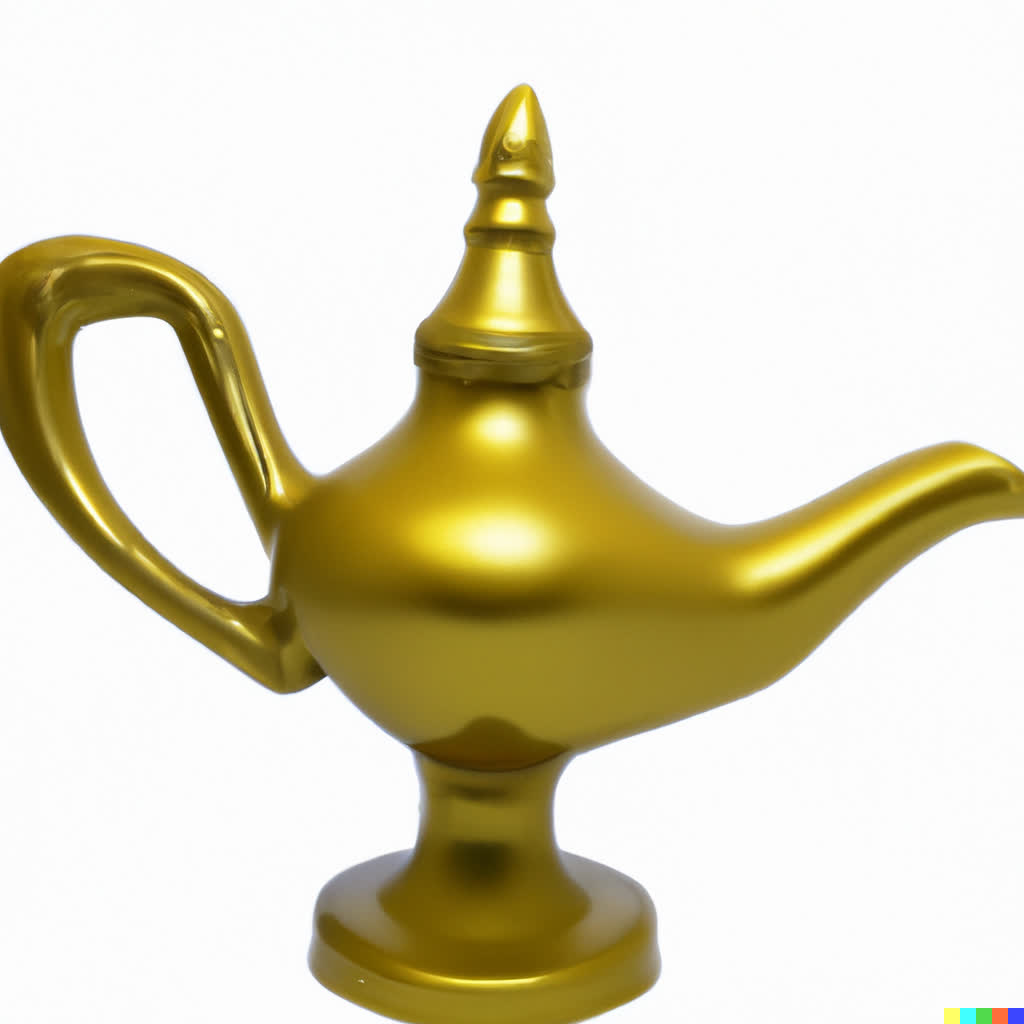} & 
    \includegraphics[width=0.09\linewidth,height=0.09\linewidth]{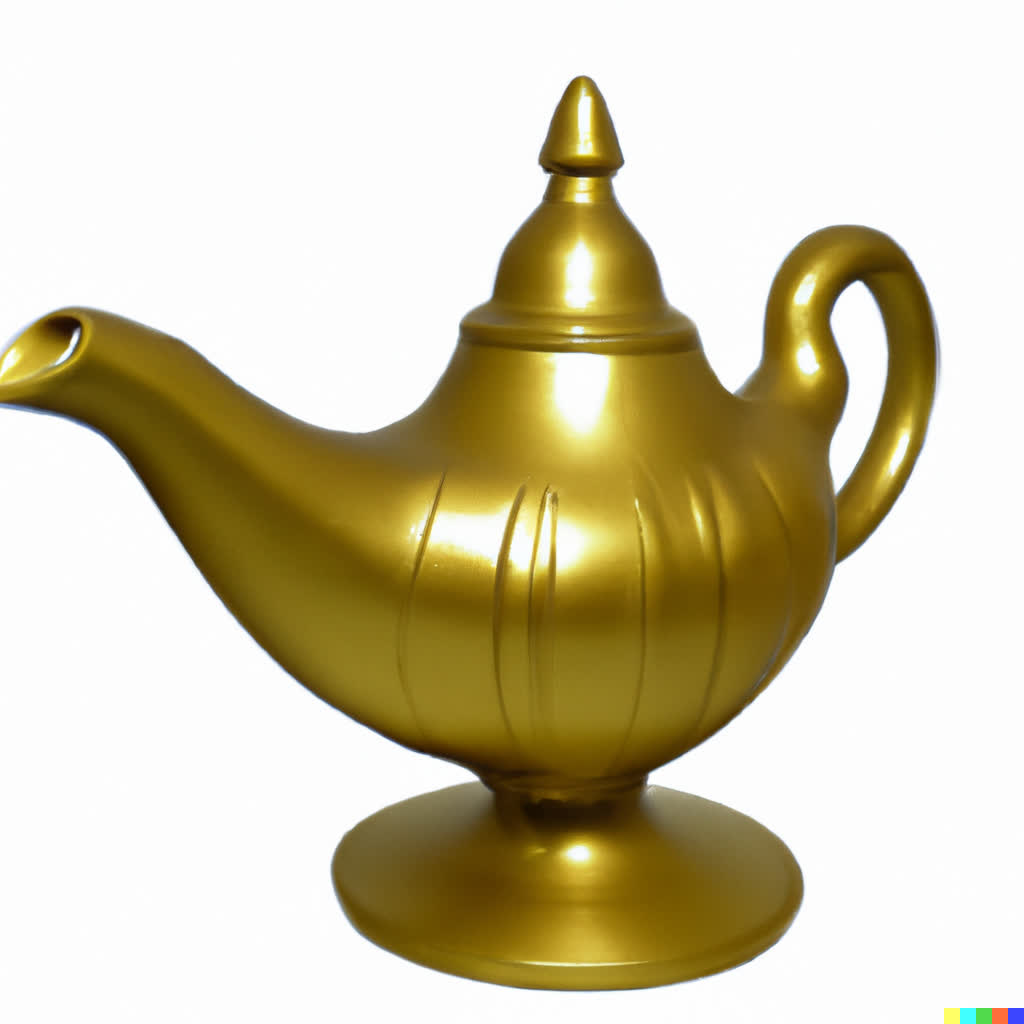} & 
    \includegraphics[width=0.09\linewidth,height=0.09\linewidth]{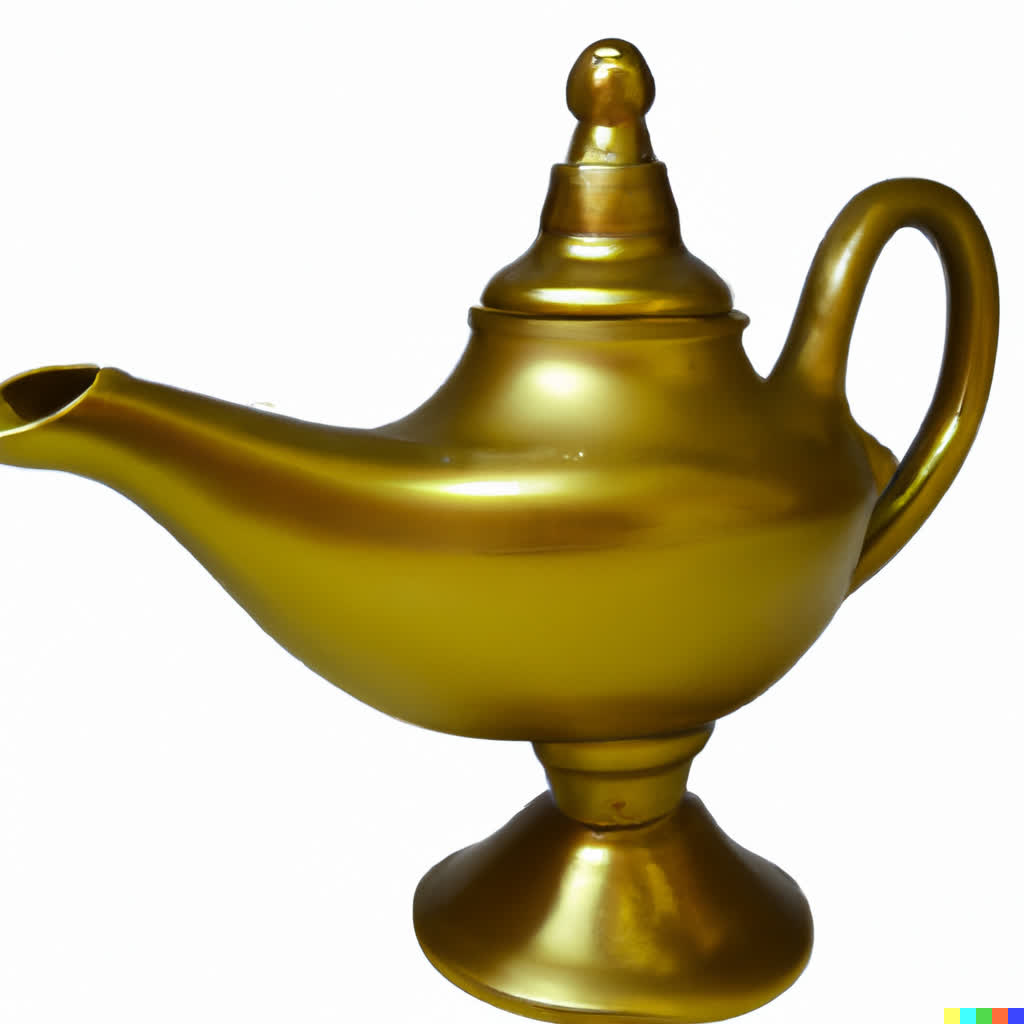} \\[5pt]
    
    \raisebox{0.045\linewidth}{\footnotesize\begin{tabular}{c@{}c@{}c@{}c@{}} DALLE-2 \\ (Long \\ Captions) \end{tabular}}  &
    \includegraphics[width=0.09\linewidth,height=0.09\linewidth]{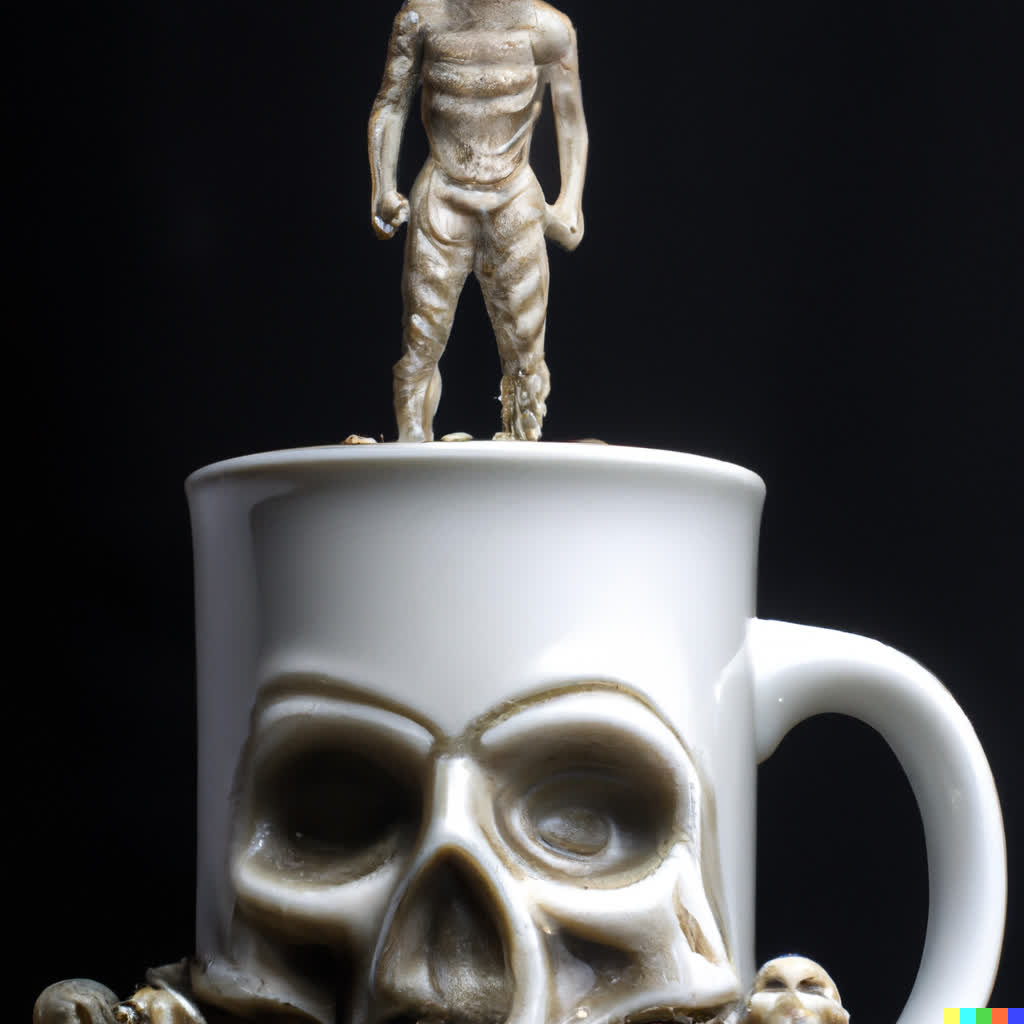} & 
    \includegraphics[width=0.09\linewidth,height=0.09\linewidth]{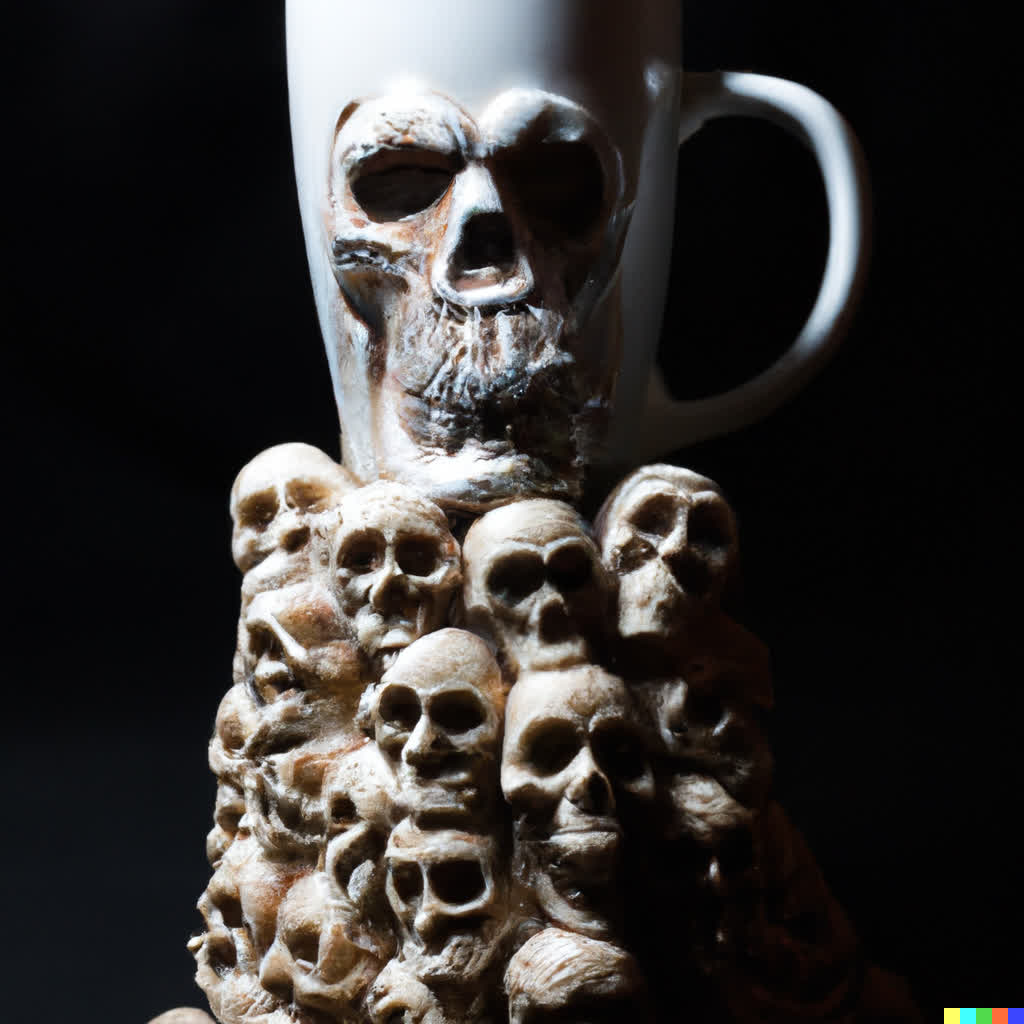} & 
    \includegraphics[width=0.09\linewidth,height=0.09\linewidth]{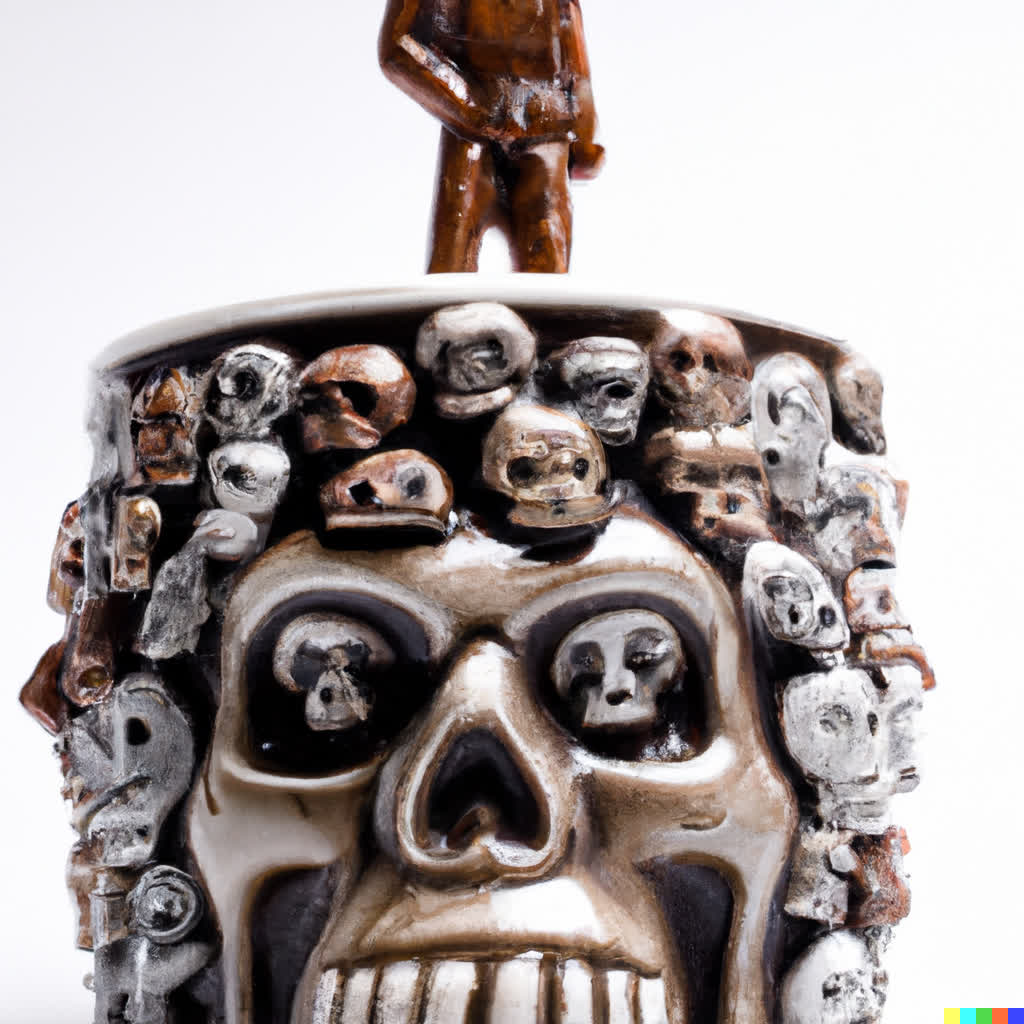} & 
    \includegraphics[width=0.09\linewidth,height=0.09\linewidth]{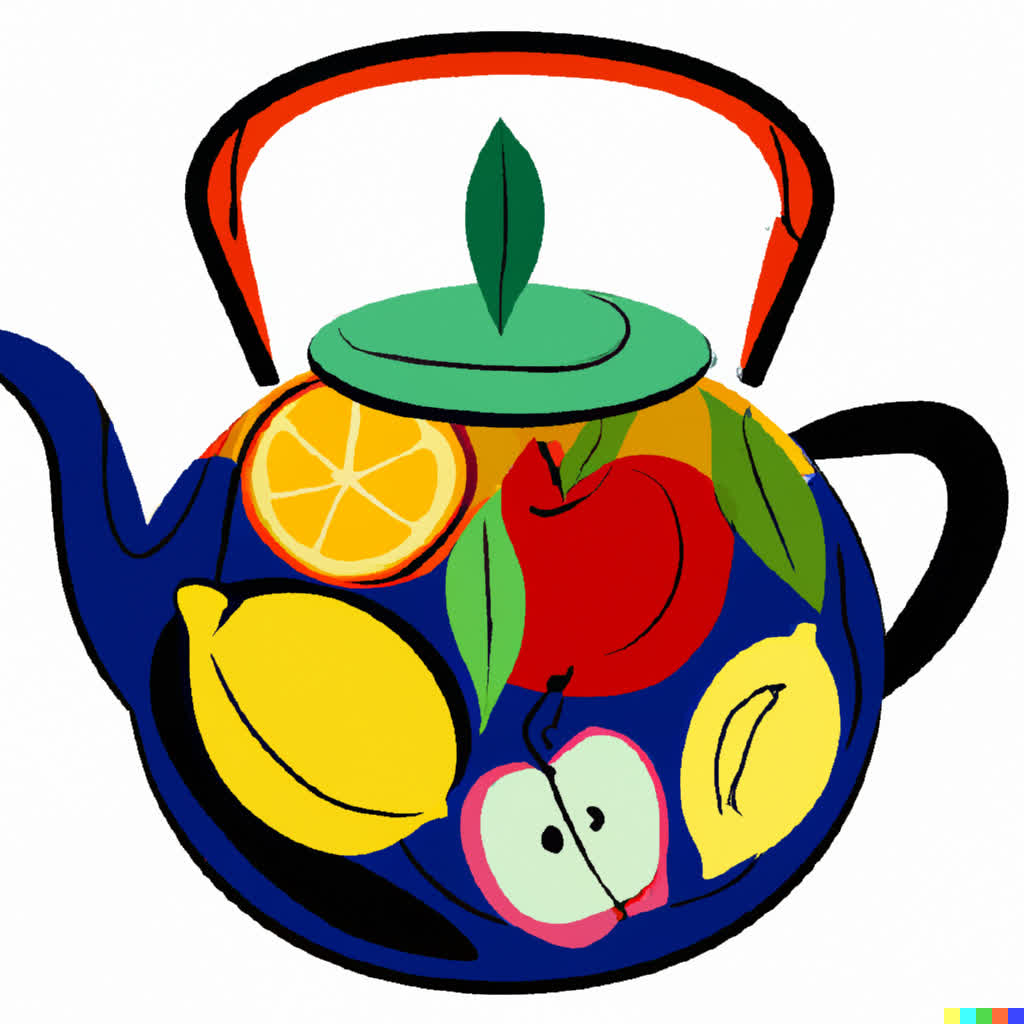} & 
    \includegraphics[width=0.09\linewidth,height=0.09\linewidth]{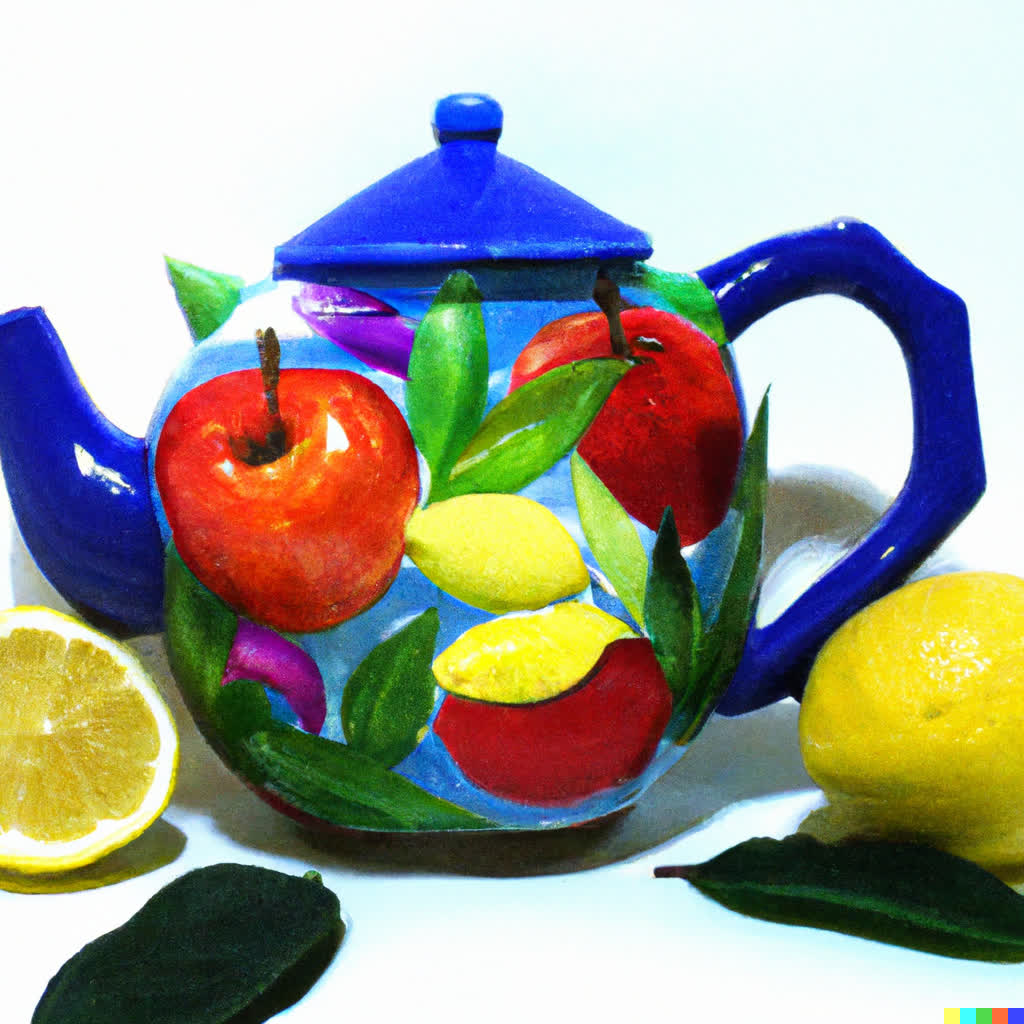} &
    \includegraphics[width=0.09\linewidth,height=0.09\linewidth]{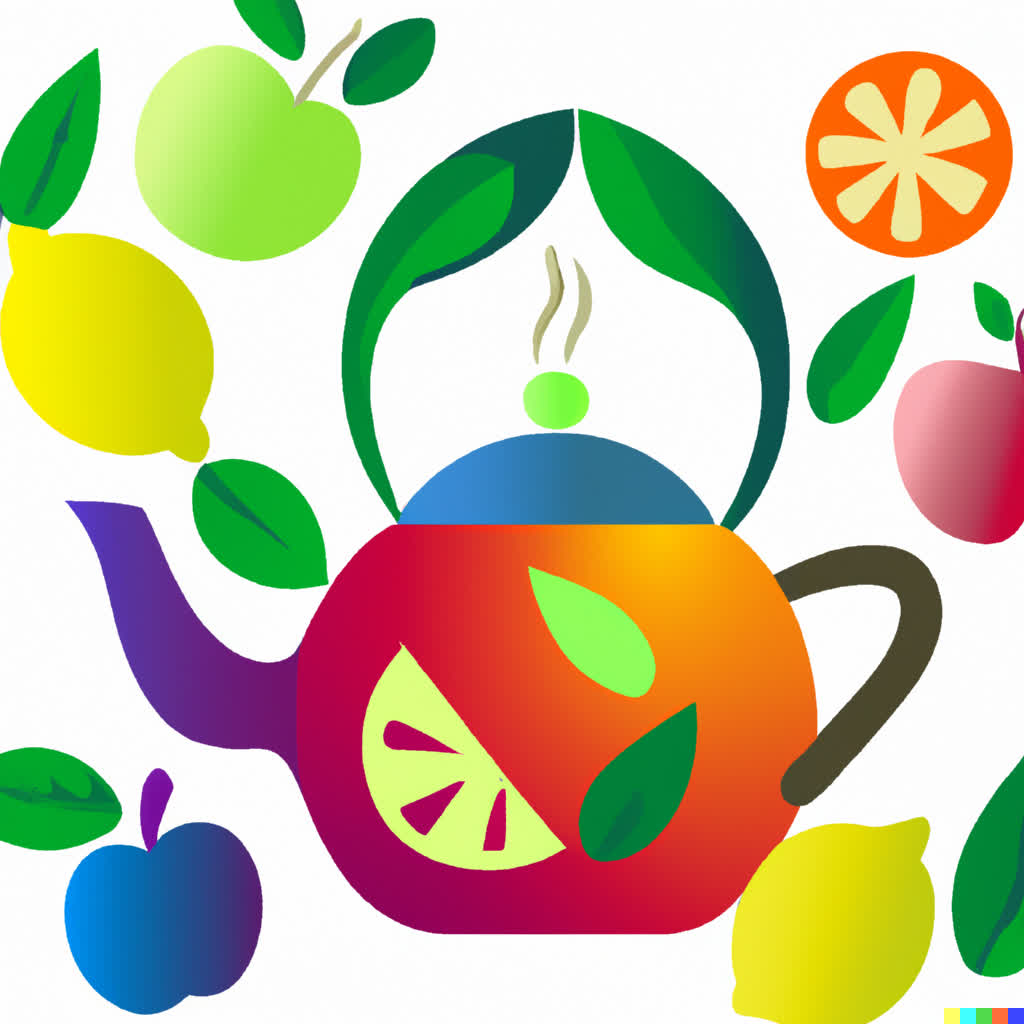} &
    \includegraphics[width=0.09\linewidth,height=0.09\linewidth]{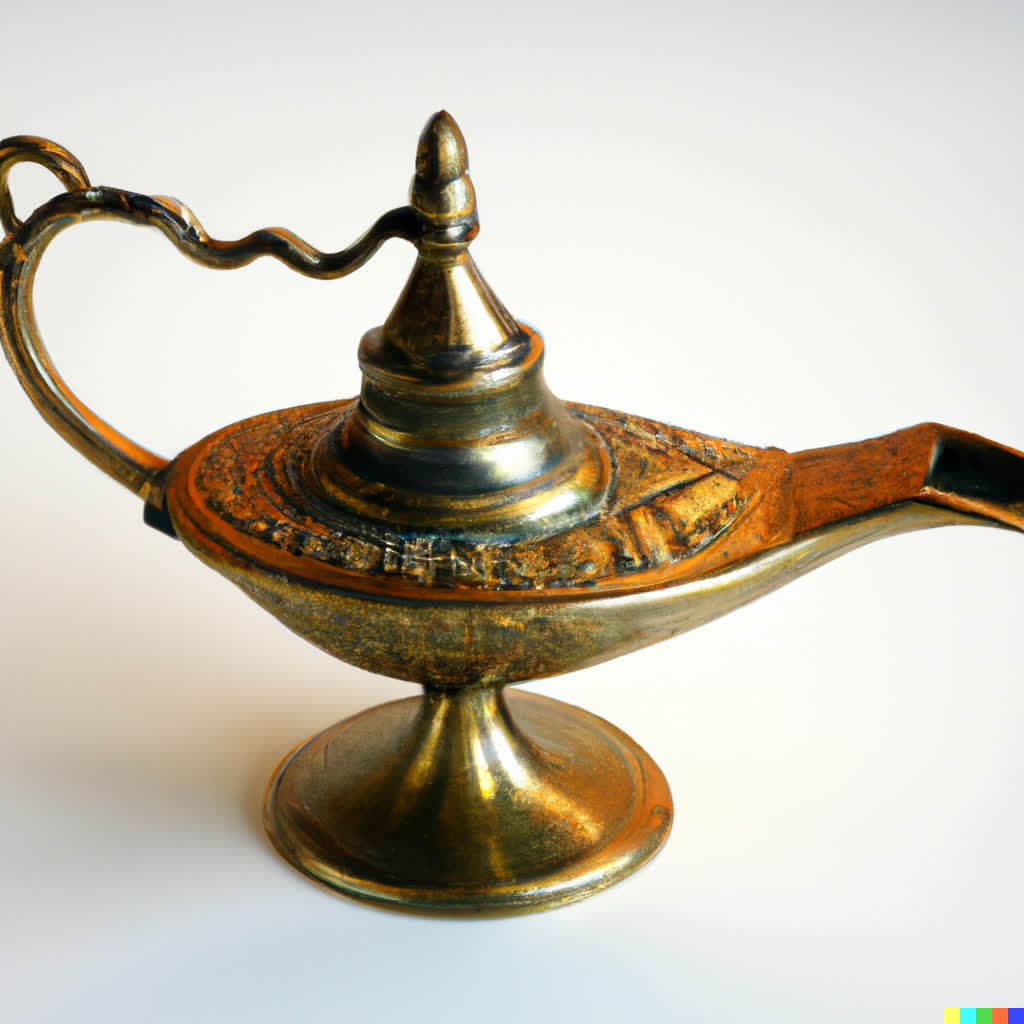} & 
    \includegraphics[width=0.09\linewidth,height=0.09\linewidth]{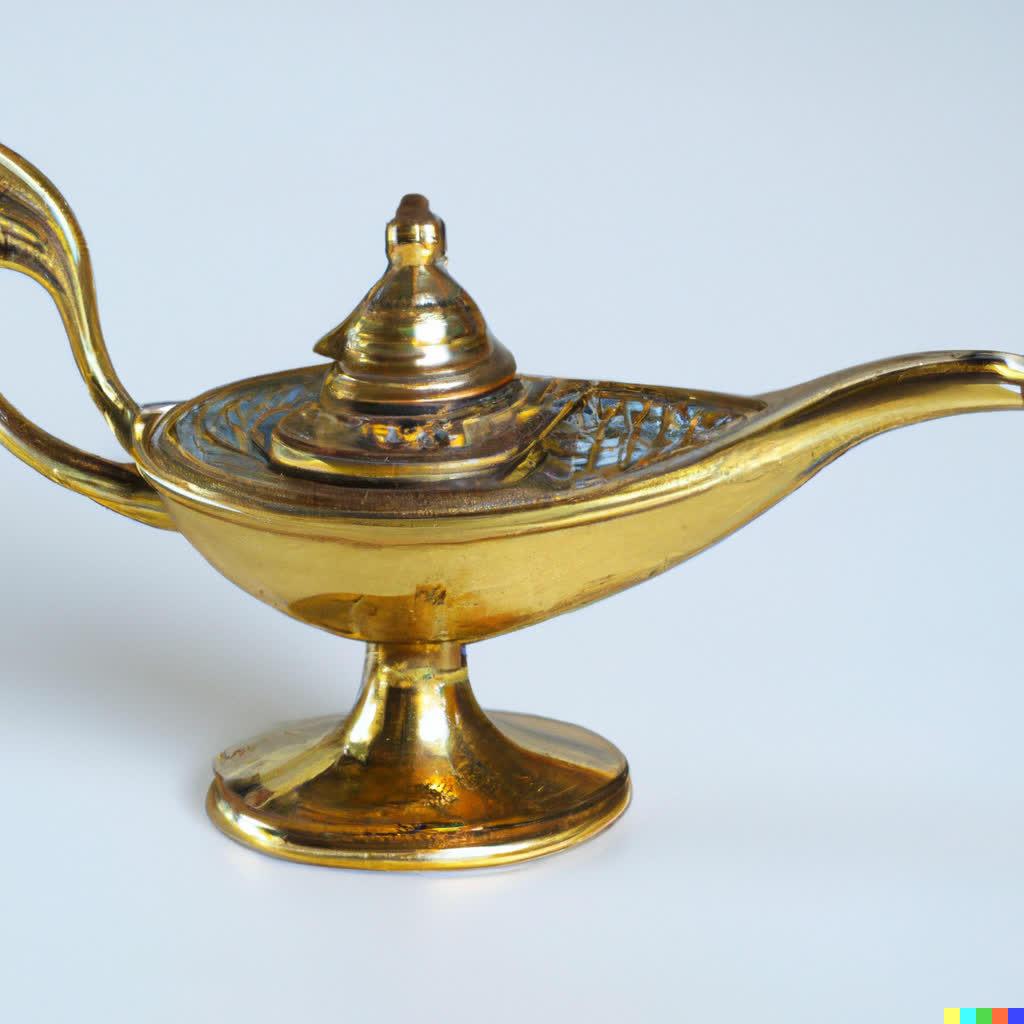} & 
    \includegraphics[width=0.09\linewidth,height=0.09\linewidth]{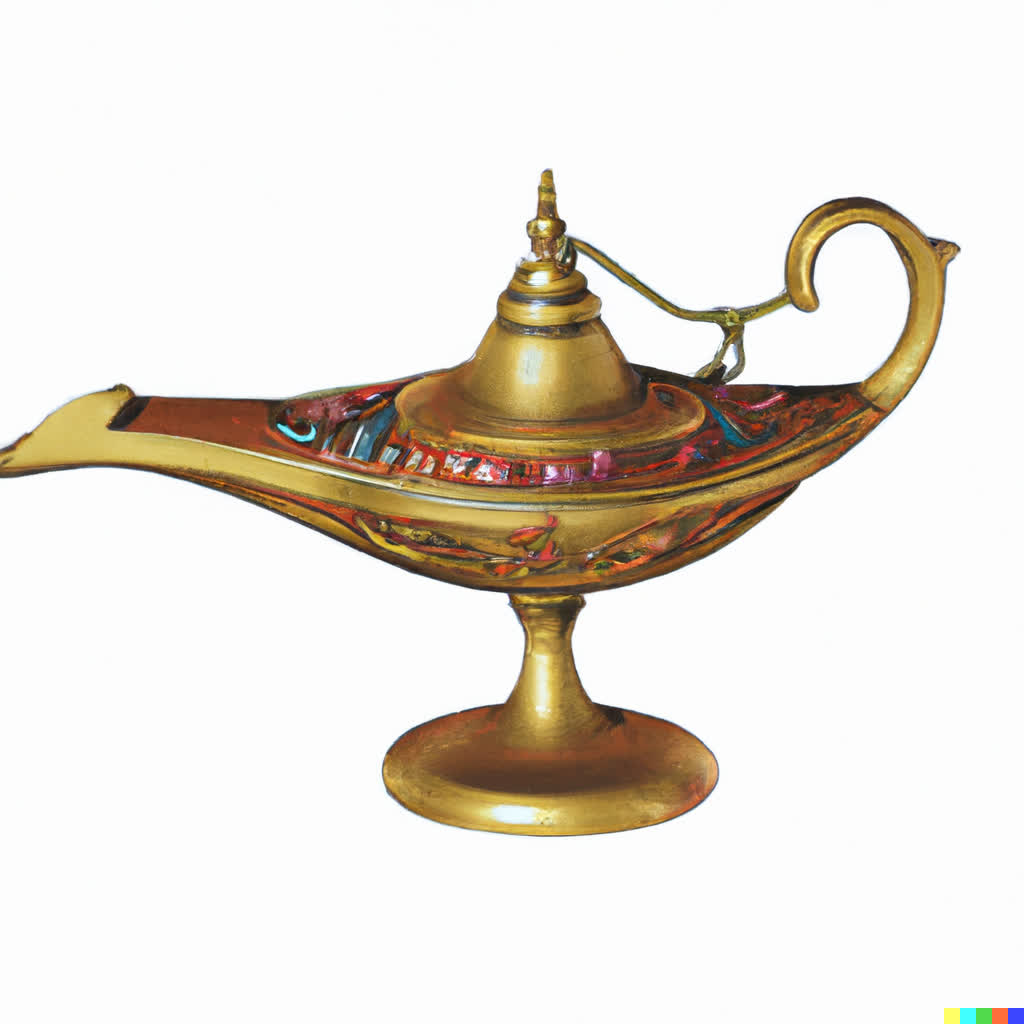} \\
    
    \raisebox{0.045\linewidth}{\footnotesize\begin{tabular}{c@{}c@{}c@{}c@{}} LDM \\ (Long \\ Captions) \end{tabular}}  &
    \includegraphics[width=0.09\linewidth,height=0.09\linewidth]{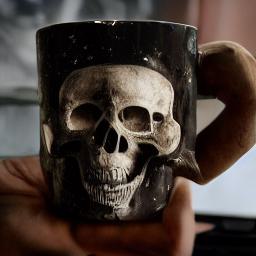} & 
    \includegraphics[width=0.09\linewidth,height=0.09\linewidth]{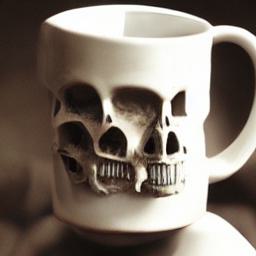} & 
    \includegraphics[width=0.09\linewidth,height=0.09\linewidth]{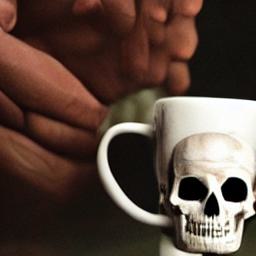} & 
    \includegraphics[width=0.09\linewidth,height=0.09\linewidth]{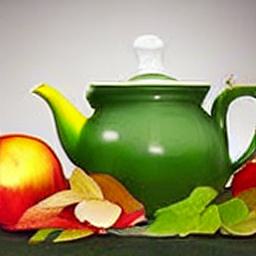} & 
    \includegraphics[width=0.09\linewidth,height=0.09\linewidth]{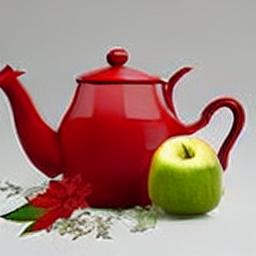} &
    \includegraphics[width=0.09\linewidth,height=0.09\linewidth]{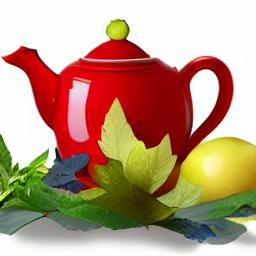} &
    \includegraphics[width=0.09\linewidth,height=0.09\linewidth]{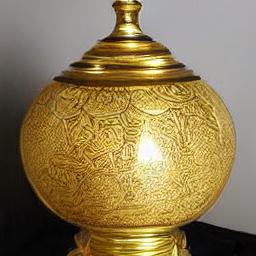} & 
    \includegraphics[width=0.09\linewidth,height=0.09\linewidth]{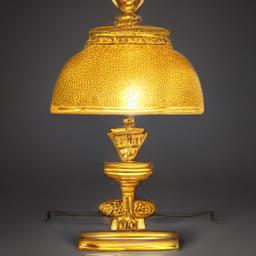} & 
    \includegraphics[width=0.09\linewidth,height=0.09\linewidth]{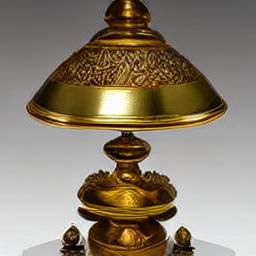} \\[-2pt]

    &
    \multicolumn{3}{c}{\tiny\begin{tabular}{c@{}c@{}} A mug having many skulls at the bottom and \\ sculpture of a man at the top of it. \end{tabular}} & 
    \multicolumn{3}{c}{\tiny\begin{tabular}{c@{}c@{}} A tea pot, with green, red, blue, yellow, an apple \\ with leaves and a lemon with leaves. \end{tabular}} &
    \multicolumn{3}{c}{\tiny\begin{tabular}{c@{}c@{}} An oil lamp, like Aladdin's lamp, in gold metal, \\ with a design around the pedestal base and lid. \end{tabular}} \\[10pt]
    
    \raisebox{0.045\linewidth}{\footnotesize\begin{tabular}{c@{}c@{}c@{}c@{}} LDM \\ (Short \\ Captions) \end{tabular}}  &
    \includegraphics[width=0.09\linewidth,height=0.09\linewidth]{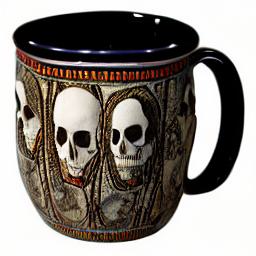} & 
    \includegraphics[width=0.09\linewidth,height=0.09\linewidth]{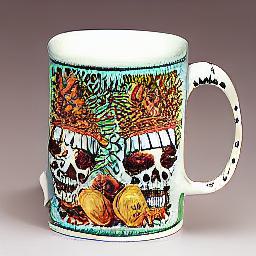} & 
    \includegraphics[width=0.09\linewidth,height=0.09\linewidth]{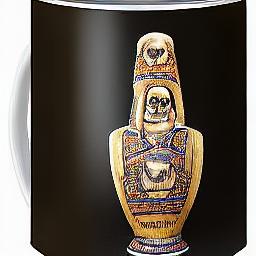} & 
    \includegraphics[width=0.09\linewidth,height=0.09\linewidth]{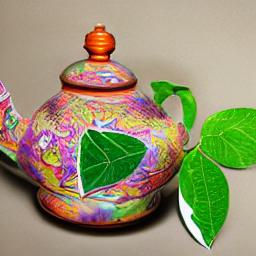} & 
    \includegraphics[width=0.09\linewidth,height=0.09\linewidth]{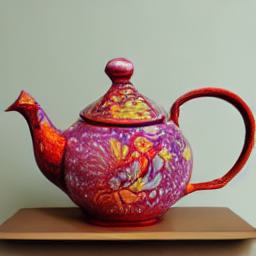} &
    \includegraphics[width=0.09\linewidth,height=0.09\linewidth]{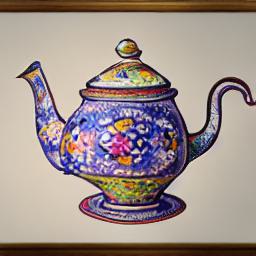} &
    \includegraphics[width=0.09\linewidth,height=0.09\linewidth]{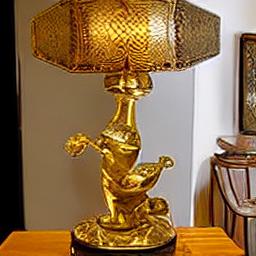} & 
    \includegraphics[width=0.09\linewidth,height=0.09\linewidth]{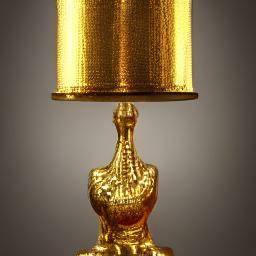} & 
    \includegraphics[width=0.09\linewidth,height=0.09\linewidth]{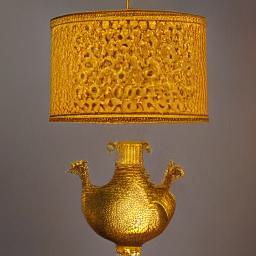} \\[-2pt]

    &
    \multicolumn{3}{c}{\tiny\begin{tabular}{c@{}c@{}c@{}} Mug or vase featuring grouchy faced \\ native standing on skulls \end{tabular}} & 
    \multicolumn{3}{c}{\tiny\begin{tabular}{c@{}c@{}c@{}} A colorful tea pot having leaf prints \end{tabular}} &
    \multicolumn{3}{c}{\tiny\begin{tabular}{c@{}c@{}c@{}} A gold genie lamp with a long spout and slim body. \end{tabular}} \\[5pt]

    \end{tabular}
    
    \end{tabular}
    
    }
    \caption{Object variations generated 
    using our method, the CLIP-based reconstruction of DALLE-2~\citep{ramesh2022hierarchical}, and human captions of varying lengths.
    Our method generates variations which are typically more faithful to the original subject.}
    \label{fig:user_sentences_comp} 

\end{figure}

%% file: resources/figures/generation.tex
\begin{figure}[!hbt]
    \centering
    \setlength{\abovecaptionskip}{6.5pt}
    \setlength{\belowcaptionskip}{-3.5pt}
    \setlength{\tabcolsep}{0.55pt}
    \renewcommand{\arraystretch}{1.0}
    {\scriptsize
    \begin{tabular}{c@{\hskip 5pt} c@{\hskip 5pt} c c c c}
    
        \begin{tabular}{c c}
            \includegraphics[width=0.09\linewidth,height=0.09\linewidth]{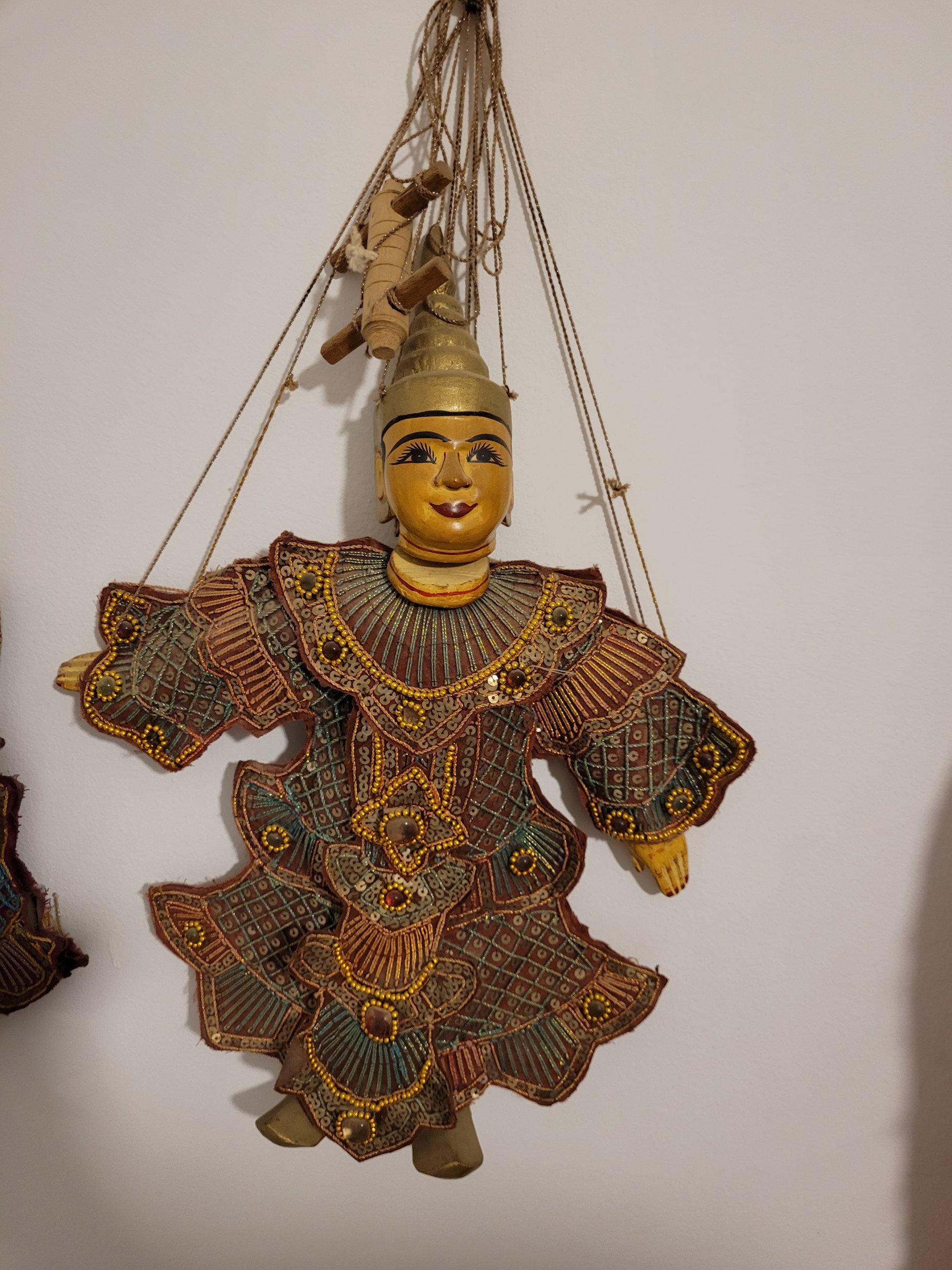} & 
            \includegraphics[width=0.09\linewidth,height=0.09\linewidth]{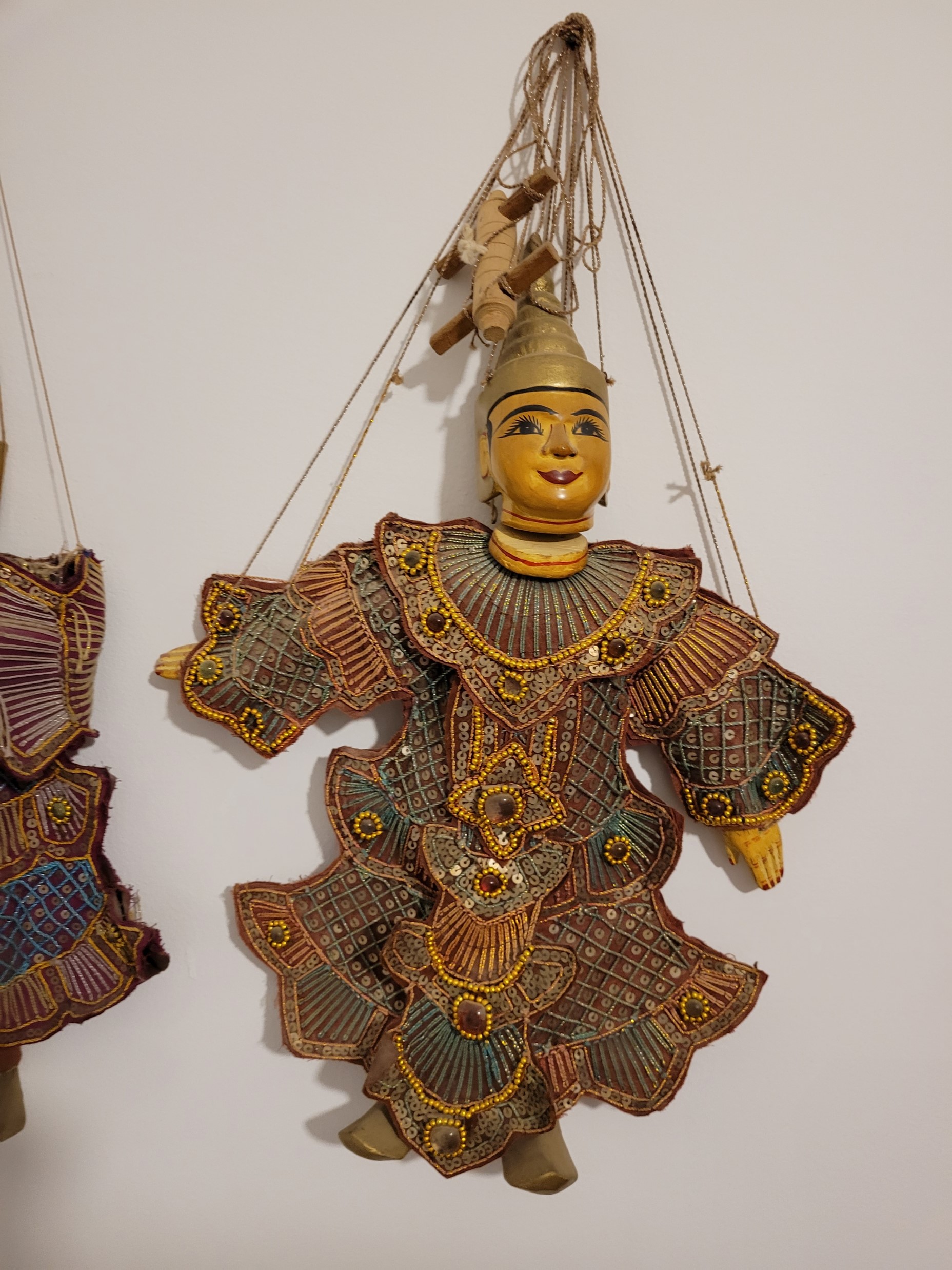} \\
            \multicolumn{2}{c}{\includegraphics[width=0.09\linewidth,height=0.09\linewidth]{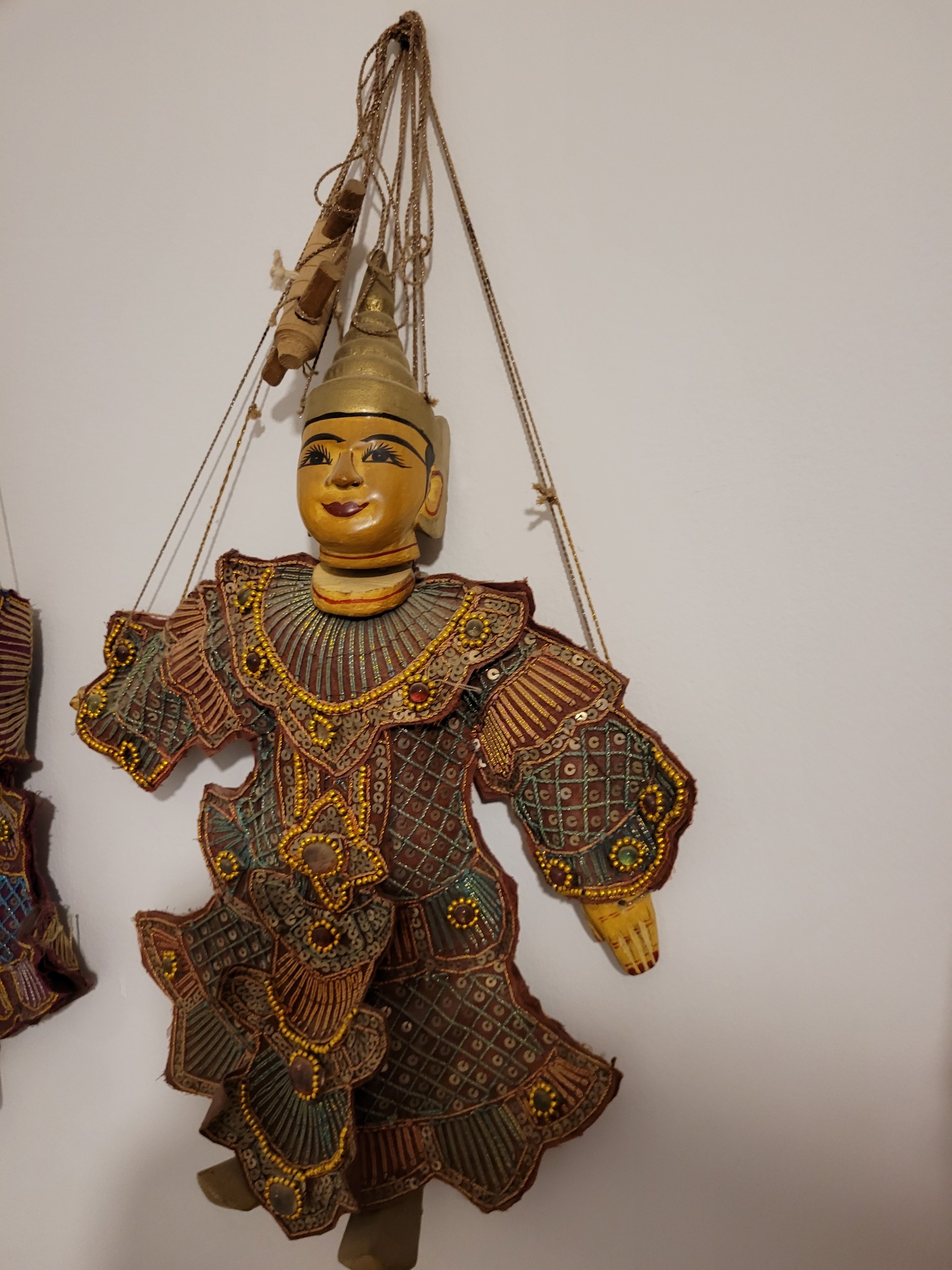}}
        \end{tabular}
        
        &
        $\rightarrow$
        &
        \begin{tabular}{c}
        \includegraphics[width=0.184\linewidth]{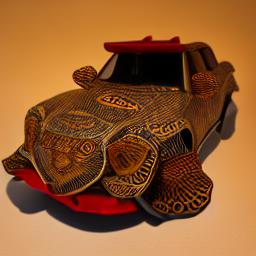}
        \end{tabular} &
        \begin{tabular}{c}
        \includegraphics[width=0.184\linewidth]{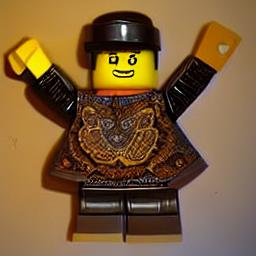} 
        \end{tabular} &
        \begin{tabular}{c}
        \includegraphics[width=0.184\linewidth]{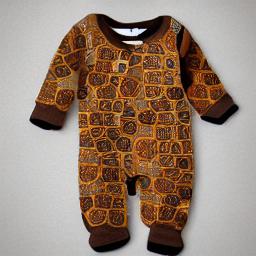} 
        \end{tabular} &
        \begin{tabular}{c}
        \includegraphics[width=0.184\linewidth]{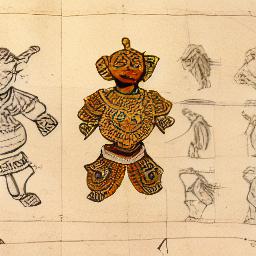}
        \end{tabular} \\
        
        {\footnotesize Input samples} & & {\begin{tabular}{c@{}c@{}c@{}c@{}} ``\pholdercolor{} sports car" \end{tabular}} & {\begin{tabular}{c@{}c@{}c@{}c@{}} ``\pholdercolor{} made of lego" \end{tabular}} & {\begin{tabular}{c@{}c@{}c@{}c@{}} ``\pholdercolor{} onesie" \end{tabular}} & {\begin{tabular}{c@{}c@{}c@{}c@{}} ``da Vinci sketch of \pholdercolor" \end{tabular}} \\ \\

        \begin{tabular}{c c}
            \includegraphics[width=0.09\linewidth,height=0.09\linewidth]{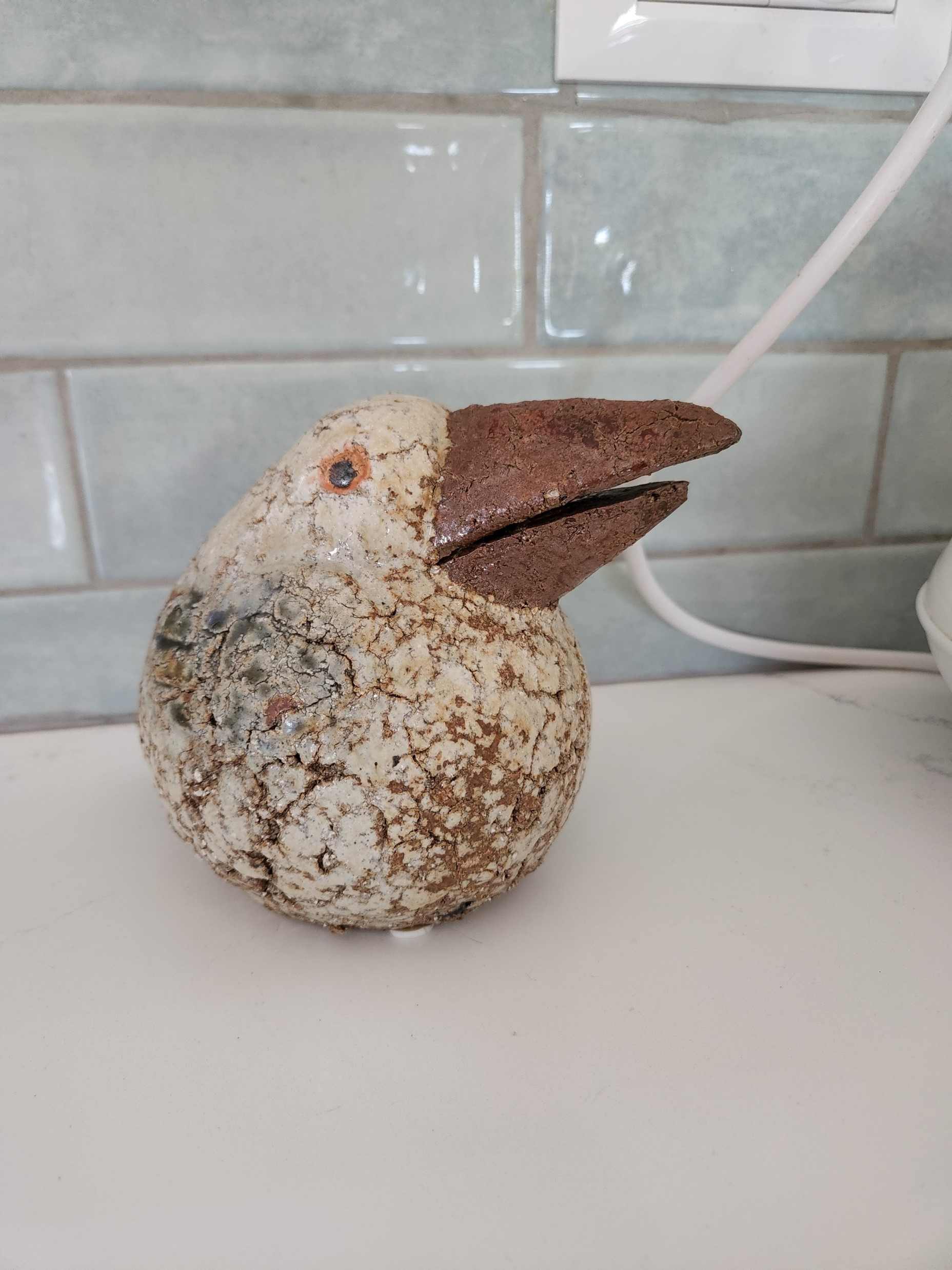} & 
            \includegraphics[width=0.09\linewidth,height=0.09\linewidth]{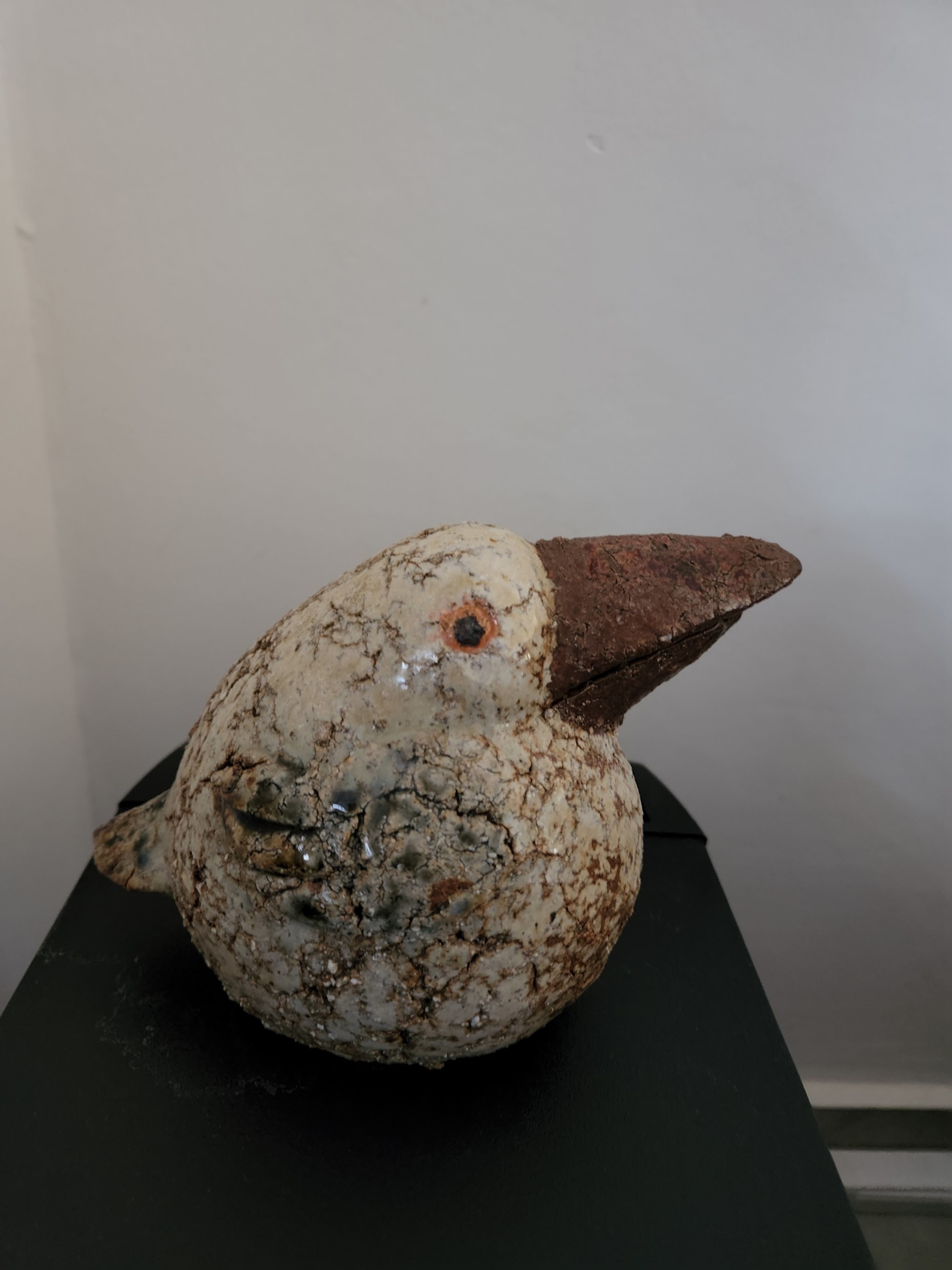} \\
            \includegraphics[width=0.09\linewidth,height=0.09\linewidth]{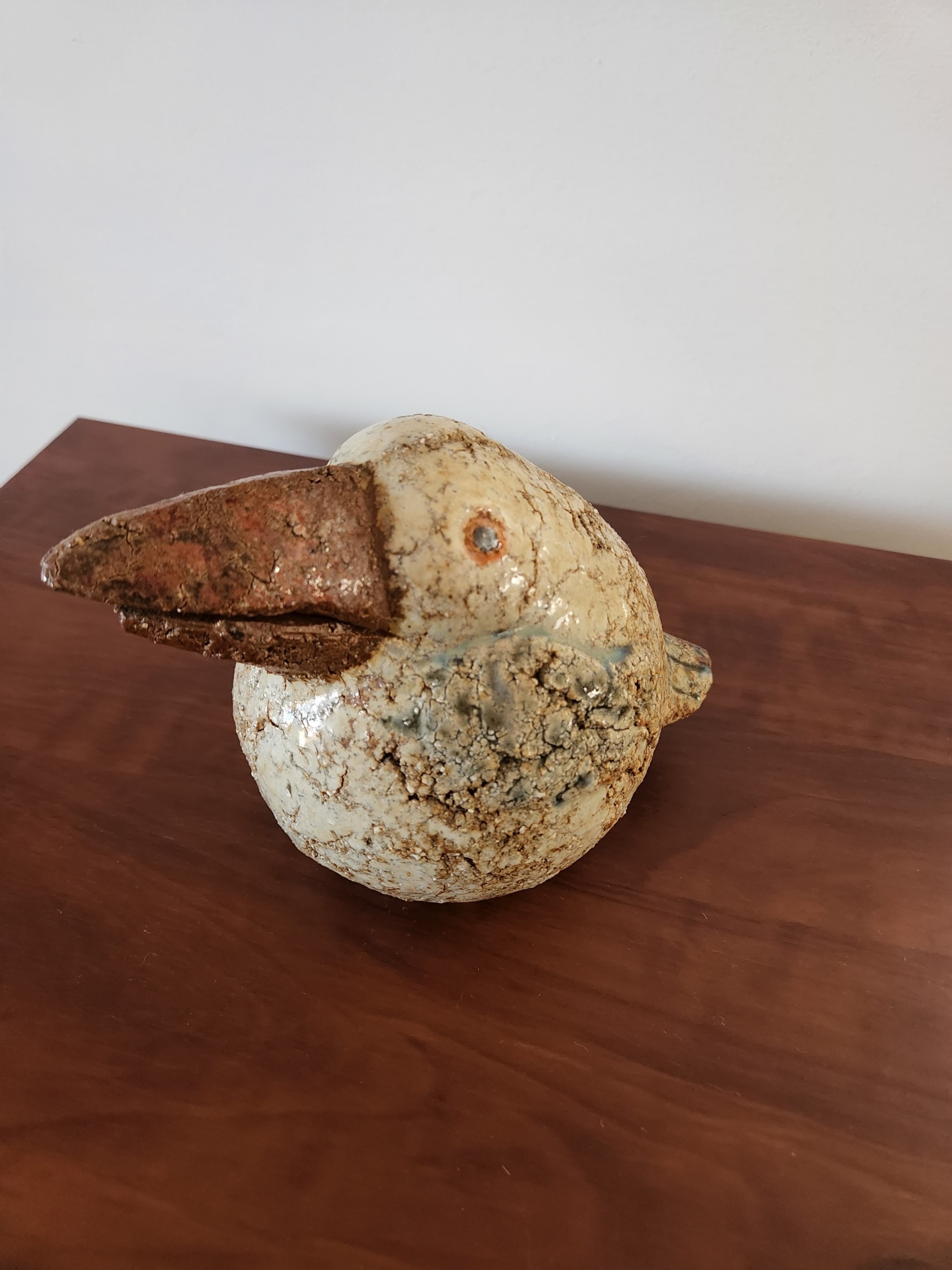} & 
            \includegraphics[width=0.09\linewidth,height=0.09\linewidth]{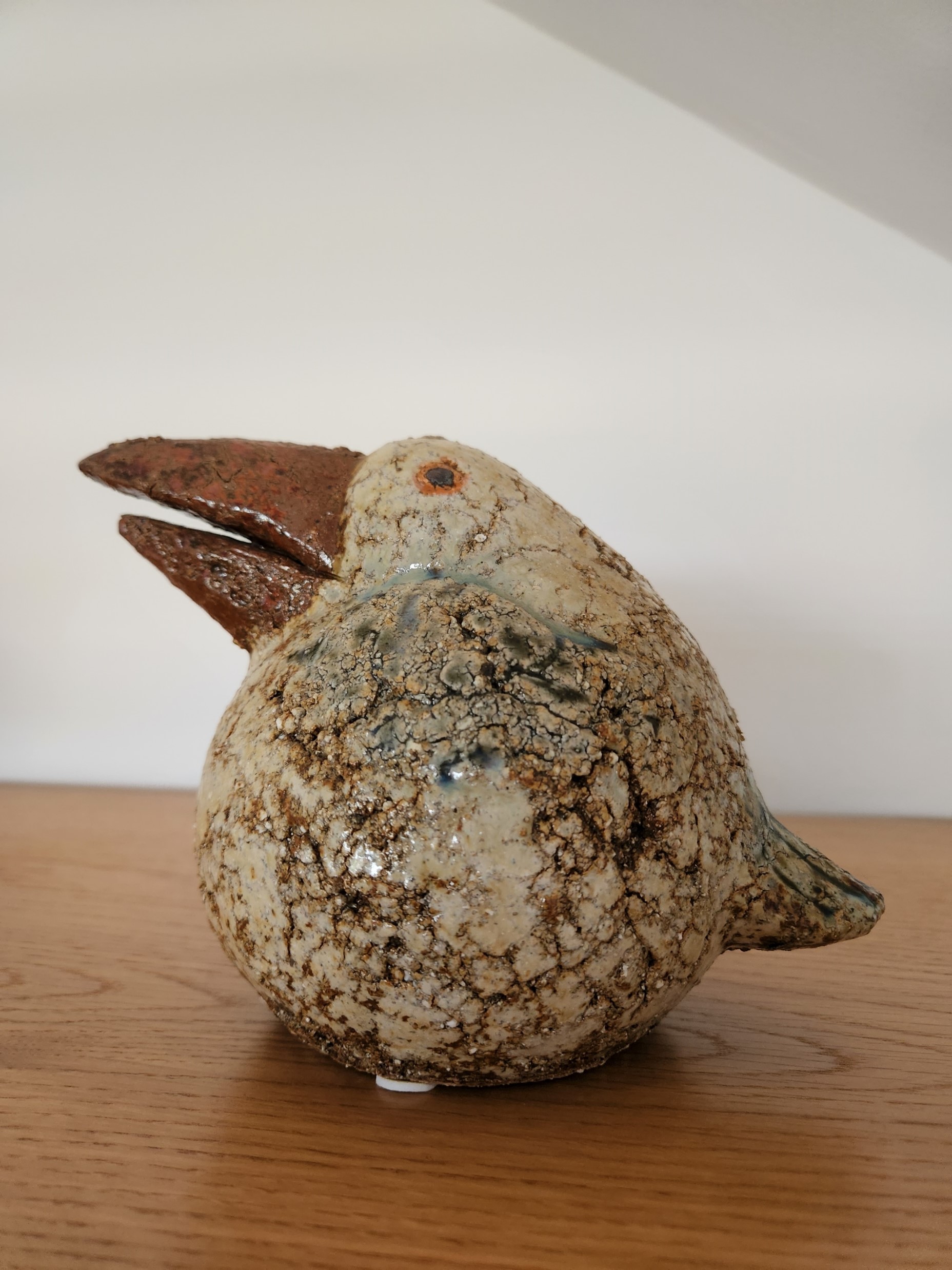}
        \end{tabular}
        
        &
        $\rightarrow$
        &
        \begin{tabular}{c}
        \includegraphics[width=0.184\linewidth]{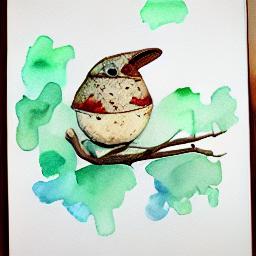}
        \end{tabular} &
        \begin{tabular}{c}
        \includegraphics[width=0.184\linewidth]{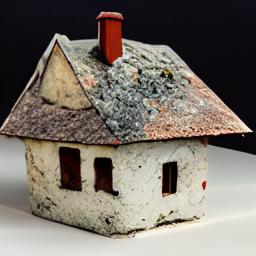} 
        \end{tabular} &
        \begin{tabular}{c}
        \includegraphics[width=0.184\linewidth]{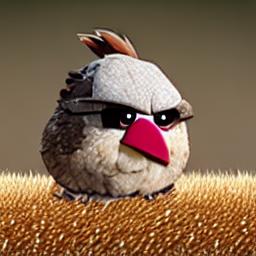} 
        \end{tabular} &
        \begin{tabular}{c}
        \includegraphics[width=0.184\linewidth]{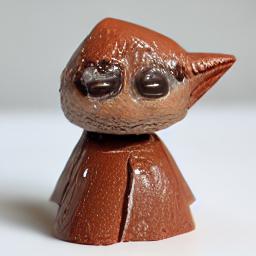}
        \end{tabular} \\
        
        {\footnotesize Input samples} & & {\begin{tabular}{c@{}c@{}c@{}c@{}} ``Watercolor painting of \pholdercolor \\ on a branch" \end{tabular}} & {\begin{tabular}{c@{}c@{}c@{}c@{}} ``A house in the style of \pholdercolor" \end{tabular}} & {\begin{tabular}{c@{}c@{}c@{}c@{}} ``Grainy photo of \pholdercolor \\ in angry birds" \end{tabular}} & {\begin{tabular}{c@{}c@{}c@{}c@{}} ``\pholdercolor{} made of chocolate" \end{tabular}} \\ \\

        \begin{tabular}{c c}
            \includegraphics[width=0.09\linewidth,height=0.09\linewidth]{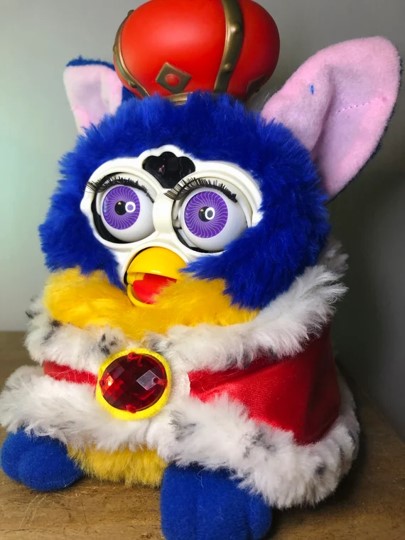} & 
            \includegraphics[width=0.09\linewidth,height=0.09\linewidth]{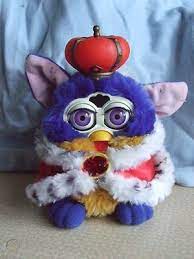} \\
            \includegraphics[width=0.09\linewidth,height=0.09\linewidth]{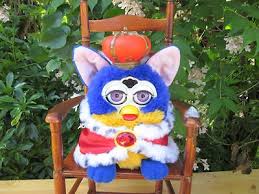} & 
            \includegraphics[width=0.09\linewidth,height=0.09\linewidth]{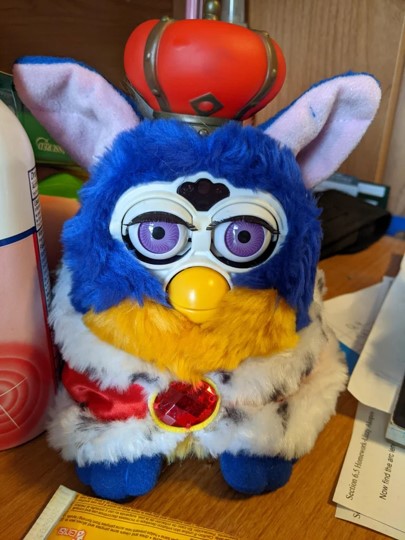}
        \end{tabular}
        
        &
        $\rightarrow$
        &
        \begin{tabular}{c}
        \includegraphics[width=0.184\linewidth]{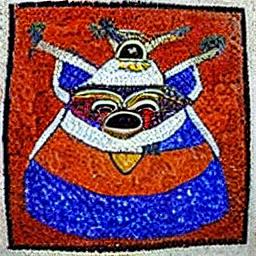}
        \end{tabular} &
        \begin{tabular}{c}
        \includegraphics[width=0.184\linewidth]{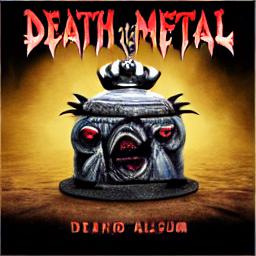} 
        \end{tabular} &
        \begin{tabular}{c}
        \includegraphics[width=0.184\linewidth]{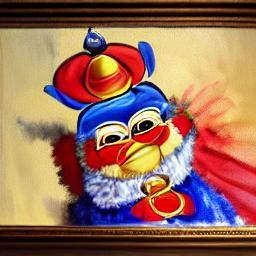} 
        \end{tabular} &
        \begin{tabular}{c}
        \includegraphics[width=0.184\linewidth]{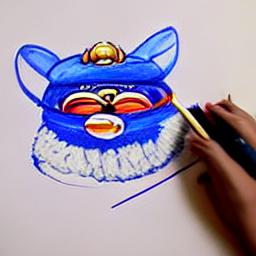}
        \end{tabular} \\
        
        {\footnotesize Input samples} & & {\begin{tabular}{c@{}c@{}c@{}c@{}} ``A mosaic depicting \pholdercolor" \end{tabular}} & {\begin{tabular}{c@{}c@{}c@{}c@{}} ``Death metal album cover \\ featuring \pholdercolor" \end{tabular}} & {\begin{tabular}{c@{}c@{}c@{}c@{}} ``Masterful oil painting of \pholdercolor \\ hanging on the wall" \end{tabular}} & {\begin{tabular}{c@{}c@{}c@{}c@{}} ``An artist drawing a \pholdercolor" \end{tabular}} \\ \\

        \begin{tabular}{c c}
            \includegraphics[width=0.09\linewidth,height=0.09\linewidth]{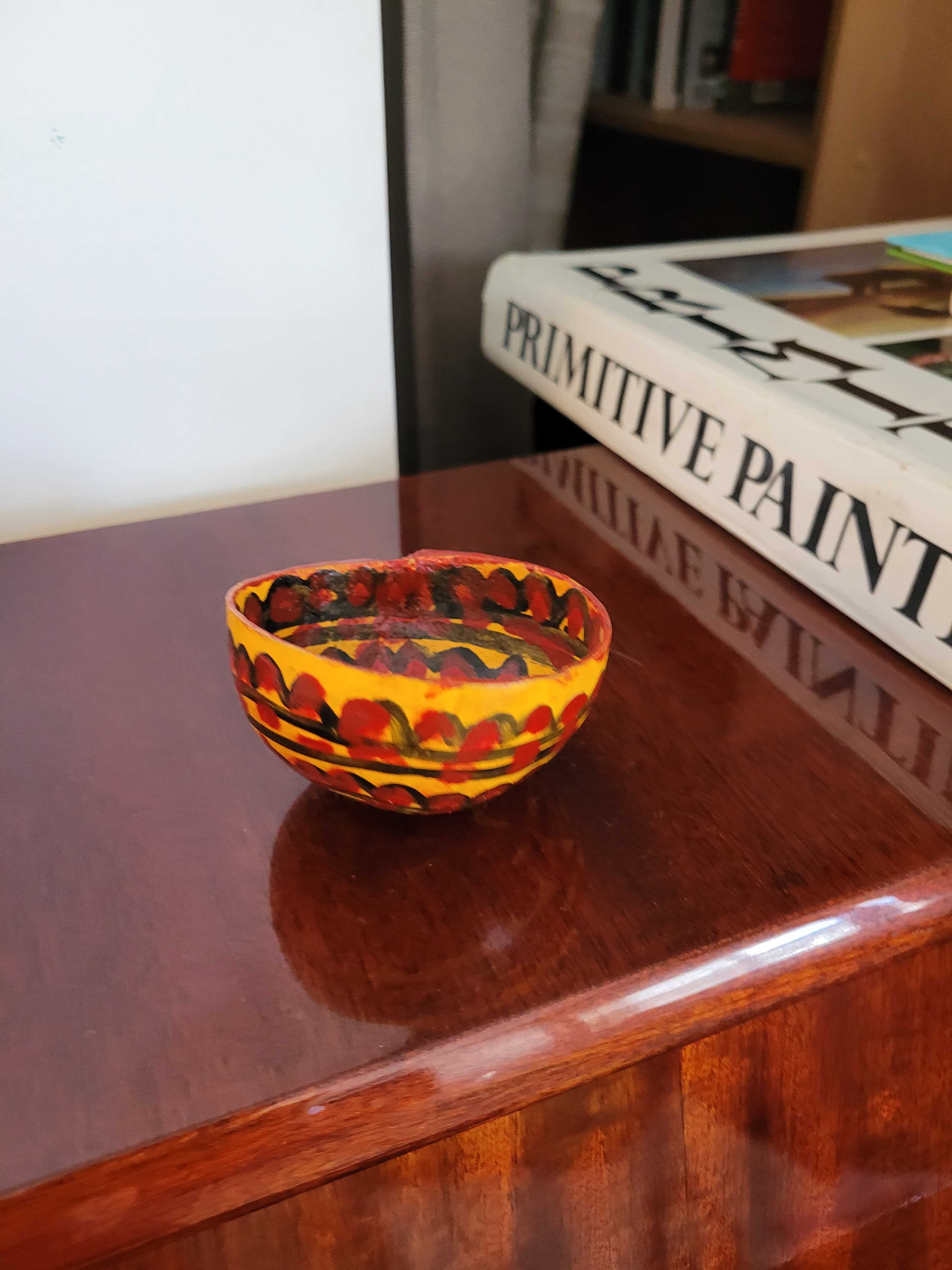} & 
            \includegraphics[width=0.09\linewidth,height=0.09\linewidth]{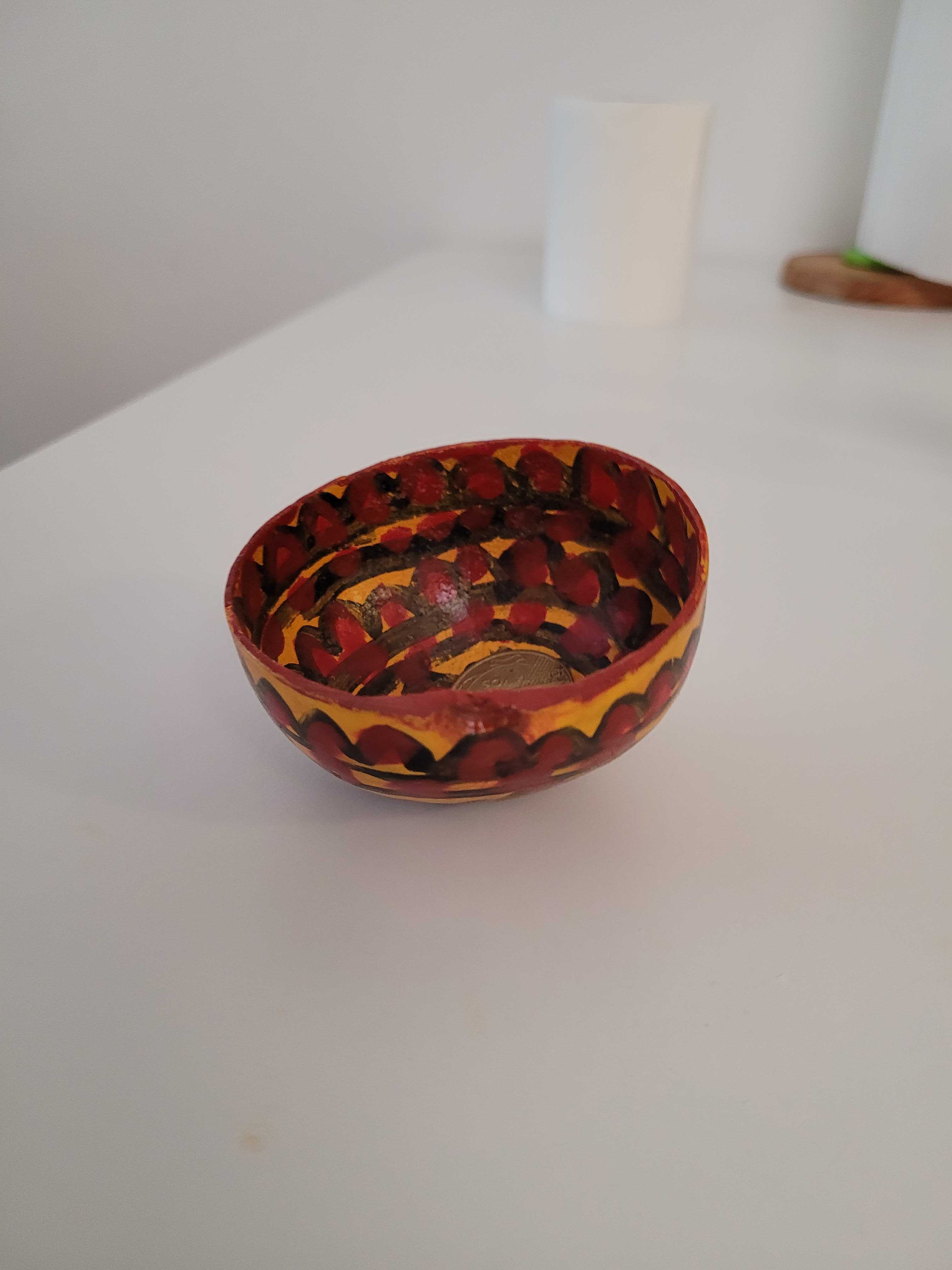} \\
            \includegraphics[width=0.09\linewidth,height=0.09\linewidth]{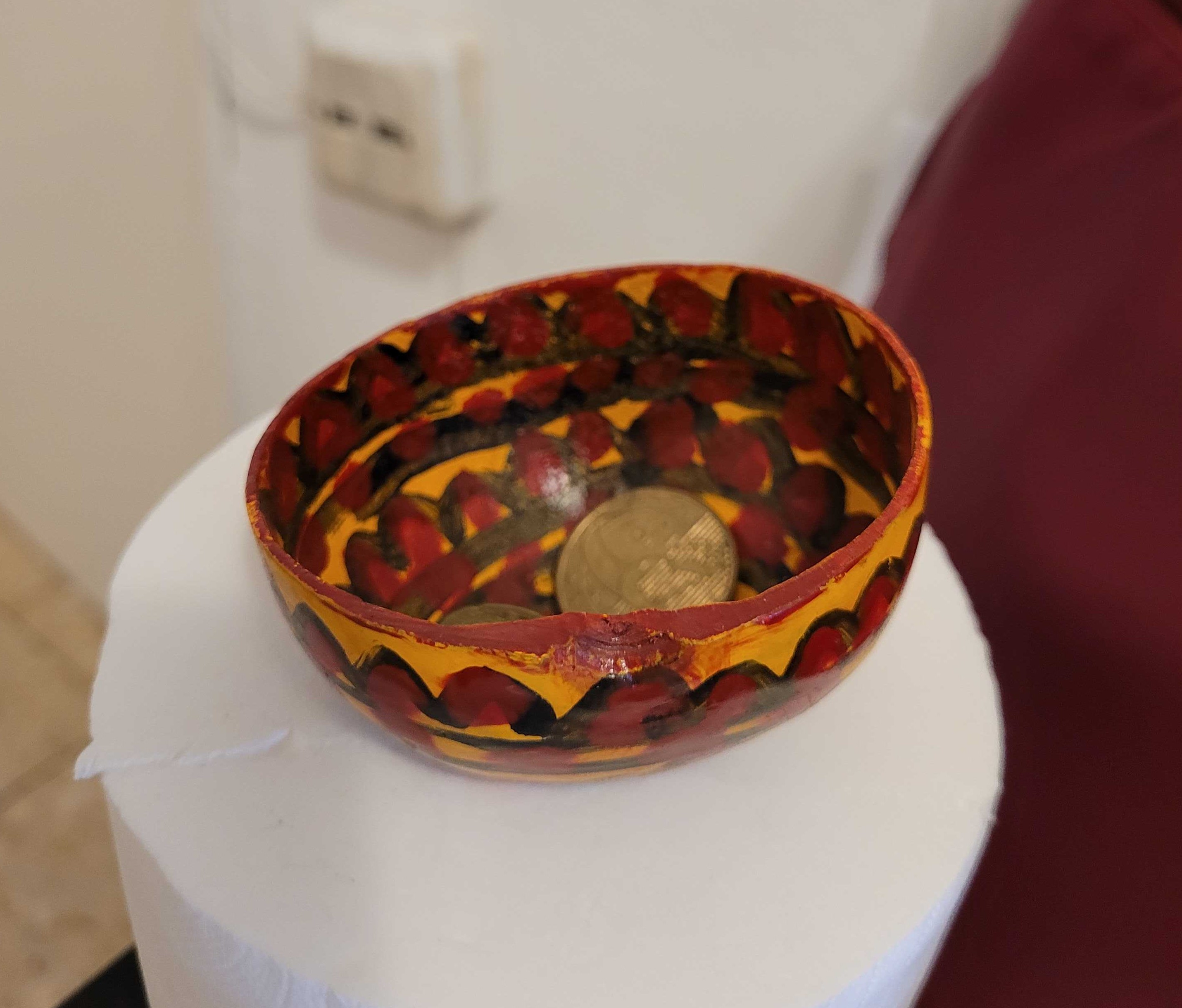} & 
            \includegraphics[width=0.09\linewidth,height=0.09\linewidth]{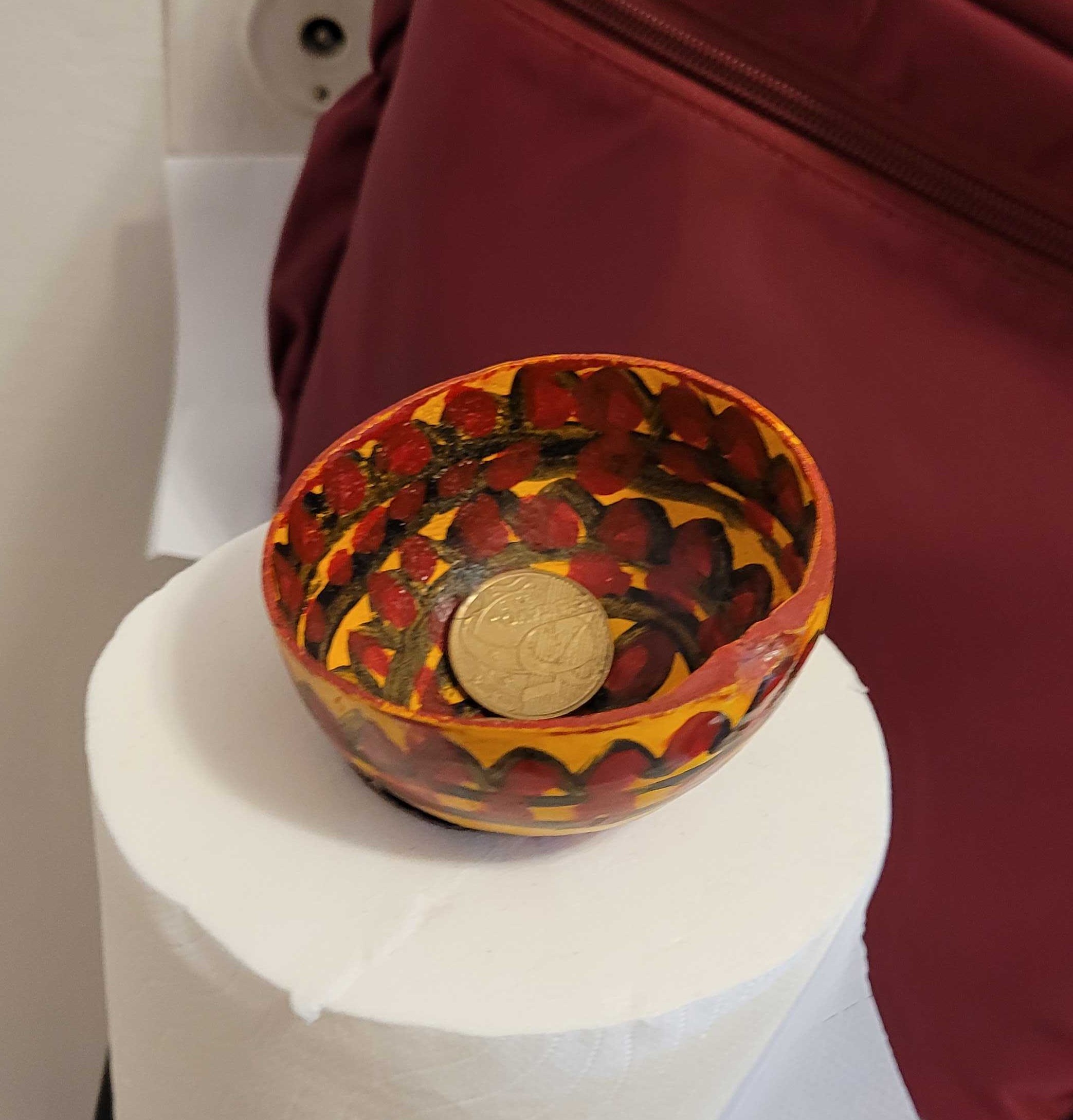}
        \end{tabular}
        
        &
        $\rightarrow$
        &
        \begin{tabular}{c}
        \includegraphics[width=0.184\linewidth]{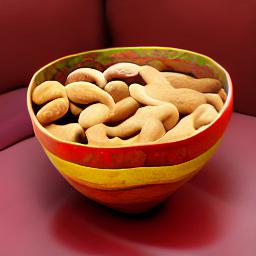}
        \end{tabular} &
        \begin{tabular}{c}
        \includegraphics[width=0.184\linewidth]{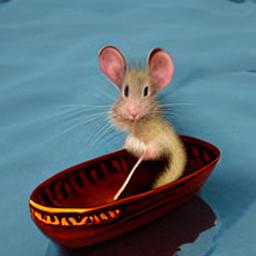} 
        \end{tabular} &
        \begin{tabular}{c}
        \includegraphics[width=0.184\linewidth]{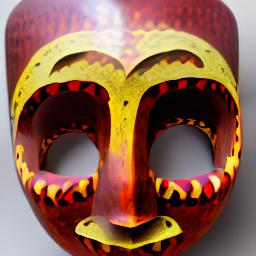} 
        \end{tabular} &
        \begin{tabular}{c}
        \includegraphics[width=0.184\linewidth]{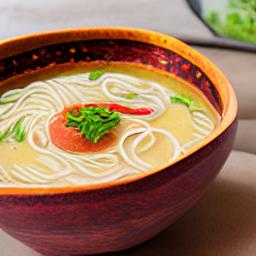}
        \end{tabular} \\
        
        {\footnotesize Input samples} & & {\begin{tabular}{c@{}c@{}c@{}c@{}} ``A photo of \pholdercolor{} full \\ of cashew nuts" \end{tabular}} & {\begin{tabular}{c@{}c@{}c@{}c@{}} ``A mouse using \pholdercolor \\ as a boat" \end{tabular}} & {\begin{tabular}{c@{}c@{}c@{}c@{}} ``A photo of a \\ \pholdercolor{} mask" \end{tabular}} & {\begin{tabular}{c@{}c@{}c@{}c@{}} ``Ramen soup served in \pholdercolor" \end{tabular}} \\ \\
        
        \begin{tabular}{c c}
            \includegraphics[width=0.09\linewidth,height=0.09\linewidth]{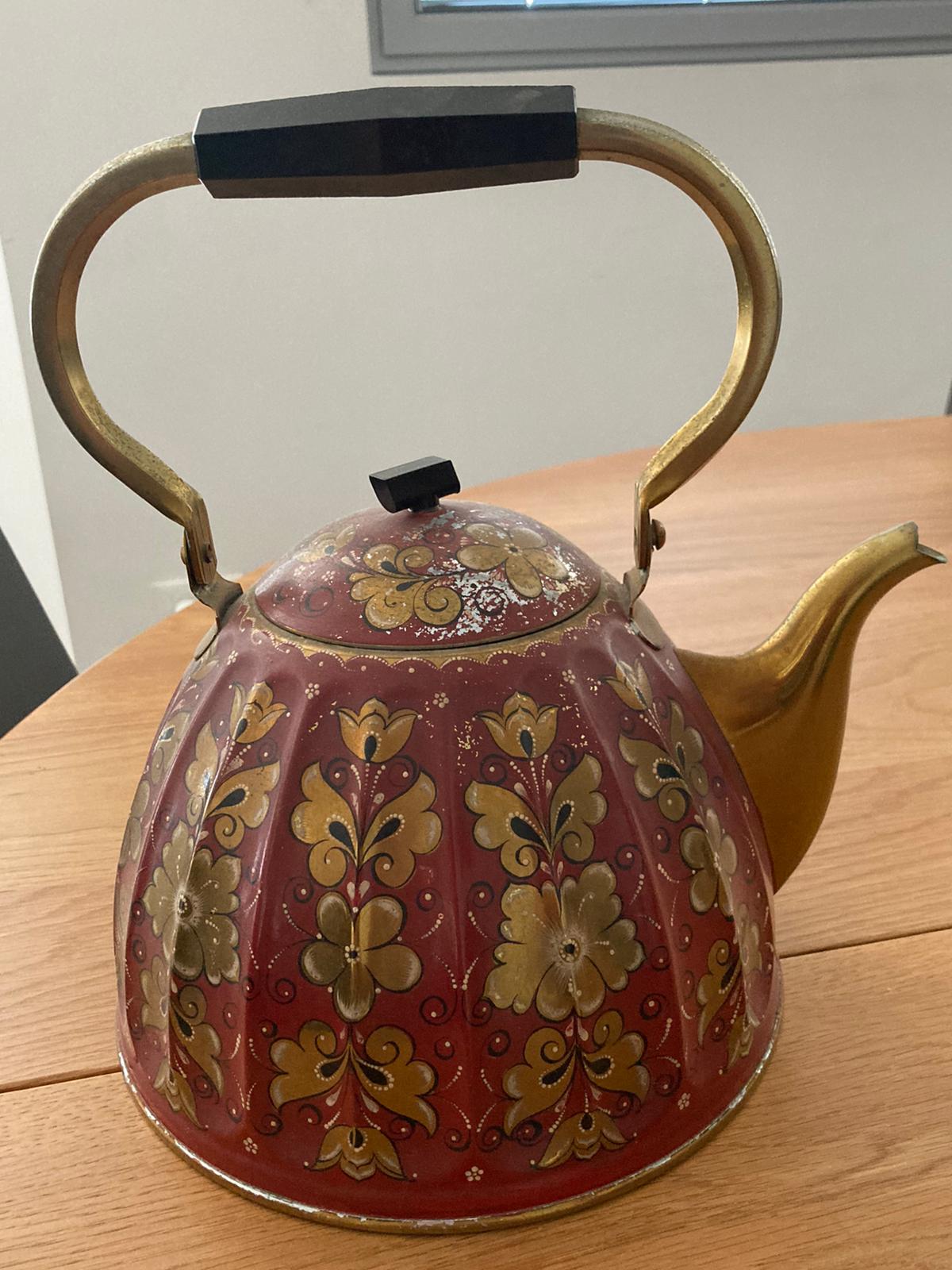} & 
            \includegraphics[width=0.09\linewidth,height=0.09\linewidth]{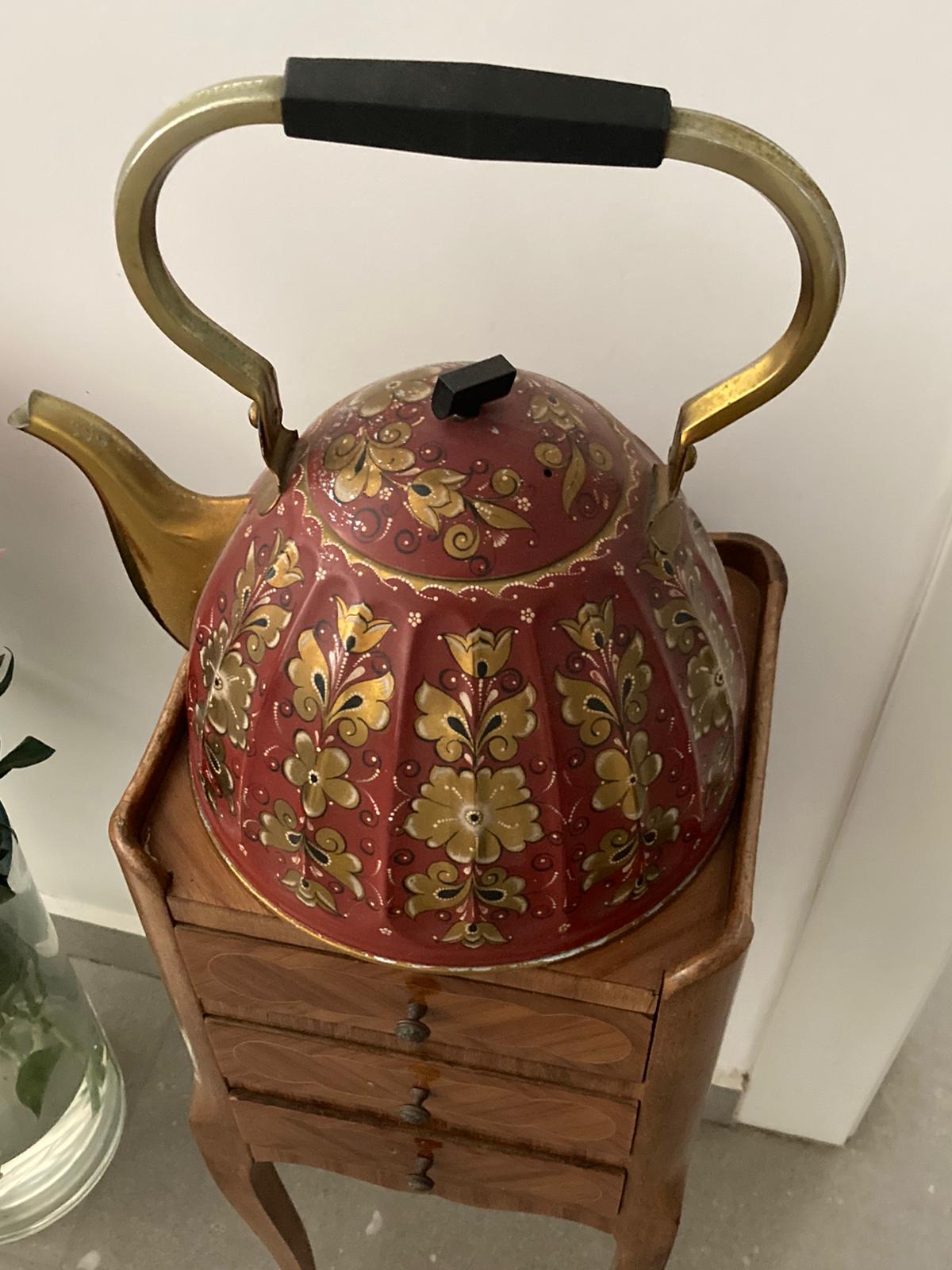} \\
            \multicolumn{2}{c}{\includegraphics[width=0.09\linewidth,height=0.09\linewidth]{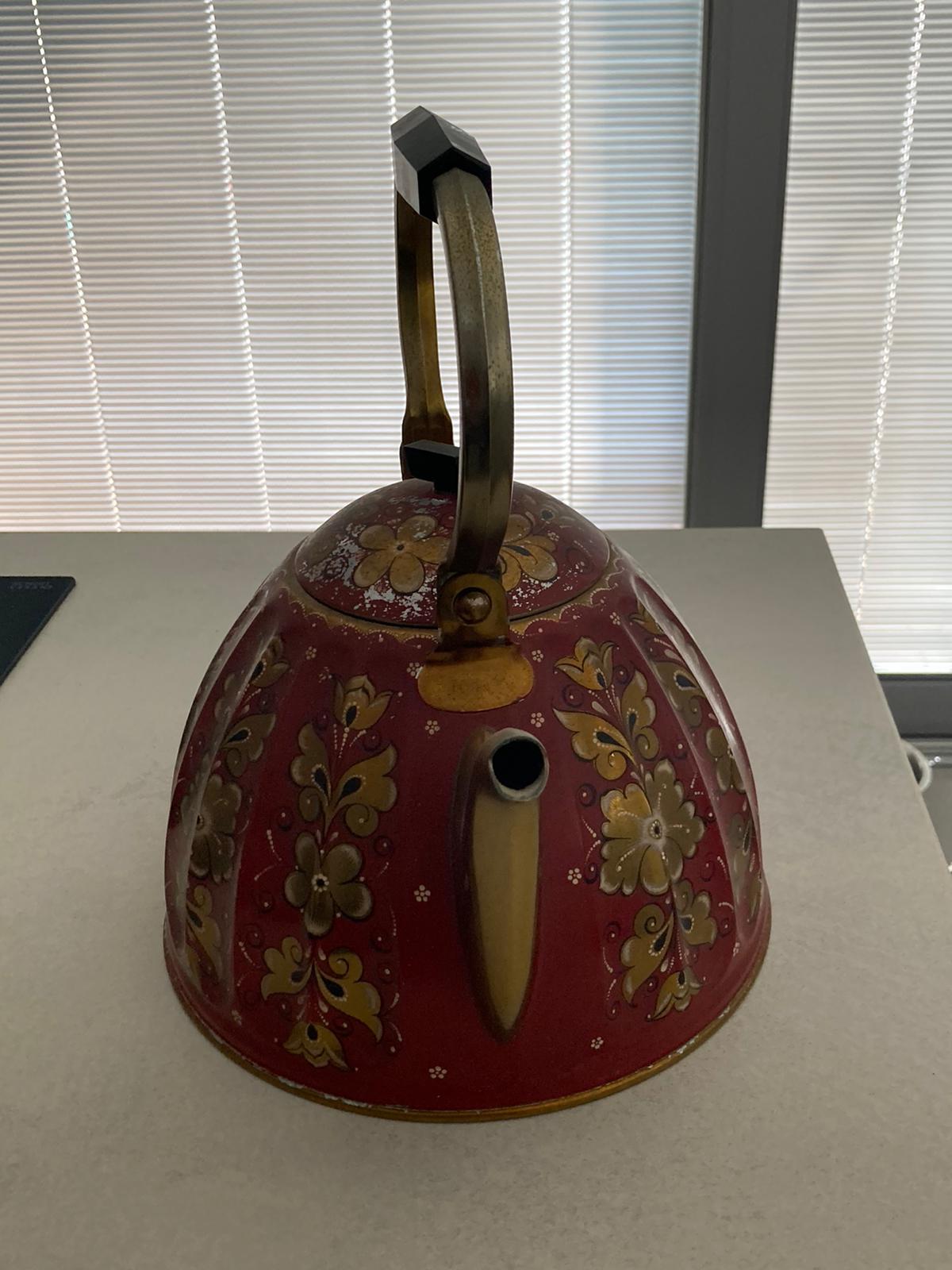}}
        \end{tabular}
        &
        $\rightarrow$
        &
        \begin{tabular}{c}
        \includegraphics[width=0.184\linewidth]{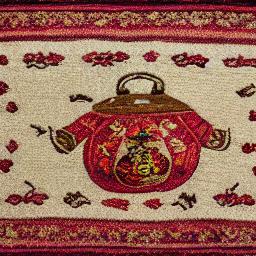} 
        \end{tabular} &
        \begin{tabular}{c}
        \includegraphics[width=0.184\linewidth]{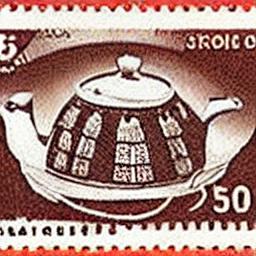} 
        \end{tabular} &
        \begin{tabular}{c}
        \includegraphics[width=0.184\linewidth]{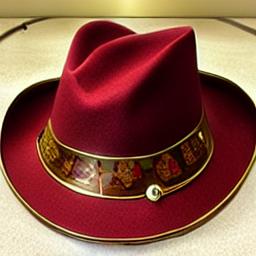} 
        \end{tabular} &
        \begin{tabular}{c}
        \includegraphics[width=0.184\linewidth]{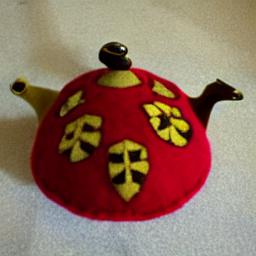} 
        \end{tabular} \\
        
        {\footnotesize Input samples} & & {\begin{tabular}{c@{}c@{}c@{}c@{}} ``A carpet with \\ \pholdercolor{} embroidery" \end{tabular}} & {\begin{tabular}{c@{}c@{}c@{}c@{}} ``A \pholdercolor{} stamp" \end{tabular}} & {\begin{tabular}{c@{}c@{}c@{}c@{}} ``\pholdercolor{} fedora" \end{tabular}} & {\begin{tabular}{c@{}c@{}c@{}c@{}} ``Felt \pholdercolor" \end{tabular}} \\

    \end{tabular}}
    \caption{Additional text-guided personalized generation results. In each row, we shows exemplars from the image set representing the concept (left), and novel compositions using the pseudo-word derived from these samples (right).}
    \label{fig:conditional_gen} 
\end{figure}

%% file: resources/figures/baseline_comparisons.tex
\begin{figure}[!hbt]
    \centering
    \setlength{\abovecaptionskip}{6.5pt}
    \setlength{\belowcaptionskip}{-3.5pt}
    \setlength{\tabcolsep}{0.55pt}
    \renewcommand{\arraystretch}{1.0}
    {

    \begin{tabular}{c}

    \begin{tabular}{c@{\hskip 5pt}  c c c c@{\hskip 10pt}  c c c c}
    & \multicolumn{4}{c}{Clock} & \multicolumn{4}{c}{Mug} \\
    \raisebox{0.045\linewidth}{\footnotesize\begin{tabular}{c@{}c@{}c@{}c@{}} Input \\ samples \end{tabular}} & \includegraphics[width=0.1\linewidth,height=0.1\linewidth]{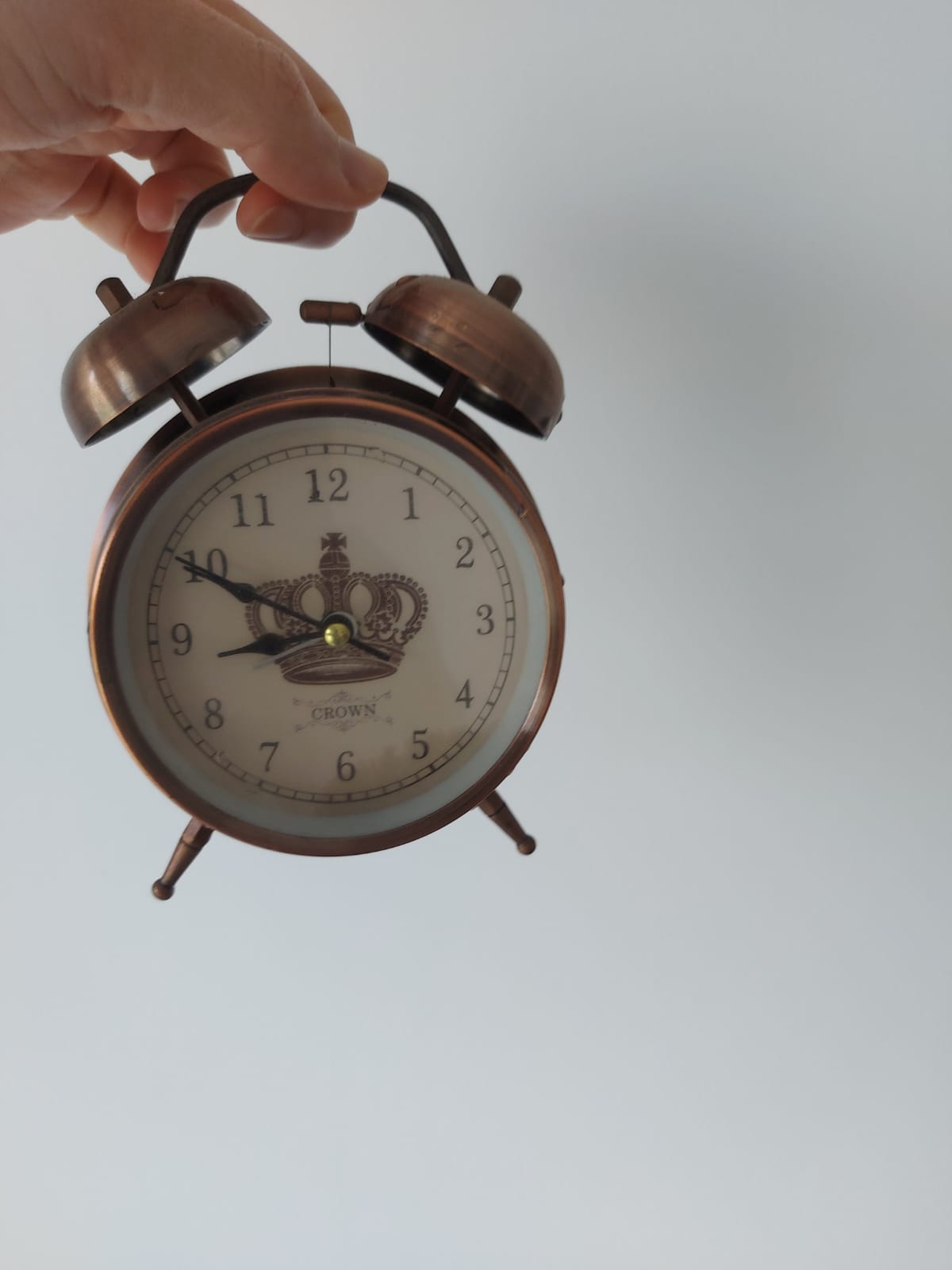} & 
    \includegraphics[width=0.1\linewidth,height=0.1\linewidth]{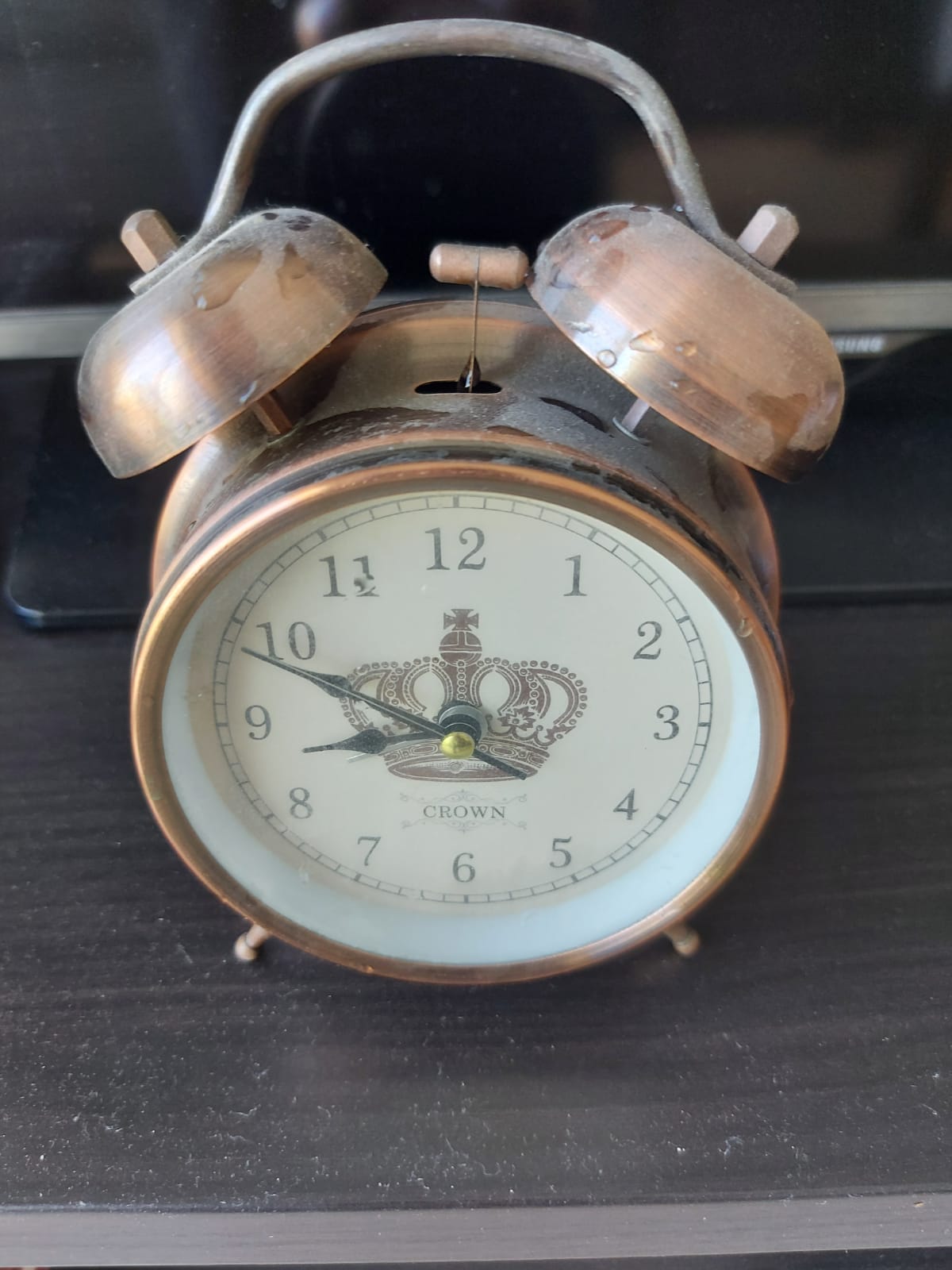} & 
    \includegraphics[width=0.1\linewidth,height=0.1\linewidth]{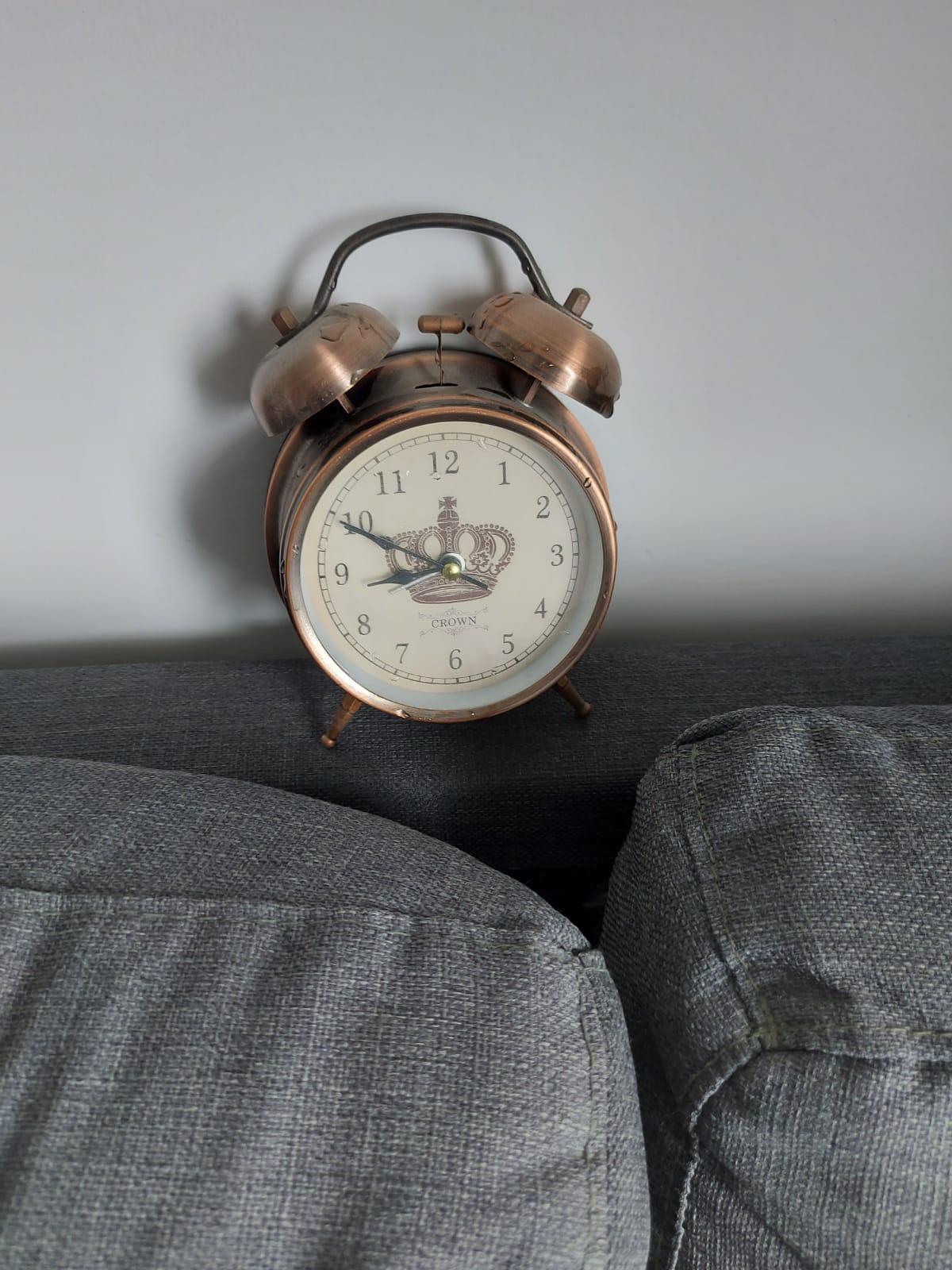} & 
    \includegraphics[width=0.1\linewidth,height=0.1\linewidth]{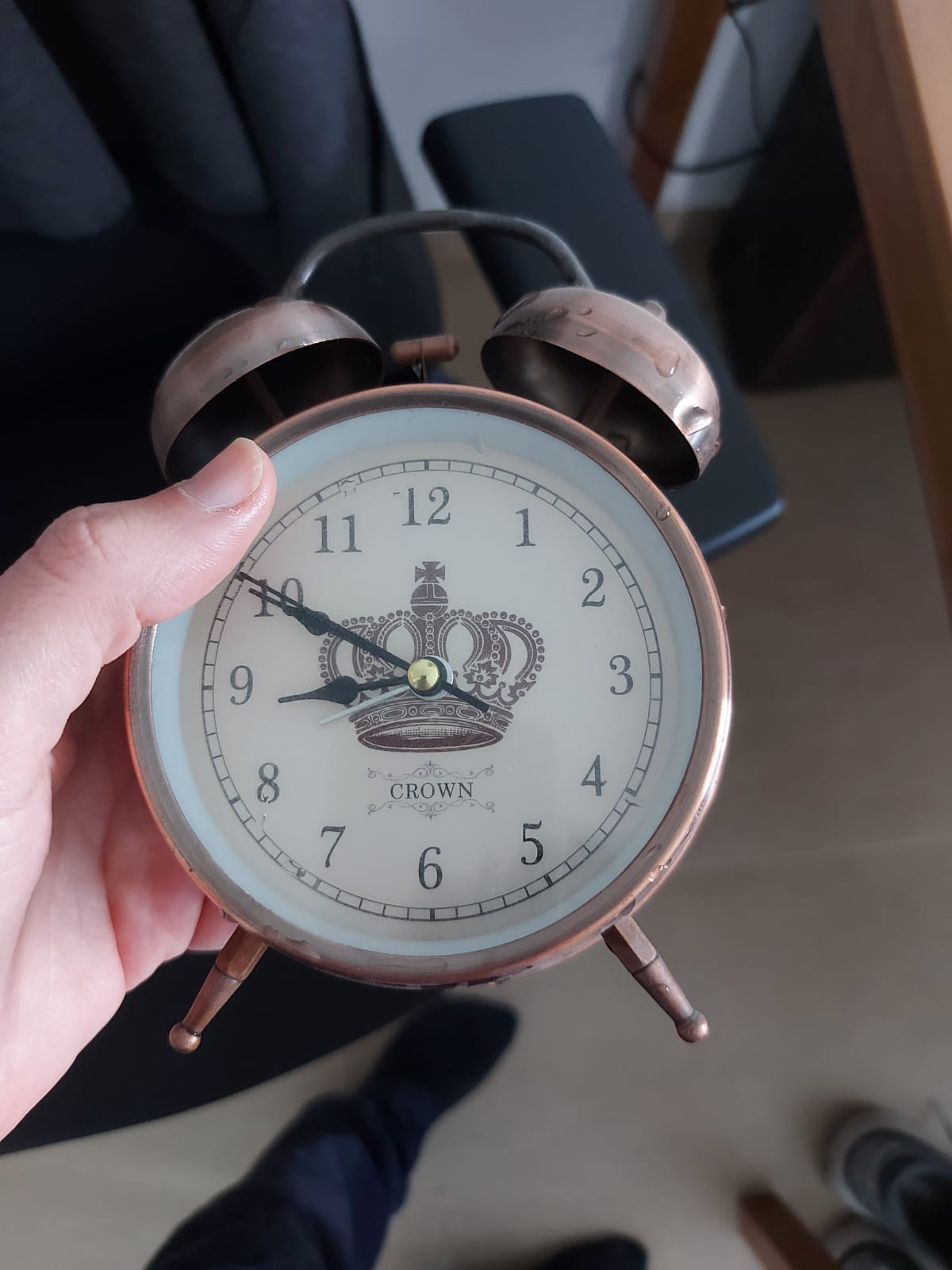} &
    \includegraphics[width=0.1\linewidth,height=0.1\linewidth]{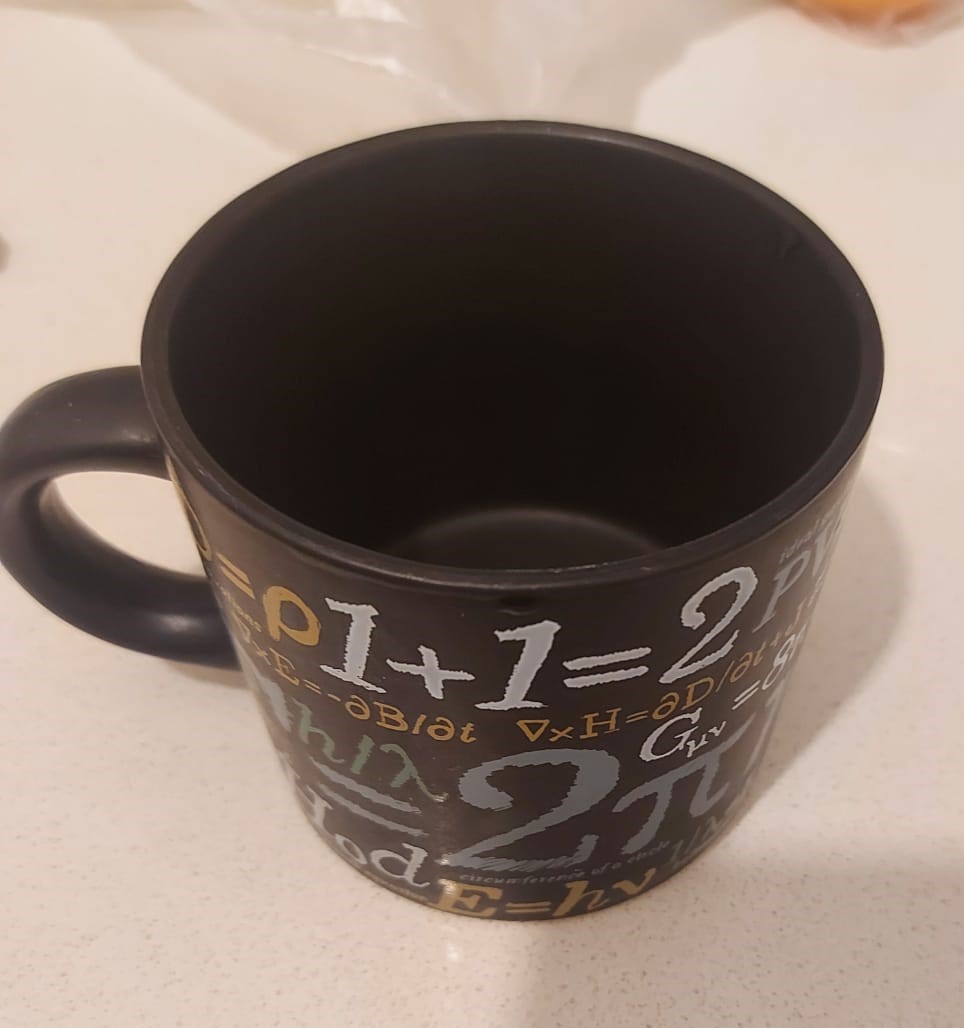} & 
    \includegraphics[width=0.1\linewidth,height=0.1\linewidth]{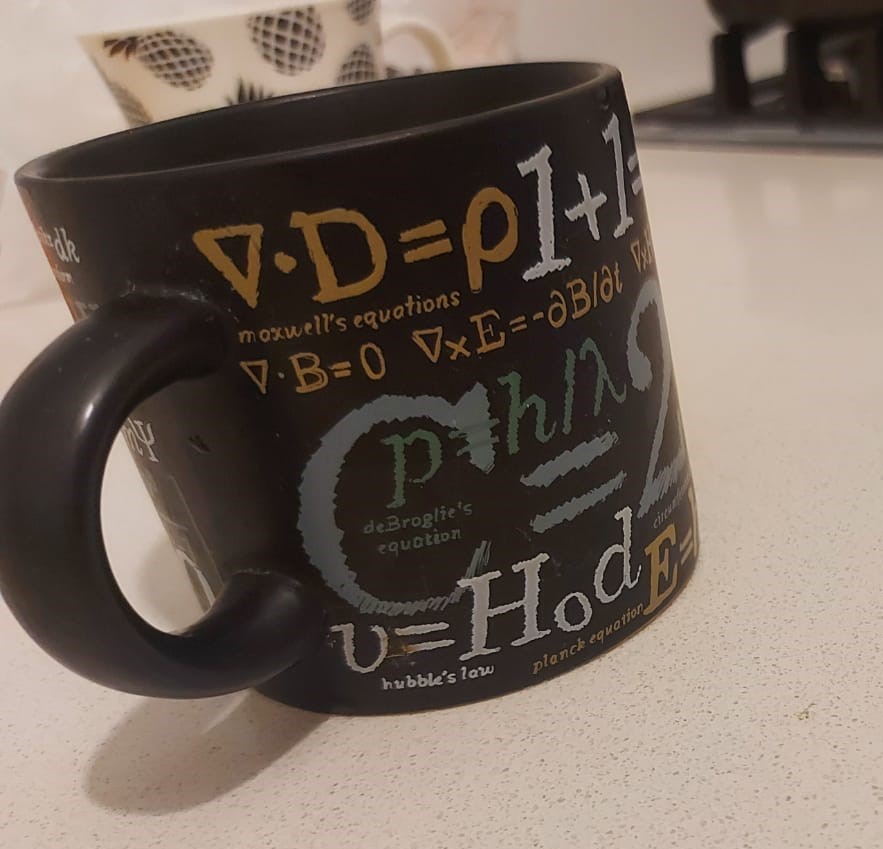} &
    \includegraphics[width=0.1\linewidth,height=0.1\linewidth]{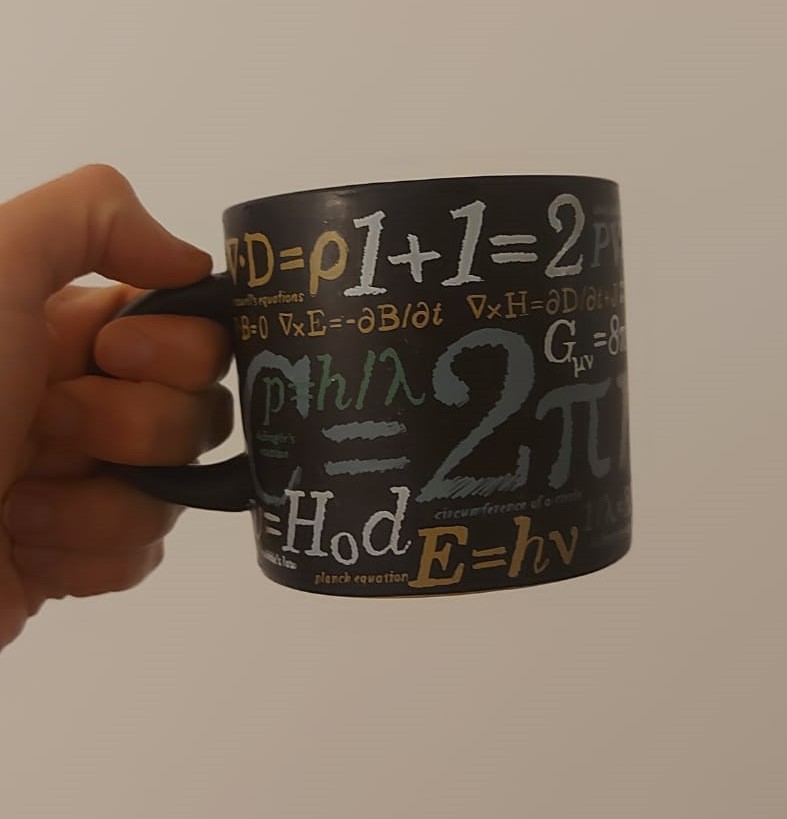} & 
    \includegraphics[width=0.1\linewidth,height=0.1\linewidth]{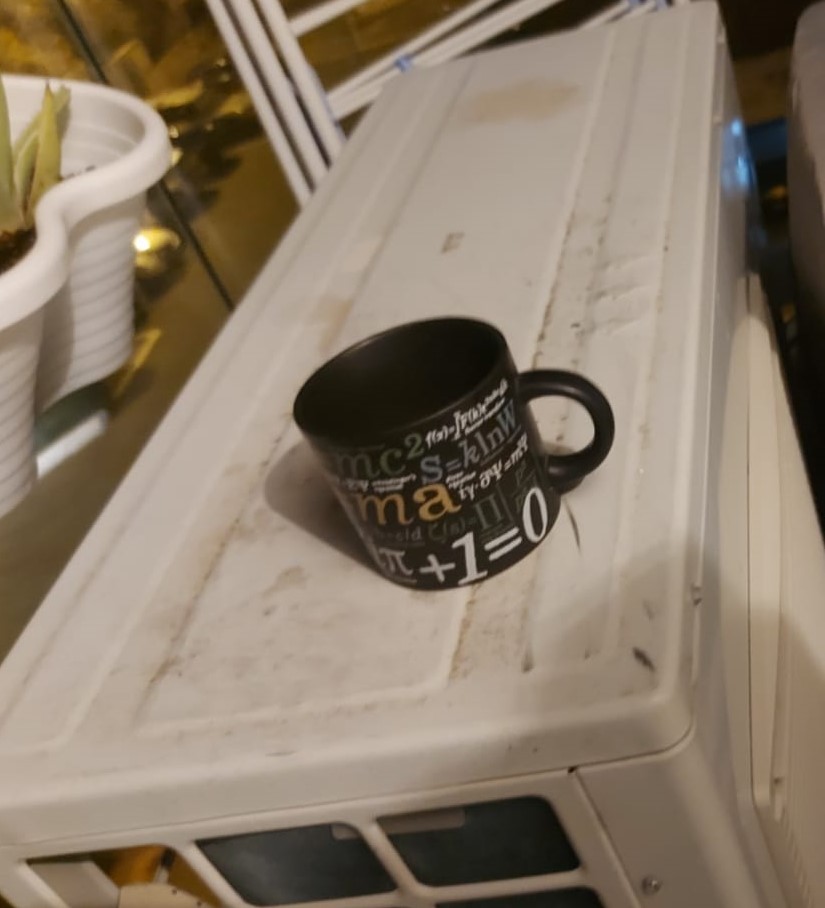} \\ \\
    
    \raisebox{0.045\linewidth}{\tiny\begin{tabular}{c@{}c@{}c@{}c@{}} PALAVRA \\ (VQGAN+CLIP) \end{tabular}}  &
    \includegraphics[width=0.1\linewidth]{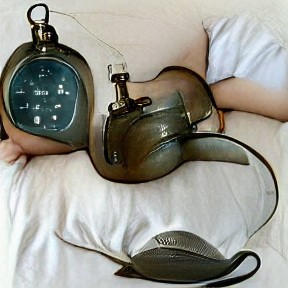} &
    \includegraphics[width=0.1\linewidth]{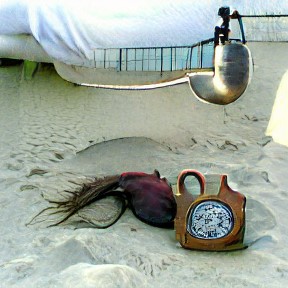} &
    \includegraphics[width=0.1\linewidth]{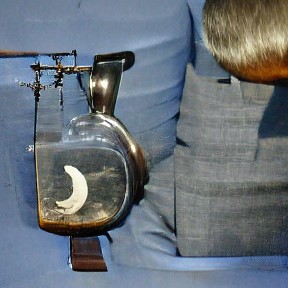} &
    \includegraphics[width=0.1\linewidth]{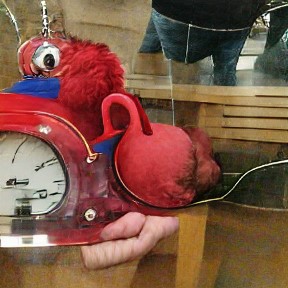} &
    \includegraphics[width=0.1\linewidth]{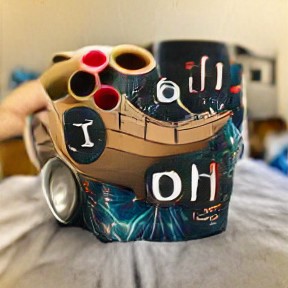} &
    \includegraphics[width=0.1\linewidth]{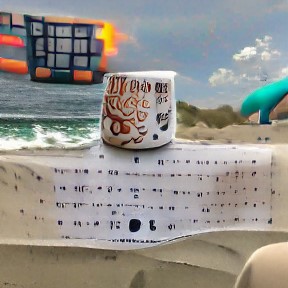} &
    \includegraphics[width=0.1\linewidth]{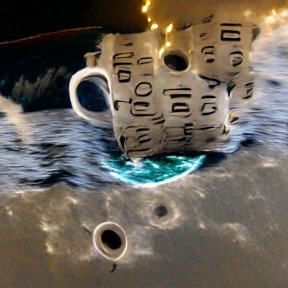} &
    \includegraphics[width=0.1\linewidth]{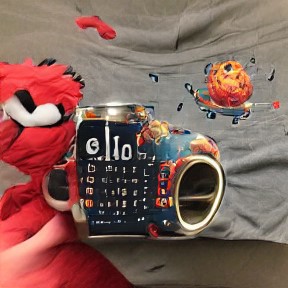} \\
    
    \raisebox{0.045\linewidth}{\tiny\begin{tabular}{c@{}c@{}c@{}c@{}} PALAVRA \\ (Guided Diffusion) \end{tabular}}  &
    \includegraphics[width=0.1\linewidth]{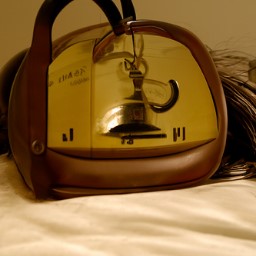} &
    \includegraphics[width=0.1\linewidth]{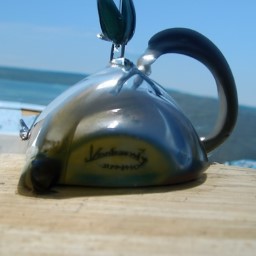} &
    \includegraphics[width=0.1\linewidth]{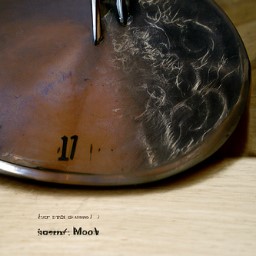} &
    \includegraphics[width=0.1\linewidth]{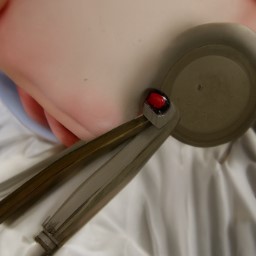} &
    \includegraphics[width=0.1\linewidth]{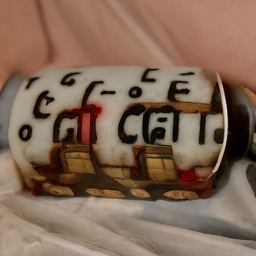} &
    \includegraphics[width=0.1\linewidth]{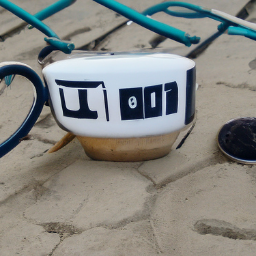} &
    \includegraphics[width=0.1\linewidth]{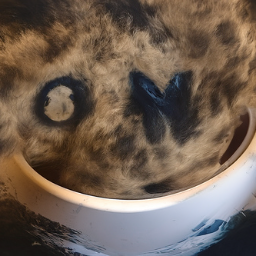} &
    \includegraphics[width=0.1\linewidth]{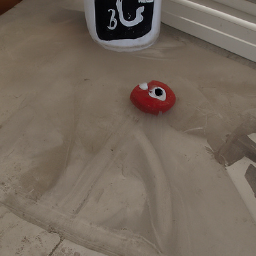} \\

    \raisebox{0.045\linewidth}{\tiny\begin{tabular}{c@{}c@{}c@{}c@{}} VQGAN+CLIP \\ (Text + Image Loss) \end{tabular}}  &
    \includegraphics[width=0.1\linewidth]{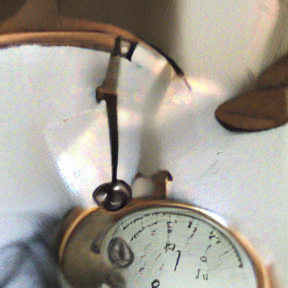} &
    \includegraphics[width=0.1\linewidth]{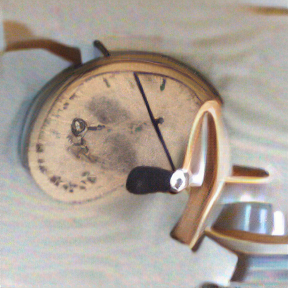} &
    \includegraphics[width=0.1\linewidth]{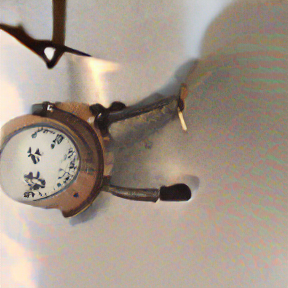} &
    \includegraphics[width=0.1\linewidth]{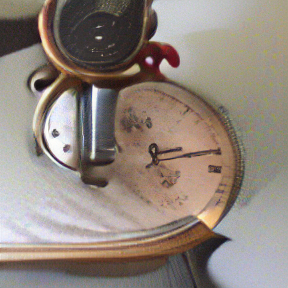} &
    \includegraphics[width=0.1\linewidth]{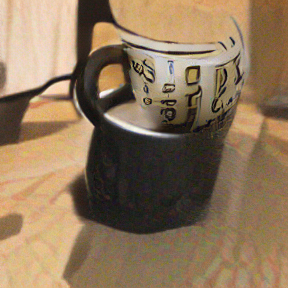} &
    \includegraphics[width=0.1\linewidth]{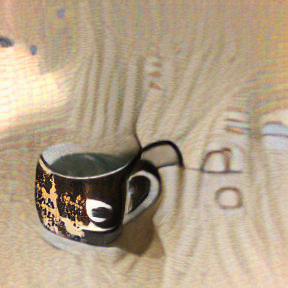} &
    \includegraphics[width=0.1\linewidth]{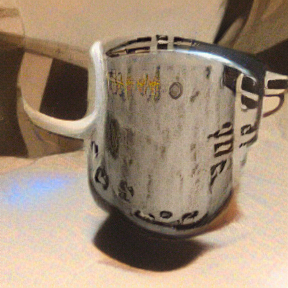} &
    \includegraphics[width=0.1\linewidth]{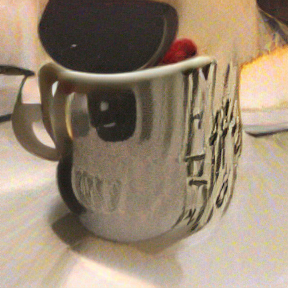} \\
    
    \raisebox{0.045\linewidth}{\tiny\begin{tabular}{c@{}c@{}c@{}c@{}} Guided Diffusion \\ (Text Loss + Init Image) \end{tabular}}  &
    \includegraphics[width=0.1\linewidth]{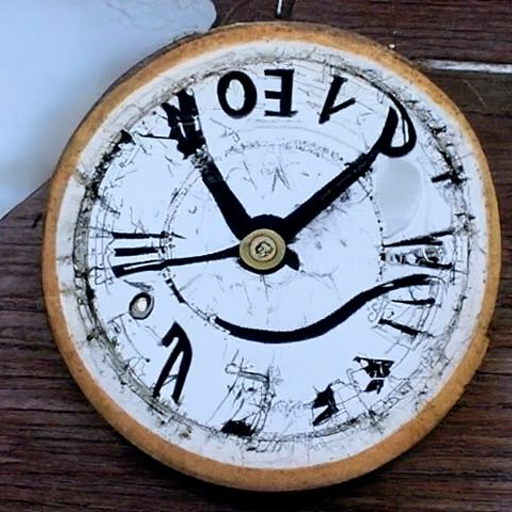} &
    \includegraphics[width=0.1\linewidth]{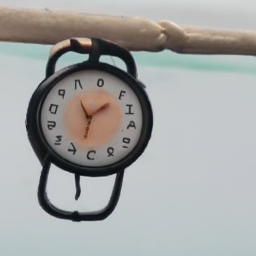} &
    \includegraphics[width=0.1\linewidth]{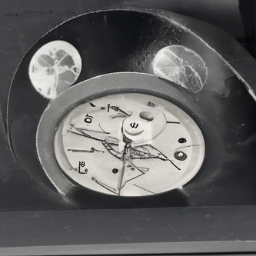} &
    \includegraphics[width=0.1\linewidth]{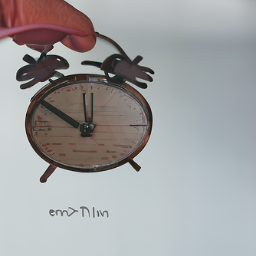} &
    \includegraphics[width=0.1\linewidth]{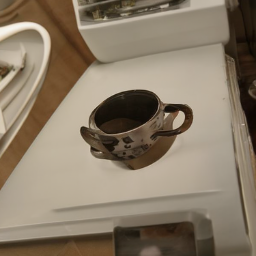} &
    \includegraphics[width=0.1\linewidth]{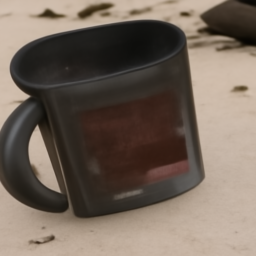} &
    \includegraphics[width=0.1\linewidth]{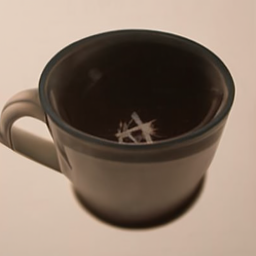} &
    \includegraphics[width=0.1\linewidth]{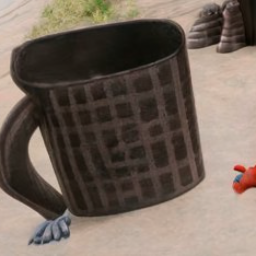} \\
    
    \raisebox{0.045\linewidth}{\footnotesize\begin{tabular}{c@{}c@{}c@{}c@{}} Ours \end{tabular}}  &
    \includegraphics[width=0.1\linewidth]{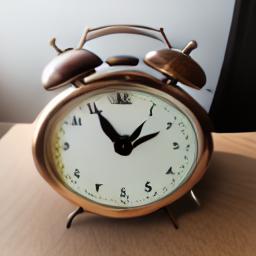} &
    \includegraphics[width=0.1\linewidth]{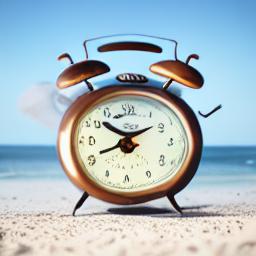} &
    \includegraphics[width=0.1\linewidth]{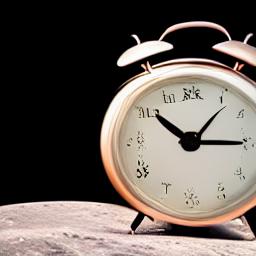} &
    \includegraphics[width=0.1\linewidth]{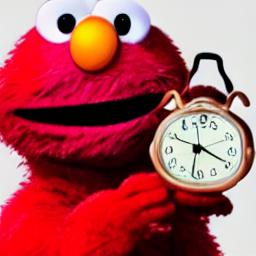} &
    \includegraphics[width=0.1\linewidth]{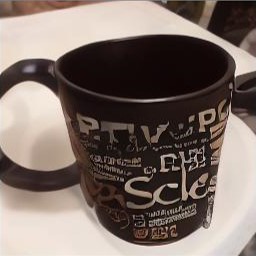} &
    \includegraphics[width=0.1\linewidth]{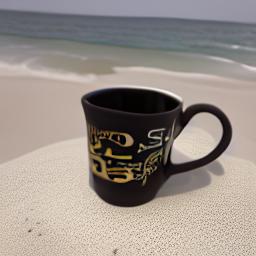} &
    \includegraphics[width=0.1\linewidth]{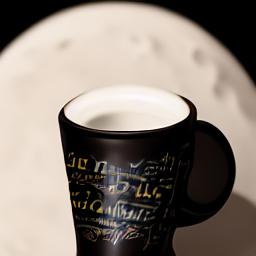} &
    \includegraphics[width=0.1\linewidth]{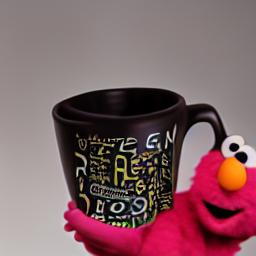} \\

    &
    {\tiny\begin{tabular}{c@{}c@{}c@{}c@{}} ``A photo of \pholdercolor" \end{tabular}} &
    {\tiny\begin{tabular}{c@{}c@{}c@{}c@{}} ``A photo of \pholdercolor{} \\ on the beach" \end{tabular}} &
    {\tiny\begin{tabular}{c@{}c@{}c@{}c@{}} ``A photo of \pholdercolor{} \\ on the moon" \end{tabular}} &
    {\tiny\begin{tabular}{c@{}c@{}c@{}c@{}} ``Elmo holding \\ a \pholdercolor" \end{tabular}\hskip 1pt} &
    {\tiny\begin{tabular}{c@{}c@{}c@{}c@{}} \hskip 1pt ``A photo of \pholdercolor" \end{tabular}} &
    {\tiny\begin{tabular}{c@{}c@{}c@{}c@{}} ``A photo of \pholdercolor{} \\ on the beach" \end{tabular}} &
    {\tiny\begin{tabular}{c@{}c@{}c@{}c@{}} ``A photo of \pholdercolor{} \\ on the moon" \end{tabular}} &
    {\tiny\begin{tabular}{c@{}c@{}c@{}c@{}} ``Elmo holding \\ a \pholdercolor" \end{tabular}} \\

    \end{tabular}
    
    \end{tabular}
    
    }
    \caption{Comparisons to alternative personalized creation approaches. Our model can more accurately preserve the subject, and can reason over both the novel embedding and the rest of the caption.}
    \label{fig:baseline_comp} 
\end{figure}

%% file: resources/figures/styles.tex
\begin{figure}[!hbt]
    \centering
    \setlength{\abovecaptionskip}{6.5pt}
    \setlength{\belowcaptionskip}{-3.5pt}
    \setlength{\tabcolsep}{0.55pt}
    \renewcommand{\arraystretch}{1.0}
    {\scriptsize
    \begin{tabular}{c@{\hskip 5pt} c@{\hskip 5pt} c c c c}
    
        \begin{tabular}{c c}
            \includegraphics[width=0.09\linewidth,height=0.09\linewidth]{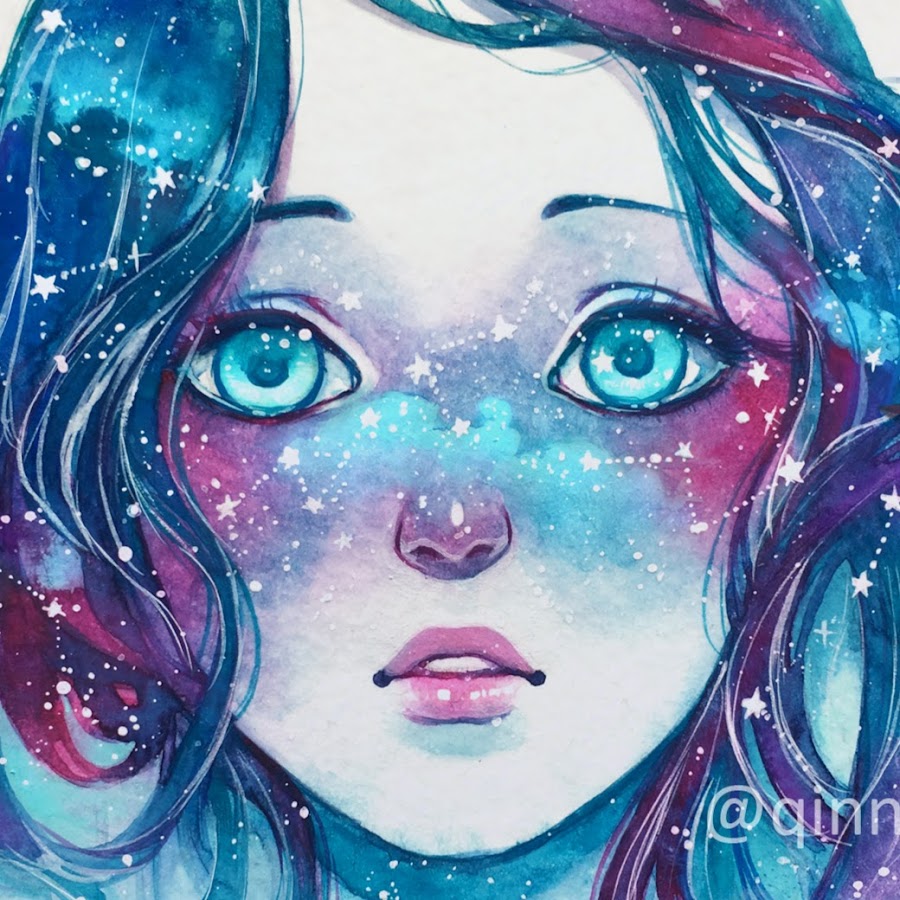} & 
            \includegraphics[width=0.09\linewidth,height=0.09\linewidth]{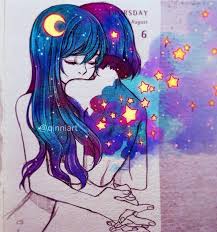} \\
            \includegraphics[width=0.09\linewidth,height=0.09\linewidth]{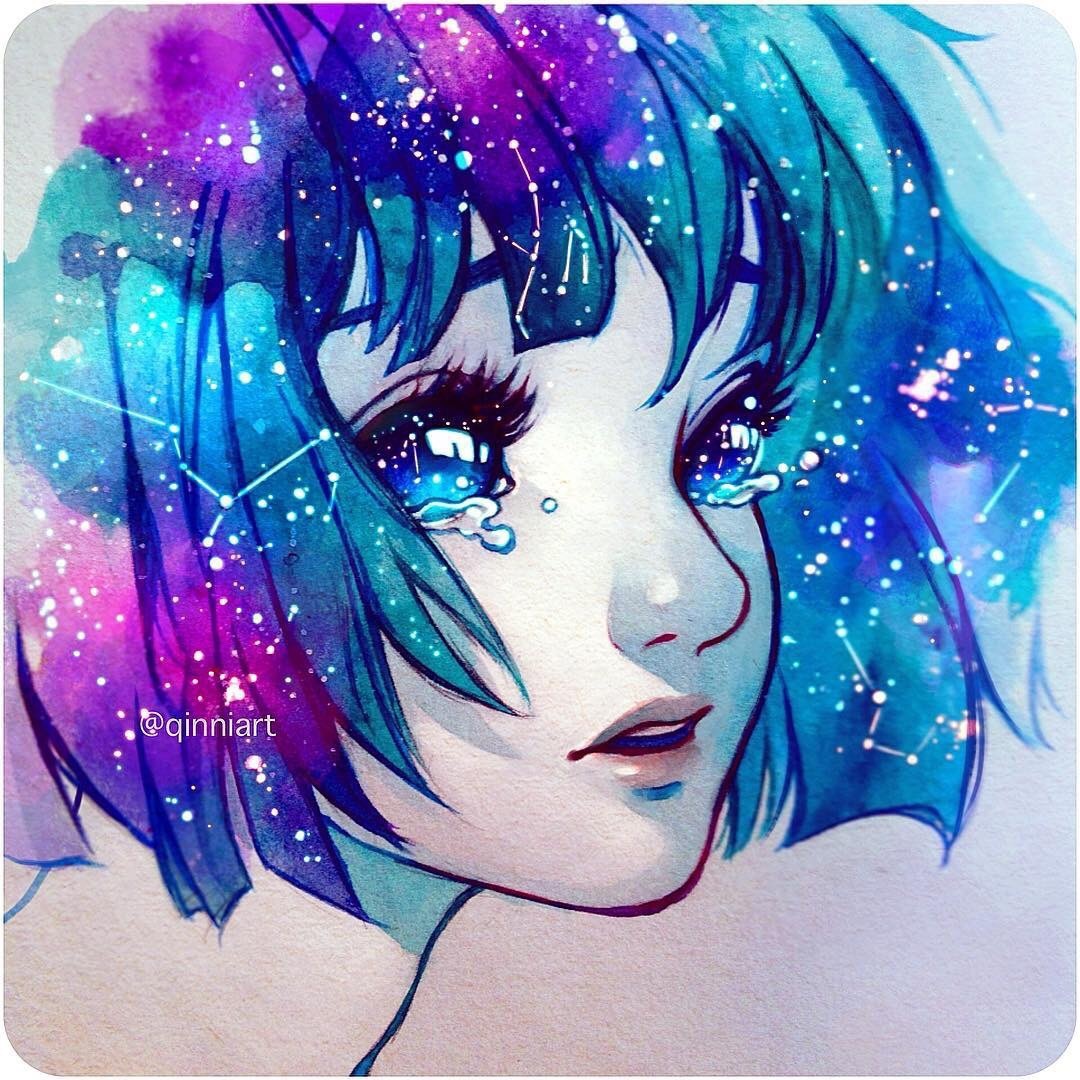} & 
            \includegraphics[width=0.09\linewidth,height=0.09\linewidth]{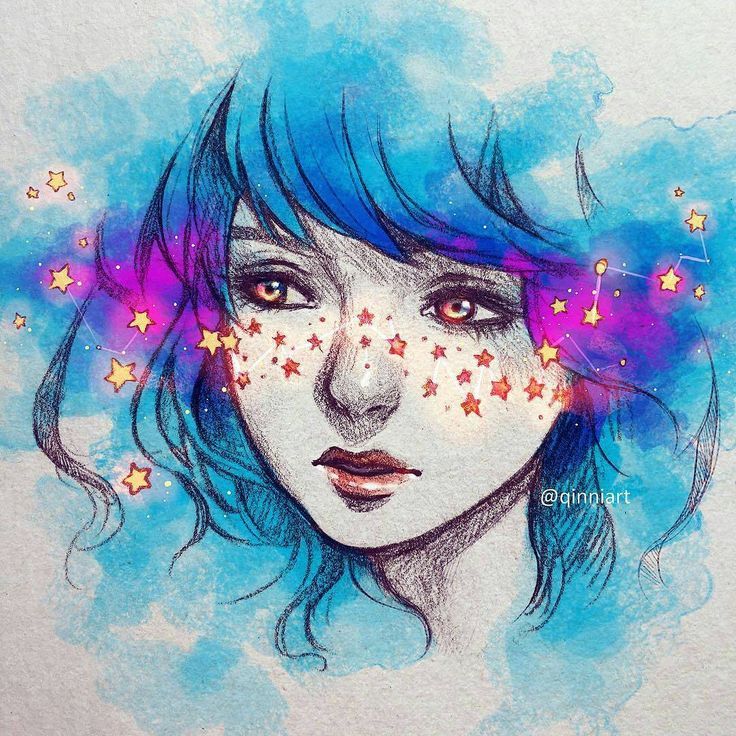}
        \end{tabular}
        
        &
        $\rightarrow$
        &
        \begin{tabular}{c}
        \includegraphics[width=0.184\linewidth]{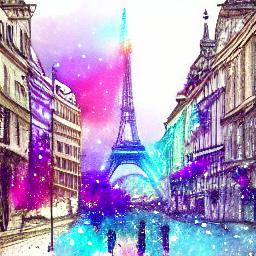} 
        \end{tabular} &
        \begin{tabular}{c}
        \includegraphics[width=0.184\linewidth]{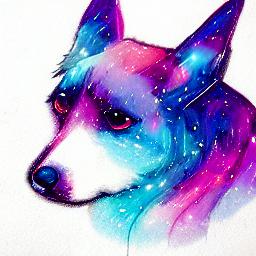} 
        \end{tabular} &
        \begin{tabular}{c}
        \includegraphics[width=0.184\linewidth]{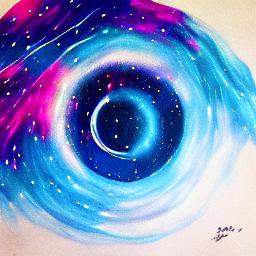}
        \end{tabular} & 
        \begin{tabular}{c}
        \includegraphics[width=0.184\linewidth]{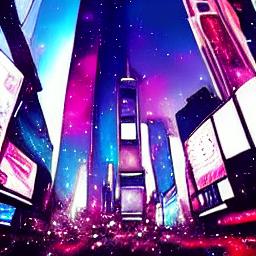}
        \end{tabular} \\

        \begin{tabular}{c c}
            \includegraphics[width=0.09\linewidth,height=0.09\linewidth]{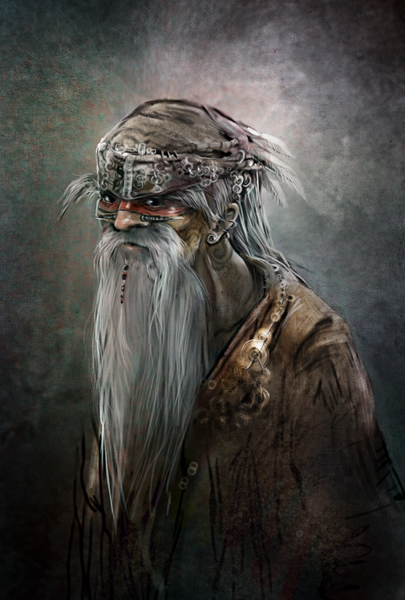} & 
            \includegraphics[width=0.09\linewidth,height=0.09\linewidth]{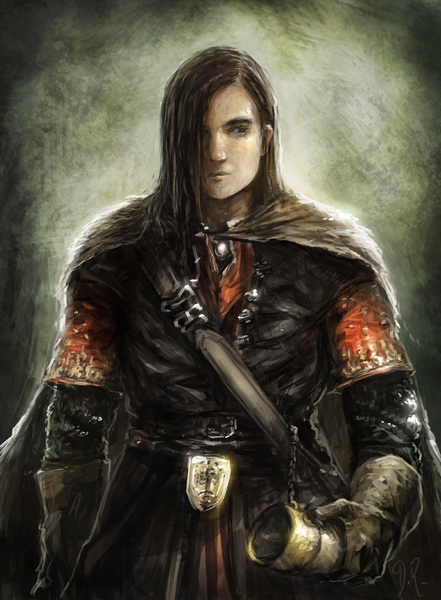} \\
            \multicolumn{2}{c}{\includegraphics[width=0.09\linewidth,height=0.09\linewidth]{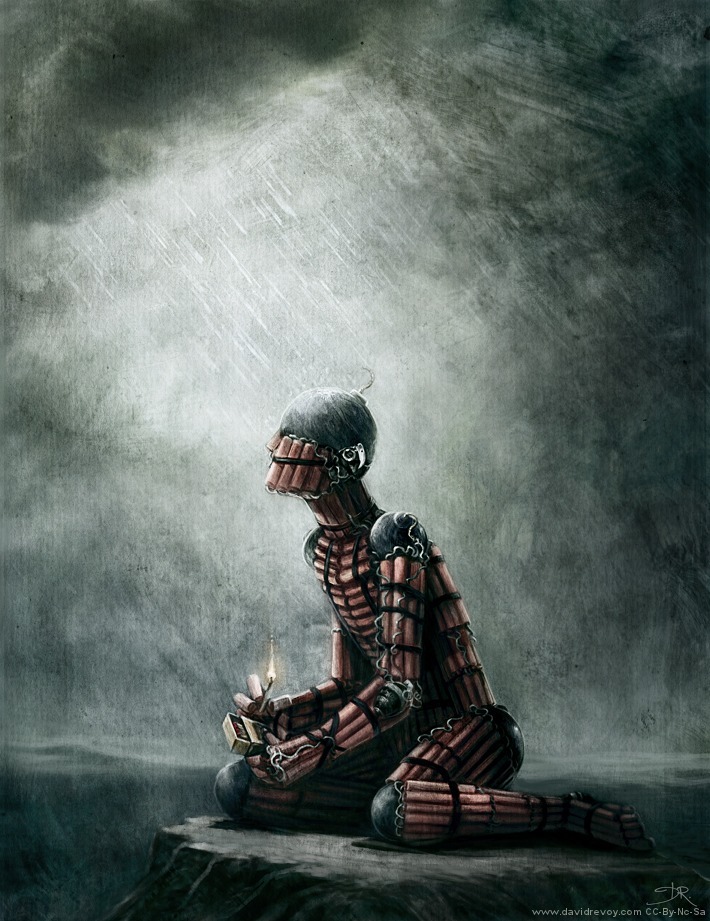}}
        \end{tabular}
        
        &
        $\rightarrow$
        &
        \begin{tabular}{c}
        \includegraphics[width=0.184\linewidth]{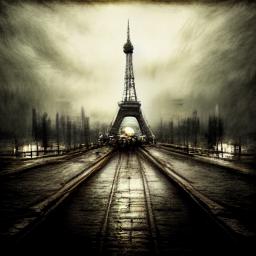}
        \end{tabular} &
        \begin{tabular}{c}
        \includegraphics[width=0.184\linewidth]{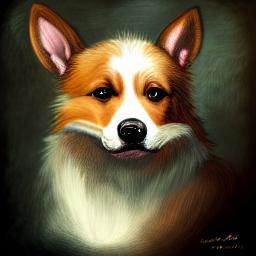} 
        \end{tabular} &
        \begin{tabular}{c}
        \includegraphics[width=0.184\linewidth]{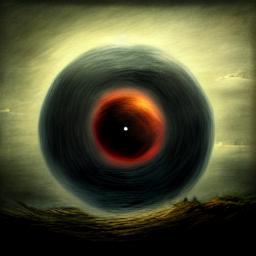} 
        \end{tabular} &
        \begin{tabular}{c}
        \includegraphics[width=0.184\linewidth]{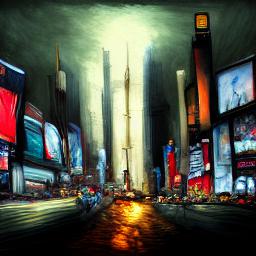}
        \end{tabular} \\
        
        {\tiny Input samples} & & {\begin{tabular}{c@{}c@{}c@{}c@{}} ``The streets of Paris \\ in the style of \pholdercolor" \end{tabular}} & {\begin{tabular}{c@{}c@{}c@{}c@{}} ``Adorable corgi \\ in the style of \pholdercolor" \end{tabular}} & {\begin{tabular}{c@{}c@{}c@{}c@{}} ``Painting of a black hole \\ in the style of \pholdercolor" \end{tabular}} & {\begin{tabular}{c@{}c@{}c@{}c@{}} ``Times square \\ in the style of \pholdercolor" \end{tabular}} \\ \\

    \end{tabular}}
    \caption{The textual-embedding space can represent more abstract concepts, including styles. This allows us to discover words which can be used for style-guided generation. Image credits: \href{https://www.deviantart.com/qinni}{@QinniArt} (top), \href{https://commons.wikimedia.org/wiki/User:Deevad}{@David Revoy} (bottom). Image reproduction authorized for non-commercial use only.
    }
    \label{fig:styles} 
\end{figure}

%% file: resources/figures/compositions.tex
\begin{figure}[!hbt]
    \centering
    \setlength{\abovecaptionskip}{6.5pt}
    \setlength{\belowcaptionskip}{-3.5pt}
    \setlength{\tabcolsep}{0.55pt}
    \renewcommand{\arraystretch}{1.0}
    
    \begin{tabular}{c}
    
    \begin{tabular}{c@{\hskip 10pt}c@{\hskip 10pt}c@{\hskip 10pt}c}
    
    \begin{tabular}{c c}
    \includegraphics[width=0.1\linewidth,height=0.1\linewidth]{resources/images/training_sets/qinni/1.jpg} & \includegraphics[width=0.1\linewidth,height=0.1\linewidth]{resources/images/training_sets/qinni/2.jpg} \\[-2pt]
    \includegraphics[width=0.1\linewidth,height=0.1\linewidth]{resources/images/training_sets/qinni/5.jpg} & \includegraphics[width=0.1\linewidth,height=0.1\linewidth]{resources/images/training_sets/qinni/4.jpg} \\[-2pt]
    \multicolumn{2}{c}{{\color[HTML]{EF5675}$S_{style}$}}
    \end{tabular} &
    
    \begin{tabular}{c c}
    \includegraphics[width=0.1\linewidth,height=0.1\linewidth]{resources/images/training_sets/clock/1.jpeg} &
    \includegraphics[width=0.1\linewidth,height=0.1\linewidth]{resources/images/training_sets/clock/3.jpeg} \\[-2pt]
    \includegraphics[width=0.1\linewidth,height=0.1\linewidth]{resources/images/training_sets/clock/4.jpeg} &
    \includegraphics[width=0.1\linewidth,height=0.1\linewidth]{resources/images/training_sets/clock/5.jpeg} \\[-2pt]
    \multicolumn{2}{c}{{\color[HTML]{7A5195}$S_{clock}$}}
    \end{tabular} &
    
    \begin{tabular}{c c c c}
    \includegraphics[width=0.1\linewidth,height=0.1\linewidth]{resources/images/training_sets/rainbow_cat/2.jpeg} &
    \includegraphics[width=0.1\linewidth,height=0.1\linewidth]{resources/images/training_sets/rainbow_cat/3.jpeg} \\[-2pt]
    \multicolumn{2}{c}{\includegraphics[width=0.1\linewidth,height=0.1\linewidth]{resources/images/training_sets/rainbow_cat/6.jpeg}} \\[-2pt]
    \multicolumn{2}{c}{{\color[HTML]{FFA600}$S_{cat}$}}
    \end{tabular} &
    
    \begin{tabular}{c c c}
    \includegraphics[width=0.1\linewidth,height=0.1\linewidth]{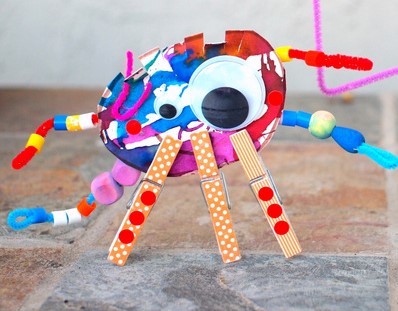} & \includegraphics[width=0.1\linewidth,height=0.1\linewidth]{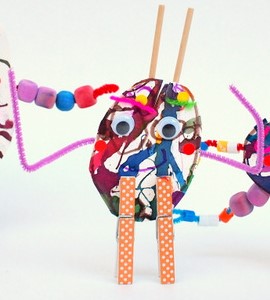} \\[-2pt]
    \multicolumn{2}{c}{\includegraphics[width=0.1\linewidth,height=0.1\linewidth]{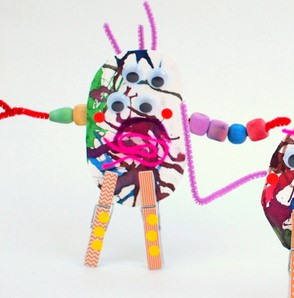}} \\[-2pt]
    \multicolumn{2}{c}{{\color[HTML]{003F5C}$S_{craft}$}}
    \end{tabular}
    
    \end{tabular} \\[-2pt] \\[-2pt]
    
    \begin{tabular}{c c c c c c}
    \includegraphics[width=0.16\linewidth,height=0.16\linewidth]{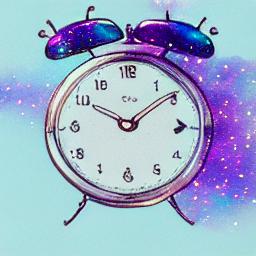} &
    \includegraphics[width=0.16\linewidth,height=0.16\linewidth]{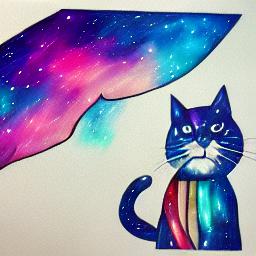} &
    \includegraphics[width=0.16\linewidth,height=0.16\linewidth]{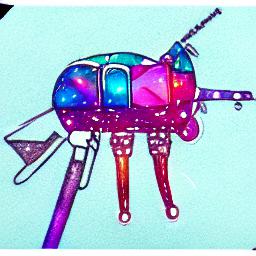} &
    \includegraphics[width=0.16\linewidth,height=0.16\linewidth]{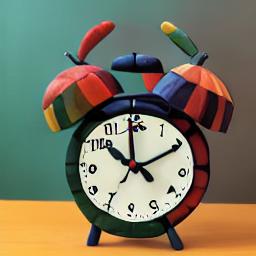} &
    \includegraphics[width=0.16\linewidth,height=0.16\linewidth]{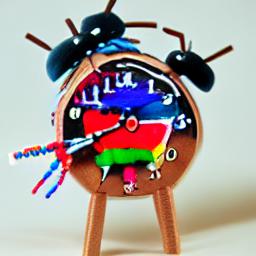} &
    \includegraphics[width=0.16\linewidth,height=0.16\linewidth]{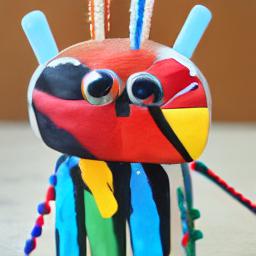} \\
    
    {\scriptsize\begin{tabular}{c@{}c@{}c@{}c@{}} ``Photo of {\color[HTML]{7A5195}$S_{clock}$} \\ in the style of {\color[HTML]{EF5675}$S_{style}$}" \end{tabular}} &
    {\scriptsize\begin{tabular}{c@{}c@{}c@{}c@{}} ``Photo of {\color[HTML]{FFA600}$S_{cat}$} \\ in the style of {\color[HTML]{EF5675}$S_{style}$}" \end{tabular}} &
    {\scriptsize\begin{tabular}{c@{}c@{}c@{}c@{}} ``Photo of {\color[HTML]{003F5C}$S_{craft}$} \\ in the style of {\color[HTML]{EF5675}$S_{style}$}" \end{tabular}} &
    {\scriptsize\begin{tabular}{c@{}c@{}c@{}c@{}} ``Photo of {\color[HTML]{7A5195}$S_{clock}$} \\ in the style of {\color[HTML]{FFA600}$S_{cat}$}" \end{tabular}} &
    {\scriptsize\begin{tabular}{c@{}c@{}c@{}c@{}} ``Photo of {\color[HTML]{7A5195}$S_{clock}$} \\ in the style of {\color[HTML]{003F5C}$S_{craft}$}" \end{tabular}} &
    {\scriptsize\begin{tabular}{c@{}c@{}c@{}c@{}} ``Photo of {\color[HTML]{FFA600}$S_{cat}$} \\ in the style of {\color[HTML]{003F5C}$S_{craft}$}" \end{tabular}}
    \end{tabular}
    
    \end{tabular}
    
    \caption{Compositional generation using two learned pseudo-words. The model is able to combine the semantics of two concepts when using a prompt that combines them both. It is limited in its ability to reason over more complex relational prompts, such as placing two concepts side-by-side. Image credits: \href{https://www.deviantart.com/qinni}{@QinniArt} (left), \href{https://www.pinkstripeysocks.com/p/about.html}{@Leslie Manlapig} (right). Reproductions authorized for non-commercial / non-print use respectively.}
    \label{fig:compositions}
    
\end{figure}

%% file: resources/figures/bias.tex
\begin{figure}[t]
    \centering
    \begin{tabular}{c@{\hskip 5pt}c}
         \includegraphics[width=0.47\linewidth]{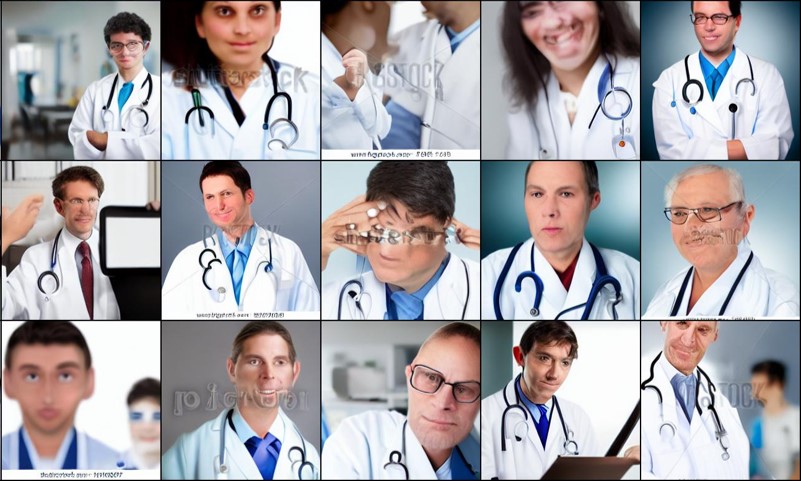} & \includegraphics[width=0.47\linewidth]{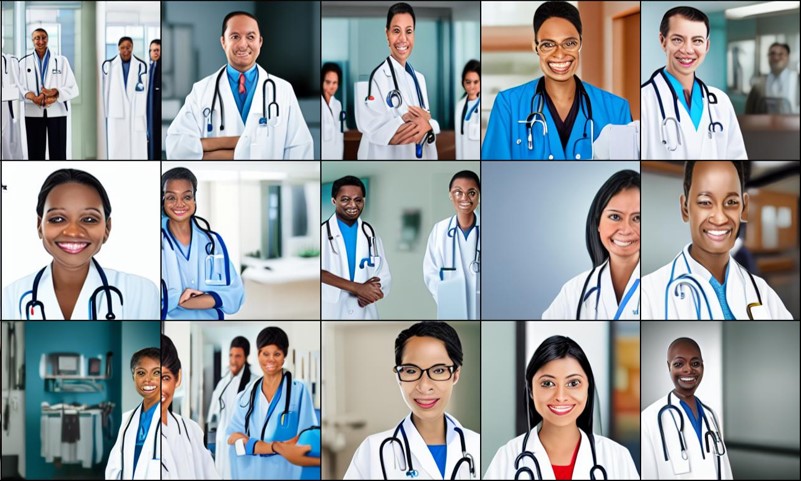} \\
         ``A stock photo of a doctor" (Base model) & ``A photo of \pholdercolor" (Ours)
    \end{tabular}
    \caption{Bias Reduction. Uncurated samples synthesized with pretrained biased embeddings (left) and our \textit{de}biased embeddings (right). Our approach can be used to reduce bias by learning new pseudo-words for known concepts. These can be optimized using small datasets, which can be carefully curated for diversity.}
    \label{fig:bias}
\end{figure}

%% file: resources/figures/belnded_diffusion.tex
\begin{figure}[!hbt]
    \centering
    \setlength{\tabcolsep}{0.55pt}
    \renewcommand{\arraystretch}{1.0}
    {
    \fontsize{8pt}{8pt}
    
    \begin{tabular}{c@{\hskip 5pt} c c c c@{\hskip 10pt}  c c c c}
    \begin{tabular}{c}
    \raisebox{0.015\linewidth}{\footnotesize\begin{tabular}{c@{}c@{}c@{}c@{}} Input \\ Samples \end{tabular}} \end{tabular} & 
    \multicolumn{4}{c}{
    \begin{tabular}{c c c c}
    \includegraphics[width=0.1\linewidth,height=0.1\linewidth]{resources/images/training_sets/headless_statue/1.jpeg} & 
    \includegraphics[width=0.1\linewidth,height=0.1\linewidth]{resources/images/training_sets/headless_statue/2.jpeg} & 
    \includegraphics[width=0.1\linewidth,height=0.1\linewidth]{resources/images/training_sets/headless_statue/3.jpeg} & 
    \includegraphics[width=0.1\linewidth,height=0.1\linewidth]{resources/images/training_sets/headless_statue/4.jpeg} 
    \end{tabular} } &
    \multicolumn{4}{c}{
    \begin{tabular}{c c c}
    \includegraphics[width=0.1\linewidth,height=0.1\linewidth]{resources/images/training_sets/rainbow_cat/2.jpeg} & 
    \includegraphics[width=0.1\linewidth,height=0.1\linewidth]{resources/images/training_sets/rainbow_cat/3.jpeg} &
    \includegraphics[width=0.1\linewidth,height=0.1\linewidth]{resources/images/training_sets/rainbow_cat/6.jpeg}
    \end{tabular}
    }
    \\

    \raisebox{0.09\linewidth}{\footnotesize\begin{tabular}{c@{}c@{}c@{}c@{}} Target Image \\ With Mask \end{tabular}} & 
    \multicolumn{2}{c}{
    \includegraphics[width=0.2\linewidth]{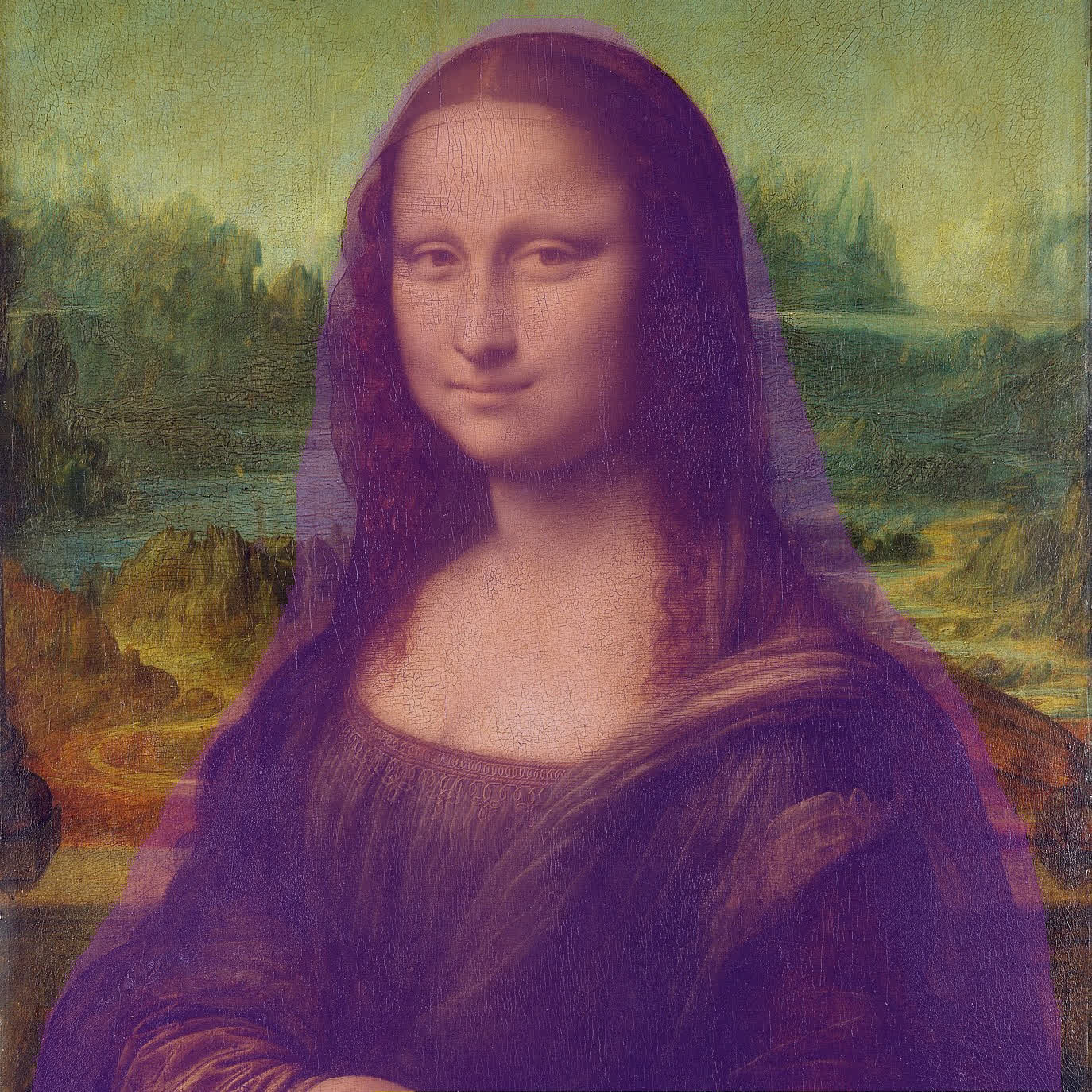}
    } &
    \multicolumn{2}{c}{
    \includegraphics[width=0.2\linewidth]{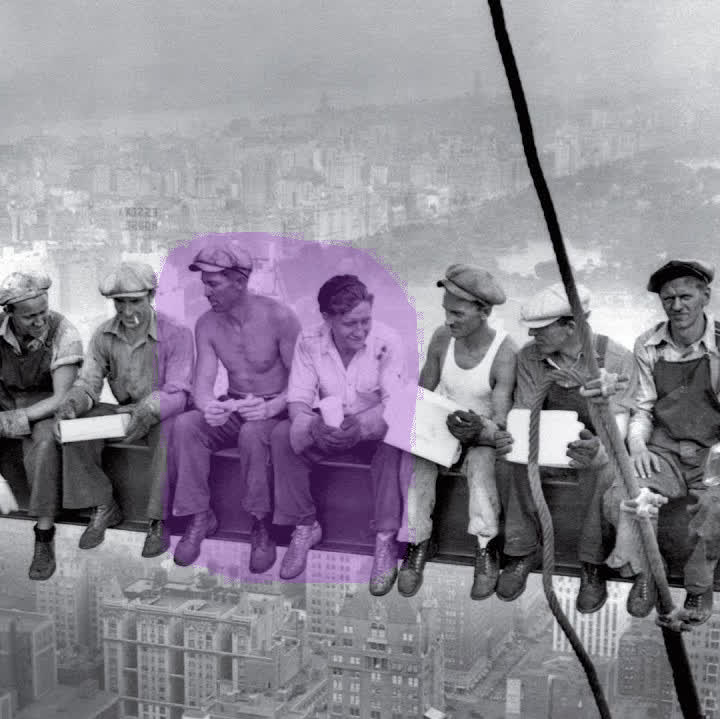}
    } &
    \multicolumn{2}{c}{
    \includegraphics[width=0.2\linewidth]{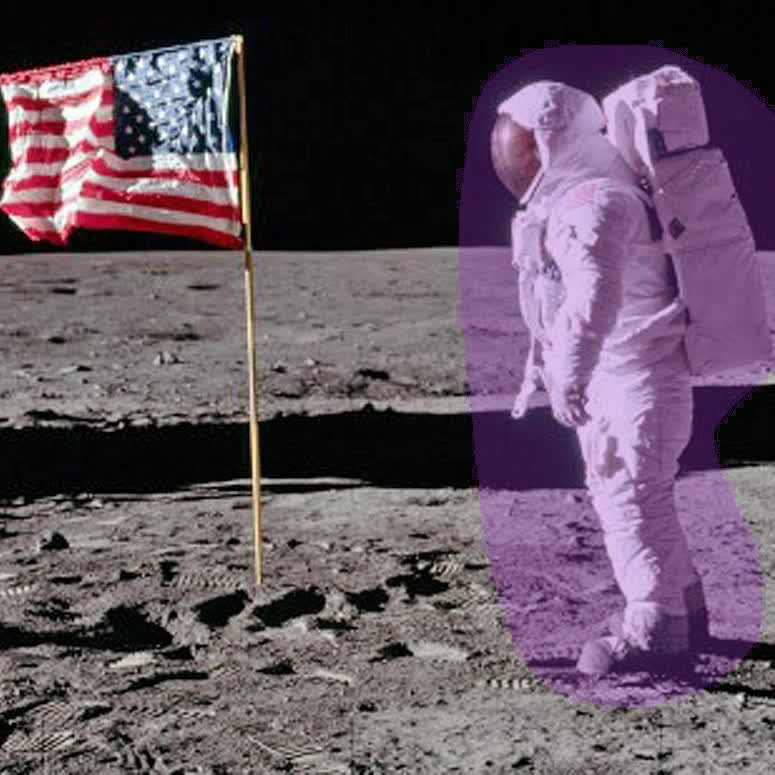}
    } &
    \multicolumn{2}{c}{
    \includegraphics[width=0.2\linewidth]{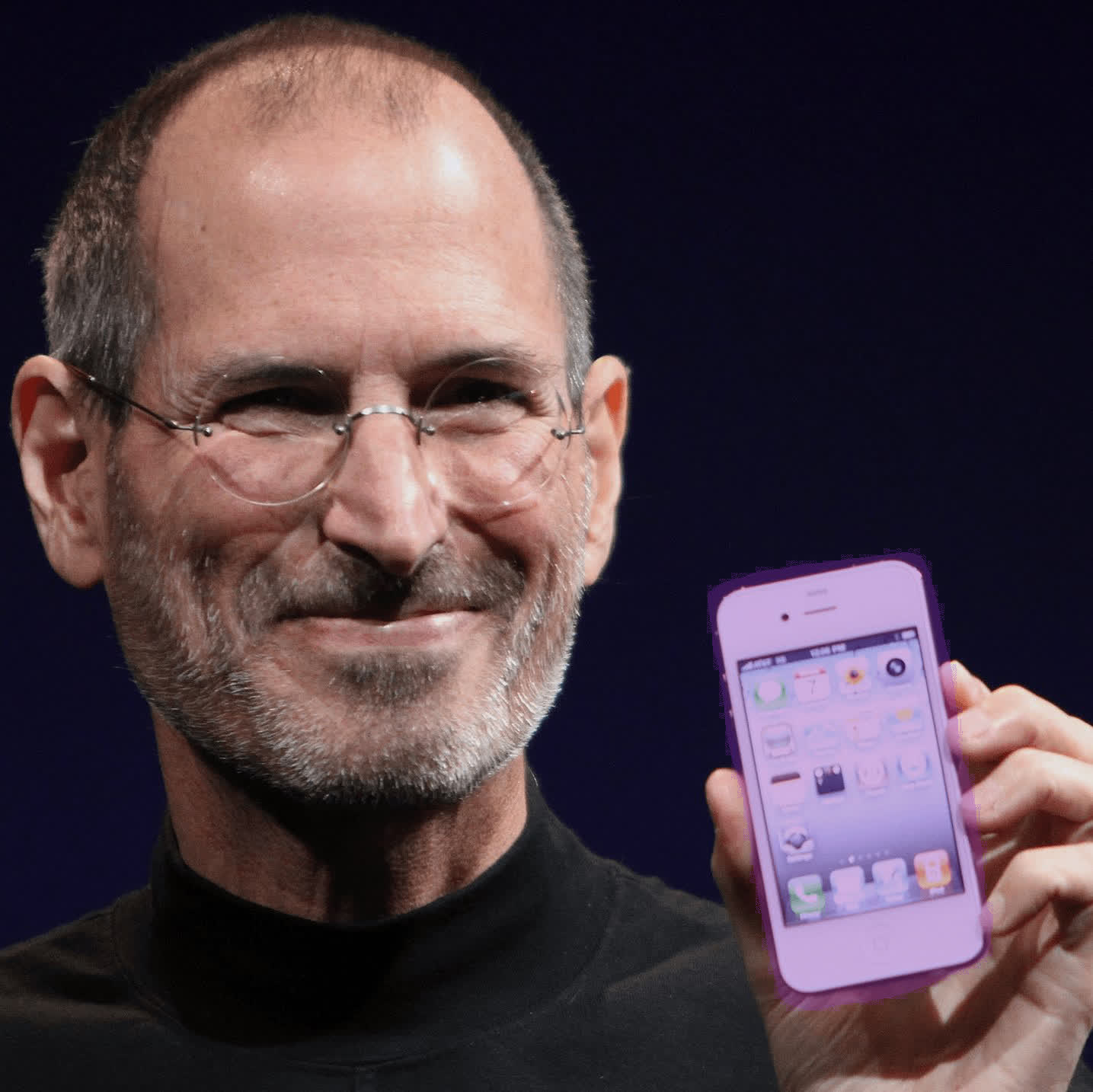}
    }
    \\

    \raisebox{0.09\linewidth}{\footnotesize\begin{tabular}{c@{}c@{}c@{}c@{}} Output \\ Image \end{tabular}} & 
    \multicolumn{2}{c}{
    \includegraphics[width=0.2\linewidth]{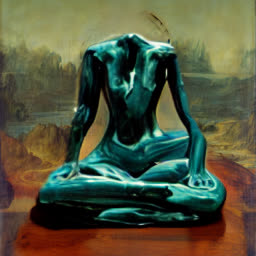}
    } &
    \multicolumn{2}{c}{
    \includegraphics[width=0.2\linewidth]{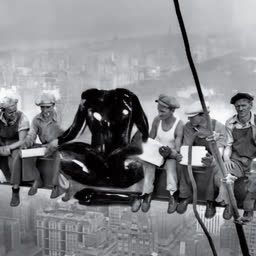}
    } &
    \multicolumn{2}{c}{
    \includegraphics[width=0.2\linewidth]{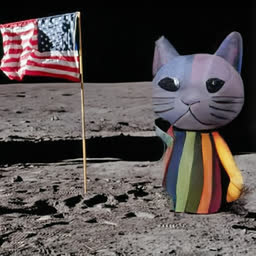}
    } &
    \multicolumn{2}{c}{
    \includegraphics[width=0.2\linewidth]{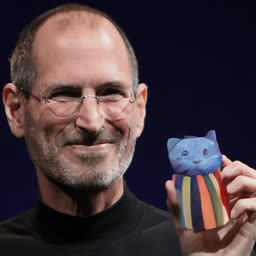}
    }
    \\

    & 
    \multicolumn{2}{c}{
    {\footnotesize\begin{tabular}{c@{}c@{}c@{}c@{}} ``An oil painting \\ of \pholdercolor" \end{tabular}}
    } &
    \multicolumn{2}{c}{
    {\footnotesize\begin{tabular}{c@{}c@{}c@{}c@{}} ``A black and white \\ photo of \pholdercolor" \end{tabular}}
    } &
    \multicolumn{2}{c}{
    {\footnotesize\begin{tabular}{c@{}c@{}c@{}c@{}} ``A \pholdercolor" \end{tabular}}
    } & 
    \multicolumn{2}{c}{
    {\footnotesize\begin{tabular}{c@{}c@{}c@{}c@{}} ``A \pholdercolor" \end{tabular}}
    } \\
    
    \end{tabular}}
    \caption{Our words can be used with downstream models that build on LDM. Here, we perform localized image editing using Blended Latent Diffusion~\citep{avrahami2022blendedlatent}}
    \label{fig:blended} 
\end{figure}

%% file: analysis.tex
\section{Quantitative analysis}
\label{sec:analysis}

Inversion into an uncharted latent space provides us with a wide range of possible design choices. Here, we examine these choices in light of the GAN inversion literature and discover that many core premises (such as a distortion-editability tradeoff~\citep{tov2021designing,zhu2020improved}) also exist in the textual embedding space. However, our analysis reveals that many of the solutions typically used in GAN inversion fail to generalize to this space, and are often unhelpful or actively harmful.

\subsection{Evaluation metrics}

To analyze the quality of latent space embeddings, we consider two fronts: reconstruction and editability.
First, we wish to gauge our ability to replicate the target concept. As our method produces variations on the concept and not a specific image, we measure similarity by considering semantic CLIP-space distances. Specifically, for each concept, we generate a $64$ of images using the prompt: ``A photo of $S_*$". Our reconstruction score is then the average pair-wise CLIP-space cosine-similarity between the generated images and the images of the concept-specific training set.

Second, we want to evaluate our ability to modify the concepts using textual prompts. To this end, we produce a set of images using prompts of varying difficulty and settings. These range from background modifications (``A photo of $S_*$ on the moon"), to style changes (``An oil painting of $S_*$"), and a compositional prompt (``Elmo holding a $S_*$"). 

For each prompt, we synthesize $64$ samples using $50$ DDIM steps, calculate the average CLIP-space embedding of the samples, and compute their cosine similarity with the CLIP-space embedding of the textual prompts, where we omit the placeholder $S_*$ (\ie ``A photo of on the moon"). Here, a higher score indicates better editing capability and more faithfulness to the prompt itself. Note that our method does not involve the direct optimization of the CLIP-based objective score and, as such, is not sensitive to the adversarial scoring flaws outlined by~\citet{nichol2021glide}.

\subsection{Evaluation setups}

We evaluate the embedding space using a set of experimental setups inspired by GAN inversion:

\paragraph{Extended latent spaces}
Following~\citet{abdal2019image2stylegan}, we consider an extended, multi-vector latent space. In this space, $S_*$ is embedded into multiple learned embeddings, an approach that is equivalent to describing the concept through multiple learned pseudo-words. We consider an extension to two and three pseudo-words (denoted $2-word$ and $3-word$, respectively). This setup aims to alleviate the potential bottleneck of a single embedding vector to enable more accurate reconstructions.

\paragraph{Progressive extensions}
We follow~\citet{tov2021designing} and consider a progressive multi-vector setup. Here, we begin training with a single embedding vector, introduce a second vector following $2,000$ training steps, and a third vector after $4,000$ steps. In this scenario, we expect the network to focus on the core details first, and then leverage the additional pseudo-words to capture finer details.

\paragraph{Regularization}
\citet{tov2021designing} observed that latent codes in the space of a GAN have increased editability when they lie closer to the code distribution which was observed during training. Here, we investigate a similar scenario by introducing a regularization term that aims to keep the learned embedding close to existing words. In practice, we minimize the L2 distance of the learned embedding to the embedding of a coarse descriptor of the object (\eg ``sculpture" and ``cat" for the images in \Cref{fig:teaser}).

\paragraph{Per-image tokens}
Moving beyond GAN-based approaches, we investigate a novel scheme where we introduce unique, per-image tokens into our inversion approach. Let $\{x_i\}_{i=1}^n$ be the set of input images. Rather than optimizing a single word vector shared across all images, we introduce both a universal placeholder, $S_*$, and an additional placeholder unique to each image, $\{S_i\}_{i=1}^n$, associated with a unique embedding $v_i$. We then compose sentences of the form ``A photo of $S_*$ with $S_i$", where every image is matched to sentences containing its own, unique string. We jointly optimize over both $S_*$ and $\{S_i\}_{i=1}^n$, using \Cref{eq:v_opt}. The intuition here is that the model should prefer to encode the shared information (\ie the concept) in the shared code $S_*$ while relegating per-image details such as the background to $S_i$. 

\paragraph{Human captions}
In addition to the learned-embedding setups, we compare to human-level performance using the captions outlined in \Cref{sec:image_variations}. Here, we simply replace the placeholder strings $S_*$ with the human captions, using both the short and long-caption setups. 

\paragraph{Reference setups}
To provide intuition for the scale of the results, we add two reference baselines.
First, we consider the expected behavior from a model that always produces copies of the training set, regardless of the prompt. For that, we simply use the training set itself as the ``generated sample".
Second, we consider a model that always aligns with the text prompt but ignores the personalized concept. We do so by synthesizing images using the evaluation prompts but without the pseudo-word.
We denote these setups as ``Image Only" and ``Prompt Only", respectively.

\paragraph{Textual-Inversion}
Finally, we consider our own setup, as outlined in \Cref{sec:method}. We further evaluate our model with an increased learning rate ($2e-2$, ``High-LR") and a decreased learning rate ($1e-4$, ``Low-LR").

\paragraph{Additional setups}
In the supplementary, we consider two additional setups for inversion: a pivotal tuning approach~\citep{roich2021pivotal,bau2020semantic}, where the model itself is optimized to improve reconstruction, and DALLE-2~\citep{ramesh2022hierarchical}'s bipartite inversion process. We further analyze the effect of the image-set size on reconstruction and editability.

\subsection{Results}

\input{resources/figures/quantitative_evaluation}

Our evaluation results are summarized in \Cref{fig:quant_eval}(a). 
We highlight four observations of particular interest:
First, the semantic reconstruction quality of our method and many of the baselines is comparable to simply sampling random images from the training set. 
Second, the single-word method achieves comparable reconstruction quality, and considerably improved editability over all multi-word baselines. These points outline the impressive flexibility of the textual embedding space, showing that it can serve to capture new concepts with a high degree of accuracy while using only a single pseudo-word.

Third, we observe that our baselines outline a distortion-editability trade-off curve, where embeddings that lie closer to the true word distribution (\eg due to regularization, fewer pseudo-words, or a lower learning rate) can be more easily modified, but fail to capture the details of the target. In contrast, deviating far from the word distribution enables improved reconstruction at the cost of severely diminished editing capabilities. Notably, our single-embedding model can be moved along this curve by simply changing the learning rate, offering a user a degree of control over this trade-off. 

As a fourth observation, we note that the use of human descriptions for the concepts not only fails to capture their likeness, but also leads to diminished editability. We hypothesize that this is tied to the selective-similarity property outlined in \citet{paiss2022no}, where vision-and-language models tend to focus on a subset of the semantically meaningful tokens. By using long captions, we increase the chance of the model ignoring our desired setting, focusing only on the object description itself. Our model, meanwhile, uses only a single token and thus minimizes this risk.

Finally, we note that while our reconstruction scores are on par with those of randomly sampled, real images, these results should be taken with a grain of salt. Our metrics compare \textit{semantic} similarity using CLIP, which is less sensitive to shape-preservation. On this front, there remains more to be done.

\subsection{Human evaluations}

We further evaluate our models using a user study. Here, we created two questionnaires. In the first, users were provided with four images from a concept's training set, and asked to rank the results produced by five models according to their similarity to these images. In the second questionnaire, users were provided with a text describing an image context (``A photo on the beach") and asked to rank the results produced by the same models according to their similarity to the text.

We used the same target concepts and prompts as the CLIP-based evaluation and collected a total of $600$ responses to each questionnaire, for a total of $1,200$ responses. Results are shown in \Cref{fig:quant_eval}(b).

The user-study results align with the CLIP-based metrics and demonstrate a similar reconstruction-editability tradeoff. Moreover, they outline the same limitations of human-based captioning when attempting to reproduce a concept, as well as when editing it.

%% file: resources/figures/quantitative_evaluation.tex
\begin{figure}[t]
    \centering
    
    \begin{tabular}{c c}
    \includegraphics[width=0.49\textwidth]{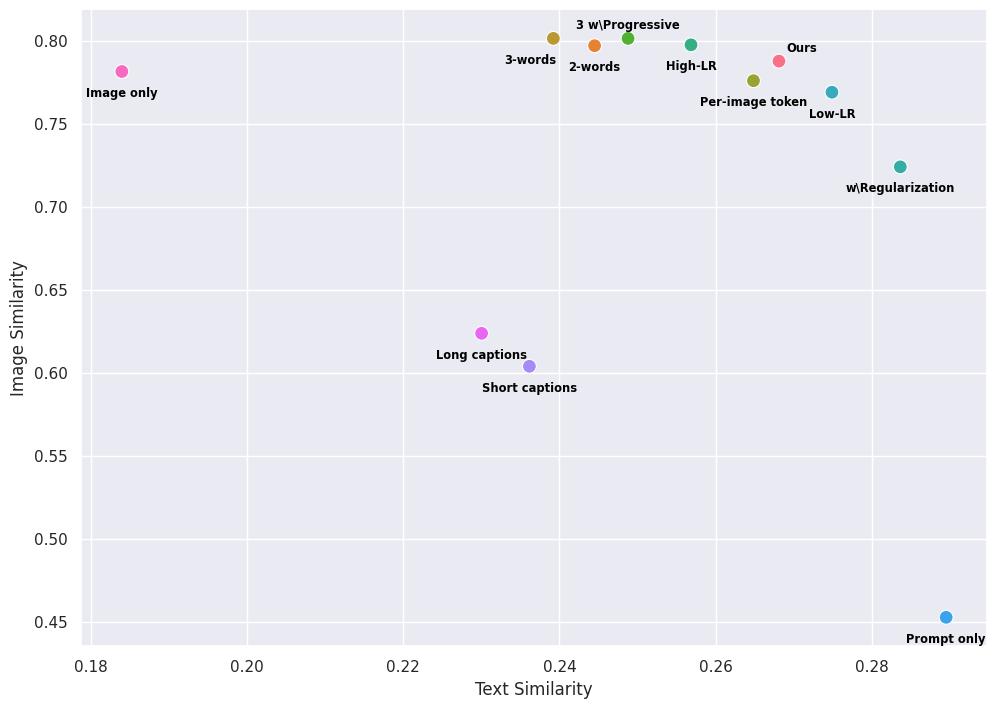} &
    \includegraphics[width=0.49\textwidth]{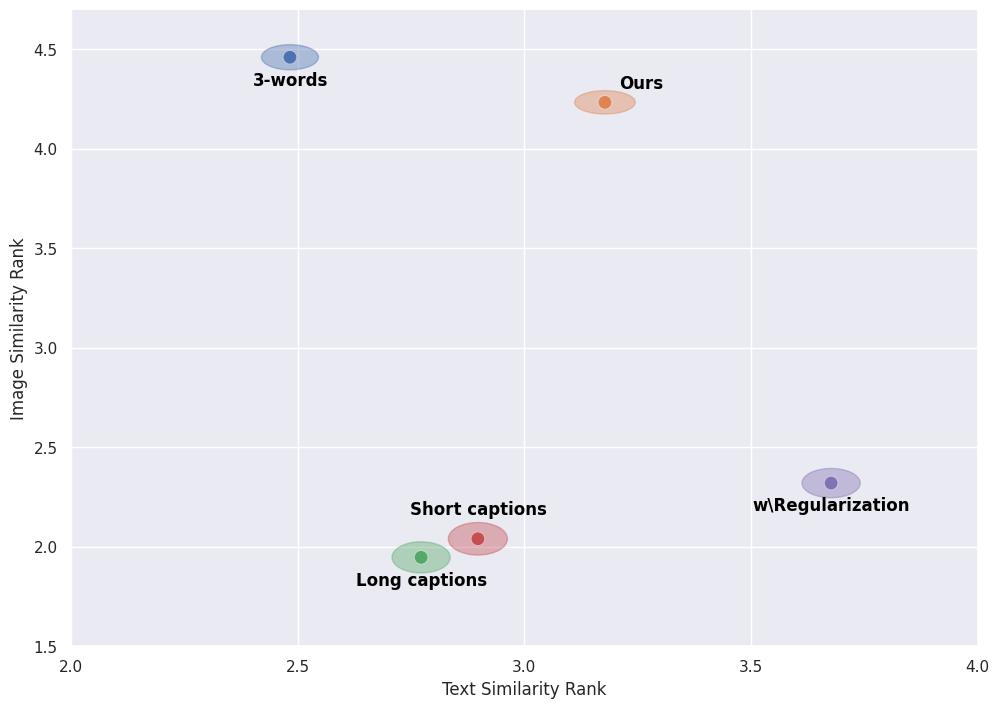} \\
    
    (a) & (b)
    \end{tabular}
    \caption{Qualitative evaluation results. (a) CLIP-based evaluations. The single-word model (ours) represents an appealing point on the distortion-editability curve, and can be moved along it by changing the learning rate. (b) User study results. These results portray a similar distortion-editability curve, and demonstrate that the CLIP-based results align with human preference. User study error bars are 95\% confidence intervals.}
    \label{fig:quant_eval}
\end{figure}

%% file: future.tex
\section{Limitations}
While our method offers increased freedom, it may still struggle with learning precise shapes, instead incorporating the ``semantic'' essence of a concept. For artistic creations, this is often enough. In the future, we hope to achieve better control over the accuracy of the reconstructed concepts, enabling users to leverage our method for tasks that require greater precision.

Another limitation of our approach is in the lengthy optimization times. Using our setup, learning a single concept requires roughly two hours. These times could likely be shortened by training an encoder to directly map a set of images to their textual embedding. We aim to explore this line of work in the future.

\section{Social impact}
Text-to-image models can be used to generate misleading content and promote disinformation. Personalized creation could allow a user to forge more convincing images of non-public individuals. However, our model does not currently preserve identity to the extent where this is a concern.

These models are further susceptible to the biases found in the training data. Examples include gender biases when portraying ``doctors'' and ``nurses'', racial biases when requesting images of scientists, and more subtle biases such as an over-representation of heterosexual couples and western traditions when prompting for a ``wedding''~\citep{mishkin2022risks}. As we build on such models, our own work may similarly exhibit biases. However, as demonstrated in \Cref{fig:bias}, our ability to more precisely describe specific concepts can also serve as a means for reducing these biases.

Finally, the ability to learn artistic styles may be misused for copyright infringement. Rather than paying an artist for their work, a user could train on their images without consent, and produce images in a similar style. While generated artwork is still easy to identify, in the future such infringement could be difficult to detect or legally pursue. However, we hope that such shortcomings are offset by the new opportunities that these tools could offer an artist, such as the ability to license out their unique style, or the ability to quickly create early prototypes for new work.

\section{Conclusions}
We introduced the task of personalized, language-guided generation, where a text-to-image model is leveraged to create images of specific concepts in novel settings and scenes.
Our approach, ``Textual Inversions", operates by \textit{inverting} the concepts into new pseudo-words within the textual embedding space of a pre-trained text-to-image model. These pseudo-words can be injected into new scenes using simple natural language descriptions, allowing for simple and intuitive modifications. In a sense, our method allows a user to leverage multi-modal information --- using a text-driven interface for ease of editing, but providing visual cues when approaching the limits of natural language.

Our approach was implemented over LDM~\citep{rombach2021highresolution}, the largest publicly available text-to-image model. However, it does not rely on any architectural details unique to their approach. As such, we believe Textual Inversions to be easily applicable to additional, larger-scale text-to-image models. There, text-to-image alignment, shape preservation, and image generation fidelity may be further improved.

We hope our approach paves the way for future personalized generation works. These could be core to a multitude of downstream applications, from providing artistic inspiration to product design. 

\paragraph{Acknowledgments} We thank Yael Vinker, Roni Paiss and Haggai Maron for reviewing early drafts and helpful suggestions. Tom Bagshaw for discussions regarding artist rights and social impacts, and Omri Avrahami for providing us with early access to Blended Latent Diffusion. This work was partially supported by Len Blavatnik and the Blavatnik family foundation, BSF (grant 2020280) and ISF (grants 2492/20 and 3441/21).

\clearpage

%% file: appendix.tex
\section{Additional inversion approaches}

In addition to the setups outlined in the core paper, we investigated two recent approaches to inversion: Bipartite DDIM-inversion~\citep{ramesh2022hierarchical,dhariwal2021diffusion} and pivotal tuning~\citep{roich2021pivotal}. Below we outline both methods and our experimental results.

\paragraph{Bipartite inversion}
\citet{dhariwal2021diffusion} demonstrated that the DDIM sampling~\citep{song2020denoising} process can be inverted through a closed-form iterative approach. Specifically, their approach can find a latent noise vector $x_T$ which will be denoised into a specific target image when the denoising process is conditioned on a given code $c_\theta(y)$. In \citep{ramesh2022hierarchical}, they further demonstrate that when the conditioning code is an output of CLIP, one can later modify this code using text-derived directions in CLIP's multi-modal embedding space, while keeping the initial noise, $x_T$, fixed. This induces semantic changes in the image while maintaining the general structure of the original object.

\input{resources/figures/pti_and_bipartite}

Here, we investigate a similar approach. However, rather than modifying the conditioning code $c_\theta(y)$ directly, we change the conditioning text $y$. Specifically, we first find an appropriate pseudo-word for our target concept. Then, we find $x_T$ for a given image of the concept using the text ``A photo of \pholdercolor" and the closed-form solution of~\citet{dhariwal2021diffusion}. Finally, we modify the conditioning text but keep $x_T$ frozen. The results are shown in \Cref{fig:advanced_inversion} (left). Here, we observe that when using LDM's typical guidance~\citep{ho2021classifier} scales ($5$-$10$), the denoiser network is unable to maintain the original object's structure through prompt changes. When reducing the guidance scale, the outline of the original image becomes visible. However, alignment with the prompt is poor.

Such guidance-dependent structure drift has also been demonstrated for GLIDE~\citep{nichol2021glide}. However, this effect is reduced in DALL-E2~\citep{ramesh2022hierarchical} (their Figure 9). Notably, state-of-the-art models~\citep{saharia2022photorealistic,ramesh2022hierarchical} typically employ guidance scales ($\sim2$) which are significantly lower than LDM's --- within the range where we observe structure preservation, but no prompt-matching. This gives us hope that a bipartite inversion would allow better shape preservation in more powerful generative models.

\paragraph{Pivotal Tuning}
In the field of GAN inversion, it has been shown~\citep{roich2021pivotal,semantic2019bau} that one may largely avoid the reconstruction-editability tradeoff using a two-stage optimization process. First, an image is inverted into ``pivot'' code in a well-behaved region of the latent space, using standard optimization. This typically results in a highly editable code, but with poor identity preservation. As a second step, the generator is fine-tuned so that the first step's pivot code will more accurately reproduce the inverted image. It was further demonstrated that such localized tuning can maintain the appealing properties of the latent space and retain similar latent-editing capabilities.

Here, we investigate a similar approach in order to improve reconstruction. We first optimize a pseudo-word using our baseline method. Then, we fine-tune the generator such that sentences of the form ``A photo of \pholdercolor" will better reconstruct the concept-specific training set images.

Our initial investigation reveals that \naive applications of this approach lead to improved shape preservation, but also to a severe collapse of editing at high guidance scales. See \Cref{fig:advanced_inversion} (right) for examples.

However, a more involved application of this same principle (\eg by combining it with a similar process to the bipartite-inversion outlined below, or by tuning around results produced with higher guidance scales) might overcome these issues. We leave such investigation to future work. 
\clearpage

\input{resources/figures/set_size_eval}

\section{Effect of training set size}

We investigated the effect of the concept's training set size on the results. Specifically, we consider the headless sculpture object of \Cref{fig:teaser} (top row). We inverted the object using our standard model but sweeped over dataset sizes ranging from a single image to $25$ samples. For ease of comparison, we further report the image-only, prompt-only, and human caption based scores for the same single object. The results are shown in ~\Cref{fig:set_size_eval}.

Using additional images leads to optimized embeddings which reside farther away from real word embeddings, harming editability. Our method operates best when provided with $~5$ images.

\section{Additional results}

\input{resources/figures/generation_sup}

We provide additional results of personalized generation using our method. In \Cref{fig:gen_sup} we show additional text-guided synthesis results.

In \Cref{fig:uncurated_photo} we show large-scale galleries of uncurated results generated with the prompt ``A photo of \pholdercolor". In \Cref{fig:uncurated_prompt_1,fig:uncurated_prompt_2} we provide large-scale galleries of uncurated results generated with a wide assortment of prompts. These are intended to provide a sense of the quality of images produced and cherry-picking involved when generating the samples in the core paper. Note that these results also contain demonstrations of typical failure cases, such as difficult relational prompts (\Cref{fig:uncurated_prompt_1}, rows $2$, $5$).

\input{resources/figures/uncurated_photo}

\input{resources/figures/uncurated_prompt_1}
\input{resources/figures/uncurated_prompt_2}

\clearpage

\section{Training prompt templates}
Below we provide the list of text templates used when optimizing a pseudo-word:

\begin{itemize}
    \item ``a photo of a \pholder.",
    \item ``a rendering of a \pholder.",
    \item ``a cropped photo of the \pholder.",
    \item ``the photo of a \pholder.",
    \item ``a photo of a clean \pholder.",
    \item ``a photo of a dirty \pholder.",
    \item ``a dark photo of the \pholder.",
    \item ``a photo of my \pholder.",
    \item ``a photo of the cool \pholder.",
    \item ``a close-up photo of a \pholder.",
    \item ``a bright photo of the \pholder.",
    \item ``a cropped photo of a \pholder.",
    \item ``a photo of the \pholder.",
    \item ``a good photo of the \pholder.",
    \item ``a photo of one \pholder.",
    \item ``a close-up photo of the \pholder.",
    \item ``a rendition of the \pholder.",
    \item ``a photo of the clean \pholder.",
    \item ``a rendition of a \pholder.",
    \item ``a photo of a nice \pholder.",
    \item ``a good photo of a \pholder.",
    \item ``a photo of the nice \pholder.",
    \item ``a photo of the small \pholder.",
    \item ``a photo of the weird \pholder.",
    \item ``a photo of the large \pholder.",
    \item ``a photo of a cool \pholder.",
    \item ``a photo of a small \pholder.",
\end{itemize}

%% file: resources/figures/pti_and_bipartite.tex
\begin{figure}[!hbt]
    \centering
    \setlength{\tabcolsep}{0.55pt}
    \renewcommand{\arraystretch}{1.0}
    {
    
    \begin{tabular}{c}
    
    \begin{tabular}{@{\hskip 10pt}c@{\hskip 10pt}c c c}
    \begin{tabular}{c}
    \raisebox{0.145\linewidth}{\begin{tabular}{c@{}c@{}c@{}c@{}} Input \\ Samples \end{tabular}}
    \end{tabular} & 
    \includegraphics[width=0.13\linewidth,height=0.13\linewidth]{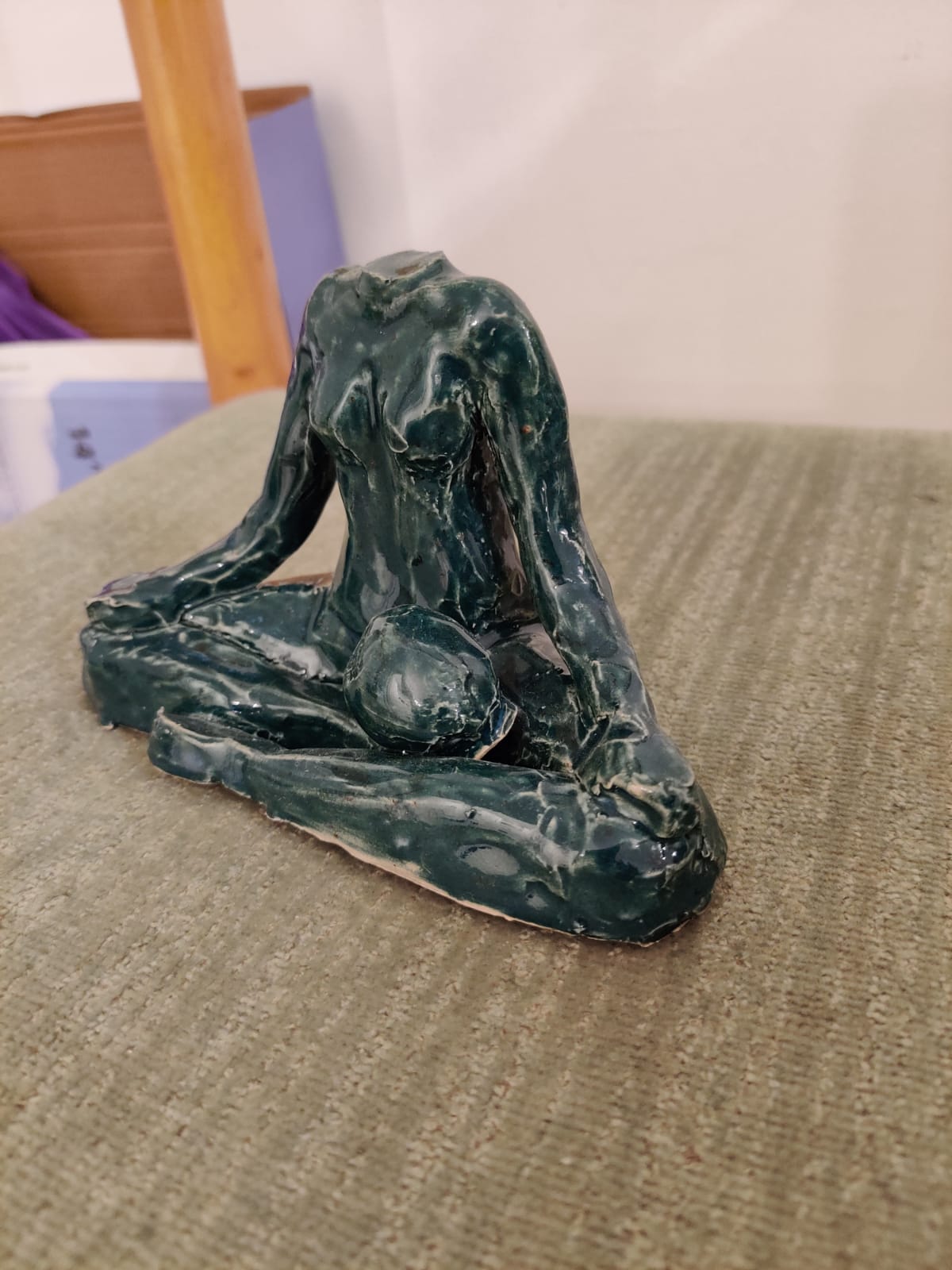} & 
    \includegraphics[width=0.13\linewidth,height=0.13\linewidth]{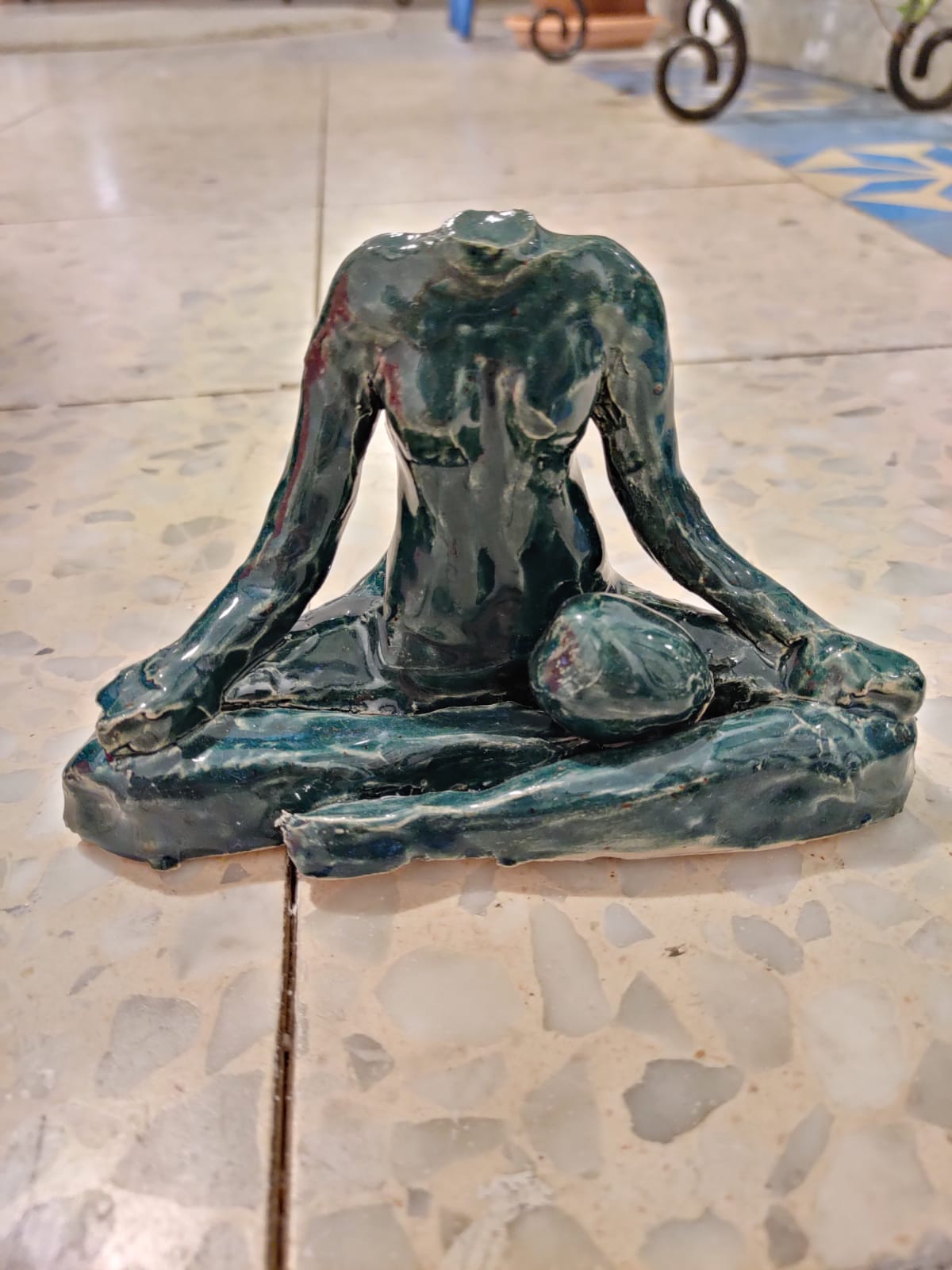} & 
    \includegraphics[width=0.13\linewidth,height=0.13\linewidth]{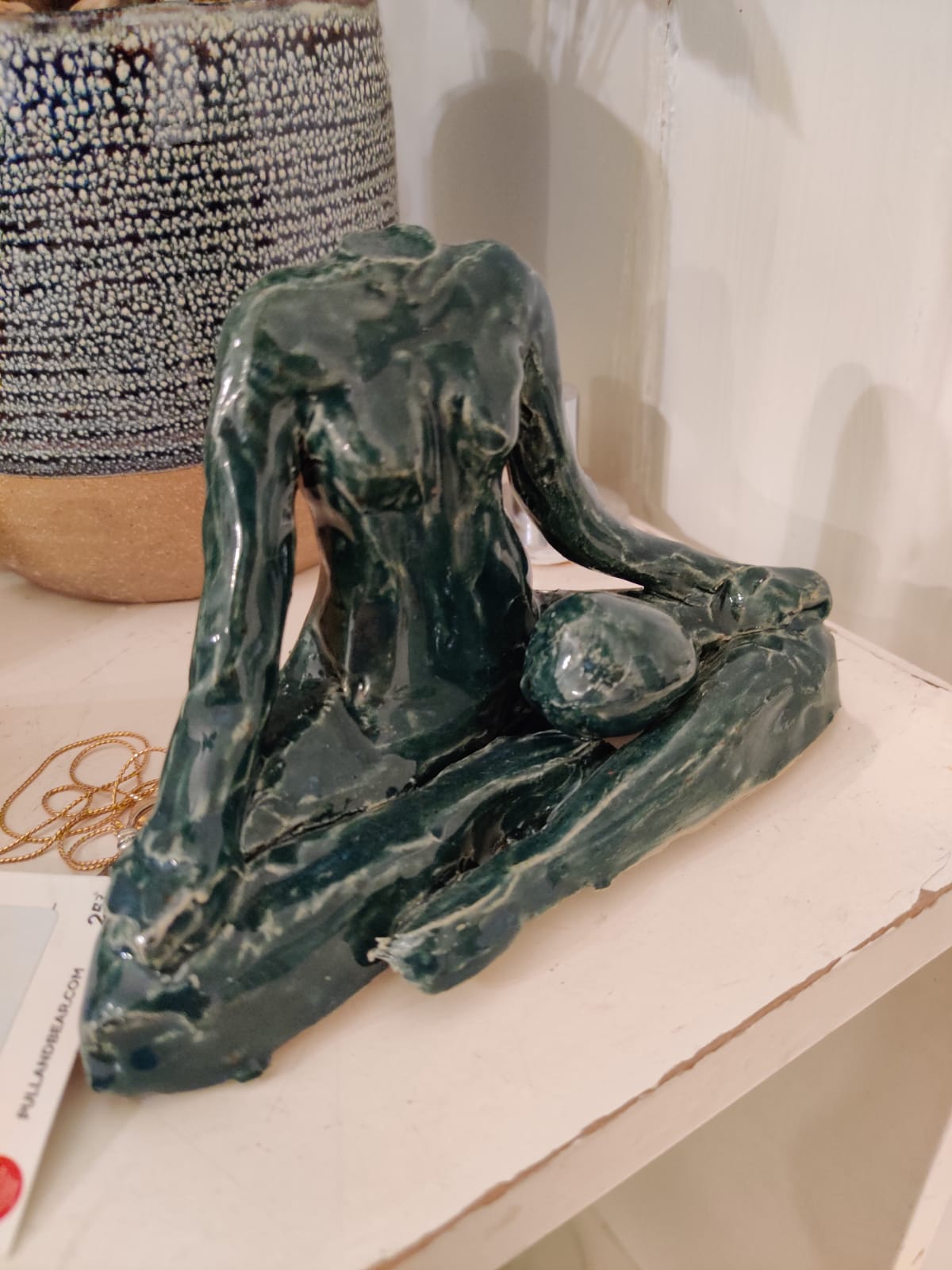}
    \end{tabular} \\[-25pt]
    
    \begin{tabular}{c@{\hskip 20pt}c}
    
    \begin{tabular}{c@{\hskip 10pt}c c c}
    & \multicolumn{3}{c}{Bipartite Inversion} \\
    \raisebox{0.055\linewidth}{Reconstruction} &
    \includegraphics[width=0.13\linewidth,height=0.13\linewidth]{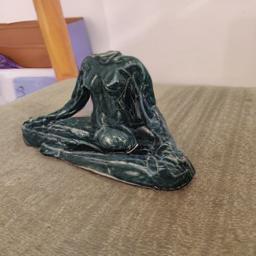} & 
    \includegraphics[width=0.13\linewidth,height=0.13\linewidth]{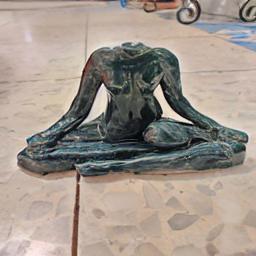} & 
    \includegraphics[width=0.13\linewidth,height=0.13\linewidth]{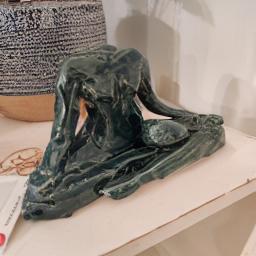} \\
    
    \raisebox{0.055\linewidth}{$s = 1$} &
    \includegraphics[width=0.13\linewidth,height=0.13\linewidth]{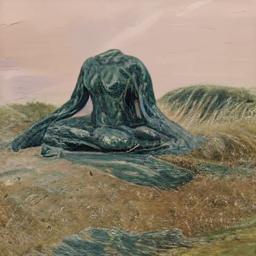} & 
    \includegraphics[width=0.13\linewidth,height=0.13\linewidth]{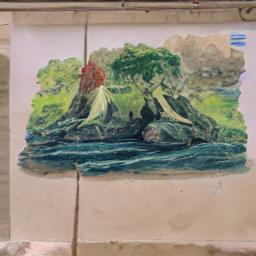} & 
    \includegraphics[width=0.13\linewidth,height=0.13\linewidth]{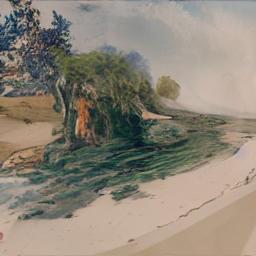} \\
    
    \raisebox{0.055\linewidth}{$s = 2$} &
    \includegraphics[width=0.13\linewidth,height=0.13\linewidth]{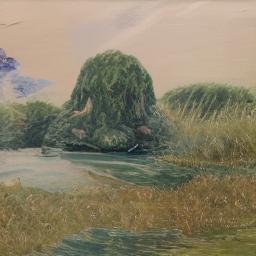} & 
    \includegraphics[width=0.13\linewidth,height=0.13\linewidth]{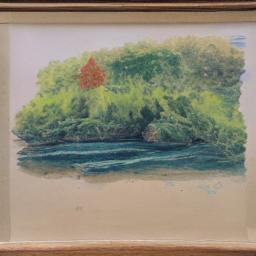} & 
    \includegraphics[width=0.13\linewidth,height=0.13\linewidth]{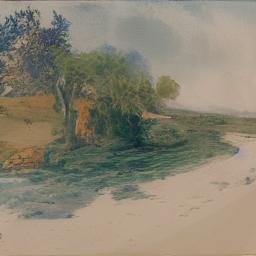} \\
    
    \raisebox{0.055\linewidth}{$s = 5$} &
    \includegraphics[width=0.13\linewidth,height=0.13\linewidth]{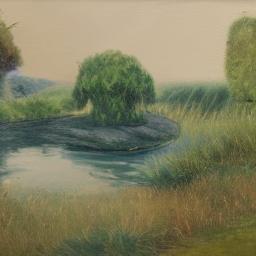} & 
    \includegraphics[width=0.13\linewidth,height=0.13\linewidth]{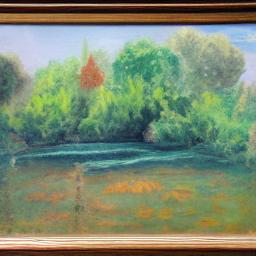} & 
    \includegraphics[width=0.13\linewidth,height=0.13\linewidth]{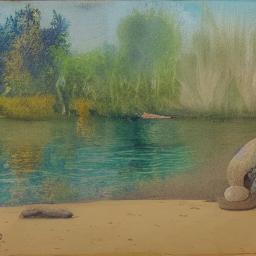} \\
    & \multicolumn{3}{c}{\begin{tabular}{c@{}c@{}c@{}c@{}} ``A painting of \pholdercolor \\ sitting next to a serene pond" \end{tabular} }
    \end{tabular} &
    
    \begin{tabular}{c c c}
    \multicolumn{3}{c}{Pivotal Tuning} \\
    \includegraphics[width=0.13\linewidth,height=0.13\linewidth]{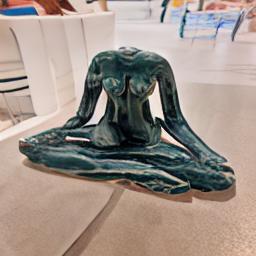} & 
    \includegraphics[width=0.13\linewidth,height=0.13\linewidth]{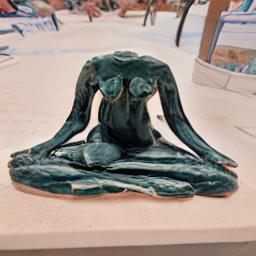} & 
    \includegraphics[width=0.13\linewidth,height=0.13\linewidth]{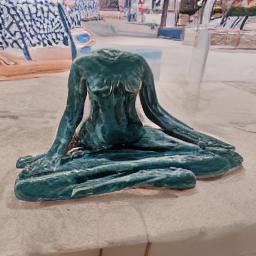} \\
    
    \includegraphics[width=0.13\linewidth,height=0.13\linewidth]{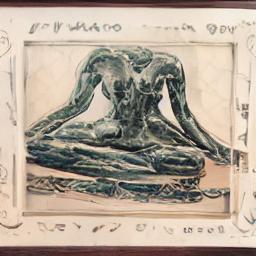} & 
    \includegraphics[width=0.13\linewidth,height=0.13\linewidth]{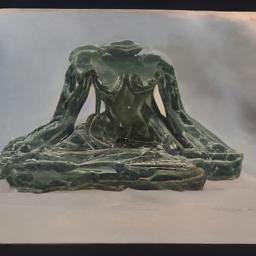} & 
    \includegraphics[width=0.13\linewidth,height=0.13\linewidth]{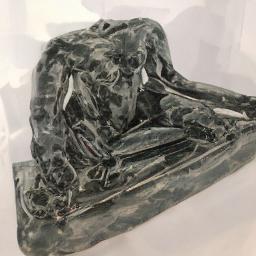} \\
    
    \includegraphics[width=0.13\linewidth,height=0.13\linewidth]{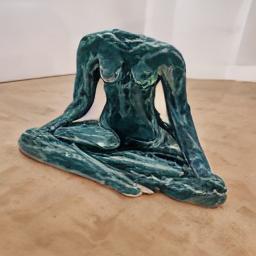} & 
    \includegraphics[width=0.13\linewidth,height=0.13\linewidth]{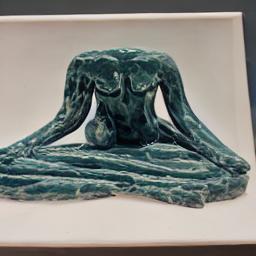} & 
    \includegraphics[width=0.13\linewidth,height=0.13\linewidth]{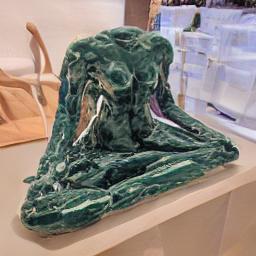} \\
    
    \includegraphics[width=0.13\linewidth,height=0.13\linewidth]{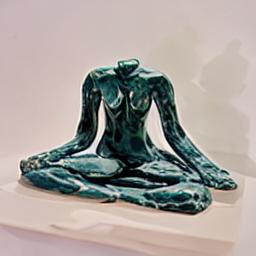} & 
    \includegraphics[width=0.13\linewidth,height=0.13\linewidth]{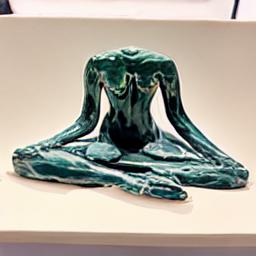} & 
    \includegraphics[width=0.13\linewidth,height=0.13\linewidth]{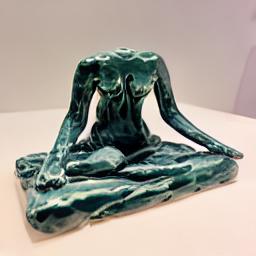} \\
    \multicolumn{3}{c}{\begin{tabular}{c@{}c@{}c@{}c@{}} ``A watercolor painting of \pholdercolor" \\ \quad \end{tabular}} 
    \end{tabular}
    
    \end{tabular}
    \end{tabular}
    }
    \caption{Advanced inversion results using Bipartite Inversion~\citep{ramesh2022hierarchical} (left) and Pivotal Tuning~\citep{roich2021pivotal} (right). $s$ is the guidance scale. Reconstructions were obtained using the prompt ``A photo of \pholdercolor". Bipartite inversion allows for more accurate reconstructions without modifying the model, but their structure is lost for complex prompts in high guidance scales. Pivotal tuning improves shapes at the cost of visual artifacts, and fail to adhere to simple prompts at high guidance scales. Note that the pivotal tuning results use a random noise input, while the bipartite results use a fixed noise for each column.}
    \label{fig:advanced_inversion} 
\end{figure}

%% file: resources/figures/set_size_eval.tex
\begin{wrapfigure}[15]{r}{0.56\linewidth}
\vspace{-41pt}
    \centering
    \includegraphics[width=0.54\textwidth]{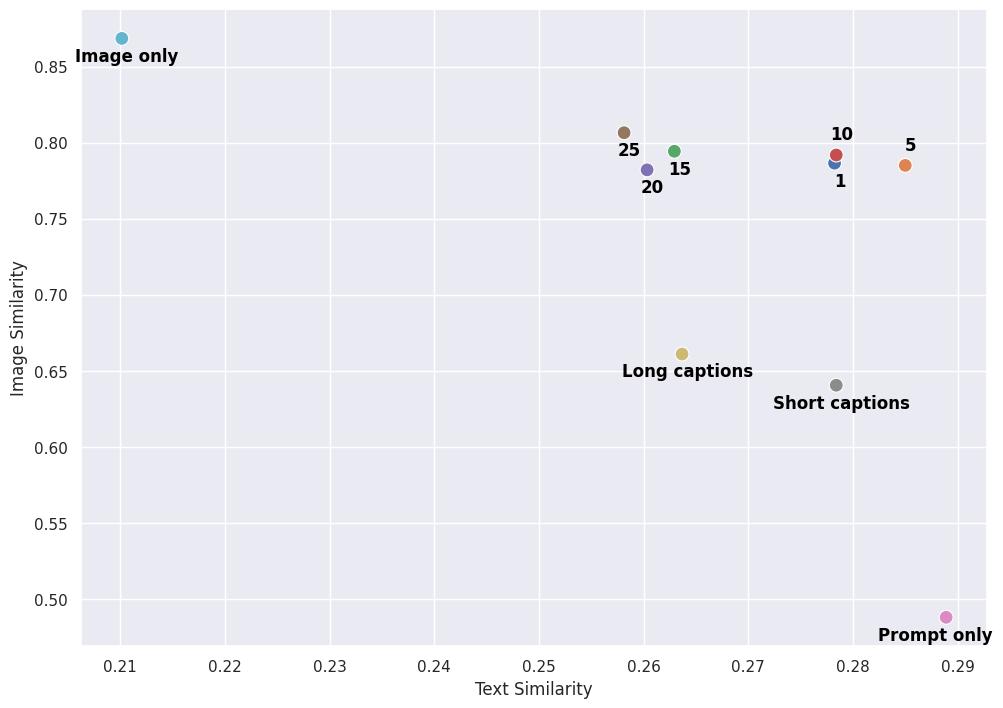} 
    \vspace{-3pt}
    \caption{Quantitative evaluation of the effects of the training set size. Significant increases to dataset sizes leads to larger deviation from the real-word distribution. This impacts editability and offers paltry improvement in reconstruction. Our approach shows the best results with $\sim5$ images.}
    \label{fig:set_size_eval}
\end{wrapfigure}

%% file: resources/figures/generation_sup.tex
\begin{figure}[!hbt]
    \centering
    \setlength{\abovecaptionskip}{6.5pt}
    \setlength{\belowcaptionskip}{-3.5pt}
    \setlength{\tabcolsep}{0.55pt}
    \renewcommand{\arraystretch}{1.0}
    {\scriptsize
    \begin{tabular}{c@{\hskip 5pt} c@{\hskip 5pt} c c c c}
    
        \begin{tabular}{c c}
            \includegraphics[width=0.09\linewidth,height=0.09\linewidth]{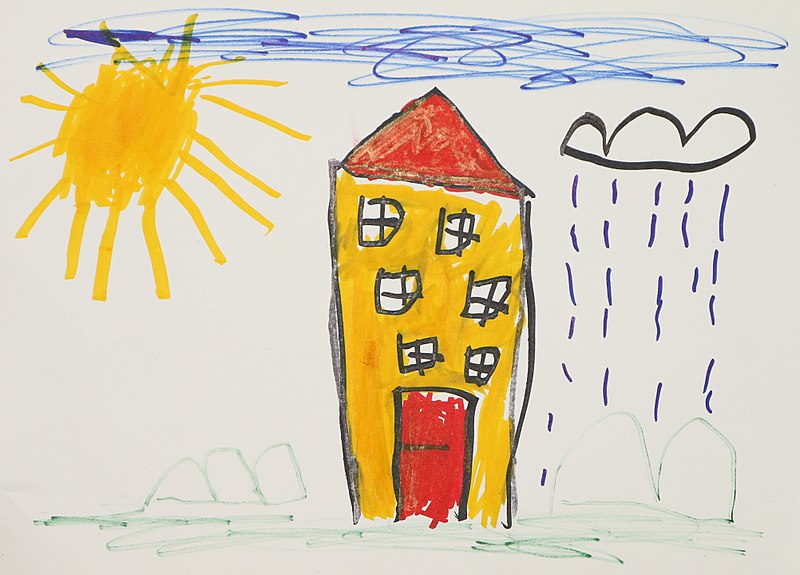} & 
            \includegraphics[width=0.09\linewidth,height=0.09\linewidth]{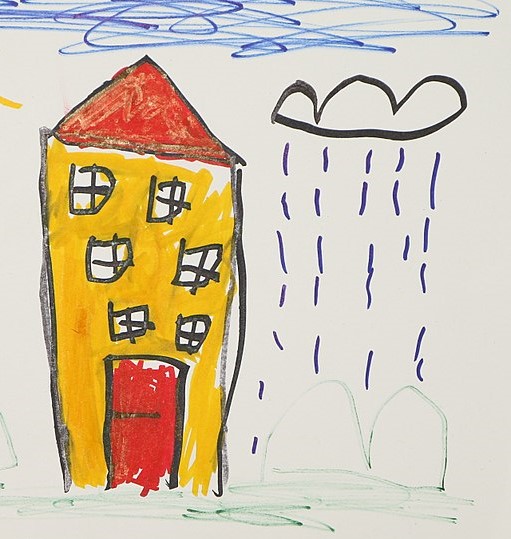} \\
            \multicolumn{2}{c}{\includegraphics[width=0.09\linewidth,height=0.09\linewidth]{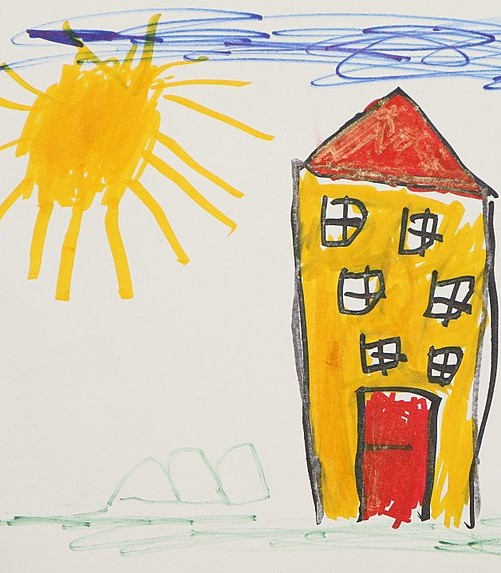}}
        \end{tabular}
        
        &
        $\rightarrow$
        &
        \begin{tabular}{c}
        \includegraphics[width=0.184\linewidth]{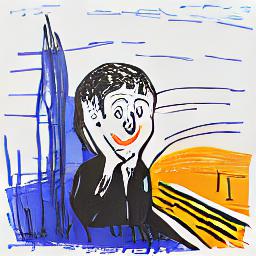}
        \end{tabular} &
        \begin{tabular}{c}
        \includegraphics[width=0.184\linewidth]{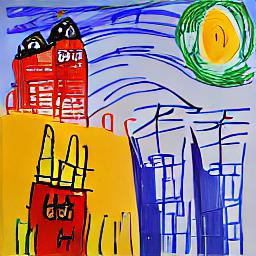} 
        \end{tabular} &
        \begin{tabular}{c}
        \includegraphics[width=0.184\linewidth]{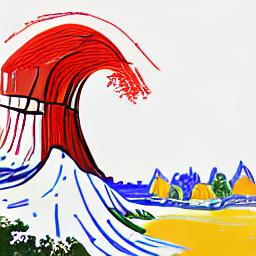} 
        \end{tabular} &
        \begin{tabular}{c}
        \includegraphics[width=0.184\linewidth]{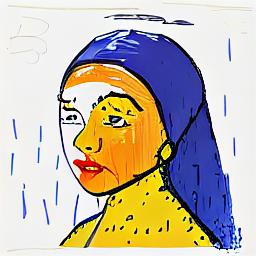}
        \end{tabular} \\
        
        {\footnotesize Input samples} & & {\begin{tabular}{c@{}c@{}c@{}c@{}} ``The Scream \\ in the style of \pholdercolor" \end{tabular}} & {\begin{tabular}{c@{}c@{}c@{}c@{}} ``\pholdercolor{} Starry Night" \end{tabular}} & {\begin{tabular}{c@{}c@{}c@{}c@{}} ``The Great Wave \\ in the style of \pholdercolor" \end{tabular}} & {\begin{tabular}{c@{}c@{}c@{}c@{}} ``Girl with a Pearl Earring \\ by Johannes Vermeer \\ in the style of \pholdercolor" \end{tabular}} \\ \\

        \begin{tabular}{c c}
            \includegraphics[width=0.09\linewidth,height=0.09\linewidth]{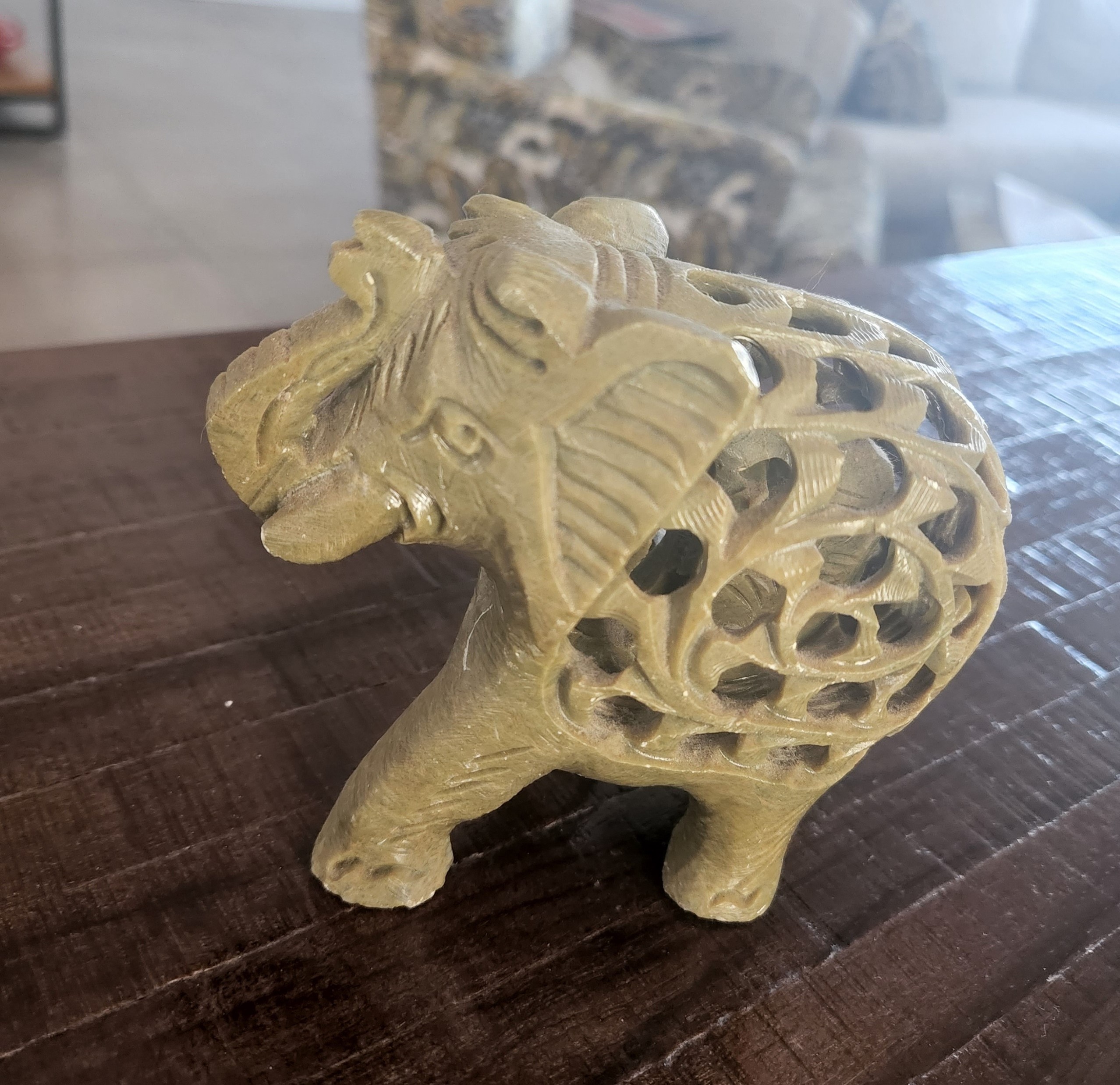} & 
            \includegraphics[width=0.09\linewidth,height=0.09\linewidth]{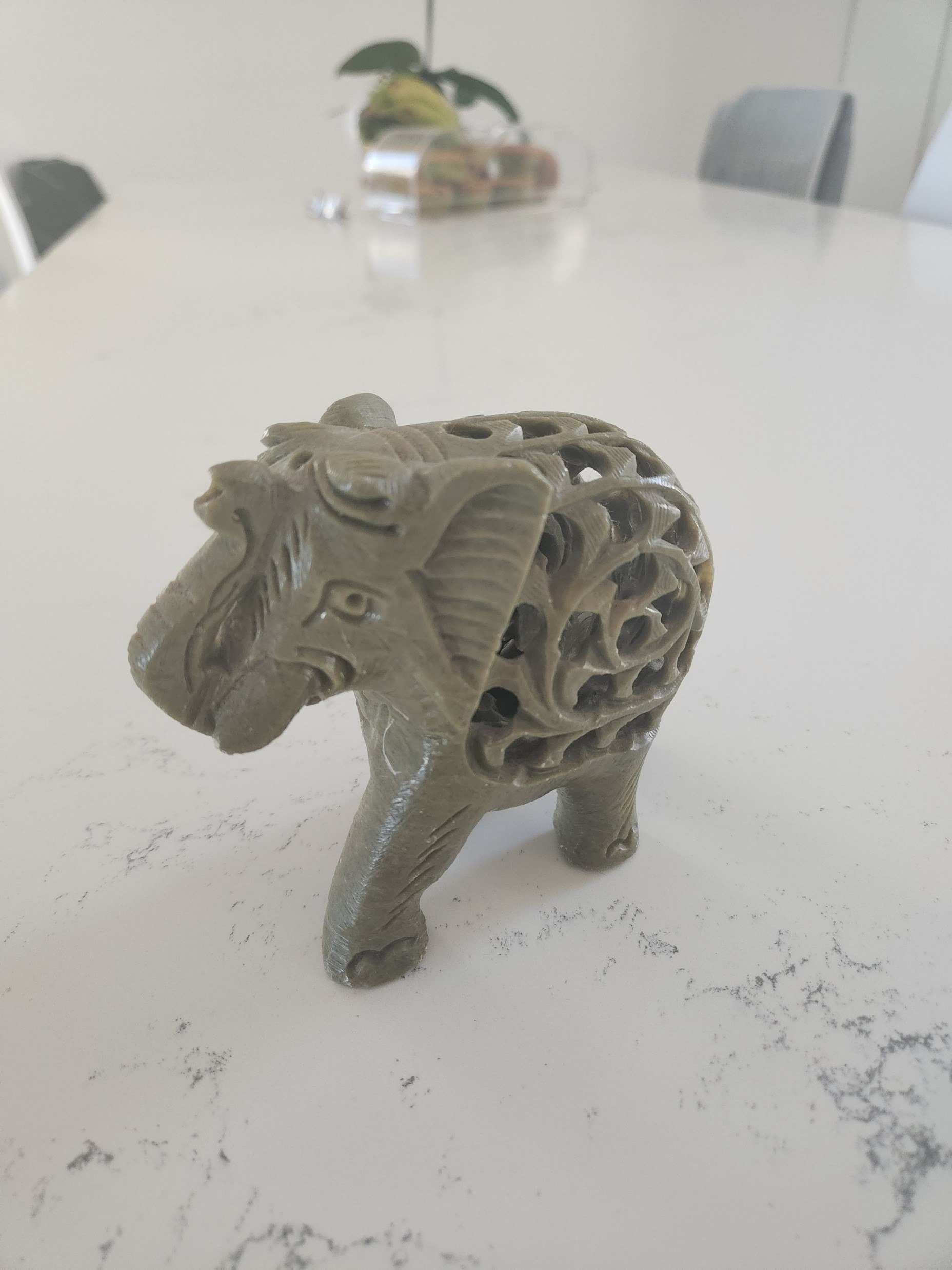} \\
            \includegraphics[width=0.09\linewidth,height=0.09\linewidth]{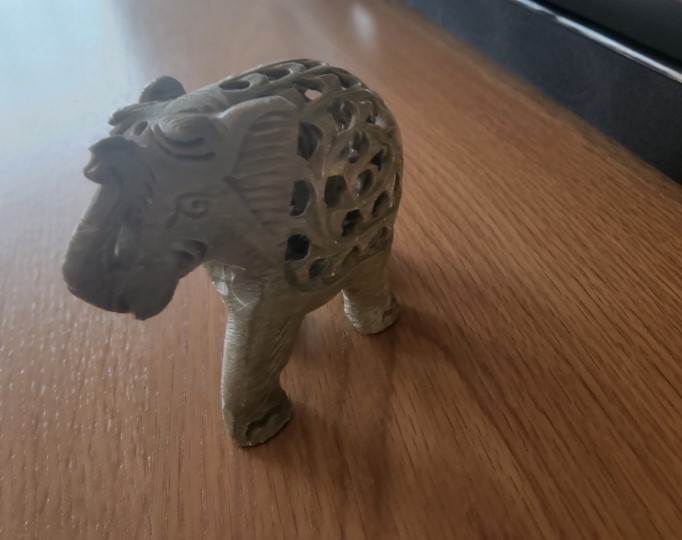} & 
            \includegraphics[width=0.09\linewidth,height=0.09\linewidth]{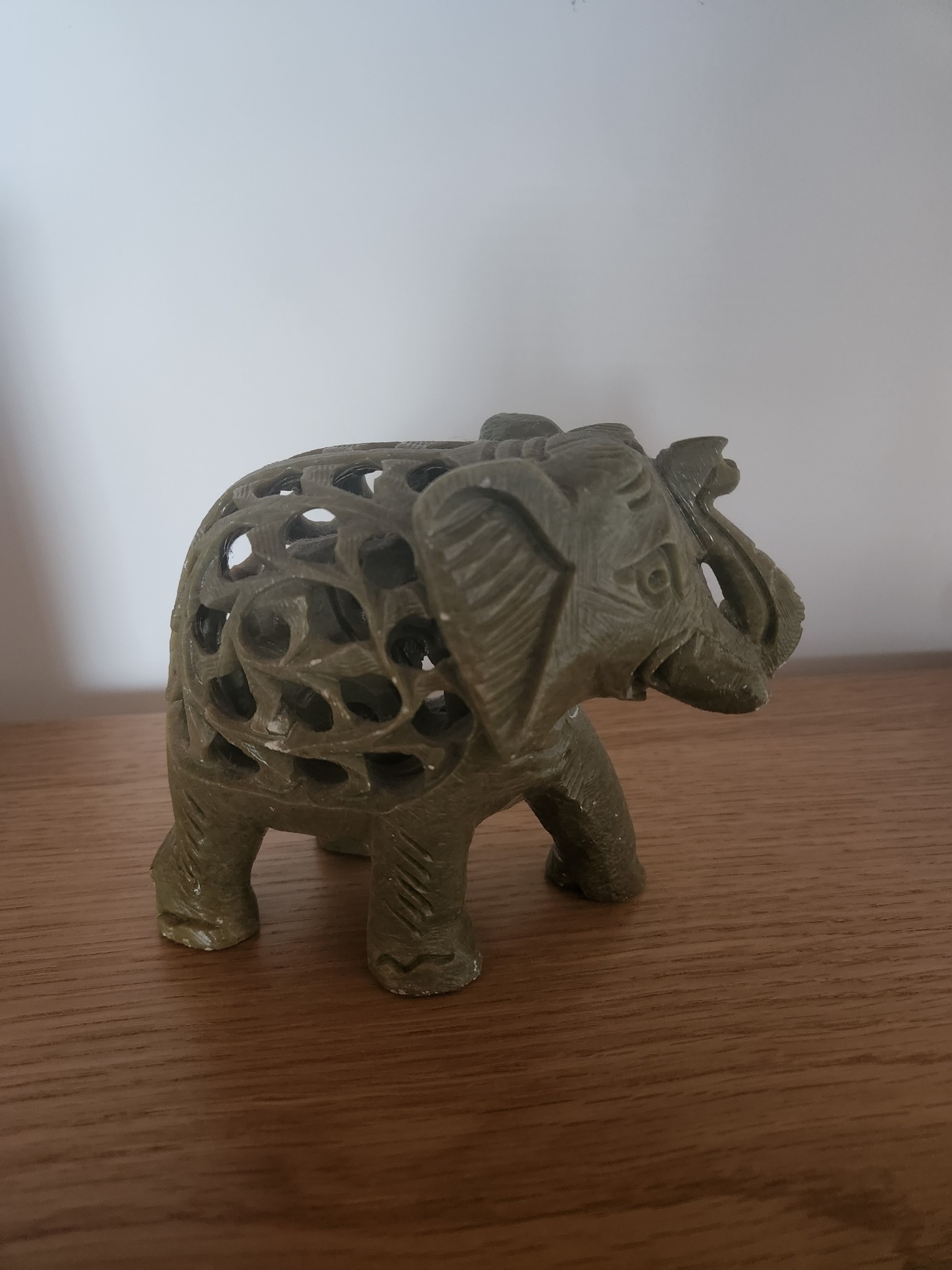}
        \end{tabular}
        
        &
        $\rightarrow$
        &
        \begin{tabular}{c}
        \includegraphics[width=0.184\linewidth]{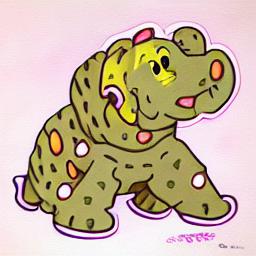}
        \end{tabular} &
        \begin{tabular}{c}
        \includegraphics[width=0.184\linewidth]{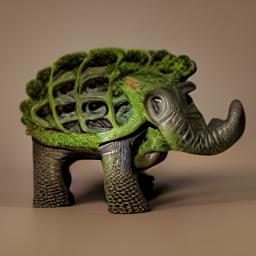} 
        \end{tabular} &
        \begin{tabular}{c}
        \includegraphics[width=0.184\linewidth]{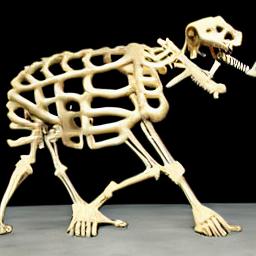} 
        \end{tabular} &
        \begin{tabular}{c}
        \includegraphics[width=0.184\linewidth]{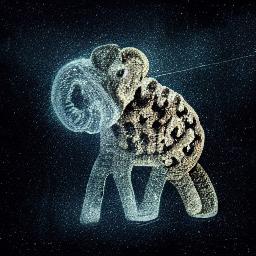}
        \end{tabular} \\
        
        {\footnotesize Input samples} & & {\begin{tabular}{c@{}c@{}c@{}c@{}} ``Anime painting of \\ a Kawaii \pholdercolor" \end{tabular}} & {\begin{tabular}{c@{}c@{}c@{}c@{}} ``A mossy \pholdercolor" \end{tabular}} & {\begin{tabular}{c@{}c@{}c@{}c@{}} ``\pholdercolor{} skeleton in the \\ National Science Museum" \end{tabular}} & {\begin{tabular}{c@{}c@{}c@{}c@{}} ``Negative space \\ painting of \pholdercolor" \end{tabular}} \\ \\

        \begin{tabular}{c c}
            \includegraphics[width=0.09\linewidth,height=0.09\linewidth]{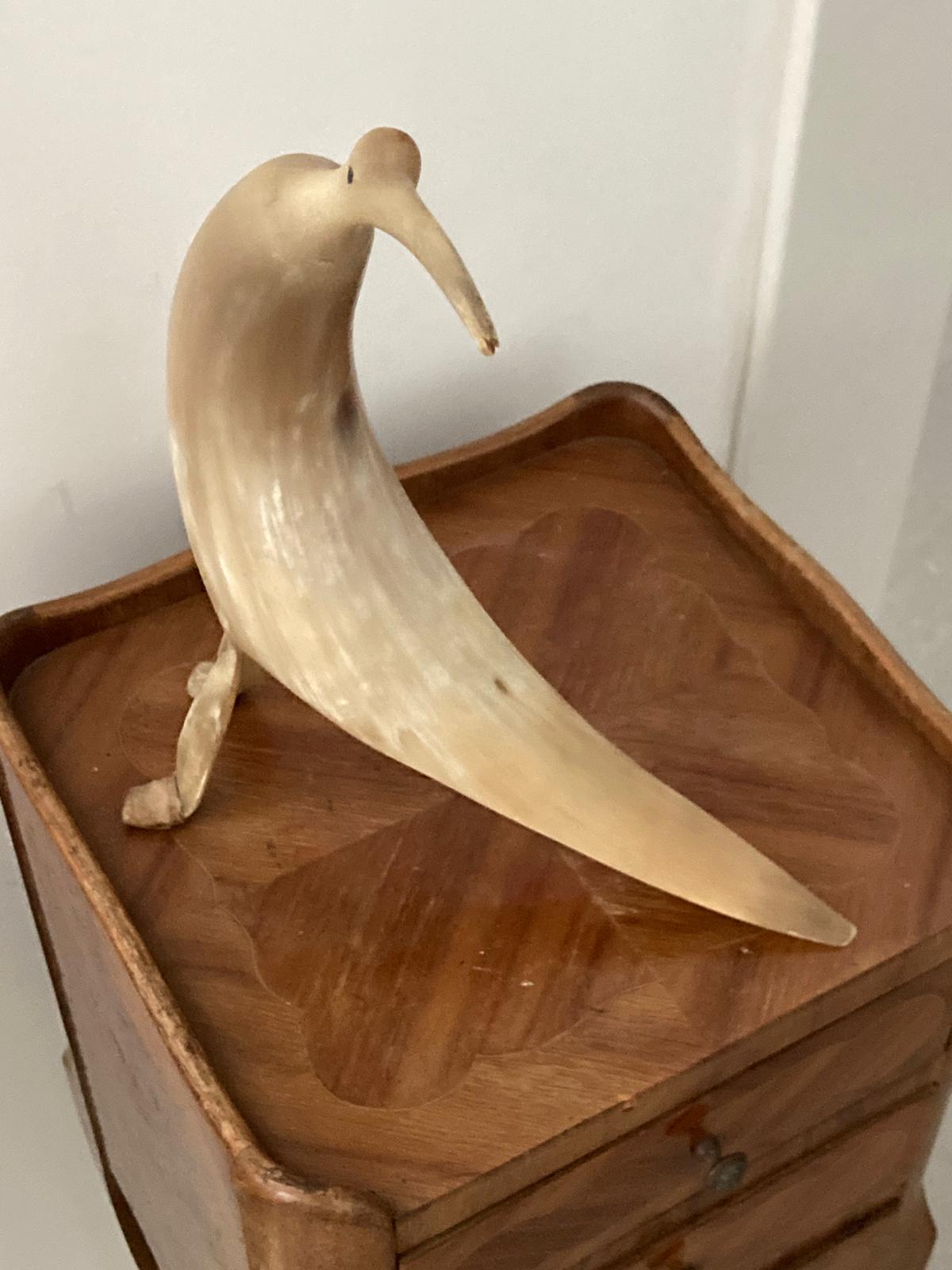} & 
            \includegraphics[width=0.09\linewidth,height=0.09\linewidth]{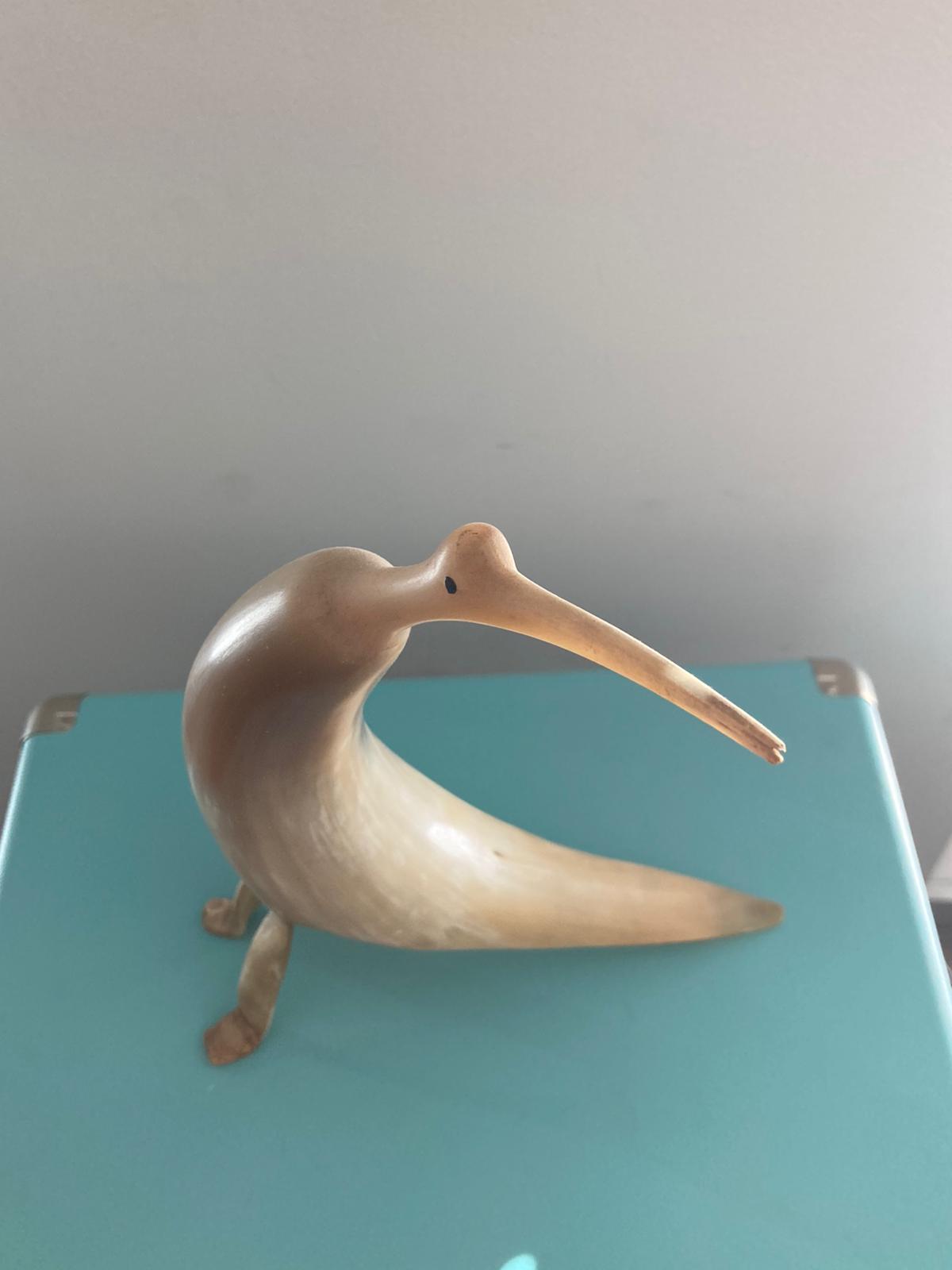} \\
            \includegraphics[width=0.09\linewidth,height=0.09\linewidth]{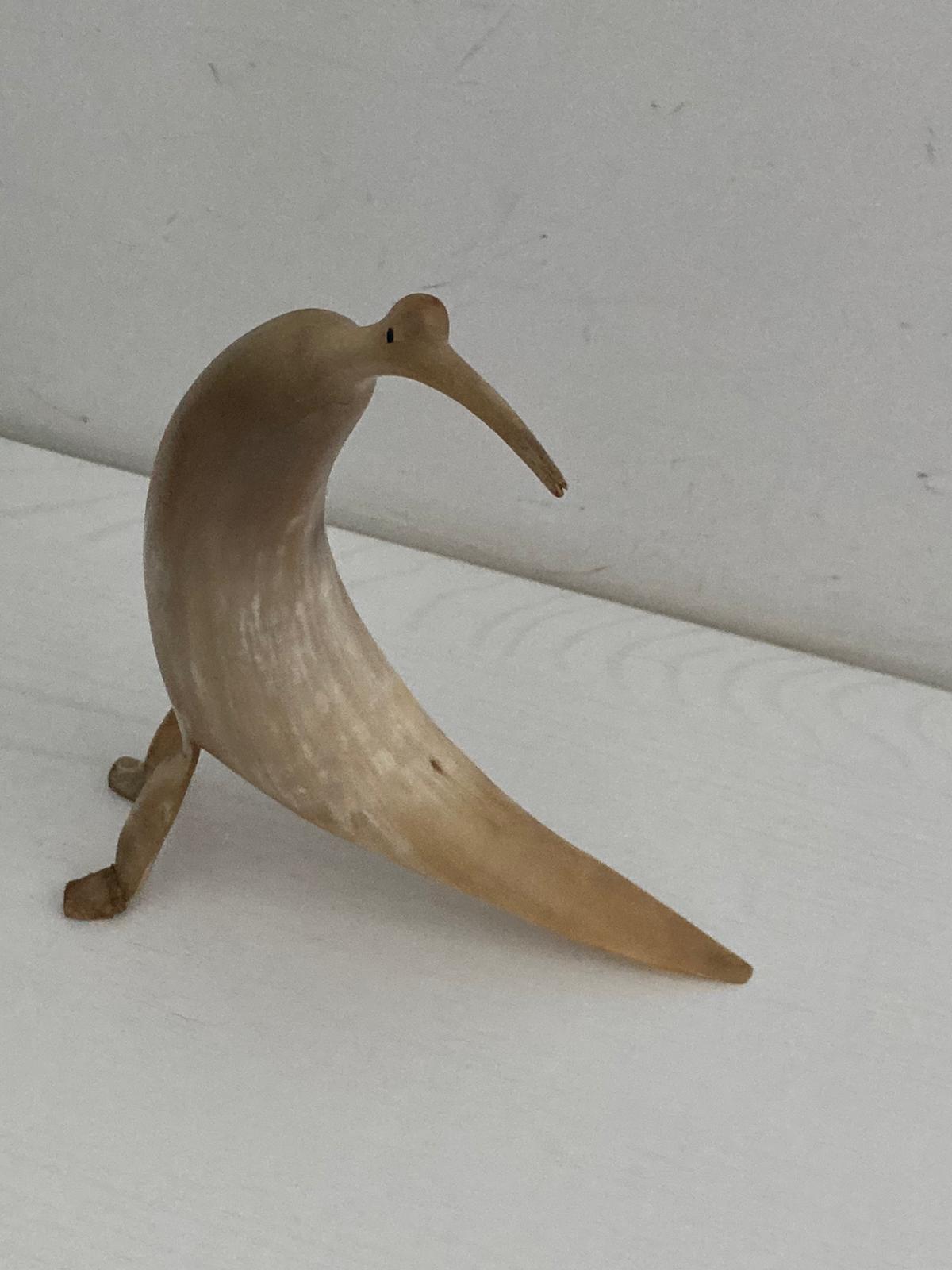} & 
            \includegraphics[width=0.09\linewidth,height=0.09\linewidth]{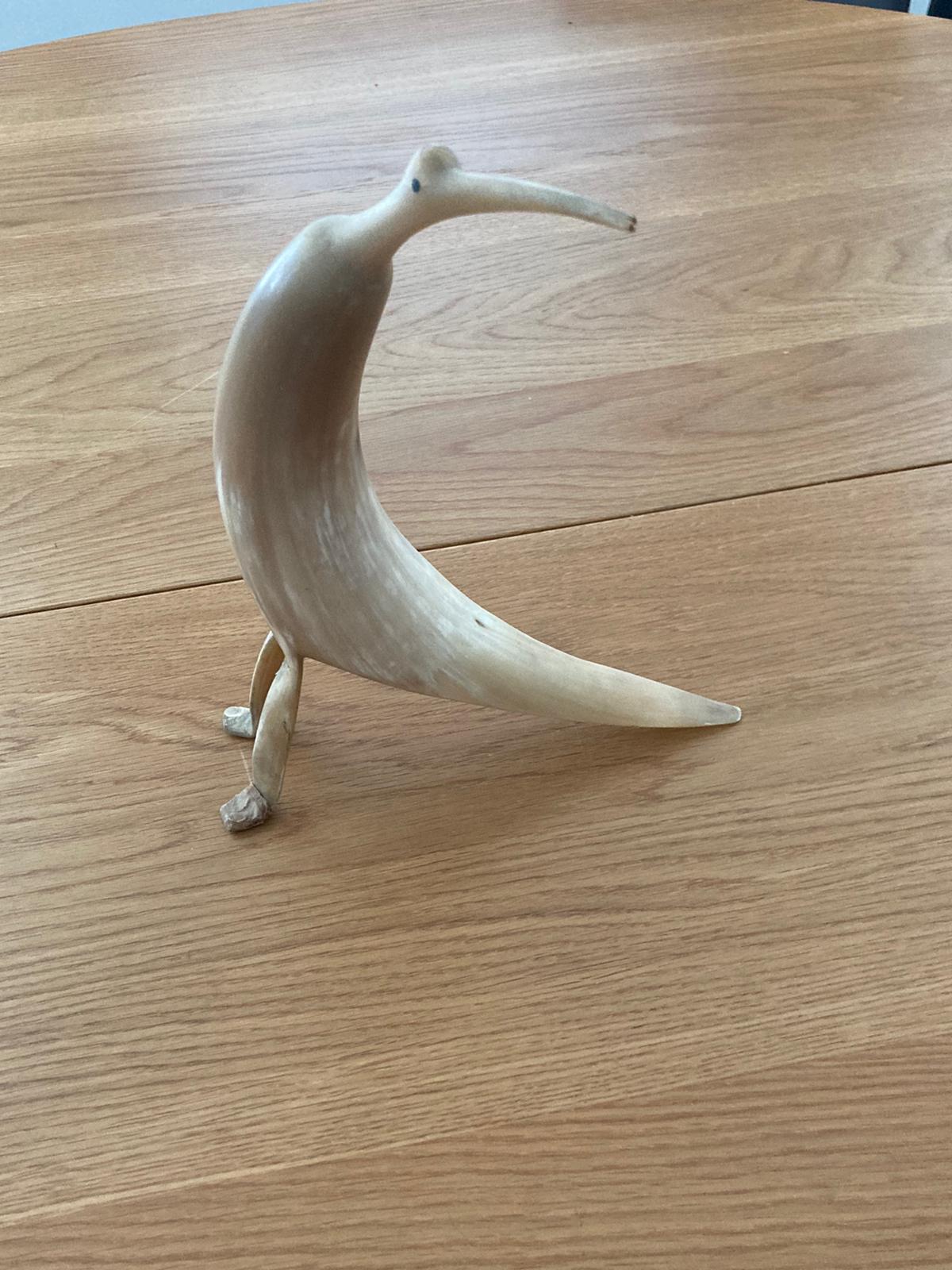}
        \end{tabular}
        
        &
        $\rightarrow$
        &
        \begin{tabular}{c}
        \includegraphics[width=0.184\linewidth]{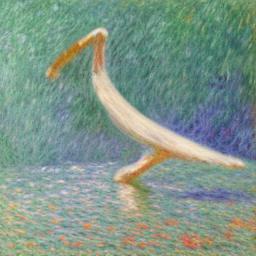}
        \end{tabular} &
        \begin{tabular}{c}
        \includegraphics[width=0.184\linewidth]{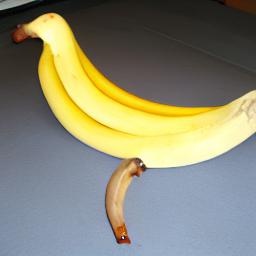} 
        \end{tabular} &
        \begin{tabular}{c}
        \includegraphics[width=0.184\linewidth]{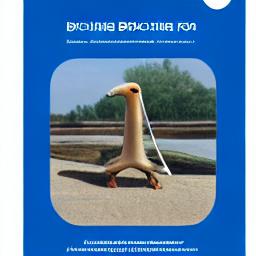} 
        \end{tabular} &
        \begin{tabular}{c}
        \includegraphics[width=0.184\linewidth]{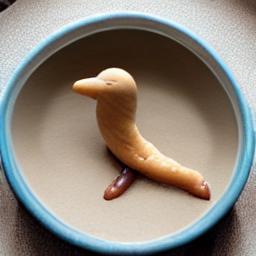}
        \end{tabular} \\
        
        {\footnotesize Input samples} & & {\begin{tabular}{c@{}c@{}c@{}c@{}} ``Painting of \pholdercolor \\ in the style of Monet" \end{tabular}} & {\begin{tabular}{c@{}c@{}c@{}c@{}} ``A banana \pholdercolor" \end{tabular}} & {\begin{tabular}{c@{}c@{}c@{}c@{}} ``Advertisement brochure \\ for \pholdercolor" \end{tabular}} & {\begin{tabular}{c@{}c@{}c@{}c@{}} ``\pholdercolor{} cookies" \end{tabular}} \\ \\

        \begin{tabular}{c c}
            \includegraphics[width=0.09\linewidth,height=0.09\linewidth]{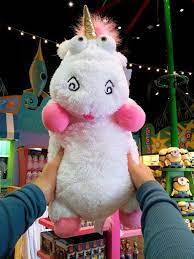} & 
            \includegraphics[width=0.09\linewidth,height=0.09\linewidth]{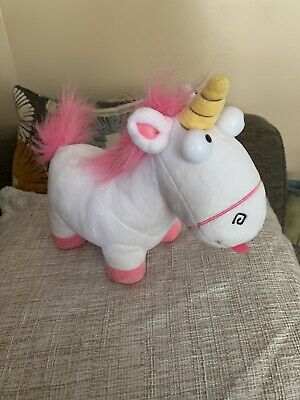} \\
            \includegraphics[width=0.09\linewidth,height=0.09\linewidth]{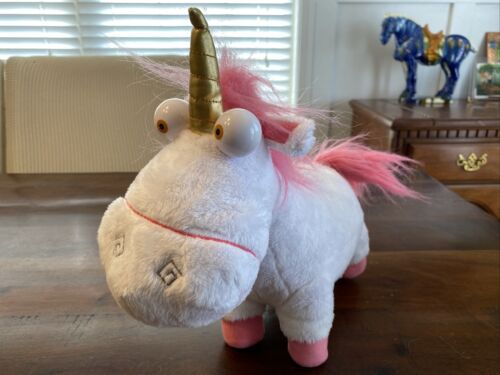} & 
            \includegraphics[width=0.09\linewidth,height=0.09\linewidth]{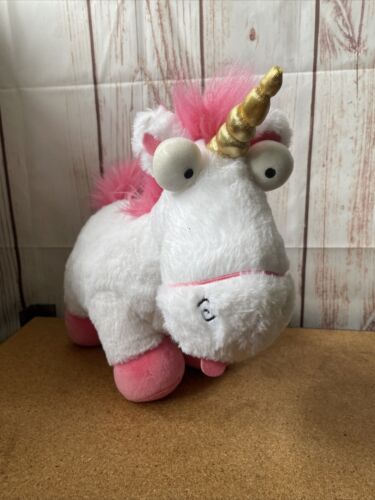}
        \end{tabular}
        
        &
        $\rightarrow$
        &
        \begin{tabular}{c}
        \includegraphics[width=0.184\linewidth]{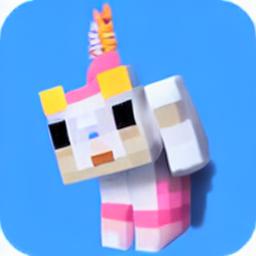}
        \end{tabular} &
        \begin{tabular}{c}
        \includegraphics[width=0.184\linewidth]{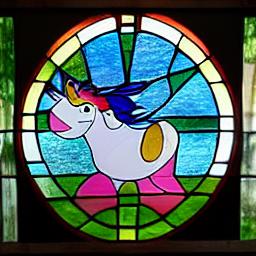} 
        \end{tabular} &
        \begin{tabular}{c}
        \includegraphics[width=0.184\linewidth]{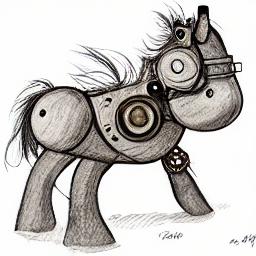} 
        \end{tabular} &
        \begin{tabular}{c}
        \includegraphics[width=0.184\linewidth]{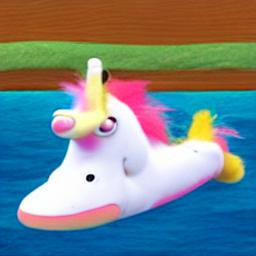}
        \end{tabular} \\
        
        {\footnotesize Input samples} & & {\begin{tabular}{c@{}c@{}c@{}c@{}} ``\pholdercolor{} in Minecraft" \end{tabular}} & {\begin{tabular}{c@{}c@{}c@{}c@{}} ``A stained glass \\ window depicting \pholdercolor" \end{tabular}} & {\begin{tabular}{c@{}c@{}c@{}c@{}} ``Sketch of a  \\ steampunk \pholdercolor" \end{tabular}} & {\begin{tabular}{c@{}c@{}c@{}c@{}} ``A \pholdercolor{} submarine" \end{tabular}} \\ \\
        
        \begin{tabular}{c c}
            \includegraphics[width=0.09\linewidth,height=0.09\linewidth]{resources/images/user_sentences/teapot_input_1.jpg} & 
            \includegraphics[width=0.09\linewidth,height=0.09\linewidth]{resources/images/user_sentences/teapot_input_2.jpg} \\
            \multicolumn{2}{c}{\includegraphics[width=0.09\linewidth,height=0.09\linewidth]{resources/images/user_sentences/teapot_input_3.jpg}}
        \end{tabular}
        &
        $\rightarrow$
        &
        \begin{tabular}{c}
        \includegraphics[width=0.184\linewidth]{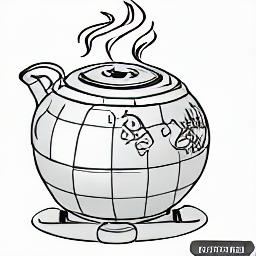} 
        \end{tabular} &
        \begin{tabular}{c}
        \includegraphics[width=0.184\linewidth]{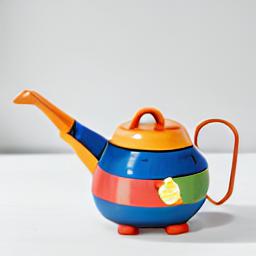} 
        \end{tabular} &
        \begin{tabular}{c}
        \includegraphics[width=0.184\linewidth]{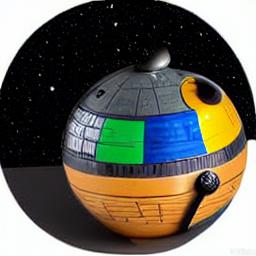} 
        \end{tabular} &
        \begin{tabular}{c}
        \includegraphics[width=0.184\linewidth]{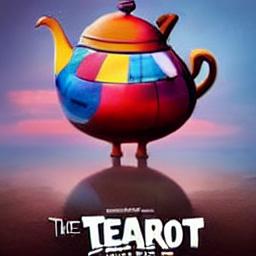} 
        \end{tabular} \\
        
        {\footnotesize Input samples} & & {\begin{tabular}{c@{}c@{}c@{}c@{}} ``Manga drawing of \\ a steaming \pholdercolor" \end{tabular}} &
        {\begin{tabular}{c@{}c@{}c@{}c@{}} ``A \pholdercolor{} watering can" \end{tabular}} &
        {\begin{tabular}{c@{}c@{}c@{}c@{}} ``\pholdercolor{} Death Star" \end{tabular}} & {\begin{tabular}{c@{}c@{}c@{}c@{}} ``A poster for the movie \\ `The Teapot' \\ starring \pholdercolor" \end{tabular}} \\

    \end{tabular}}
    \caption{Injecting user-specific concepts into new scenes. Our method can change a concept's style, composition, or use it to inspire new creations. Top row image credits: \href{https://commons.wikimedia.org/wiki/File:Wc_yellow_house_child_drawing.jpg}{@Øyvind Holmstad}.}
    \label{fig:gen_sup} 
\end{figure}

%% file: resources/figures/uncurated_photo.tex
\begin{figure}[!hbt]
    \centering
    \setlength{\abovecaptionskip}{6.5pt}
    \setlength{\belowcaptionskip}{-3.5pt}
    \setlength{\tabcolsep}{0.55pt}
    \renewcommand{\arraystretch}{1.0}
    {
    \begin{tabular}{c@{\hskip 5pt} c@{\hskip 5pt} c}
    
        \begin{tabular}{c c}
            \includegraphics[width=0.09\linewidth,height=0.09\linewidth]{resources/images/training_sets/fat_stone_bird/1.jpg} & 
            \includegraphics[width=0.09\linewidth,height=0.09\linewidth]{resources/images/training_sets/fat_stone_bird/2.jpg} \\
            \includegraphics[width=0.09\linewidth,height=0.09\linewidth]{resources/images/training_sets/fat_stone_bird/3.jpg} & 
            \includegraphics[width=0.09\linewidth,height=0.09\linewidth]{resources/images/training_sets/fat_stone_bird/4.jpg}
        \end{tabular}
        
        &
        $\rightarrow$
        &
        \begin{tabular}{c}
        \includegraphics[width=0.736\linewidth]{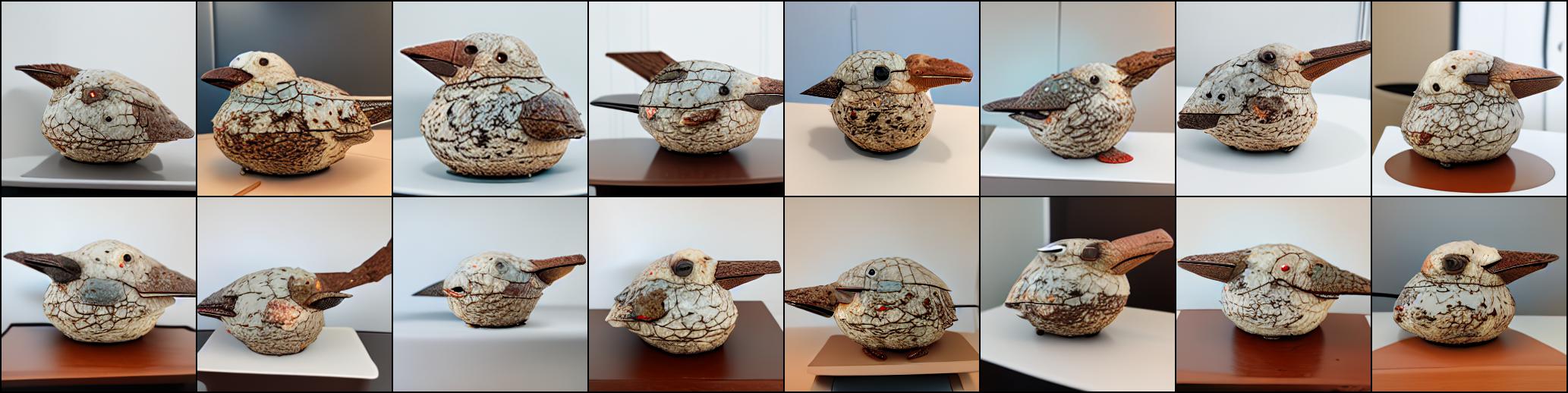}
        \end{tabular} \\
        
        {Input samples} & & {``A photo of \pholdercolor"} \\

        \begin{tabular}{c c}
            \includegraphics[width=0.09\linewidth,height=0.09\linewidth]{resources/images/training_sets/red_bowl/1.jpg} & 
            \includegraphics[width=0.09\linewidth,height=0.09\linewidth]{resources/images/training_sets/red_bowl/2.jpg} \\
            \includegraphics[width=0.09\linewidth,height=0.09\linewidth]{resources/images/training_sets/red_bowl/3.jpg} & 
            \includegraphics[width=0.09\linewidth,height=0.09\linewidth]{resources/images/training_sets/red_bowl/4.jpg}
        \end{tabular}
        
        &
        $\rightarrow$
        &
        \begin{tabular}{c}
        \includegraphics[width=0.736\linewidth]{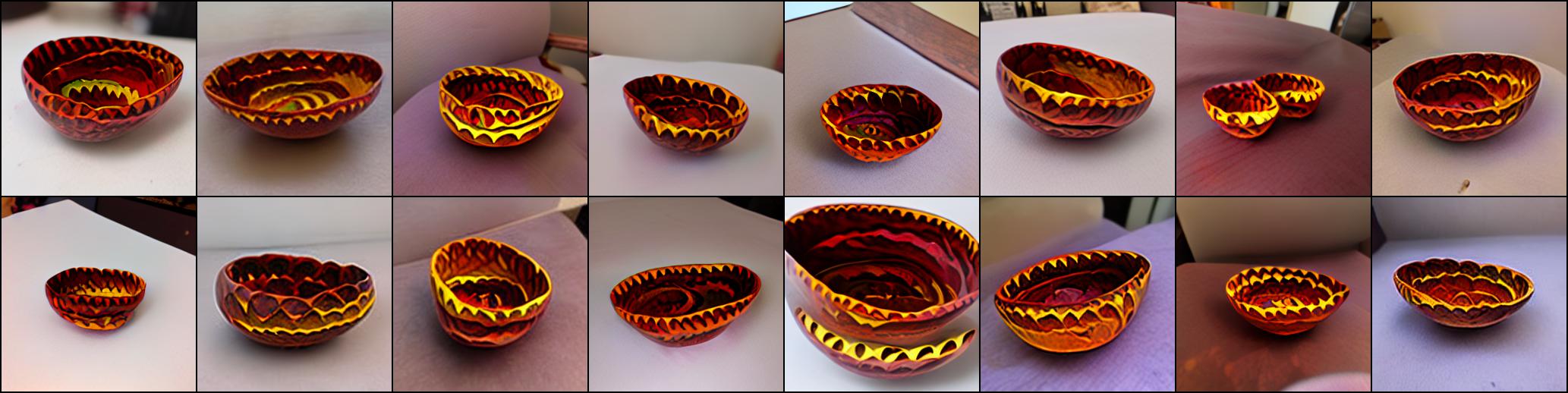}
        \end{tabular} \\
        
        {Input samples} & & {``A photo of \pholdercolor"} \\

        \begin{tabular}{c c}
            \includegraphics[width=0.09\linewidth,height=0.09\linewidth]{resources/images/training_sets/rainbow_cat/2.jpeg} & 
            \includegraphics[width=0.09\linewidth,height=0.09\linewidth]{resources/images/training_sets/rainbow_cat/3.jpeg} \\
            \multicolumn{2}{c}{\includegraphics[width=0.09\linewidth,height=0.09\linewidth]{resources/images/training_sets/rainbow_cat/6.jpeg}}
        \end{tabular}
        
        &
        $\rightarrow$
        &
        \begin{tabular}{c}
        \includegraphics[width=0.736\linewidth]{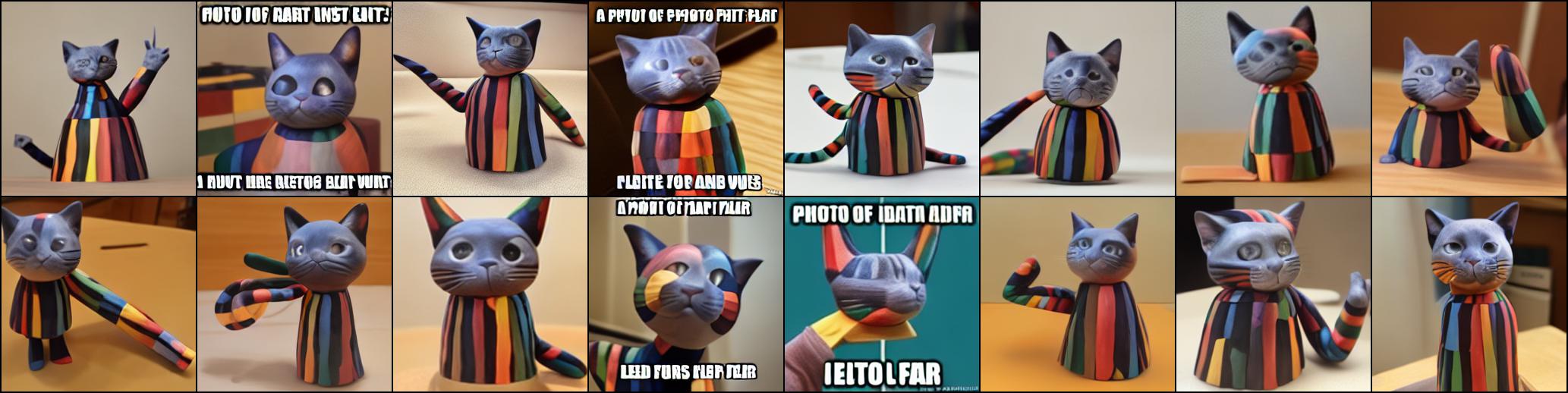}
        \end{tabular} \\
        
        {Input samples} & & {``A photo of \pholdercolor"} \\

        \begin{tabular}{c c}
            \includegraphics[width=0.09\linewidth,height=0.09\linewidth]{resources/images/training_sets/headless_statue/1.jpeg} & 
            \includegraphics[width=0.09\linewidth,height=0.09\linewidth]{resources/images/training_sets/headless_statue/2.jpeg} \\
            \includegraphics[width=0.09\linewidth,height=0.09\linewidth]{resources/images/training_sets/headless_statue/3.jpeg} & 
            \includegraphics[width=0.09\linewidth,height=0.09\linewidth]{resources/images/training_sets/headless_statue/4.jpeg}
        \end{tabular}
        
        &
        $\rightarrow$
        &
        \begin{tabular}{c}
        \includegraphics[width=0.736\linewidth]{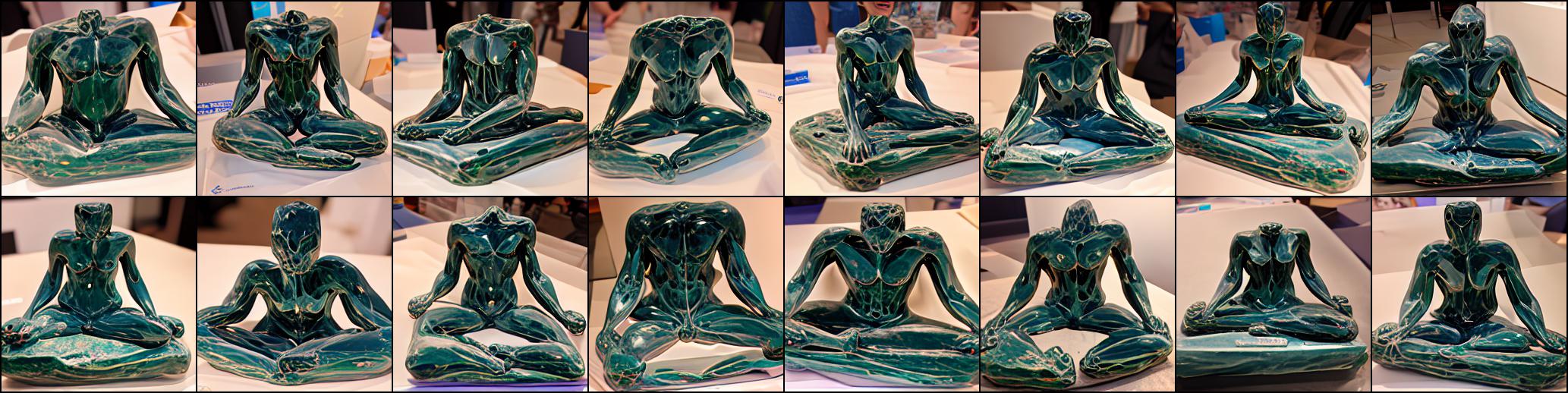}
        \end{tabular} \\
        
        {Input samples} & & {``A photo of \pholdercolor"} \\
        
        \begin{tabular}{c c}
            \includegraphics[width=0.09\linewidth,height=0.09\linewidth]{resources/images/training_sets/slim_stone_bird/1.jpeg} & 
            \includegraphics[width=0.09\linewidth,height=0.09\linewidth]{resources/images/training_sets/slim_stone_bird/2.jpeg} \\
            \includegraphics[width=0.09\linewidth,height=0.09\linewidth]{resources/images/training_sets/slim_stone_bird/3.jpeg} & 
            \includegraphics[width=0.09\linewidth,height=0.09\linewidth]{resources/images/training_sets/slim_stone_bird/4.jpeg}
        \end{tabular}
        
        &
        $\rightarrow$
        &
        \begin{tabular}{c}
        \includegraphics[width=0.736\linewidth]{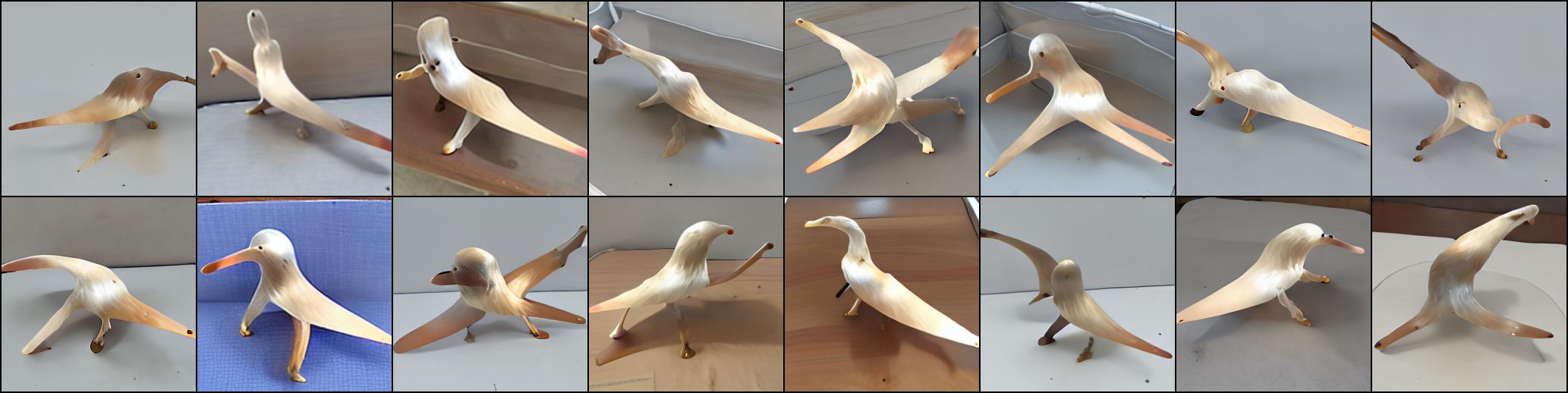}
        \end{tabular} \\
        
        {Input samples} & & {``A photo of \pholdercolor"} \\
        
        \begin{tabular}{c c}
            \includegraphics[width=0.09\linewidth,height=0.09\linewidth]{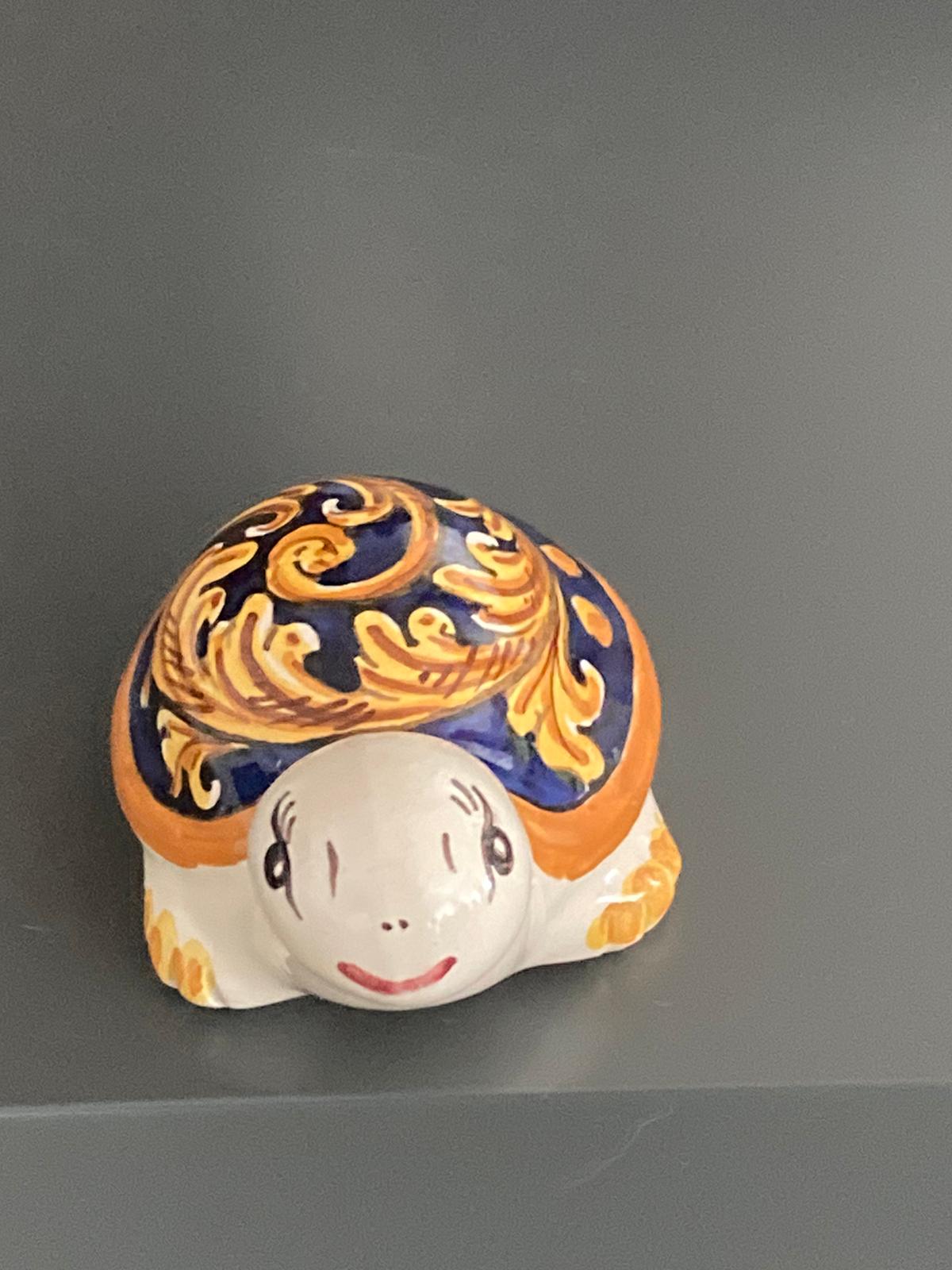} & 
            \includegraphics[width=0.09\linewidth,height=0.09\linewidth]{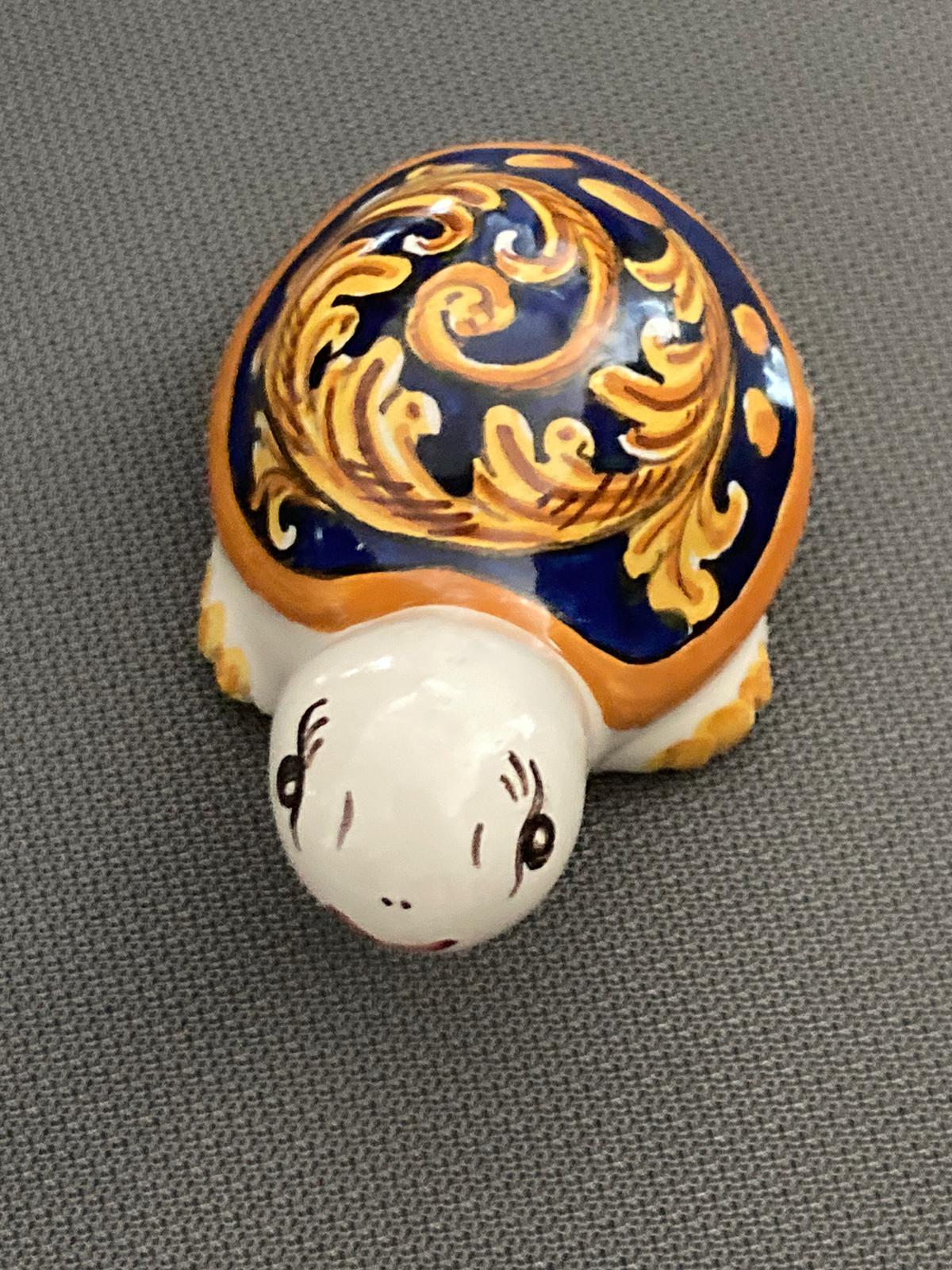} \\
            \includegraphics[width=0.09\linewidth,height=0.09\linewidth]{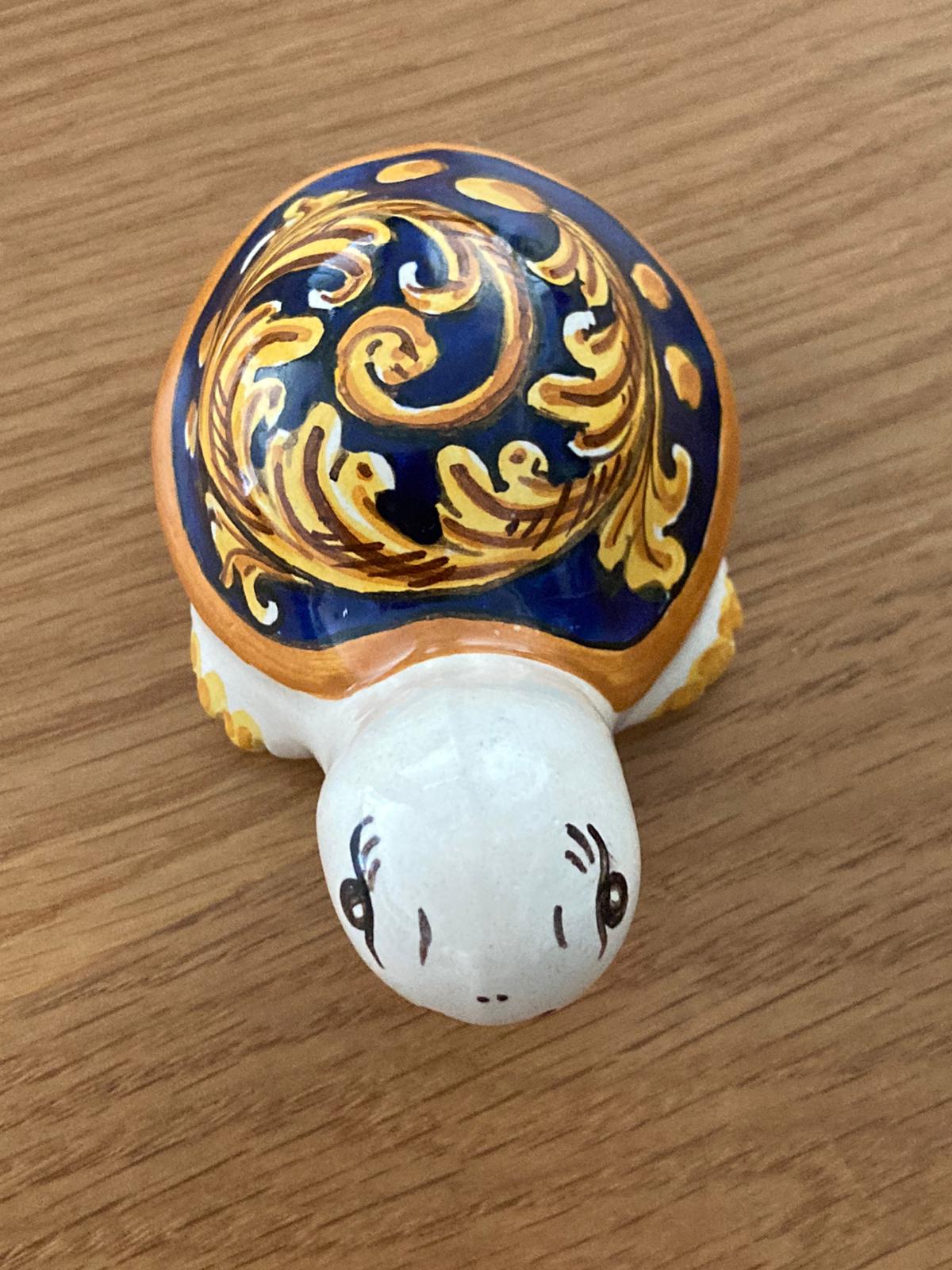} & 
            \includegraphics[width=0.09\linewidth,height=0.09\linewidth]{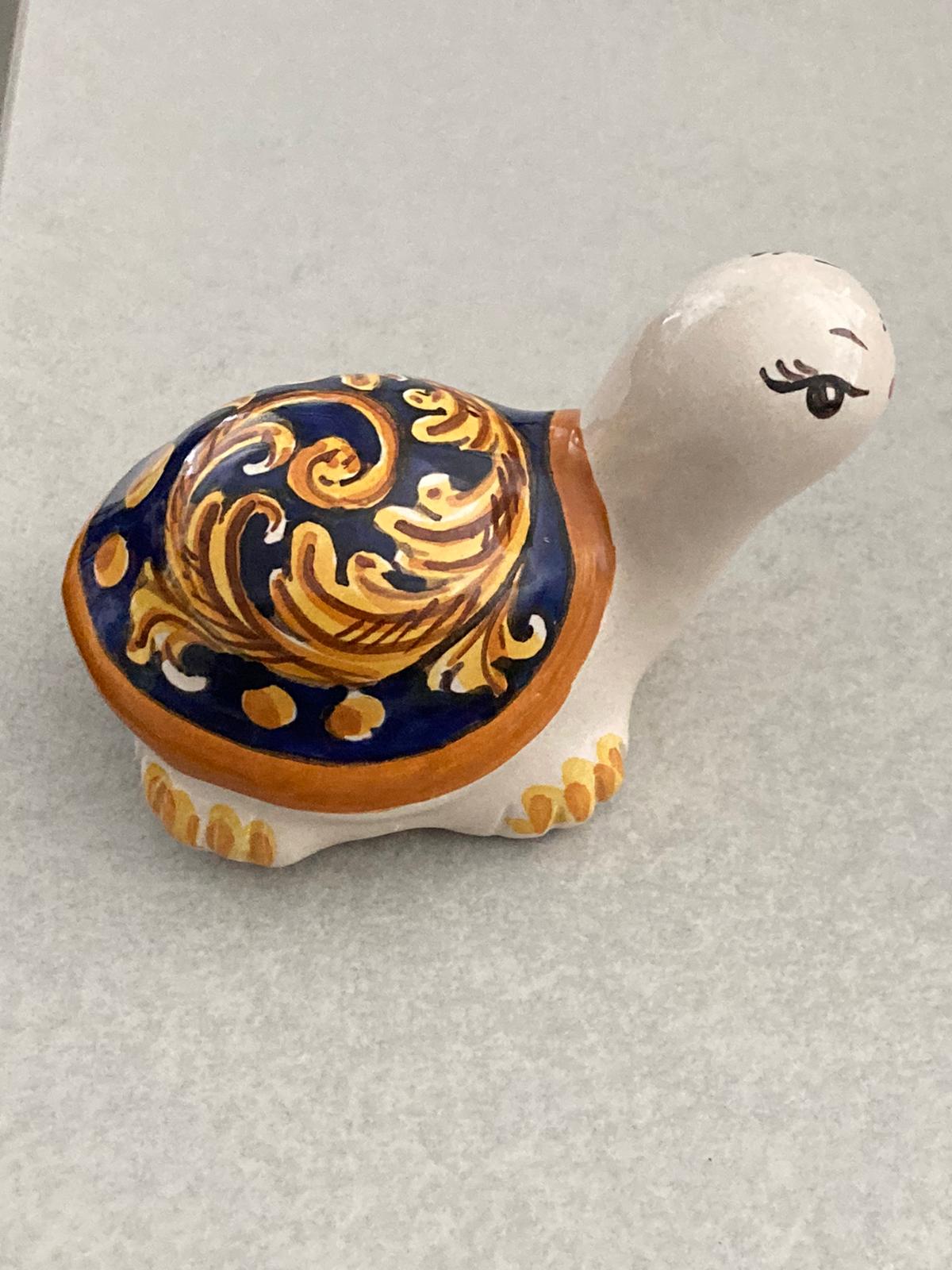}
        \end{tabular}
        
        &
        $\rightarrow$
        &
        \begin{tabular}{c}
        \includegraphics[width=0.736\linewidth]{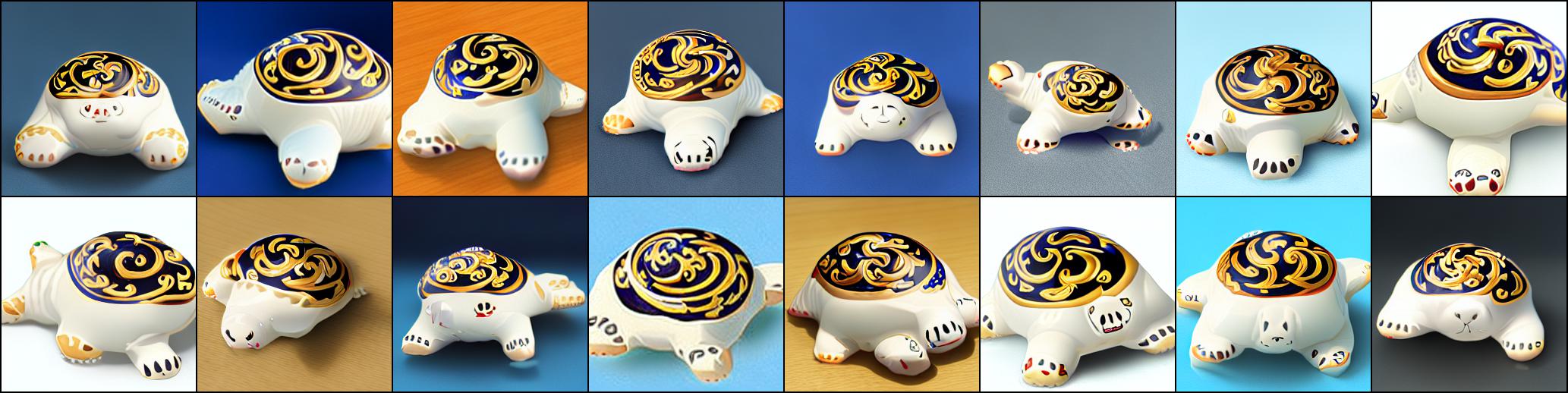}
        \end{tabular} \\
        
        {Input samples} & & {``A photo of \pholdercolor"} \\

    \end{tabular}}
    \caption{Uncurated samples of object variations created using the prompt "A photo of \pholdercolor".}
    \label{fig:uncurated_photo} 
\end{figure}

%% file: resources/figures/uncurated_prompt_1.tex
\begin{figure}[!hbt]
    \centering
    \setlength{\abovecaptionskip}{6.5pt}
    \setlength{\belowcaptionskip}{-3.5pt}
    \setlength{\tabcolsep}{0.55pt}
    \renewcommand{\arraystretch}{1.0}
    {
    \begin{tabular}{c@{\hskip 5pt} c@{\hskip 5pt} c}
    
        \multirow{2}{*}{
        \begin{tabular}{c c}
            \includegraphics[width=0.09\linewidth,height=0.09\linewidth]{resources/images/training_sets/rainbow_cat/2.jpeg} & 
            \includegraphics[width=0.09\linewidth,height=0.09\linewidth]{resources/images/training_sets/rainbow_cat/3.jpeg} \\
            \multicolumn{2}{c}{\includegraphics[width=0.09\linewidth,height=0.09\linewidth]{resources/images/training_sets/rainbow_cat/6.jpeg}} \\
            \multicolumn{2}{c}{{\color[HTML]{FFA600}$S_{cat}$}}
        \end{tabular}
        }
        &
        
        &
        \begin{tabular}{c}
        \includegraphics[width=0.736\linewidth]{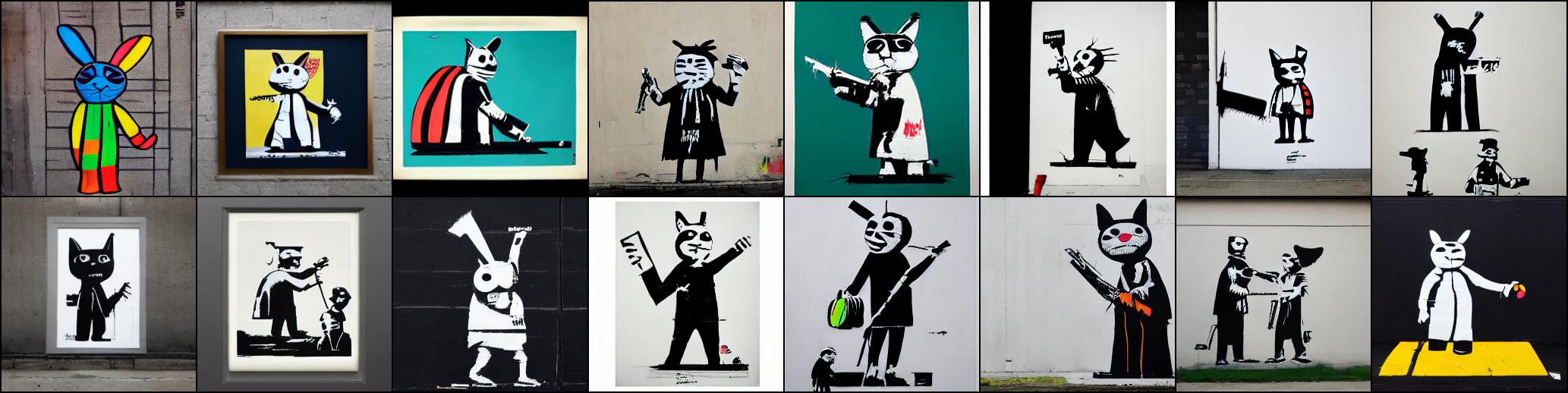}
        \end{tabular} \\
        
         & & {``Banksy art of {\color[HTML]{FFA600}$S_{cat}$}"} \\

        &
        
        &
        \begin{tabular}{c}
        \includegraphics[width=0.736\linewidth]{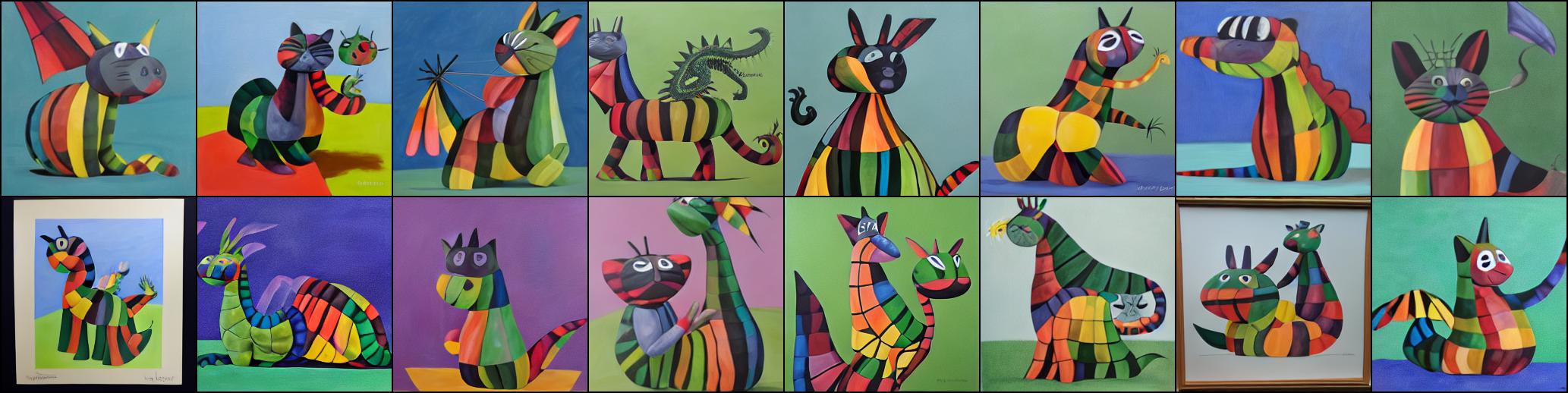}
        \end{tabular} \\
        
         & & {``Painting of a {\color[HTML]{FFA600}$S_{cat}$} riding a dragon"} \\

        \multirow{2}{*}{
        \begin{tabular}{c c}
            \includegraphics[width=0.1\linewidth,height=0.1\linewidth]{resources/images/training_sets/clock/1.jpeg} &
            \includegraphics[width=0.1\linewidth,height=0.1\linewidth]{resources/images/training_sets/clock/3.jpeg} \\[-2pt]
            \includegraphics[width=0.1\linewidth,height=0.1\linewidth]{resources/images/training_sets/clock/4.jpeg} &
            \includegraphics[width=0.1\linewidth,height=0.1\linewidth]{resources/images/training_sets/clock/5.jpeg} \\[-2pt]
            \multicolumn{2}{c}{{\color[HTML]{7A5195}$S_{clock}$}}
        \end{tabular}}
        
        &
        
        &
        \begin{tabular}{c}
        \includegraphics[width=0.736\linewidth]{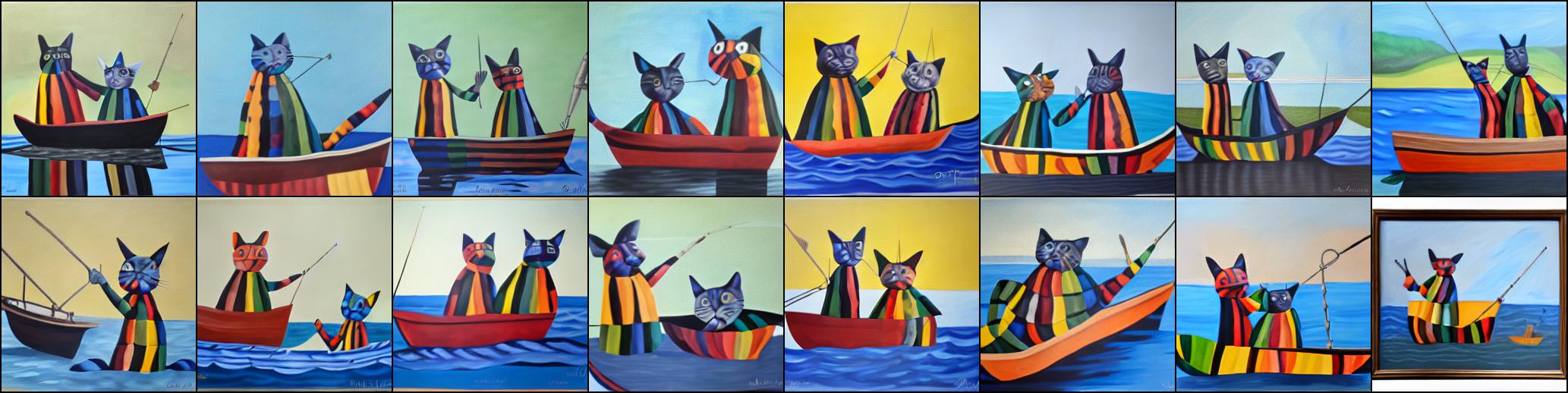}
        \end{tabular} \\
        
         & & {``Painting of two {\color[HTML]{FFA600}$S_{cat}$} fishing on a boat"} \\
        
        &
        
        &
        \begin{tabular}{c}
        \includegraphics[width=0.736\linewidth]{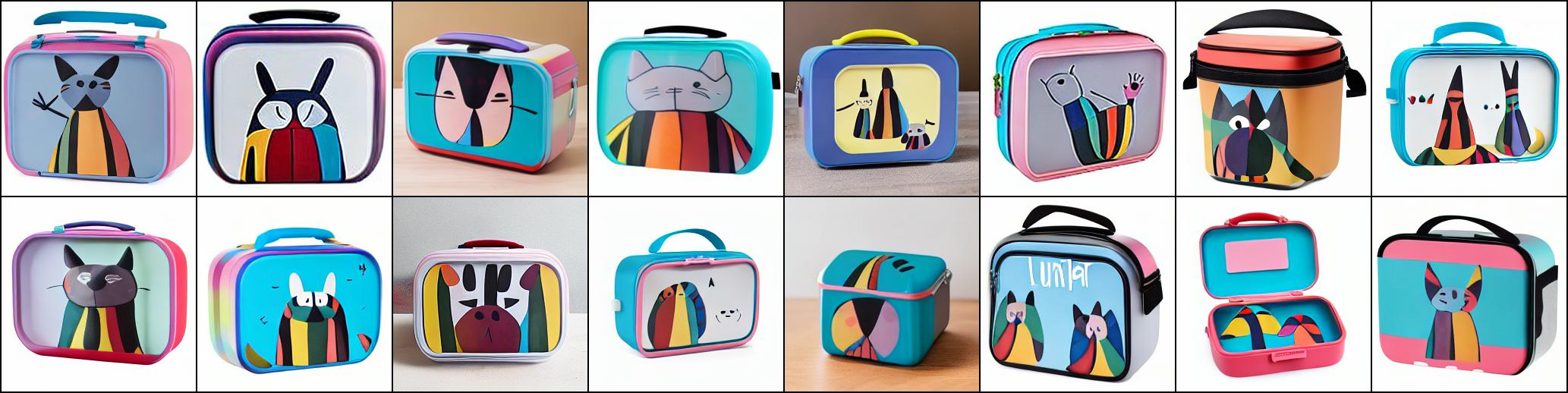}
        \end{tabular} \\
        
         & & {``A {\color[HTML]{FFA600}$S_{cat}$} themed lunchbox"} \\
        
        \multirow{2}{*}{
        \begin{tabular}{c c}
            \includegraphics[width=0.09\linewidth,height=0.09\linewidth]{resources/images/training_sets/fat_stone_bird/1.jpg} & 
            \includegraphics[width=0.09\linewidth,height=0.09\linewidth]{resources/images/training_sets/fat_stone_bird/2.jpg} \\
            \includegraphics[width=0.09\linewidth,height=0.09\linewidth]{resources/images/training_sets/fat_stone_bird/3.jpg} & 
            \includegraphics[width=0.09\linewidth,height=0.09\linewidth]{resources/images/training_sets/fat_stone_bird/4.jpg} \\
            \multicolumn{2}{c}{{{\color[HTML]{EF5675}$S_{bird}$}}}
        \end{tabular}
        }
        &
        
        &
        \begin{tabular}{c}
        \includegraphics[width=0.736\linewidth]{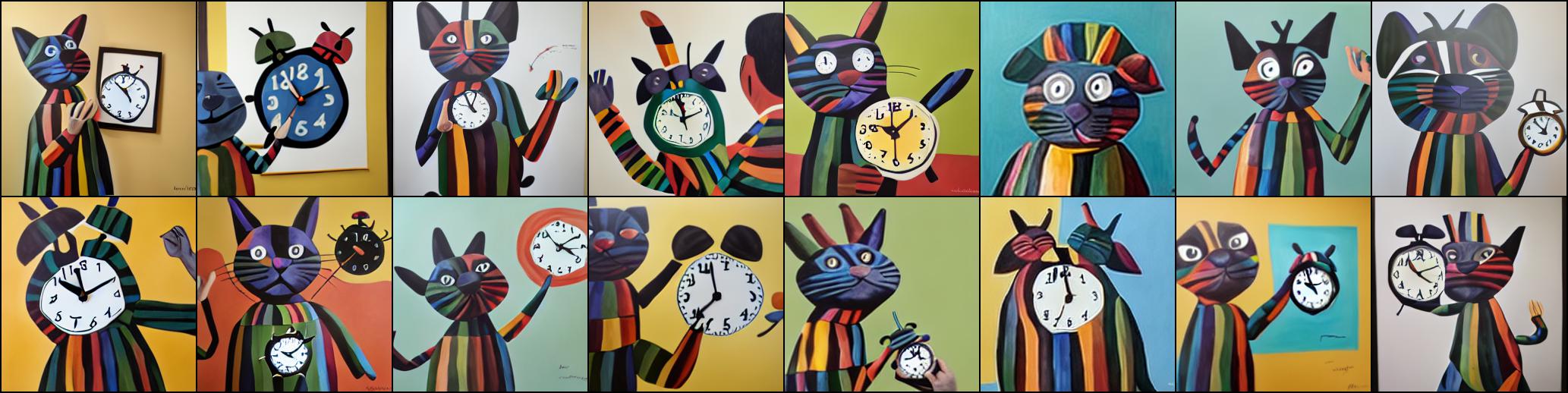}
        \end{tabular} \\
        
         & & {``A painting of {\color[HTML]{FFA600}$S_{cat}$} holding a {\color[HTML]{7A5195}$S_{clock}$}"} \\

        &
        
        &
        \begin{tabular}{c}
        \includegraphics[width=0.736\linewidth]{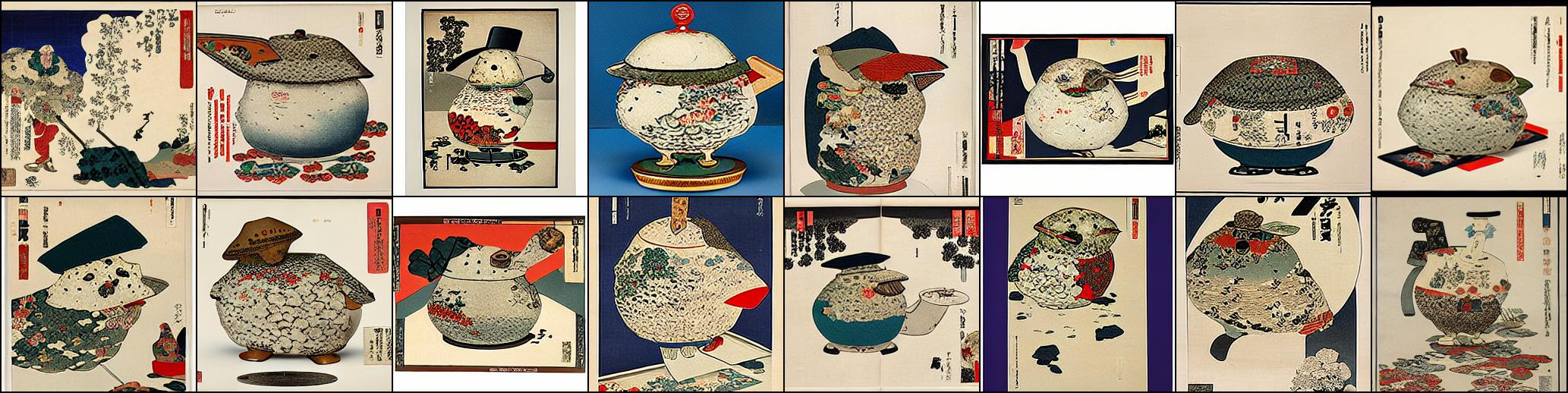}
        \end{tabular} \\
        
         & & {``Ukiyo-e painting of {\color[HTML]{EF5675}$S_{bird}$}"} \\

    \end{tabular}}
    \caption{Uncurated samples generated with context prompts. Quality and prompt-matching varies within the sample. However, we observe that a batch size of $16$ is typically sufficient to ensure several good samples.}
    \label{fig:uncurated_prompt_1} 
\end{figure}

%% file: resources/figures/uncurated_prompt_2.tex
\begin{figure}[!hbt]
    \centering
    \setlength{\abovecaptionskip}{6.5pt}
    \setlength{\belowcaptionskip}{-3.5pt}
    \setlength{\tabcolsep}{0.55pt}
    \renewcommand{\arraystretch}{1.0}
    {
    \begin{tabular}{c@{\hskip 5pt} c@{\hskip 5pt} c}
    
        \multirow{2}{*}{
        \begin{tabular}{c c}
            \includegraphics[width=0.09\linewidth,height=0.09\linewidth]{resources/images/training_sets/headless_statue/1.jpeg} & 
            \includegraphics[width=0.09\linewidth,height=0.09\linewidth]{resources/images/training_sets/headless_statue/2.jpeg} \\
            \includegraphics[width=0.09\linewidth,height=0.09\linewidth]{resources/images/training_sets/headless_statue/3.jpeg} & 
            \includegraphics[width=0.09\linewidth,height=0.09\linewidth]{resources/images/training_sets/headless_statue/4.jpeg}\\
            \multicolumn{2}{c}{{\color[HTML]{FFA600}$S_{sculpture}$}}
        \end{tabular}}
        
        &
        
        &
        \begin{tabular}{c}
        \includegraphics[width=0.736\linewidth]{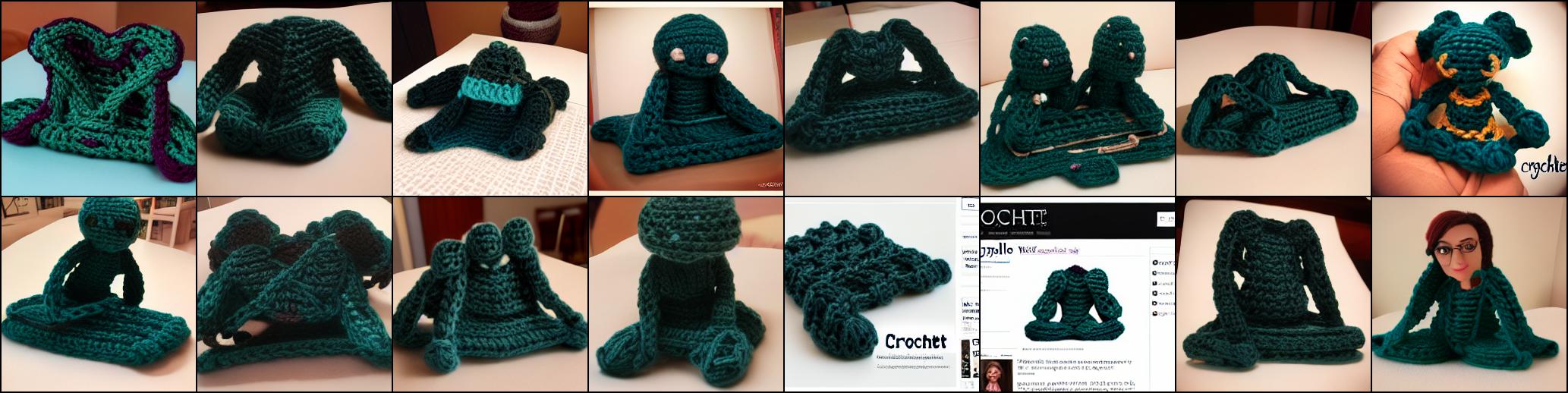}
        \end{tabular} \\
        
         & & {``Crochet {\color[HTML]{FFA600}$S_{sculpture}$}"} \\

        &
        
        &
        \begin{tabular}{c}
        \includegraphics[width=0.736\linewidth]{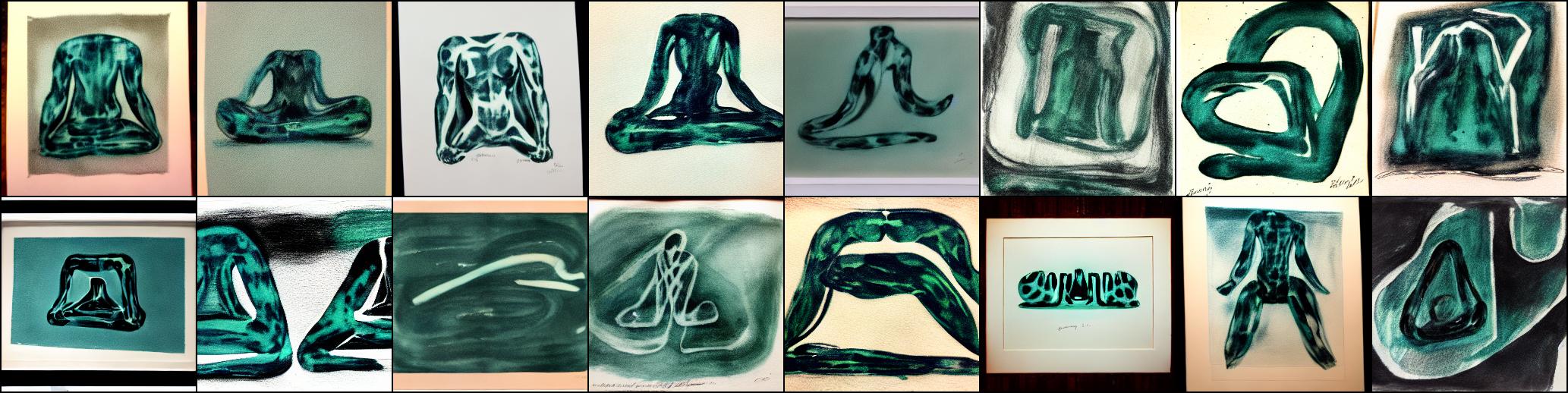}
        \end{tabular} \\
        
         & & {``{\color[HTML]{FFA600}$S_{sculpture}$} ink wash calligraphy"} \\

        \multirow{2}{*}{
        \begin{tabular}{c c}
            \includegraphics[width=0.1\linewidth,height=0.1\linewidth]{resources/images/training_sets/kid_monster/1.jpg} & \includegraphics[width=0.1\linewidth,height=0.1\linewidth]{resources/images/training_sets/kid_monster/2.jpg} \\
            \multicolumn{2}{c}{\includegraphics[width=0.1\linewidth,height=0.1\linewidth]{resources/images/training_sets/kid_monster/3.jpg}} \\
            \multicolumn{2}{c}{{\color[HTML]{003F5C}$S_{craft}$}}
        \end{tabular}}
        
        &
        
        &
        \begin{tabular}{c}
        \includegraphics[width=0.736\linewidth]{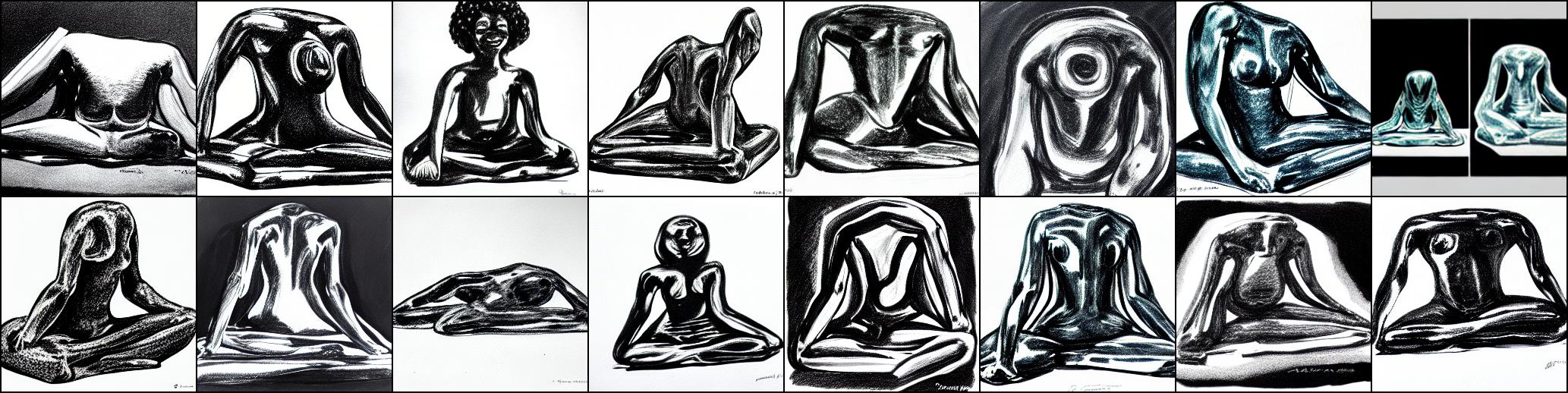}
        \end{tabular} \\
        
         & & {``A black and white sketch of {\color[HTML]{FFA600}$S_{sculpture}$}"} \\

        &
        
        &
        \begin{tabular}{c}
        \includegraphics[width=0.736\linewidth]{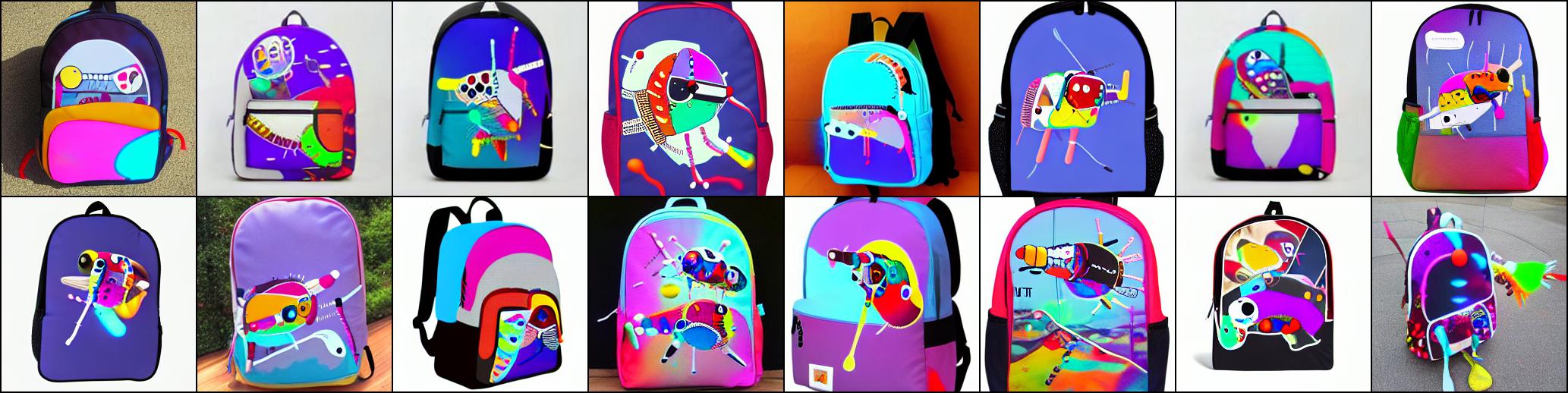}
        \end{tabular} \\
        
         & & {``A {\color[HTML]{003F5C}$S_{craft}$} backpack"} \\
        
        \multirow{2}{*}{
        \begin{tabular}{c c}
            \includegraphics[width=0.1\linewidth,height=0.1\linewidth]{resources/images/training_sets/qinni/1.jpg} & \includegraphics[width=0.1\linewidth,height=0.1\linewidth]{resources/images/training_sets/qinni/2.jpg} \\
            \includegraphics[width=0.1\linewidth,height=0.1\linewidth]{resources/images/training_sets/qinni/5.jpg} & \includegraphics[width=0.1\linewidth,height=0.1\linewidth]{resources/images/training_sets/qinni/4.jpg} \\
            \multicolumn{2}{c}{{\color[HTML]{EF5675}$S_{style}$}}
        \end{tabular}}
        
        &
        
        &
        \begin{tabular}{c}
        \includegraphics[width=0.736\linewidth]{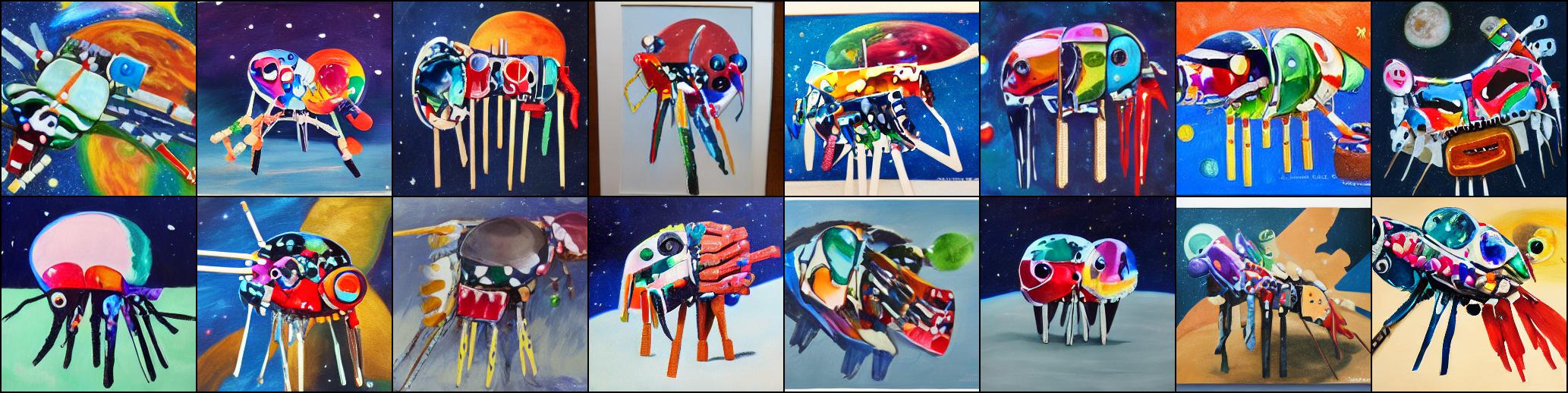}
        \end{tabular} \\
        
         & & {``A painting of {\color[HTML]{003F5C}$S_{craft}$} invading from space"} \\
        
        &
        
        &
        \begin{tabular}{c}
        \includegraphics[width=0.736\linewidth]{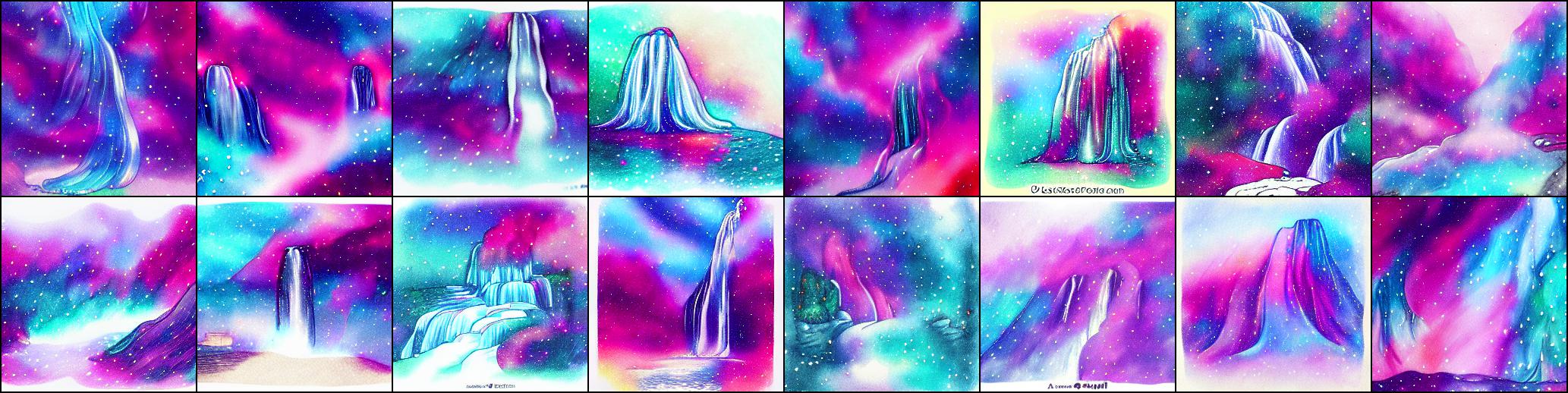}
        \end{tabular} \\
        
         & & {``A waterfall in the style of {\color[HTML]{EF5675}$S_{style}$}"} \\

    \end{tabular}}
    \caption{Additional uncurated samples generated with context prompts. Quality and prompt-matching varies within the sample. However, we observe that a batch size of $16$ is typically sufficient to ensure several good samples. Image credits: \href{https://www.deviantart.com/qinni}{@QinniArt} (bottom), authorized for non-commercial use only.}
    \label{fig:uncurated_prompt_2} 
\end{figure}